\documentclass[10pt,twocolumn,letterpaper]{article}

\usepackage[pagenumbers]{wacv} 

\usepackage{graphicx}
\usepackage{amsmath}
\usepackage{amssymb}
\usepackage{booktabs}

\usepackage{nicematrix}
\usepackage{mathtools}
\usepackage{subcaption}
\usepackage{arydshln}
\usepackage{dsfont}
\usepackage{pifont}
\usepackage{multirow}

\usepackage{makecell}

\usepackage[accsupp]{axessibility} 

\usepackage[pagebackref,breaklinks,colorlinks,bookmarksnumbered]{hyperref}
\usepackage{bookmark}

\usepackage[capitalize]{cleveref}
\crefname{section}{Sec.}{Secs.}
\Crefname{section}{Section}{Sections}
\Crefname{table}{Table}{Tables}
\crefname{table}{Tab.}{Tabs.}

\def\wacvPaperID{280} 
\def\confName{WACV}
\def\confYear{2025}

\newcommand{\notcheckmark}{\textcolor{green}{\checkmark\textcolor{red}{\makebox[-1pt][r]{\normalfont\symbol{'52}}}}}
\newcommand{\yescheckmark}{\textcolor{green}{\checkmark}}
\newcommand{\cross}{\textcolor{red}{\ding{55}}}

\begin{document}

\addtocontents{toc}{\protect\setcounter{tocdepth}{0}}

\title{VerA: Versatile Anonymization Applicable to Clinical Facial Photographs}

\author{
Majed El Helou$^{*1}$, Doruk Cetin$^{*2}$, Petar Stamenkovic$^1$, Niko Benjamin Huber$^2$, Fabio Z\"und$^1$ \\
$^1$ETH Z\"urich, Switzerland, $^2$Align Technology, Switzerland \\
{\tt\small \{majed.elhelou,petars,fabio.zund\}@ethz.ch}, {\tt\small \{dcetin,nhuber\}@aligntech.com}
}
\maketitle

\noindent\let\thefootnote\relax\footnotetext{$^*$These authors contributed equally to this work.}

\begin{abstract}
    The demand for privacy in facial image dissemination is gaining ground internationally, echoed by the proliferation of regulations such as GDPR, DPDPA, CCPA, PIPL, and APPI. While recent advances in anonymization surpass pixelation or blur methods, additional constraints to the task pose challenges.
    Largely unaddressed by current anonymization methods are clinical images and pairs of before-and-after clinical images illustrating facial medical interventions, e.g., facial surgeries or dental procedures. \\
    We present \textbf{VerA}, the first \textbf{Ver}satile \textbf{A}nonymization framework that solves two challenges in clinical applications: A) it preserves selected semantic areas (e.g., mouth region) to show medical intervention results, that is, anonymization is only applied to the areas outside the preserved area; and B) it produces anonymized images with consistent personal identity across multiple photographs, which is crucial for anonymizing photographs of the same person taken before and after a clinical intervention. 
    
    We validate our results on both single and paired anonymization of clinical images through extensive quantitative and qualitative evaluation. We also demonstrate that VerA reaches the state of the art on established anonymization tasks, in terms of photorealism and de-identification.
\end{abstract}

\vspace{-0.7cm}


\section{Introduction} \label{sec:introduction}
Emerging privacy legislation restrains medical practitioners from sharing images showing a patient's face. 
Such cases include photographs illustrating clinical results of reconstructive or restorative plastic facial surgeries, as well as dental interventions. It is therefore not possible to share results with future patients and other practitioners, or to create public datasets useful for machine learning.
Anonymization is thus necessary, yet it is challenging to achieve without severe compromise. Anonymizing by pixelation~\cite{thomas2018image} or blur~\cite{du2019efficient} is destructive and leads to unaesthetic images and a loss of information that is not permitted in clinical cases. This reduces the illustrative value, especially for interventions that contain an aesthetic aspect such as plastic surgery. 

\begin{figure}[t]
    \centering
    \includegraphics[trim={0 0 0 0},clip,width=\linewidth]{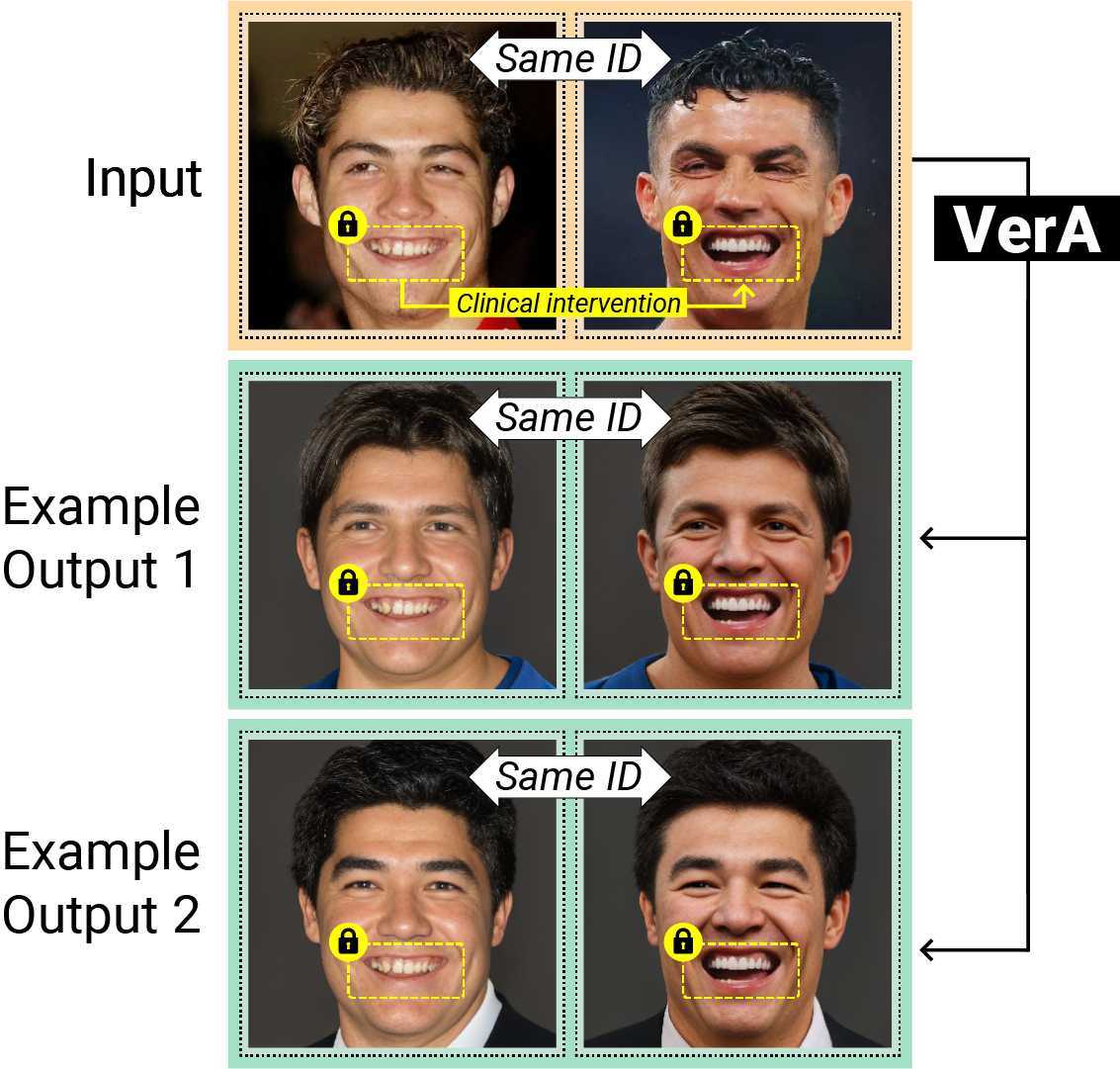}
    \caption{\textbf{Clinical paired image anonymization.} VerA anonymizes two photographs of a person before (left) and after (right) a clinical intervention in the mouth. \textbf{Top row:} the input image pair with a common identity (the same person, but different photographs) and a semantic region to preserve, which is the medically treated area. \textbf{Second and third row:}  two example outputs with each a before- (left) and an after-intervention image (right). In the example outputs, the faces except the preserved areas are completely anonymized while the persons' identities in the output images before and after the interventions are preserved.}
    \label{fig:banner}
\end{figure}

Novel image anonymization solutions emerged with the recent advances in learning-based computer vision that try to alleviate information destruction.
Some methods yield visually similar faces, either indirectly due to the chosen approach~\cite{wu2019privacy, gafni2019live}, or because they perform adversarial attacks~\cite{shan2020fawkes, yang2021towards, zhong2022opom, hu2022protecting}. 
However, most solutions use learned priors to synthesize a perceptually different face, exploiting strong prior-based hallucination~\cite{zhang2021face,elhelou2022bigprior}. 
They could yield aesthetic results, enable the use of anonymized images downstream, and the dissemination of useful datasets that would otherwise not be allowed~\cite{metz2019facial,peng2020facial}. 

A plethora of anonymization methods thus emerged. 
Some approaches inpaint~\cite{sun2018natural,hukkelaas2019deepprivacy,hukkelaas2023deepprivacy2}, or include a password~\cite{gu2020password,cao2021personalized,li2023riddle}. Others are end-to-end solutions focusing on actions~\cite{ren2018learning}, identity distance~\cite{wu2019privacy,gafni2019live}, generating new attributes~\cite{zhai2022a3gan}, or conditioning on a novel image~\cite{kuang2021effective}.
Multiple methods thus require as input a target identity~\cite{sun2018hybrid,maximov2020ciagan,kuang2021effective}. 
However, whether requiring extra input or not, all available anonymization methods focus on everyday \textbf{standard} single images with no constraints on preserving a certain area. They are therefore not readily applicable to \textbf{clinical} image anonymization where the semantic region targeted in medical intervention must be preserved (such as teeth in Fig.~\ref{fig:banner}).
Additionally, image \textit{pairs} of the same person need to be anonymized consistently for before-and-after medical results. 
These paired cases are challenging as the photographs may be taken months apart in different settings. This temporal incoherence can make it harder to achieve identity consistency across anonymized pairs.

VerA (\textbf{Ver}satile \textbf{A}nonymization) is a novel image anonymization framework solving various problems with valuable use cases. Our contributions are as follows:

\vspace{-0.1cm}

\begin{itemize}
    \setlength\itemsep{0em}
    \item To the best of our knowledge, we are the first to formulate and address the problems of clinical and paired image anonymization.
    \item We introduce a framework that exploits our novel generator, which is the first with \textit{both} high-level and semantic explicit control, and our specialized inversion tailored to our anonymization tasks.
    \item We show through extensive experiments that VerA can consistently anonymize clinical image \textit{pairs} with common \textit{identity} and preserve desired \textit{semantic} components. Our evaluation shows that VerA outperforms state-of-the-art anonymization methods in photorealism while achieving high de-identification rates.
\end{itemize}

\vspace{-0.3cm}

\begin{figure*}[t]
    \centering
    \includegraphics[width=\linewidth, trim={0 0 0 0},clip]{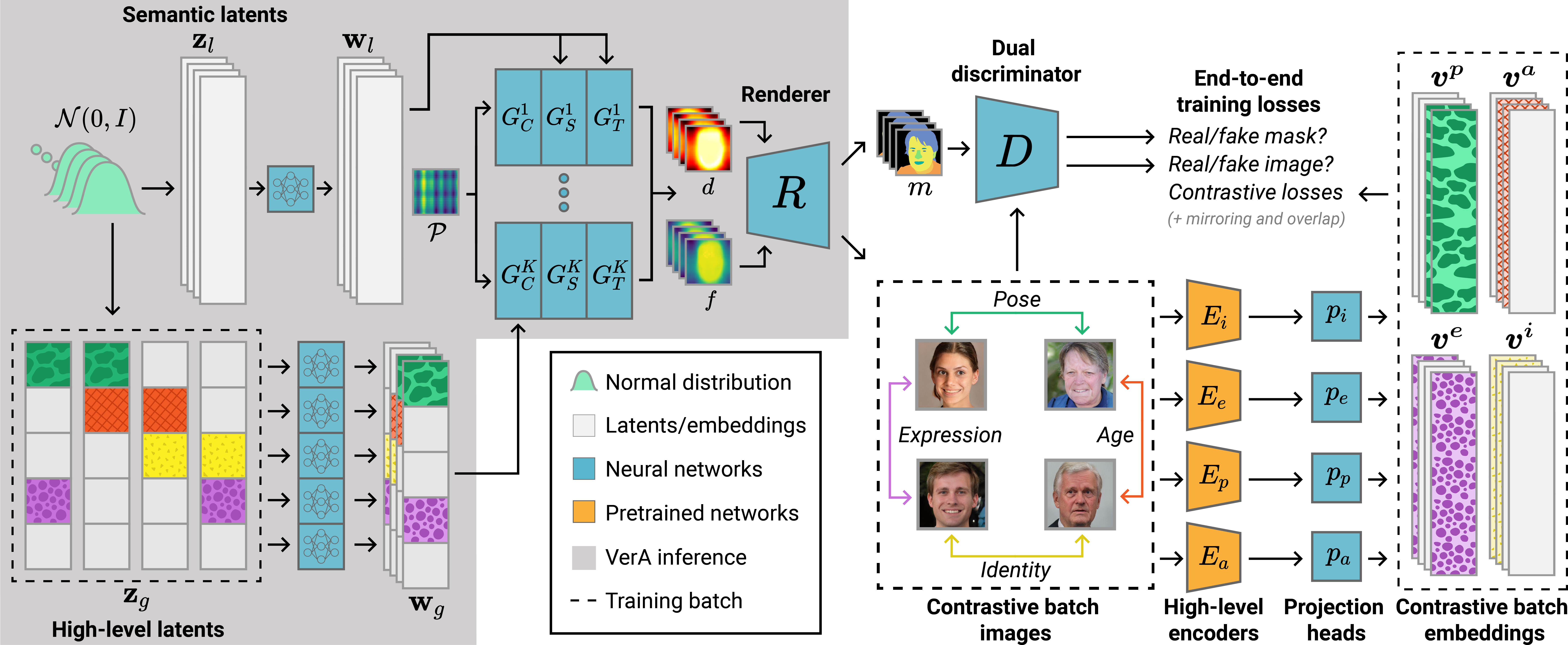}
    \caption{VerA exploits a novel controllable image generator with specialized inversion designed for anonymization tasks, unlike previous methods built on pretrained generators.
    Compared to SemanticStyleGAN~\cite{shi2022semanticstylegan} that can only control semantics, our illustrated generator has a dual latent space (\textbf{z}$_g$) with individual mappings for high-level attributes. We adapt the conditioning layers of each generator to the dual space, and design a generative contrastive learning approach to learn the high-level controls.}
    \label{fig:pipeline}
\end{figure*}

\section{Related Work} \label{sec:related}
\noindent \textbf{Image anonymization.} Multiple anonymization solutions alter attributes~\cite{li2019anonymousnet} in one or more generative latent spaces~\cite{ma2021cfa,wen2022identitydp}.
Gu~\emph{et al.}~\cite{gu2020password} transform the image with a password that conditions the de-identification, which is reversible.
PI-Net~\cite{chen2021perceptual} similarly inverts the input to the latent space where it gets manipulated. Subsequently, Cao~\emph{et al.}~\cite{cao2021personalized} propose two encoders for attribute and identity latents. The constraints imposed by the password and the reversibility limit the performance in photorealism and flexibility (for instance with pose~\cite{cao2021personalized}). 
In practice, anonymized images rarely need de-anonymization as the owners can access the originals and can share them securely with encryption.
Gafni~\emph{et al.}~\cite{gafni2019live} anonymize videos by slightly editing images in the identity direction then fusing the result with the input. With the temporal coherence in videos and fusion blur, they achieve identity consistency and few artifacts. However, resulting images are perceptually similar and are of low quality. 
CIAGAN~\cite{maximov2020ciagan} builds on image-to-image translation, guiding with an external image and face landmarks to preserve pose, and inspired a similar approach in~\cite{dall2022graph}. A$^3$GAN~\cite{zhai2022a3gan} removes identity information with suppressive convolution followed by attribute regeneration. The authors claim improved expression control compared with CIAGAN~\cite{maximov2020ciagan}.  
DeepPrivacy~\cite{hukkelaas2019deepprivacy} exploits inpainting; a bounding box delimits the face area inpainted with a U-Net~\cite{ronneberger2015u}. Building on their previous work, the authors propose full-body anonymization~\cite{hukkelaas2023realistic} and DeepPrivacy2~\cite{hukkelaas2023deepprivacy2} (DP2) that also does inpainting anonymization. While this improves photorealism, it trades-off control. The target's \textit{high-level attributes, semantics, and identity, are not controllable}. This implies that anonymizing two images of a person cannot guarantee a consistent synthetic identity. 
That is also the case of IDeudemon~\cite{wen2023divide} that perturbs the identity in a 3D latent space. Starting with a single 2D image can result in degraded synthetic images when mapping back from 3D, which IDeudemon regenerates with a pretrained GFP-GAN~\cite{wang2021towards}. 
RiDDLE~\cite{li2023riddle} trains an encryptor with reconstruction and identity losses over the latent space of a pretrained StyleGAN2~\cite{karras2020analyzing}. It achieves reversibility but degrades photorealism.
FALCO~\cite{barattin2023attribute} searches for the input's nearest neighbor in a large synthetic image set. It optimizes the neighbor's latent, over a pretrained StyleGAN2, to make the synthetic image similar to the input. This similarity is both in the FaRL~\cite{zheng2022general} feature space to preserve attributes, and in identity space up to a threshold. However, its attribute preservation results are only on par with DP2's inpainted results. 
In contrast, we control high-level attributes including identity as well as the semantics of anonymized images, to address the various anonymization objectives (Table~\ref{table:method_comparison}).

\begin{table}[t]
    \centering
    \resizebox{\linewidth}{!}{
    \begin{NiceTabular}{l|c|c|c|c|c|c|c}
    \toprule
    Method  & CIAGAN & FIT & DP2 & RiDDLE & FALCO & IDeudemon & Ours  \\
    Year \& Ref. & '20~\cite{maximov2020ciagan} & '20~\cite{gu2020password} & '23~\cite{hukkelaas2023deepprivacy2} & '23~\cite{li2023riddle} & '23~\cite{barattin2023attribute} & '23~\cite{wen2023divide} & - \\
    \midrule
    Generation & CIAGAN & FIT & DP2 & \multicolumn{2}{c}{StyleGAN2} & GFP-GAN & Ours \\
    De-ID & \yescheckmark & \yescheckmark & \yescheckmark & \yescheckmark & \yescheckmark & \yescheckmark & \yescheckmark \\
    Photorealism & \cross & \cross & \yescheckmark & \notcheckmark & \notcheckmark & \yescheckmark & \yescheckmark \\ 
    In-place anon & \cross & \cross & \yescheckmark & \notcheckmark & \cross & \yescheckmark & \yescheckmark \\
    Clinical anon & \cross & \cross & \cross & \cross & \cross &  \textcolor{red}{?} & \yescheckmark \\
    Paired anon & \cross & \cross & \cross & \cross & \cross & \cross & \yescheckmark \\
    Clinical pair & \cross & \cross & \cross & \cross & \cross & \cross & \yescheckmark \\
    \bottomrule
    \end{NiceTabular}
    }
    \caption{High-level comparison with the two most commonly referenced baselines and four most recent state-of-the-art anonymization methods. \notcheckmark: RiDDLE's weaker photorealism is likely due to the encryption over the pretrained StyleGAN2~\cite{karras2020analyzing} generator. \\\notcheckmark: FALCO has adaptive normalization that can lead to washed-out images or odd color artifacts if toggled off (supplementary). \\\textcolor{red}{?}: IDeudemon might be adaptable, but this is not addressed by the authors and their code and models are not made available.}
    \label{table:method_comparison}
\end{table}

\noindent \textbf{Image generation and inversion.} With the progress from PGGAN~\cite{karras2018progressive} to StyleGAN~\cite{karras2019style,karras2020analyzing}, a multitude of methods invert and edit images in the latent space. For inversion, ReStyle~\cite{alaluf2021restyle} encodes with iterative correction. HyperStyle~\cite{alaluf2022hyperstyle} uses HyperNetworks~\cite{ha2016hypernetworks} to edit the generator itself. Recent methods balance invertibility and editability by exploiting different latent depths and rates~\cite{xu2022transeditor,hu2022style,parmar2022spatially,wang2022high}, or human feedback~\cite{davis2022brain}. Learned latent space encoding is outside our scope. We invert by optimization, thus retaining flexibility. Encoders can improve efficiency, and most approaches are readily applicable to our solution.

\noindent \textbf{Image editing.} 
GANSpace~\cite{harkonen2020ganspace} edits images along PCA directions. Focusing on faces, InterFaceGAN~\cite{shen2020interpreting} and latent transformer~\cite{yao2021latent} search for directions to edit attributes such as age or expression, and REDs~\cite{balakrishnan2022rayleigh} performs attribute-based optimization to search for a traversal path per image. 
StyleCLIP\cite{patashnik2021styleclip} does text-guided optimization with CLIP embeddings~\cite{radford2021learning}. 
These editing methods aim to preserve identity. On the contrary, StyleID~\cite{le2022styleid} searches for disentangled identity dimensions to manipulate, while preserving other features. StyleFace~\cite{luo2022styleface} learns an identity projector to map to a space where identity is better disentangled, similar to the encoders of FICGAN~\cite{jeong2021ficgan}.
However, due to the complex latent-space entanglement, a single global editing remains sub-optimal. To mitigate this, StyleFlow~\cite{abdal2021styleflow} conditions the editing on the desired attribute. However, a more robust and interpretable approach is to disentangle the generative process. SEAN~\cite{zhu2020sean} conditions the generation on semantic masks, enabling the transfer of specific semantic components. GAN-Control~\cite{shoshan2021gan} learns attribute encoders and enforces disentanglement of their latent spaces. SemanticStyleGAN~\cite{shi2022semanticstylegan} explicitly separates the generation of each semantic component. We redesign its architecture and training to create a generator that has not only explicitly disentangled semantics but also explicit latent spaces for high-level attributes. Rather than generic editing methods, not fit for image pairs, VerA exploits our tailored generator to propose specialized anonymization solutions.

\section{Method} \label{sec:method}
VerA can anonymize standard facial images as well as clinical images that may come in pairs (before-and-after images) and require the accurate preservation of semantic components. To solve the problems of clinical and paired anonymization, we train a generator where high-level attributes and semantics are explicitly disentangled and controlled. We propose a semantics-aware inversion specialized to single and paired cases, and present our approach for versatile anonymization.  

\subsection{Controllable generator} \label{sec:method_generator}
We illustrate VerA's generator training in Fig.~\ref{fig:pipeline}.

\noindent \textbf{Model overview.} Our model builds on the rich image generation literature. The backbone builds on StyleGAN2~\cite{karras2020analyzing} with multiple semantic generators trained explicitly on semantic masks similar to SemanticStyleGAN~\cite{shi2022semanticstylegan}. In contrast to the latter, we further inter-disentangle semantic and high-level attributes, and over-ride part of the generator control with dual latent vectors. 
Our latent vector \textbf{z} is composed of \{\textbf{z}$_l$, \textbf{z}$_g$\} sampled from normal distributions, referring to local semantic and global high-level attributes. The global vector is further decomposed into non-overlapping sub-vectors \{\textbf{z}$_g^i$, \textbf{z}$_g^e$, \textbf{z}$_g^p$, \textbf{z}$_g^a$\} referring to the high-level attributes of identity, expression, pose, and age, plus an unconstrained sub-vector. We train individual MLPs for each of \{\textbf{z}$_l$, \textbf{z}$_g^i$, \textbf{z}$_g^e$, \textbf{z}$_g^p$, \textbf{z}$_g^a$\} to preserve independence, and obtain the latent codes \{\textbf{w}$_l$, \textbf{w}$_g$\} after regrouping the global sub-codes. 
Borrowing from SemanticStyleGAN~\cite{shi2022semanticstylegan}, each feature generator (one per semantic component) is a composition of three generators $G_C^k$, $G_S^k$, $G_T^k$ for coarse, structure, and texture information, $k \in [1,K]$. However, we over-ride the coarse generator's control with our high-level latents. The generators are initialized with positional Fourier features $\mathcal{P}$~\cite{rahimi2007random,zhongreconstructing}. We obtain the generated features $f^k$ and the corresponding attention maps $d^k$ by composition
\begin{equation}
    f^k, d^k = G_T^k( G_S^k( G_C^k(\mathcal{P}, \text{\textbf{w}}_g ), \text{\textbf{w}}_l^k), \text{\textbf{w}}_l^k).
\end{equation}
A cascaded network $R$ renders the final $512\times 512$ resolution image and the semantic map residual. $R$ takes as input $\sum_{k=1}^K \text{Softmax}(\{d\}_{k=1}^K)^k \odot f^k $. The adversarial training procedure and regularization losses build on~\cite{karras2020analyzing,shi2022semanticstylegan}. However, we additionally require a separate learning objective for our high-level controls that we achieve through contrastive learning as explained in what follows.

\noindent \textbf{Generative contrastive learning.} We constrain our global latent space \textbf{z}$_g$ to propose a contrastive learning over the generator. Within a batch \{\textbf{z}$_g$\}$_B$, we create contrastive pairs that share identical sub-vectors of \textbf{z}$_g$ for different high-level attributes. From this batch of latent vectors, we create a batch of synthetically generated images. We pass these images through the pretrained and frozen feature extractors of classification networks ($E$ blocks in Fig.~\ref{fig:pipeline}). The classifiers are ArcFace~\cite{deng2019arcface} for identity, ESR~\cite{siqueira2020efficient} for expression, Hopenet~\cite{ruiz2018fine} for pose, and DEX~\cite{rothe2015dex} for age. We then pass the synthetic-image feature embeddings through learnable projection heads~\cite{chen2020simple} ($p$ blocks in Fig.~\ref{fig:pipeline}), and compute a modified contrastive loss on the resulting vector batches (e.g. age batch $\{v^a\}_B$). The number of positive and negative pairs is critical in contrastive learning~\cite{awasthi2022more}. To maximize the number of positive pairs per batch, we exploit the fact that our sub-vectors within \textbf{z}$_g$ are non-overlapping, and can obtain up to $N$ positive pairs sharing different attributes per batch of size $N$. We also virtually double the number of negative pairs per batch by proposing a \textit{mirrored} contrastive loss, which extends on SimCLR~\cite{chen2020simple, saunshi2022understanding}. We do so by exploiting each of the two positive pairs as different anchors. We provide ablation studies in our supplementary on the mirroring and projection heads. Our \textit{mirroring} strategy is possible because although $\textbf{z}_g^a[\alpha]=\textbf{z}_g^a[\beta]$ for the age attribute as an example, other high-level sub-vectors in \textbf{z}$_g$ are different, for instance expression $\textbf{z}_g^e[\alpha]\neq\textbf{z}_g^e[\beta]$. This difference leads to varied synthetic results at the renderer level and hence different synthetic-image vector batches $\{v\}_B$. We define our contrastive loss for each high-level attribute over its corresponding vector batch, for instance the age vector batch $\{v^a\}_B$, as 
\begin{equation}
    - \log \frac{g(v^a[\alpha], v^a[\beta])^2}{ \sum_{\forall \gamma \neq \alpha} g(v^a[\alpha], v^a[\gamma]) * \sum_{\forall \gamma \neq \beta} g(v^a[\beta], v^a[\gamma])},
\end{equation}
where $\alpha$, $\beta$, $\gamma$, are indices within the batch \{\textbf{z}$_g$\}$_B$ such that \textbf{z}$_g^a[\alpha]=$\textbf{z}$_g^a[\beta]$, and $g$ is given by
\begin{equation}
    g(u, v) \coloneqq \exp\left(
        \frac{1}{\tau} \frac{v^T\cdot u}{ ||u||_2 \cdot ||v||_2}
    \right),
\end{equation}
where $\tau$ is a constant and $\exp$ is the exponential function. We add to the loss the sum of all contrastive losses of all attributes without tweaking their weights.

\begin{figure}[t]
    \centering
    \includegraphics[width=\linewidth]{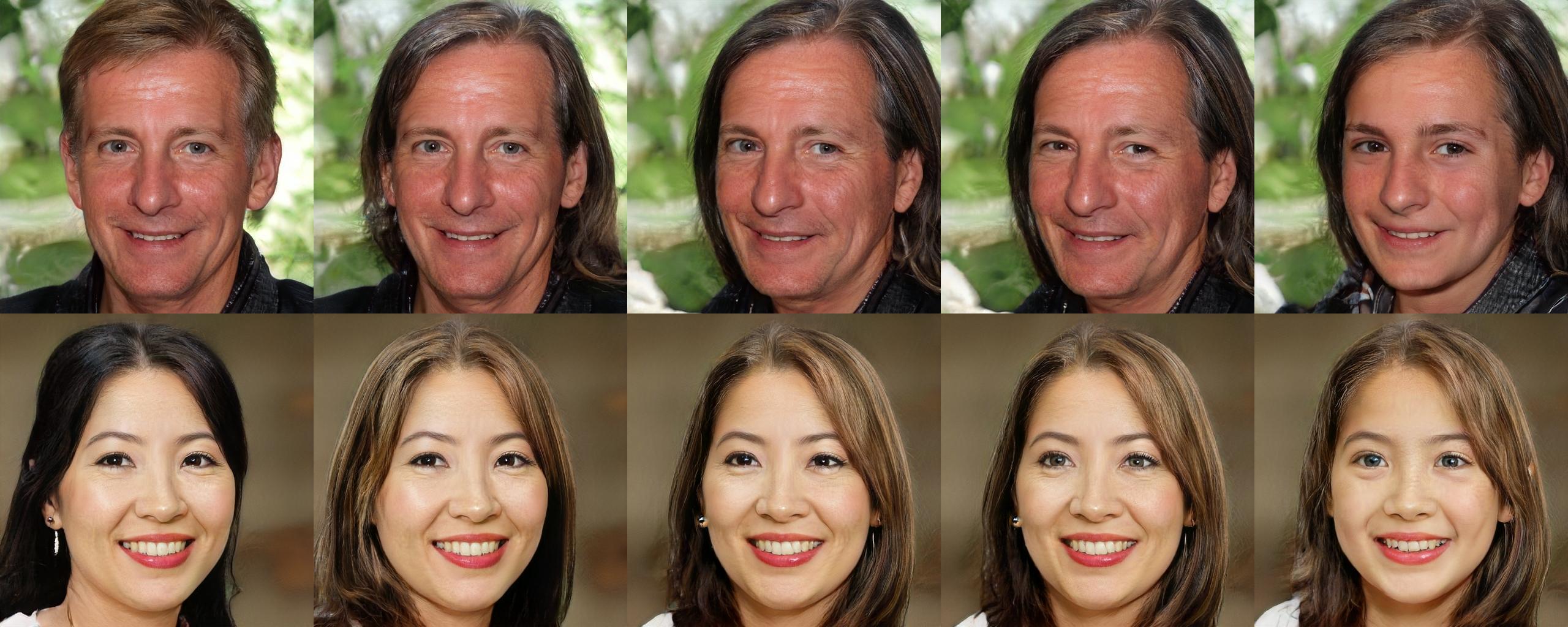}
    \caption{Illustration of accumulated semantic and high-level edits (left to right modifications: hair, pose, eyes, and age) with our trained image generator.}
    \label{fig:control_joint}
\end{figure}

\begin{figure}[t]
    \centering
    \begin{subfigure}[b]{0.185\linewidth}
        \includegraphics[clip,width=\textwidth]{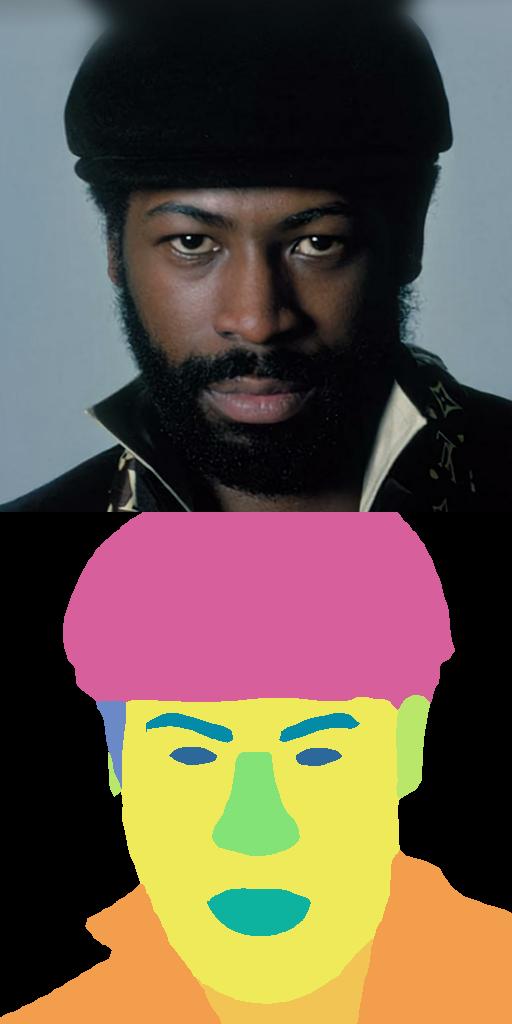}
        \caption{Input}
        \label{fig:inversion_segmentation_a}
    \end{subfigure}
    \begin{subfigure}[b]{0.37\linewidth}
        \includegraphics[clip,width=\textwidth]{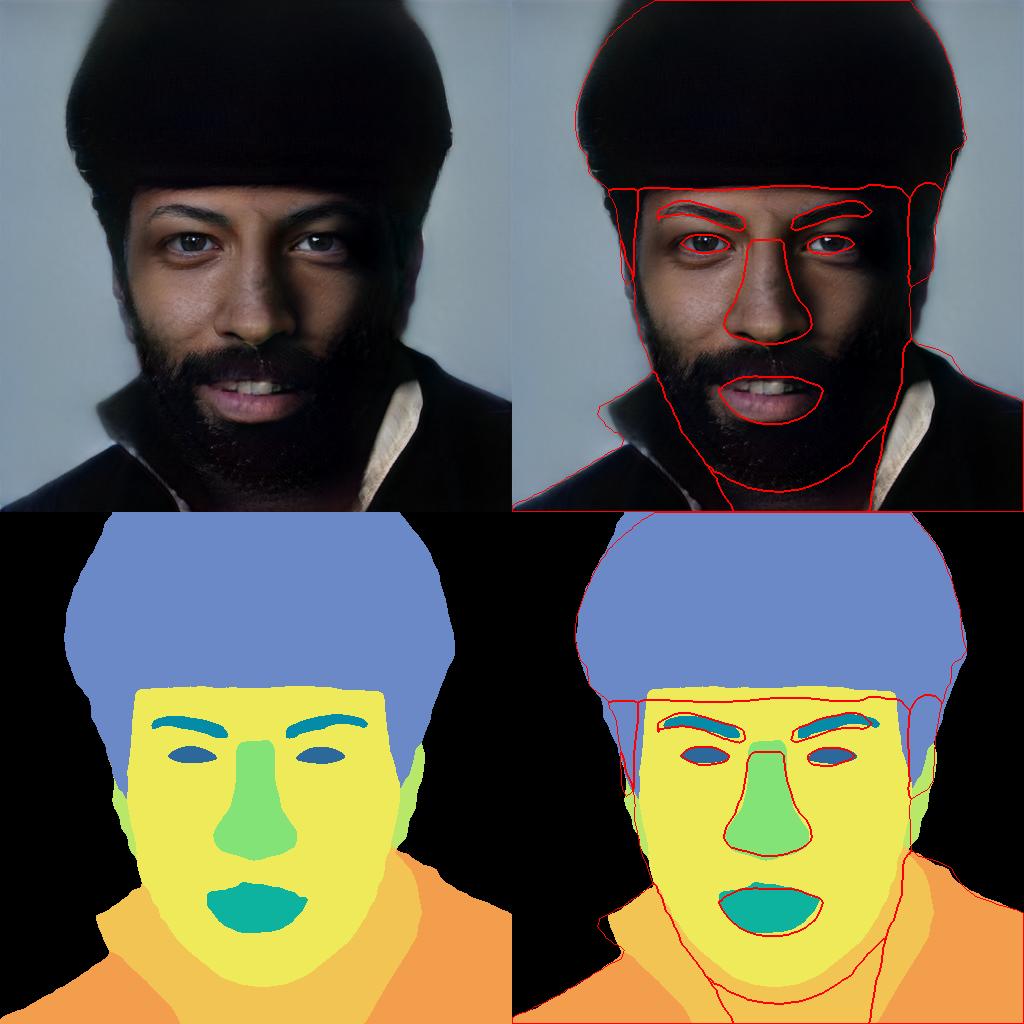}
        \caption{Inverted w/o guidance}
        \label{fig:inversion_segmentation_b}
    \end{subfigure}
    \begin{subfigure}[b]{0.37\linewidth}
        \includegraphics[clip,width=\textwidth]{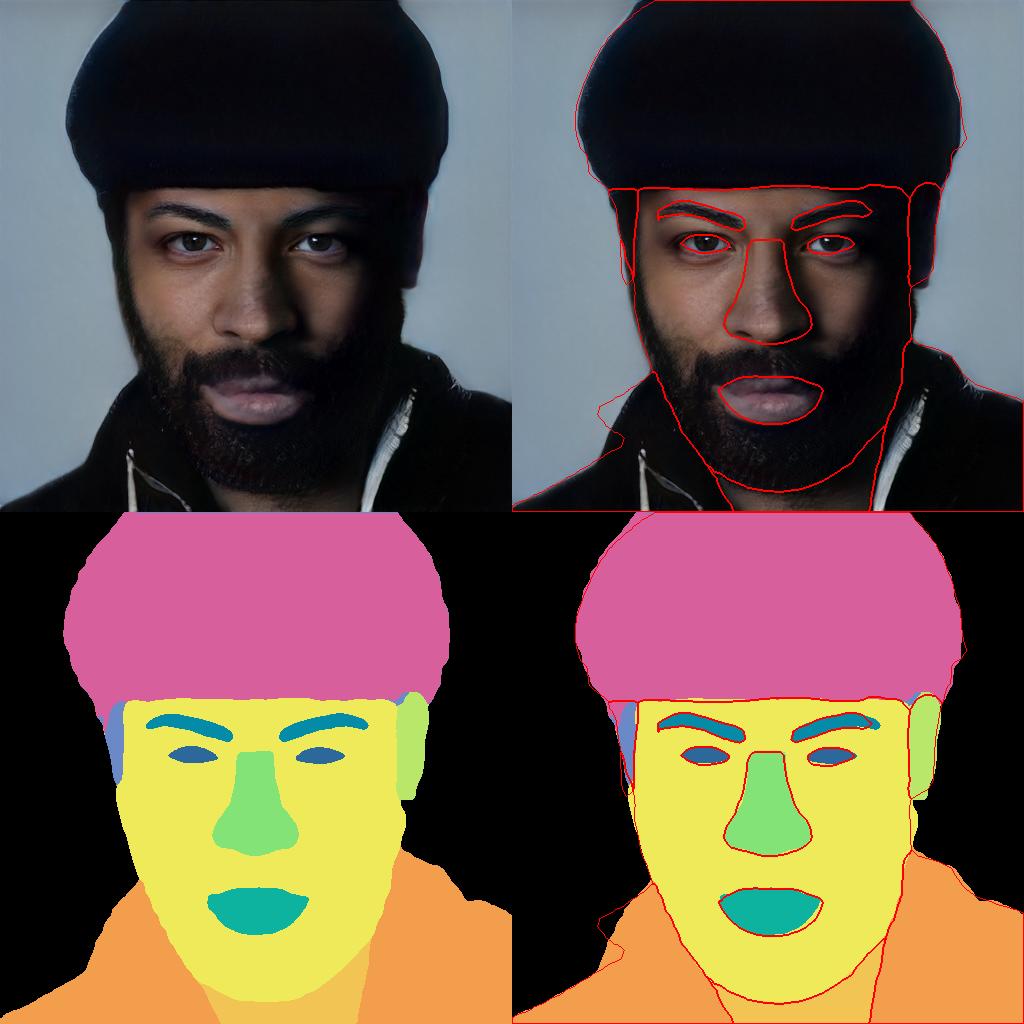}
        \caption{Inverted w/ guidance}
        \label{fig:inversion_segmentation_c}
    \end{subfigure}
    \caption{The effect of segmentation loss on inversion. (a) Input and predicted segmentation, (b) inversion results without and (c) with segmentation loss. Overlaid edges show the reference segmentation from the input (a), illustrating improved inversion.}
    \label{fig:inversion_segmentation}
\end{figure}

\subsection{Specialized inversion} \label{sec:method_inversion}
We propose a specialized inversion strategy to invert the input into the latent space that 1) exploits our disentangled latent space, 2) exploits the inherently generated semantic masks, and 3) specializes to the paired anonymization scenario. We invert into our disentangled extended space $\mathcal{W^+}$ that is composed of the global vector \textbf{w}$_g$, and $2*K$ different \textbf{w}$_l$ local vectors (one per semantic generator's structure and texture networks), for increased flexibility~\cite{abdal2019image2stylegan}. We specialize our inversion approach to our different anonymization setups, namely for single and multiple images.
 
\noindent \textbf{Single image.} For single-image anonymization, we invert by optimizing over $\mathcal{W^+}$, with our network weights frozen. We optimize a loss composed of an elastic net penalty ($\ell 1$ and $\ell 2$ losses), LPIPS~\cite{zhang2018unreasonable}, mean regularization pushing to the expected value of the latent space, and a semantic segmentation cross-entropy loss. We readily have semantic labels generated by the renderer for our synthetic images. For the input, we use a pretrained semantic segmentation model to obtain the target labels. We thus compute the cross-entropy loss over all semantic labels between the generated semantic maps and the target labels of the input. 

\noindent \textbf{Paired images.} 
In contrast to the single-image cases, we are interested in inverting multiple images of the same person. 
We use the same losses and optimization as the single-image inversion, including weights and optimization parameters. However, the particularity when inverting multiple images is that we exploit our global vector for high-level attributes to enforce that identity latents are joint across the different images, as they represent the same person.

\begin{figure}[t]
\centering
\begin{tabular}{cccc}
\hspace{-0.9cm}
\begin{tabular}{c} \includegraphics[width=0.245\linewidth]{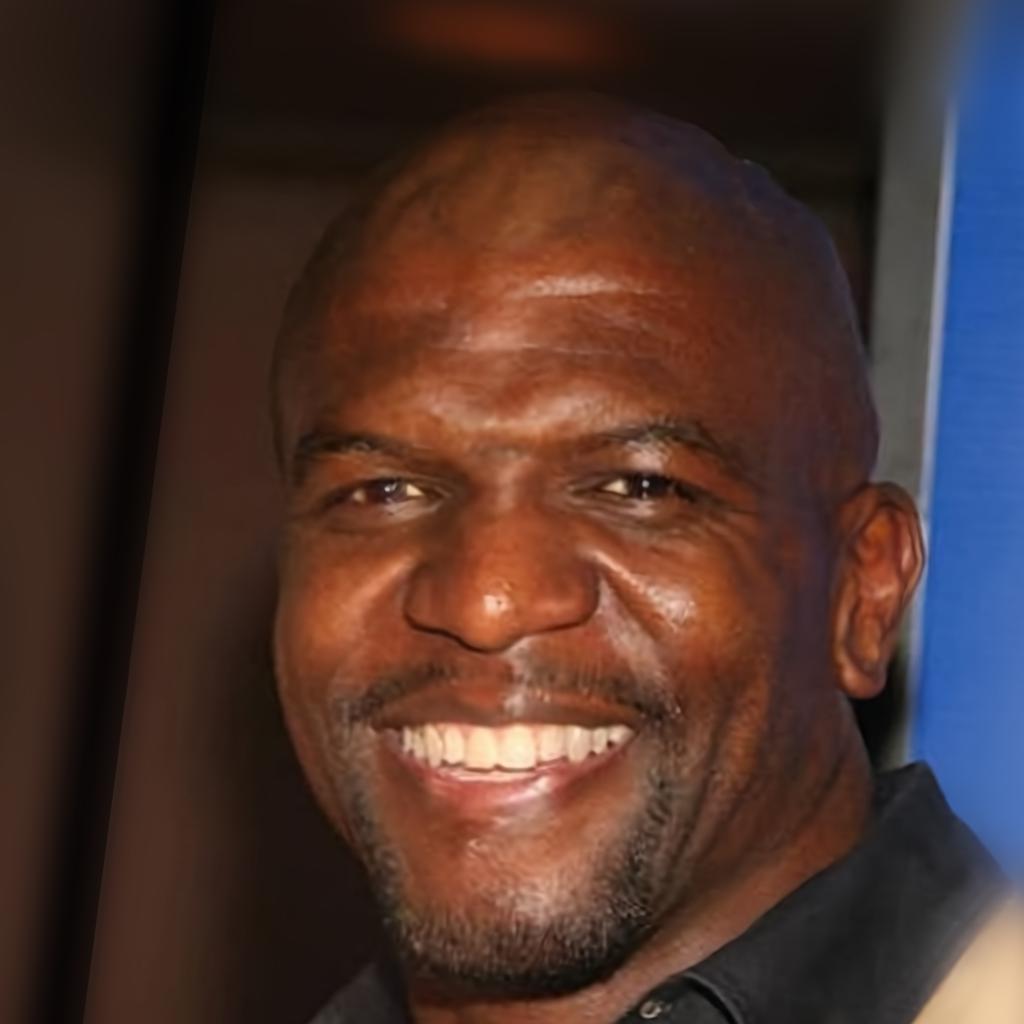} \end{tabular} \hspace{-0.9cm}
&
\begin{tabular}{c} \includegraphics[width=0.245\linewidth]{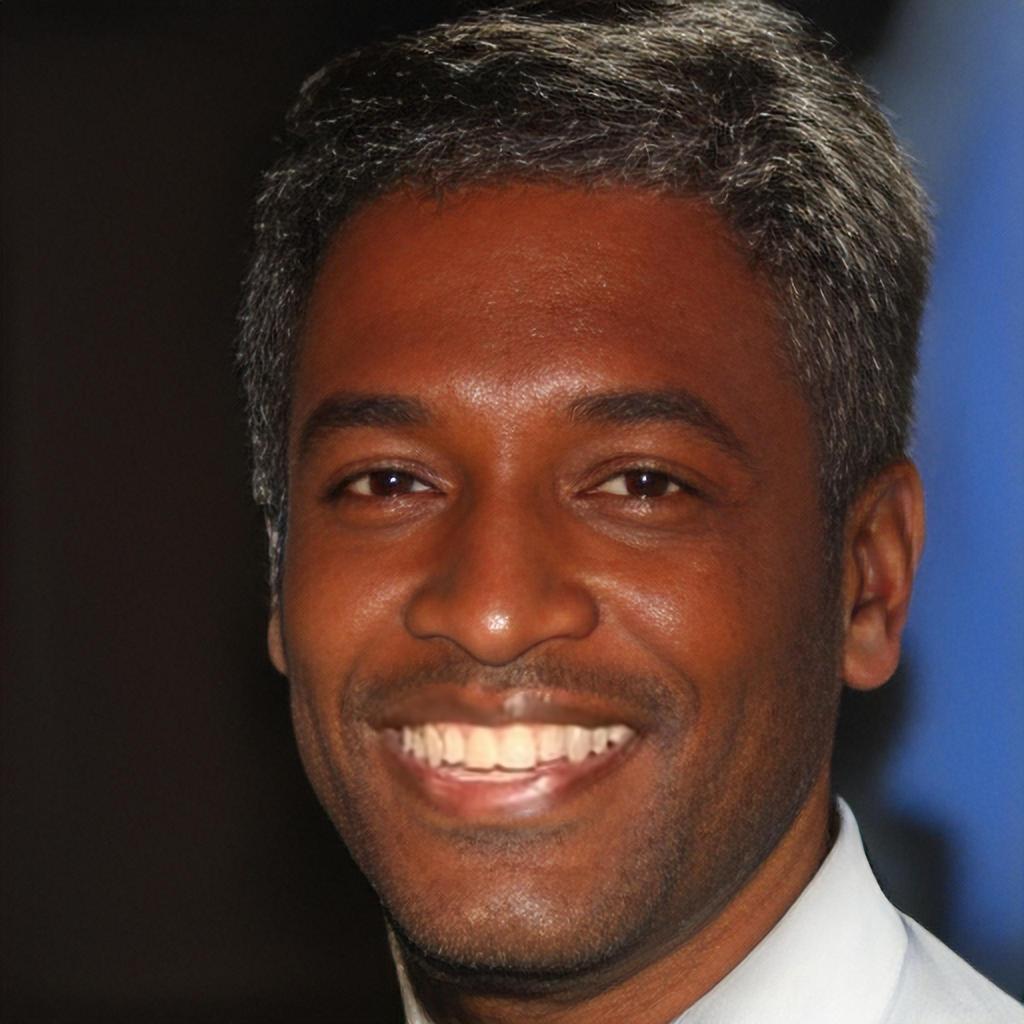} \end{tabular} \hspace{-0.9cm}
&
\begin{tabular}{c} \includegraphics[width=0.245\linewidth]{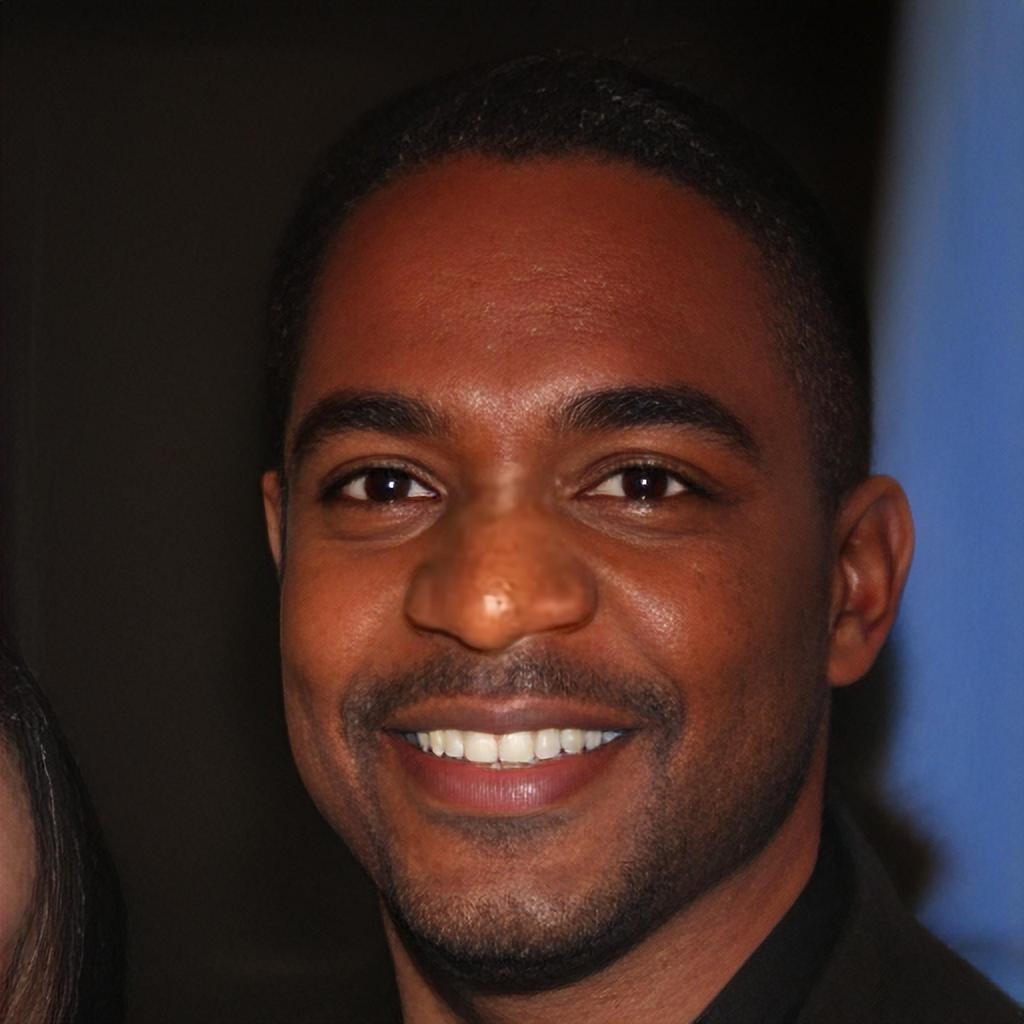} \end{tabular} \hspace{-0.9cm}
&
\begin{tabular}{c} \includegraphics[width=0.245\linewidth]{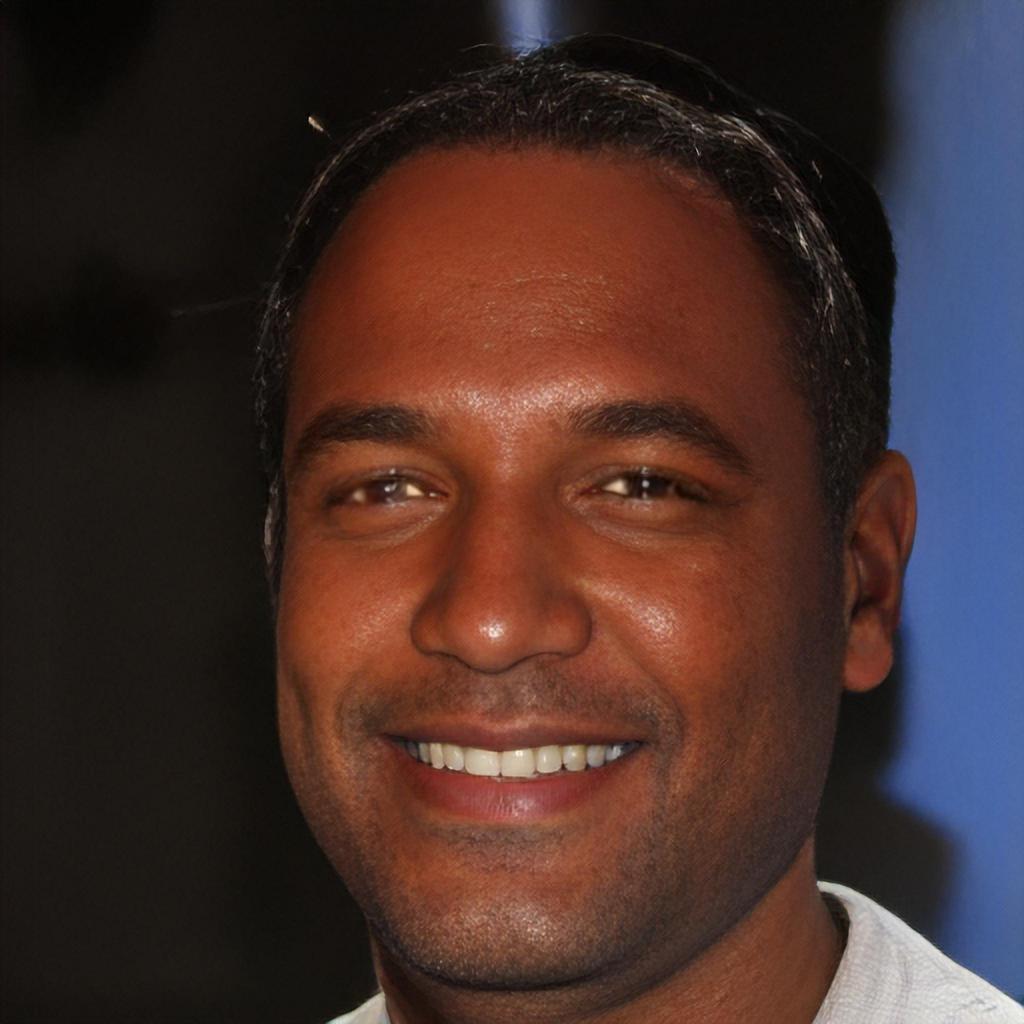} \end{tabular} \hspace{-0.9cm}
\\
\hspace{-0.9cm}
\begin{tabular}{c} \makebox[0.245\linewidth]{Input} \end{tabular} \hspace{-0.9cm}
&
\begin{tabular}{c} \makebox[0.245\linewidth]{Ours (\textbf{mouth})} \end{tabular} \hspace{-0.9cm}
&
\begin{tabular}{c} \makebox[0.245\linewidth]{Ours (\textbf{nose})} \end{tabular} \hspace{-0.9cm}
&
\begin{tabular}{c} \makebox[0.245\linewidth]{Ours (\textbf{eyes})} \end{tabular} \hspace{-0.9cm}
\\
\hspace{-0.9cm}
\begin{tabular}{c} \includegraphics[width=0.245\linewidth]{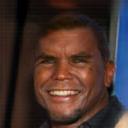} \end{tabular} \hspace{-0.9cm}
&
\begin{tabular}{c} \includegraphics[width=0.245\linewidth]{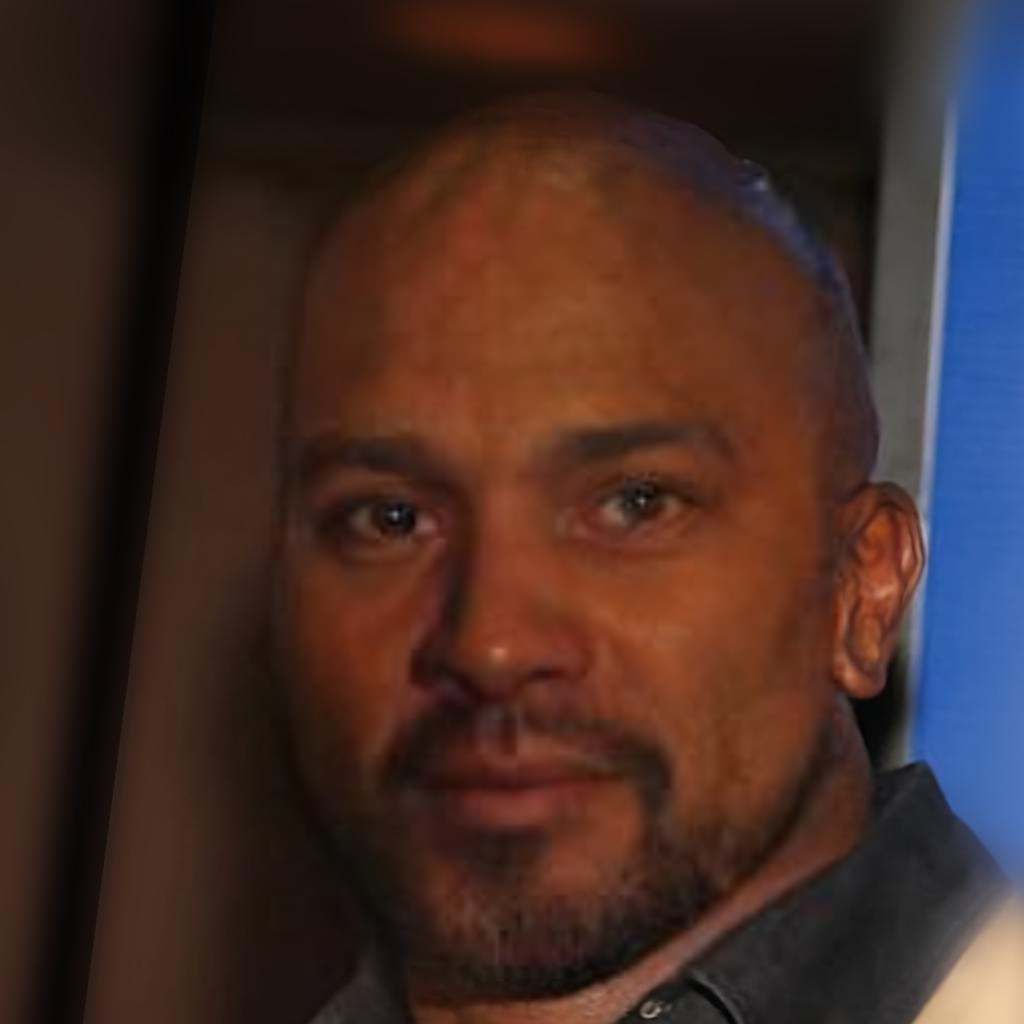} \end{tabular} \hspace{-0.9cm}
&
\begin{tabular}{c} \includegraphics[width=0.245\linewidth]{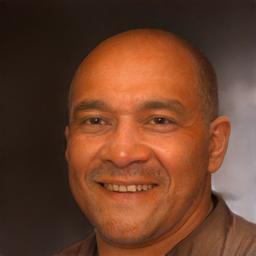} \end{tabular} \hspace{-0.9cm}
&
\begin{tabular}{c} \includegraphics[width=0.245\linewidth]{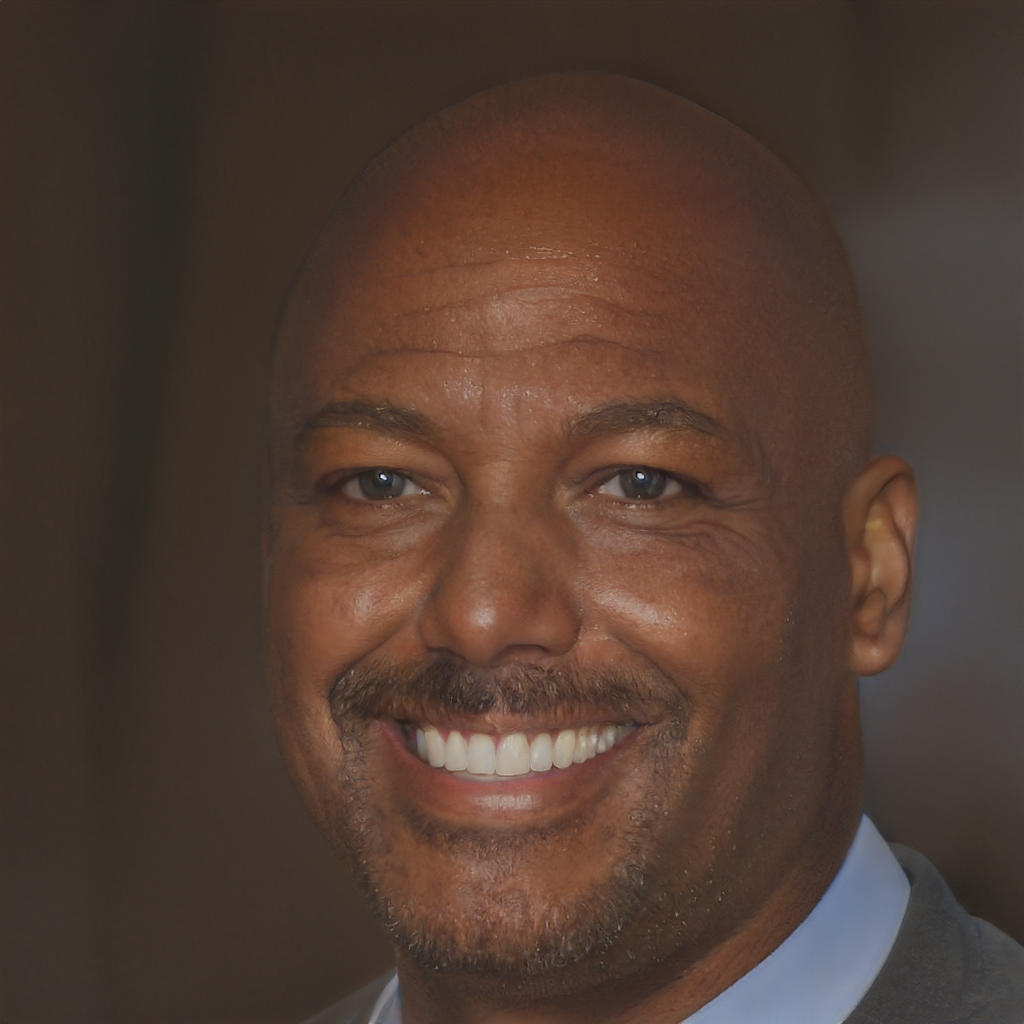} \end{tabular} \hspace{-0.9cm}
\\
\hspace{-0.9cm}
\begin{tabular}{c} \makebox[0.245\linewidth]{FIT~\cite{gu2020password}} \end{tabular} \hspace{-0.9cm}
&
\begin{tabular}{c} \makebox[0.245\linewidth]{DP2~\cite{hukkelaas2023deepprivacy2}} \end{tabular} \hspace{-0.9cm}
&
\begin{tabular}{c} \makebox[0.245\linewidth]{RiDDLE~\cite{li2023riddle}} \end{tabular} \hspace{-0.9cm}
&
\begin{tabular}{c} \makebox[0.245\linewidth]{FALCO~\cite{barattin2023attribute}} \end{tabular} \hspace{-0.9cm}
\\
\end{tabular}
\caption{Our \textit{clinical single-image} anonymization results preserving respectively the mouth, nose, or eyes, compared with state-of-the-art anonymization methods.}
\label{fig:single_semantic_compact}
\end{figure}

\subsection{Image anonymization} \label{sec:method_anonymization}

\subsubsection{Clinical anonymization}

For \textbf{single} images showing a medical intervention, we address clinical anonymization, where specific semantic components are preserved. 
We exploit the semantic disentanglement in the generator to anonymize as follows. We invert the input into our extended latent space. We randomize all semantic components except the one that must be preserved, and synthesize the anonymized image, thus preserving the desired semantic component. For clinical images, preserve pixel-level details is important, therefore, we fuse back the semantic area to be preserved from the original image. The blending works smoothly, as we also keep high-level latents unchanged, so synthesized images are consistent with original images on attributes such as pose.
We also perform a final refinement with a face-prior-based blending and correction, for more robustness (Sec.~\ref{sec:prior_processing}).
To anonymize \textbf{pairs} of before-and-after clinical images, we follow a similar strategy but using the paired-image inversion presented in Sec.~\ref{sec:method_inversion}, and apply joint semantic modification in the latent space. Furthermore, we exploit our high-level disentanglement to restrict the synthetic identity and preserve it across anonymized images. That way, we anonymize paired images of a person, such as before-and-after images of medical operations, with a consistent synthetic identity.

\begingroup
\setlength{\tabcolsep}{0pt} 
\renewcommand{\arraystretch}{1} 
\begin{figure}[t!]
\centering
\begin{tabular}{lcccc}
\multirow{2}{*}{\rotatebox[origin=c]{90}{\makebox[25mm]{CelebAHQ}}} \hspace{-0.0cm} &
\begin{tabular}{c} \includegraphics[width=0.22\linewidth]{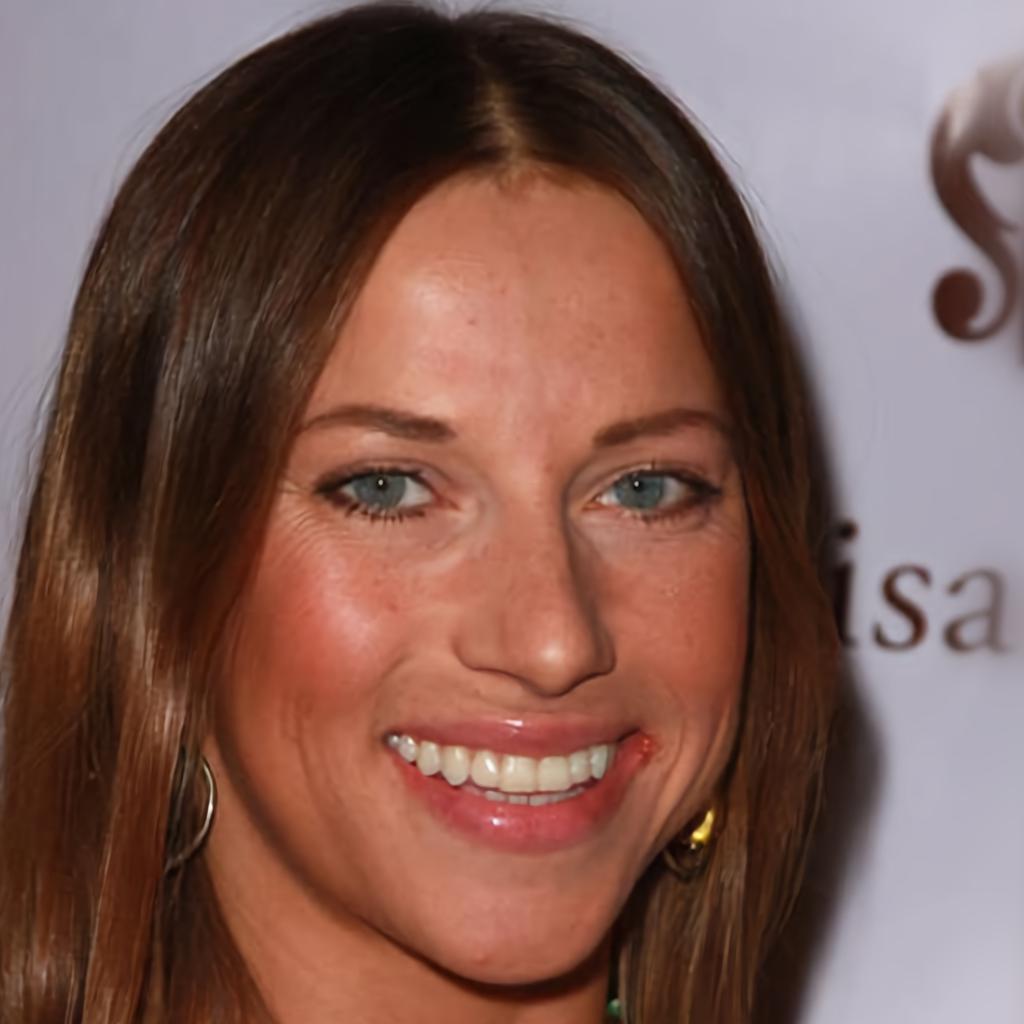} \end{tabular} \hspace{-0.0cm}
&
\begin{tabular}{c} \includegraphics[width=0.22\linewidth]{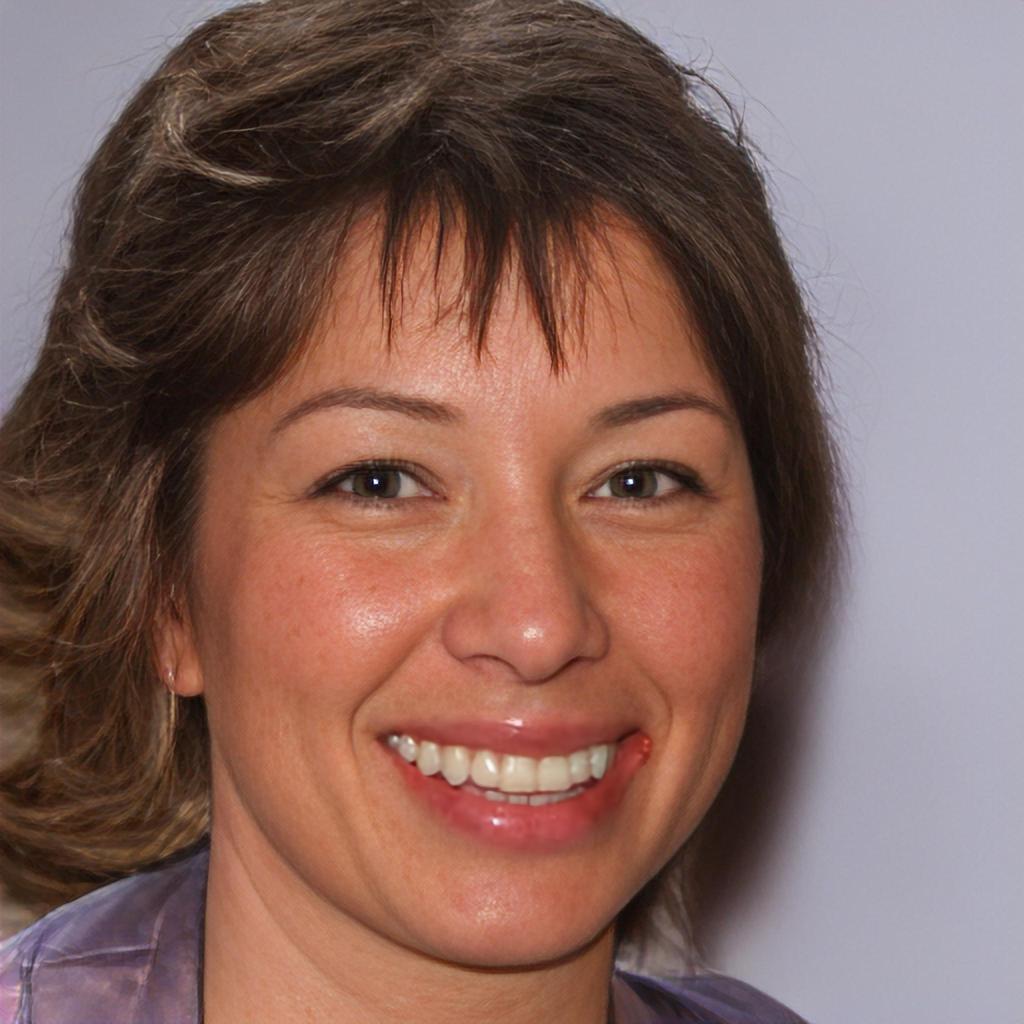} \end{tabular} \hspace{-0.0cm}
&
\begin{tabular}{c} \includegraphics[width=0.22\linewidth]{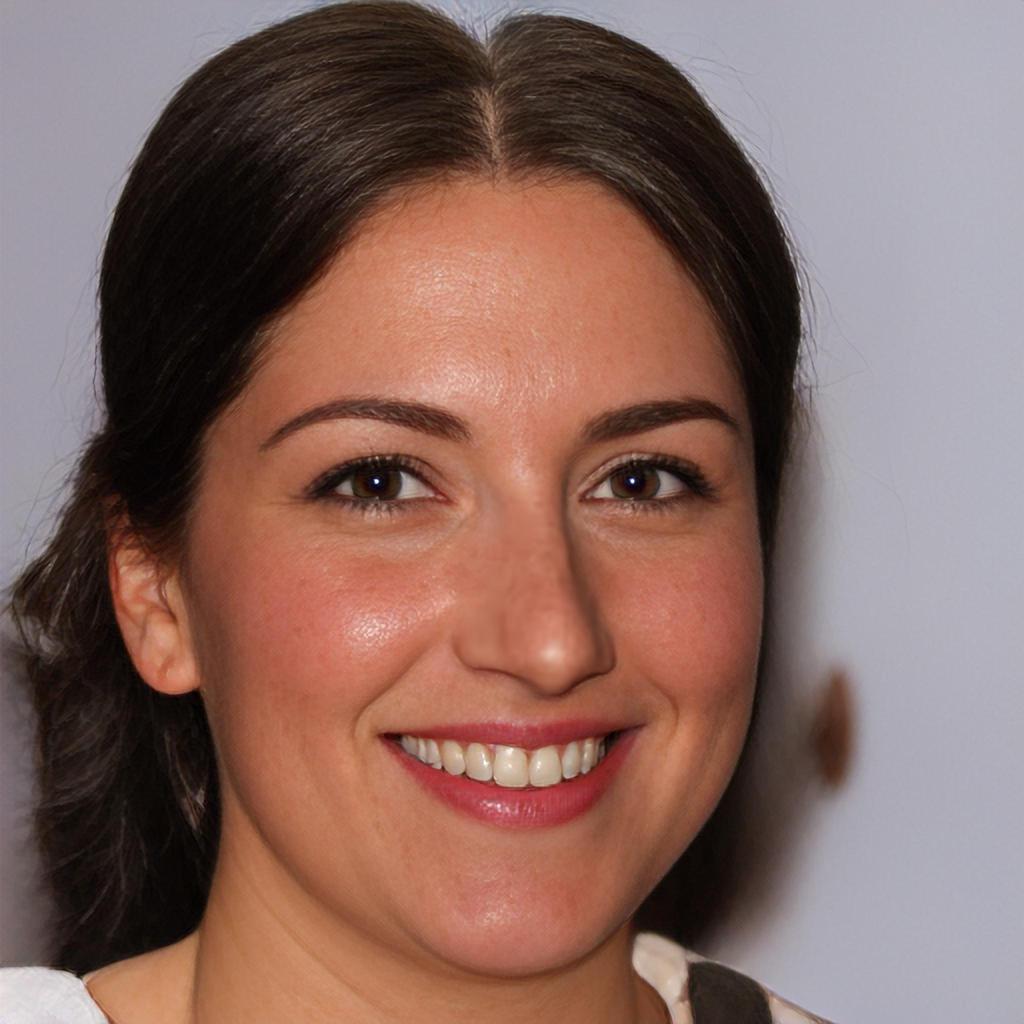} \end{tabular} \hspace{-0.0cm}
&
\begin{tabular}{c} \includegraphics[width=0.22\linewidth]{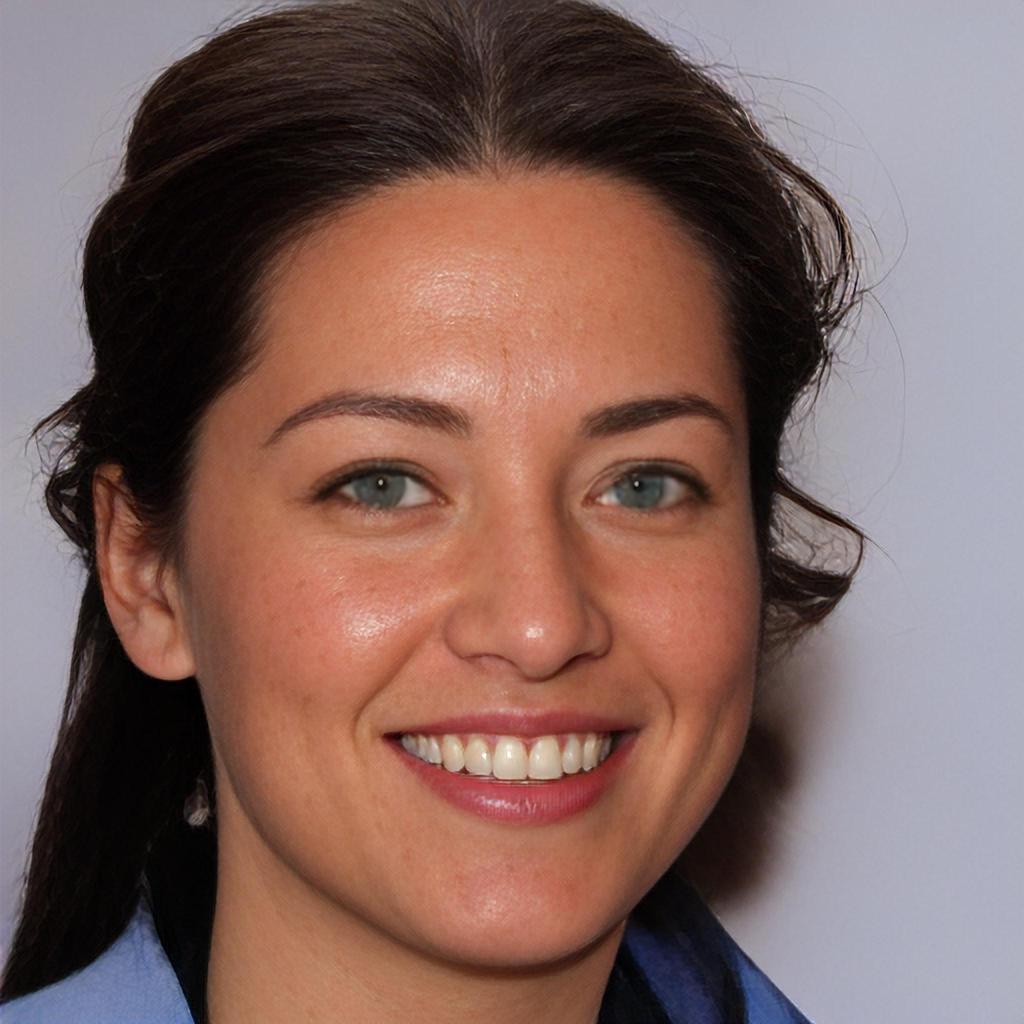} \end{tabular} \hspace{-0.0cm}
\\
\hspace{-0.0cm} &
\begin{tabular}{c} \includegraphics[width=0.22\linewidth]{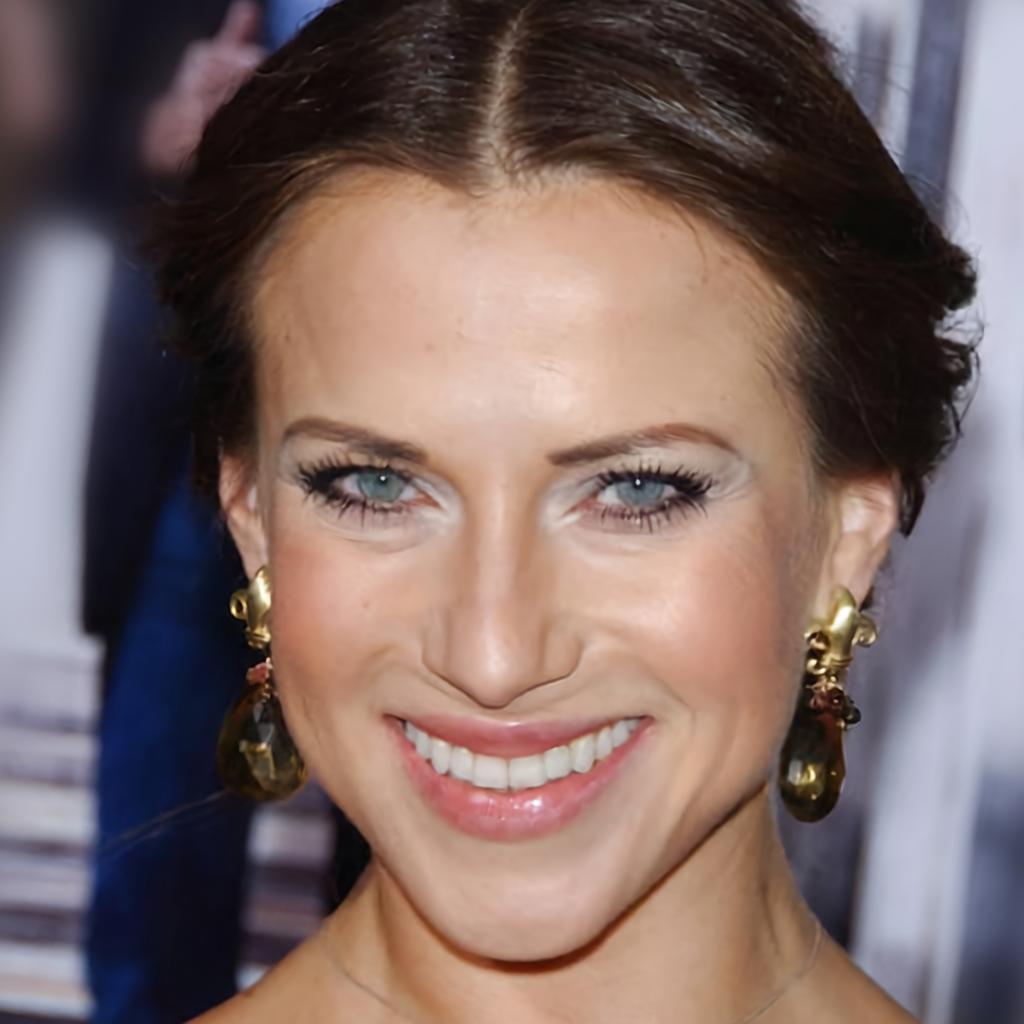} \end{tabular} \hspace{-0.0cm}
&
\begin{tabular}{c} \includegraphics[width=0.22\linewidth]{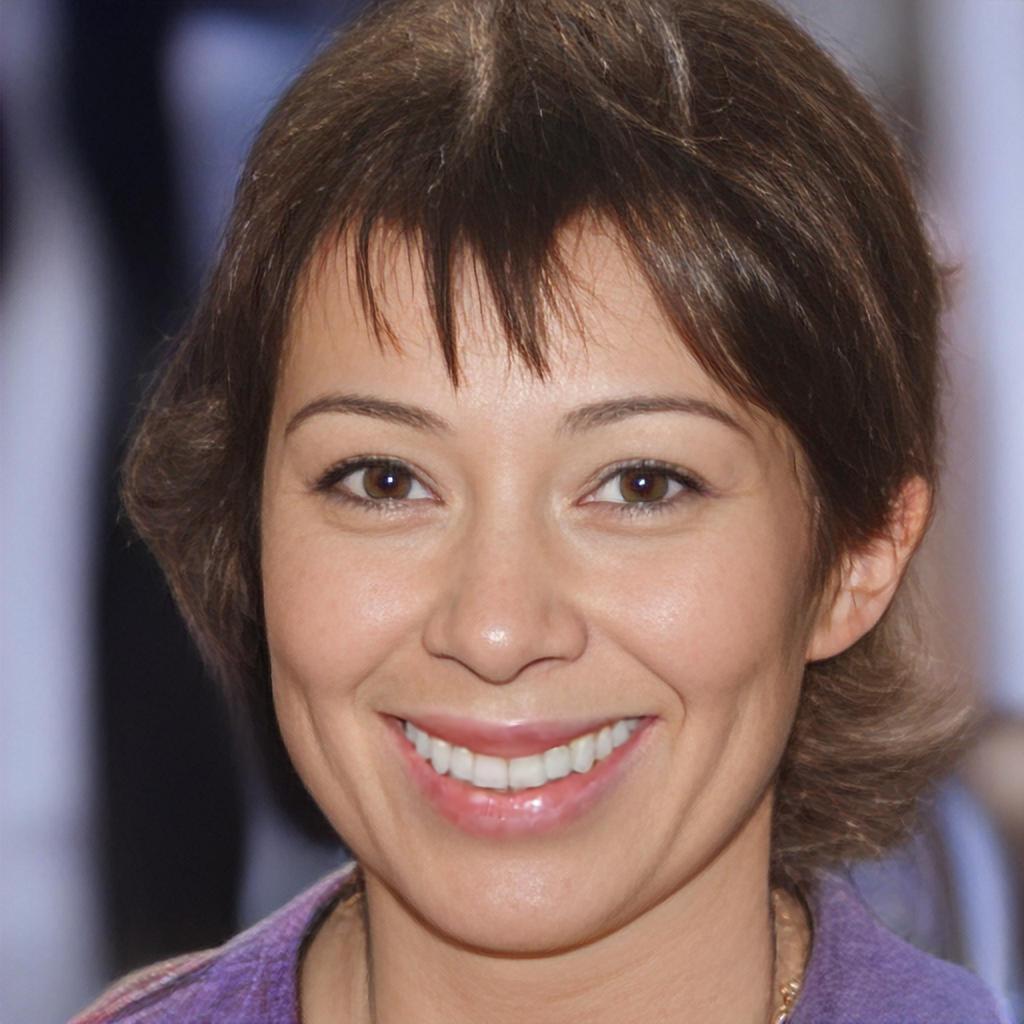} \end{tabular} \hspace{-0.0cm}
&
\begin{tabular}{c} \includegraphics[width=0.22\linewidth]{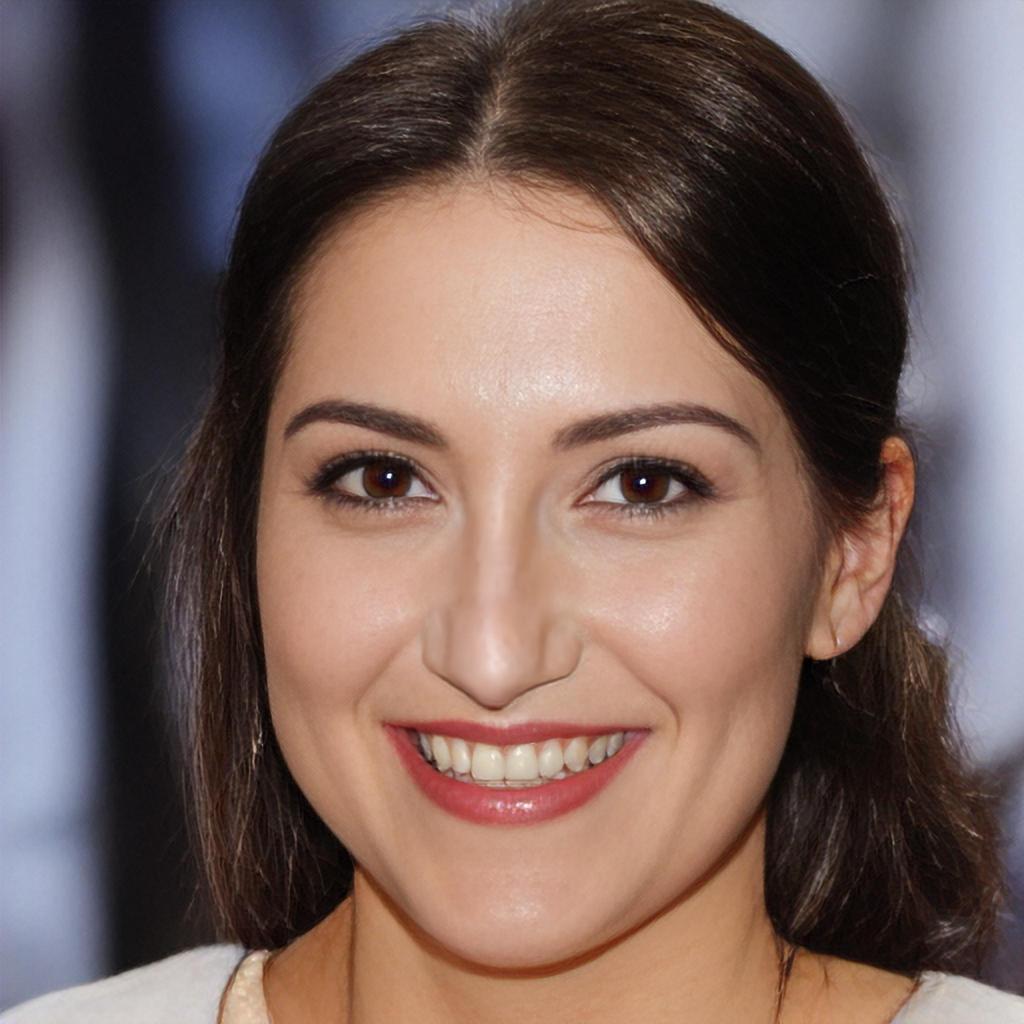} \end{tabular} \hspace{-0.0cm}
&
\begin{tabular}{c} \includegraphics[width=0.22\linewidth]{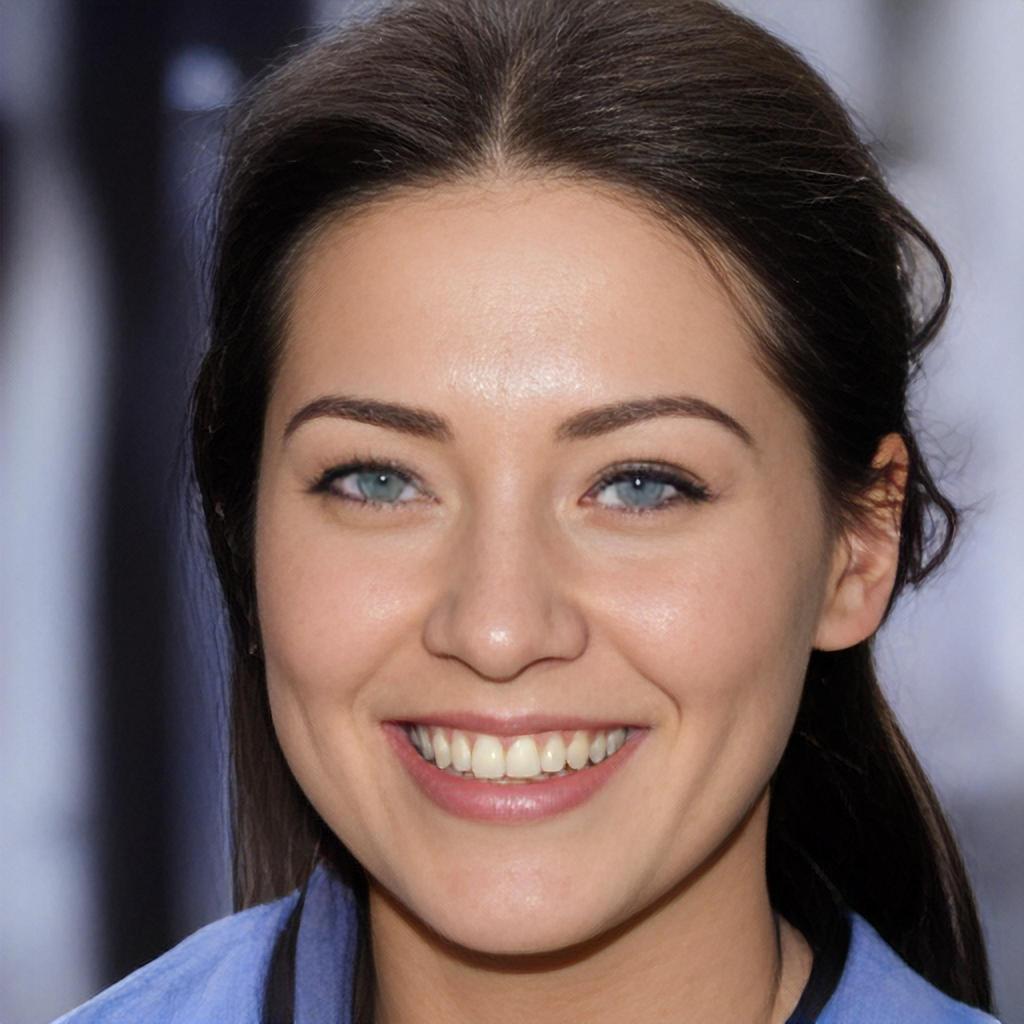} \end{tabular} \hspace{-0.0cm}
\\
\multirow{2}{*}{\rotatebox[origin=c]{90}{\makebox[20mm]{SiblingsDB}}} \hspace{-0.0cm} &
\begin{tabular}{c} \includegraphics[width=0.22\linewidth]{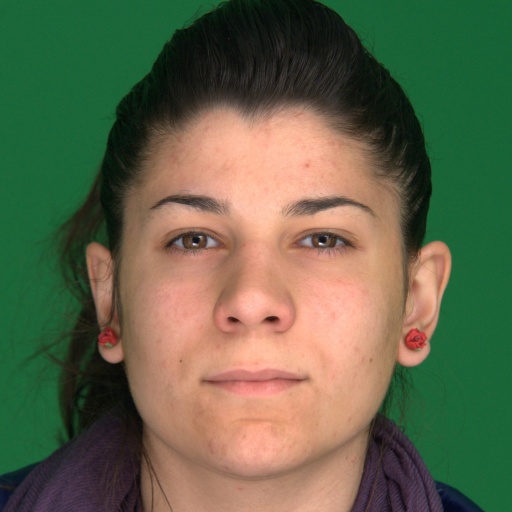} \end{tabular} \hspace{-0.0cm}
&
\begin{tabular}{c} \includegraphics[width=0.22\linewidth]{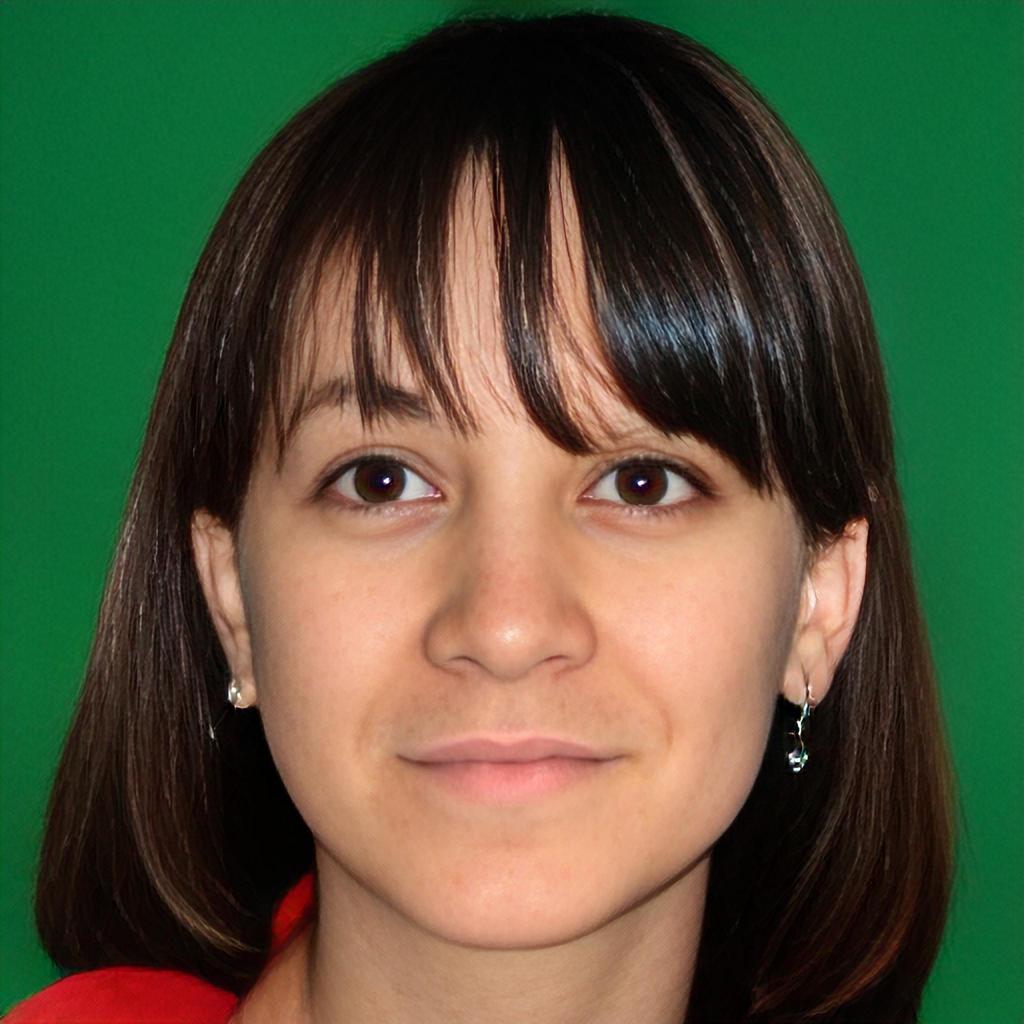} \end{tabular} \hspace{-0.0cm}
&
\begin{tabular}{c} \includegraphics[width=0.22\linewidth]{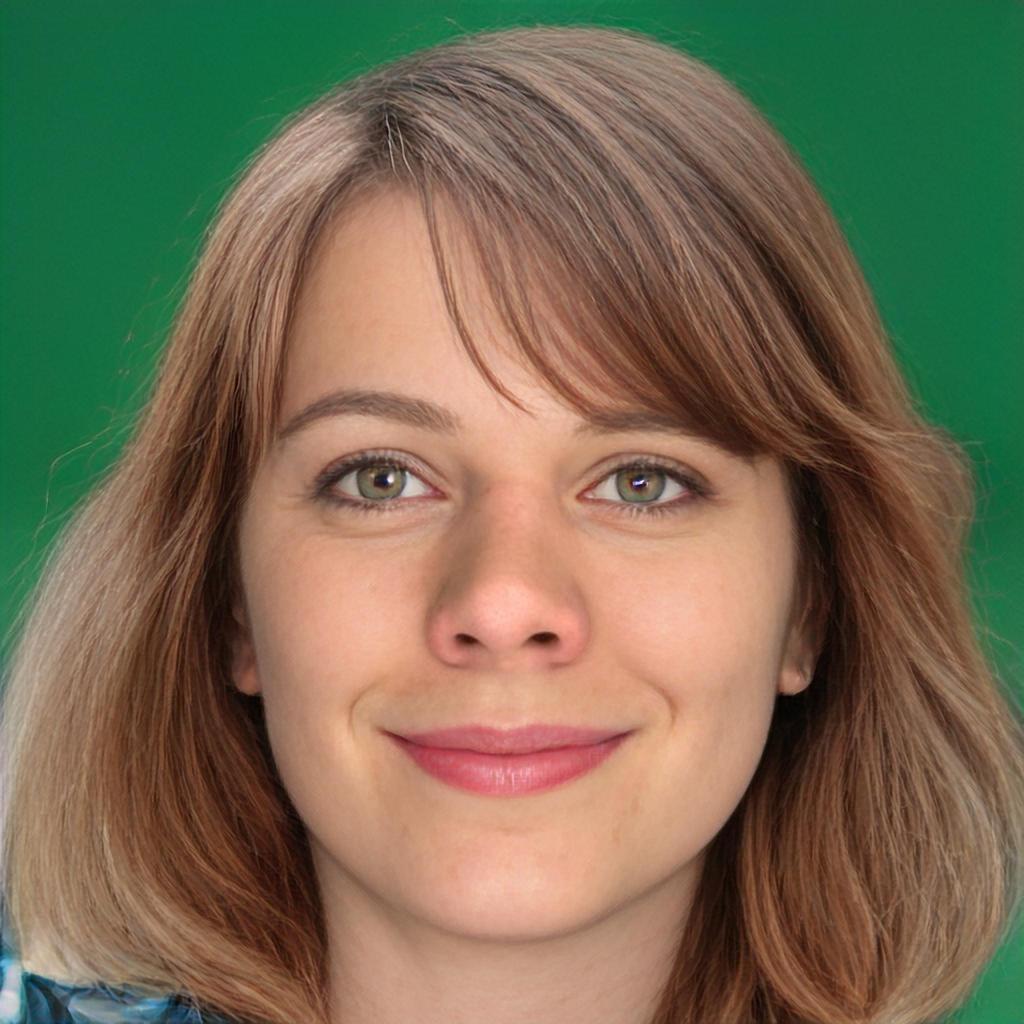} \end{tabular} \hspace{-0.0cm}
&
\begin{tabular}{c} \includegraphics[width=0.22\linewidth]{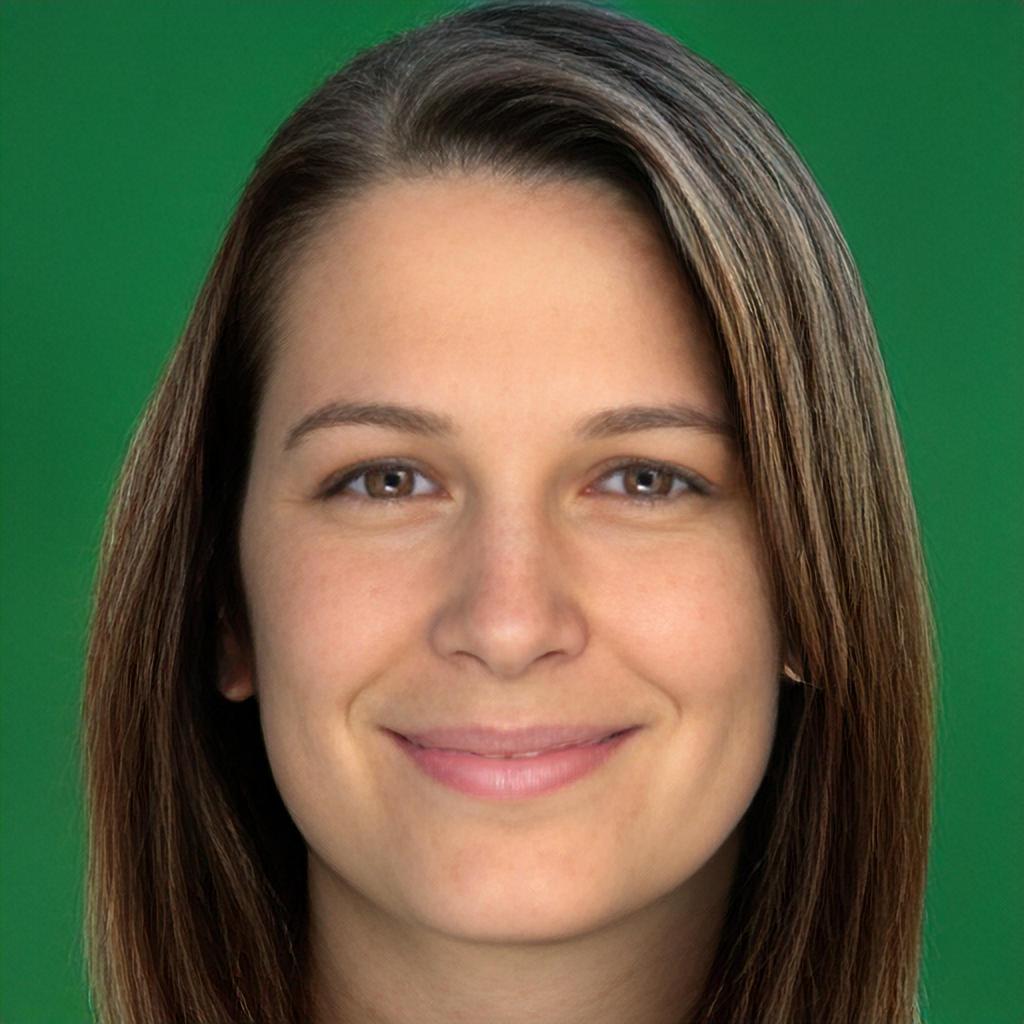} \end{tabular} \hspace{-0.0cm}
\\
\hspace{-0.0cm} &
\begin{tabular}{c} \includegraphics[width=0.22\linewidth]{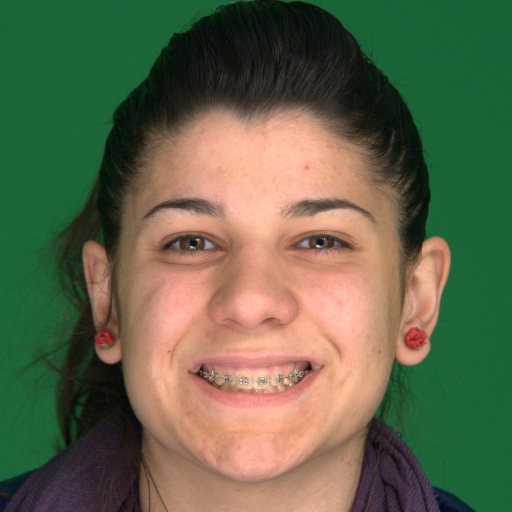} \end{tabular} \hspace{-0.0cm}
&
\begin{tabular}{c} \includegraphics[width=0.22\linewidth]{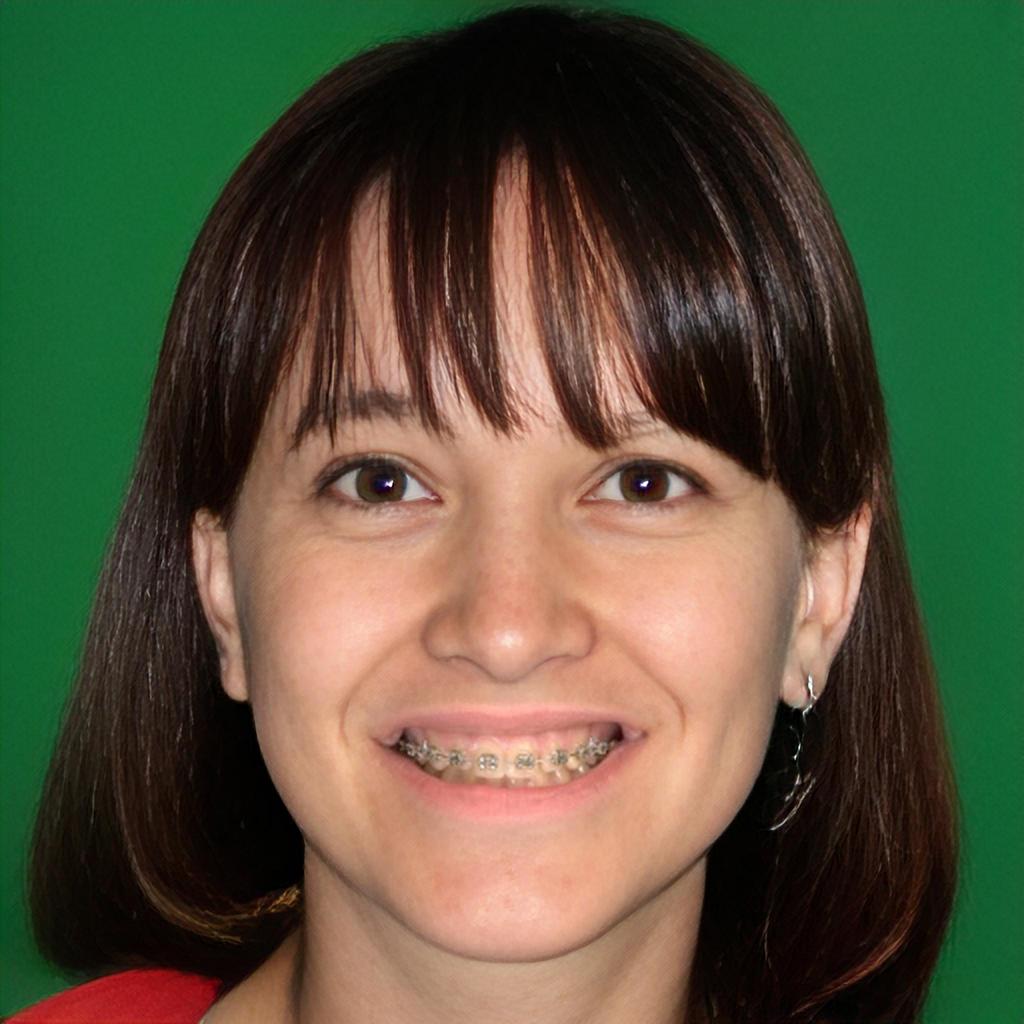} \end{tabular} \hspace{-0.0cm}
&
\begin{tabular}{c} \includegraphics[width=0.22\linewidth]{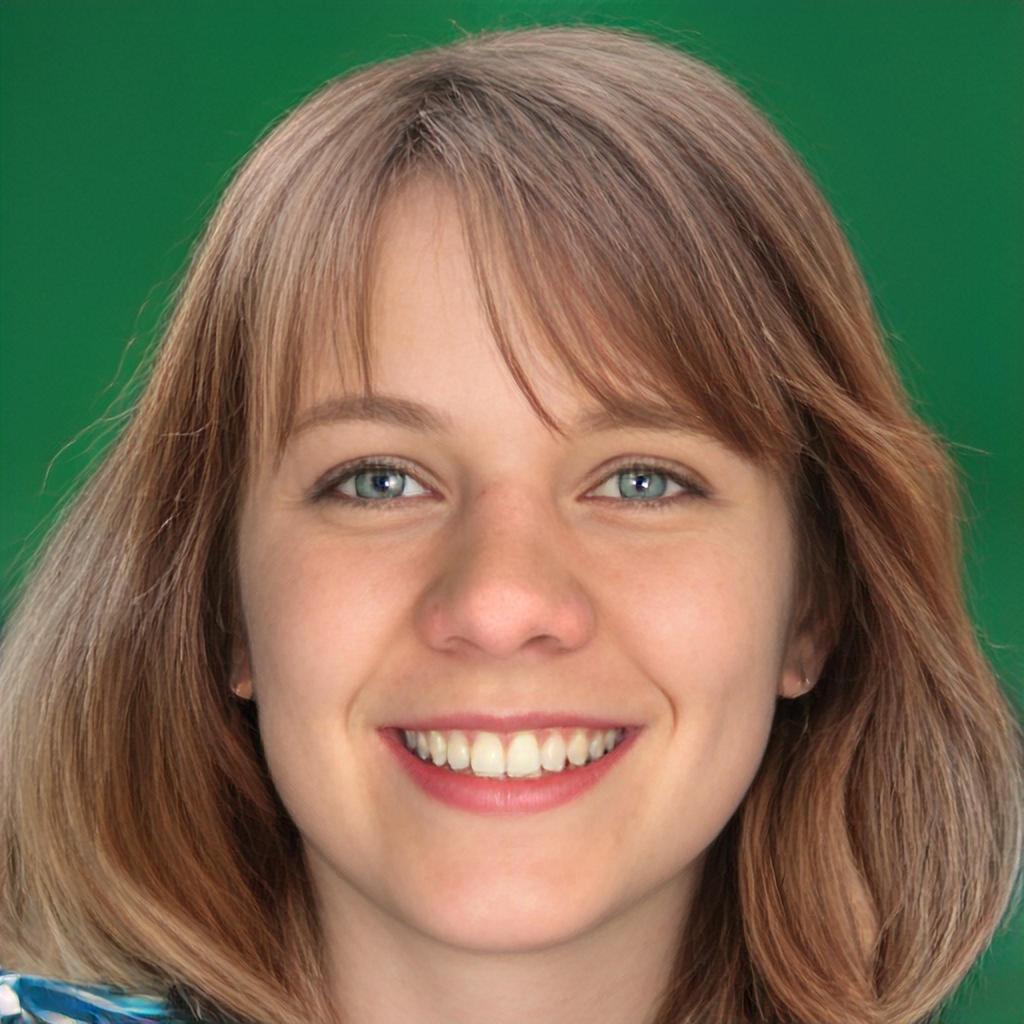} \end{tabular} \hspace{-0.0cm}
&
\begin{tabular}{c} \includegraphics[width=0.22\linewidth]{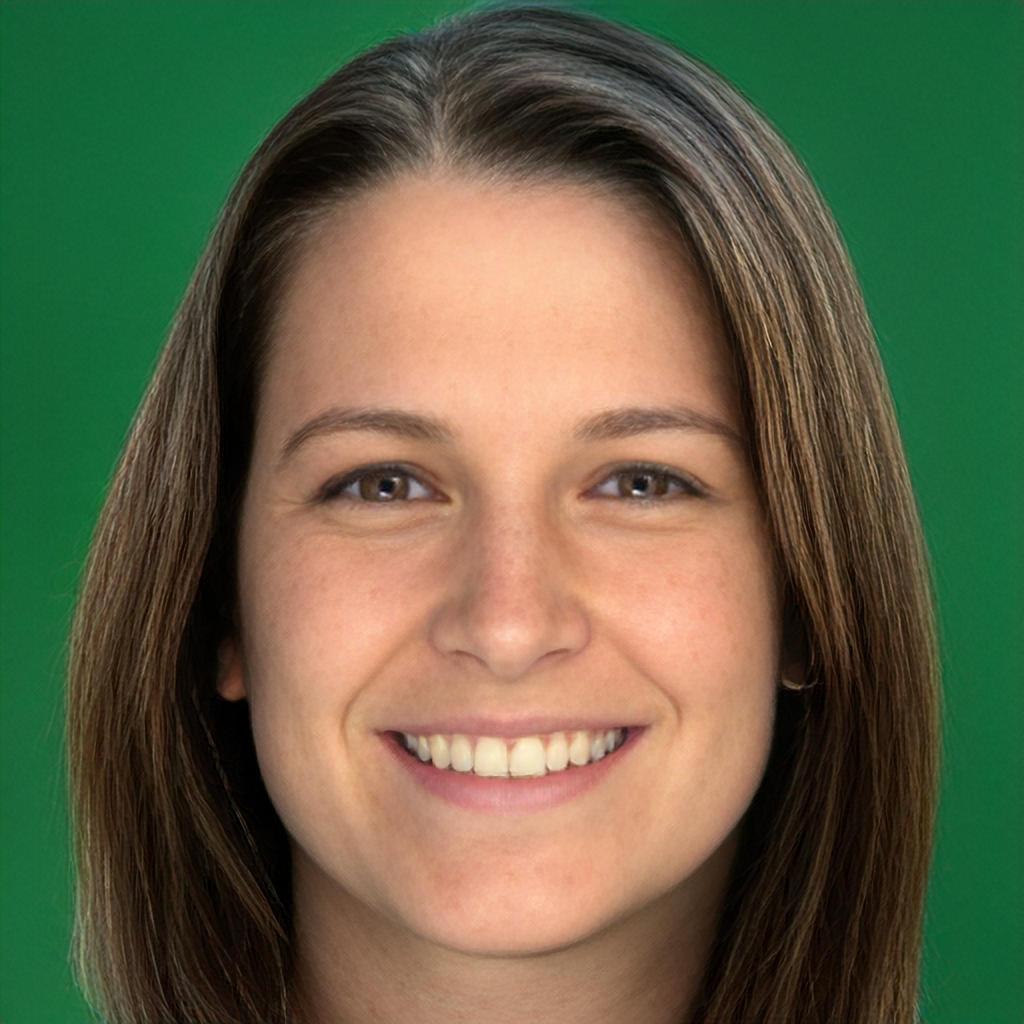} \end{tabular} \hspace{-0.0cm}
\\
\rotatebox[origin=c]{90}{\makebox[1mm]{ }} \hspace{-0.0cm} &
\begin{tabular}{c} \makebox[0.22\linewidth]{Input \textbf{pair}} \end{tabular} \hspace{-0.0cm}
&
\begin{tabular}{c} \makebox[0.22\linewidth]{Ours (\textbf{mouth})} \end{tabular} \hspace{-0.0cm}
&
\begin{tabular}{c} \makebox[0.22\linewidth]{Ours (\textbf{nose})} \end{tabular} \hspace{-0.0cm}
&
\begin{tabular}{c} \makebox[0.22\linewidth]{Ours (\textbf{eyes})} \end{tabular} \hspace{-0.0cm}
\\
\end{tabular}
\caption{Our \textit{clinical paired-image} anonymization preserving from left to right the mouth, nose, or eyes of the input person. Top: on a pair example from CelebAHQ, bottom: on a pair example taken from the standardized-capture SiblingsDB~\cite{vieira2014detecting}.}
\label{fig:paired_semantic_all}
\end{figure}
\endgroup

\subsubsection{Standard anonymization}
For standard \textbf{single} images, it is common practice to perform in-place anonymization~\cite{maximov2020ciagan,gu2020password,hukkelaas2023deepprivacy2,wen2023divide}, which enables scene preservation. In this setting, only the inside part of the face, delimited by the face contour, is anonymized. Remaining parts such as hair/neck are preserved. In-place anonymization can be performed as a special case of VerA, where the semantic area to preserve is the union of all semantic components excluding the inside of the face, i.e. hair, neck, background, clothes, ears, earrings, and hats. We keep the high-levels unchanged, as we do in clinical anonymization setting, to blend preserved components more smoothly.
We anonymize \textbf{pairs} of standard images of the same person following a similar setup as paired-image clinical anonymization, while preserving the set of semantic components mentioned above.

\begin{table*}[t]
    \centering
    \resizebox{\textwidth}{!}{%
    \begin{NiceTabular}{p{1em}l|c|c|c||c|c|c||c|c|c||c|c|c}
    \toprule
    & & \multicolumn{3}{c}{$\ell_1$ distance $\downarrow$} & \multicolumn{3}{c}{PSNR $\uparrow$} & \multicolumn{3}{c}{Semantic IoU $\uparrow$} & \multicolumn{3}{c}{Landmark offset $\downarrow$} \\
    & Method & Mouth & Nose & Eyes & Mouth & Nose & Eyes & Mouth & Nose & Eyes & Mouth & Nose & Eyes \\
    \midrule
    \multirow{6}{*}{\rotatebox[origin=c]{90}{Standard}} & CIAGAN~\cite{maximov2020ciagan}      & 34.25 & 25.97 & 52.74 & 15.08 & 17.36 & 11.65 & 0.55 & 0.58 & 0.02 & 16.84 & 17.17 & 40.97 \\
    & FIT~\cite{gu2020password}            & \underline{20.88} & \underline{15.38} & \underline{29.48} & \underline{19.26} & \underline{21.65} & \underline{16.23} & \underline{0.78} & \underline{0.84} & \underline{0.64} & \underline{8.40}  & \underline{8.07}  & \underline{7.97}  \\
    & DP2~\cite{hukkelaas2023deepprivacy2} & 41.03 & 30.96 & 53.73 & 13.47 & 15.55 & 11.38 & 0.51 & 0.60 & 0.27 & 31.37 & 33.57 & 27.75 \\
    & RiDDLE~\cite{li2023riddle}           & 33.26 & 28.54 & 41.56 & 15.30 & 16.55 & 13.57 & 0.71 & 0.76 & 0.59 & 13.32 & 19.08 & 9.97  \\
    & FALCO~\cite{barattin2023attribute}   & 32.71 & 26.61 & 40.57 & 15.52 & 17.47 & 13.69 & 0.67 & 0.79 & 0.55 & 15.51 & 15.27 & 10.60 \\
    & Ours                                 & 34.20 & 18.82 & 41.42 & 14.70 & 19.25 & 13.20 & 0.66 & 0.80 & 0.60 & 16.25 & 14.24 & 9.49  \\
    \hdashline
    \multirow{3}{*}{\rotatebox[origin=c]{90}{Clinical}} & Ours (mouth)                         & \textbf{0.22} & 18.86 & 43.21 & \textbf{51.06} & 19.30 & 12.94 & \textbf{0.92} & 0.81 & 0.57 & \textbf{6.81}  & 14.17 & 10.18 \\
    & Ours (nose)                          & 33.94 & \textbf{0.17} & 42.51 & 14.76 & \textbf{51.33} & 13.04 & 0.67 & \textbf{0.93} & 0.59 & 16.38 & \textbf{5.67}  & 9.63  \\
    & Ours (eyes)                          & 34.58 & 18.90 & \textbf{0.48} & 14.64 & 19.28 & \textbf{40.82} & 0.65 & 0.80 & \textbf{0.83} & 17.61 & 14.70 & \textbf{6.53}  \\
    \bottomrule
    \end{NiceTabular}
    }
    \caption{Semantic preservation results, in terms of content ($\ell_1$, PSNR) and area (IoU, landmarks). We note two main observations: standard anonymization approaches destroy all semantic components that may need to be preserved in clinical images, and our clinical anonymization successfully preserves the desired component while also flexibly modifying non-blocked components as much as the baselines.}
    \label{table:semantic_preservation}
\end{table*}

\subsection{Prior-based processing} \label{sec:prior_processing}
\noindent \textbf{Prior-based blending.} We exploit a learned face prior to improve the quality of our semantic blending. 
To that end, we use a pretrained inpainting network, trained on facial data. Specifically, we use a mask-aware transformer (MAT)~\cite{li2022mat} trained on FFHQ~\cite{karras2019style}. To delimit the blending area, we define our mask $m_{inp}$
\begin{equation}
    m_{inp} \coloneqq
    \mathds{1} \left ( ( m_{sem}^{real} \cup m_{sem}^{syn} ) \circledast k_{\mathcal{N}} > \eta \right )
    - m_{sem}^{real},
\end{equation}
where $m_{sem}^{real}$ is the binary mask indicating the semantic area of interest that we preserve from the real image, $m_{sem}^{syn}$ is the mask of the same semantic area in our synthetic image, which could slightly deviate from the original, $k_{\mathcal{N}}$ is a Gaussian blur kernel and $\eta$ is a fixed threshold determining final distances. With this, we guarantee the preservation of the original semantic area, and a smooth blending based on a robust face prior for the small surrounding area. 

\noindent \textbf{Prior-based correction.} We ward off potential residual artifacts with a final correction step. We exploit the face prior learned by a CodeFormer~\cite{zhou2022towards} model and pass the final anonymized images through it with frozen weights.

\noindent We provide ablation over both steps in our supplementary.

\section{Experimental Evaluation} \label{sec:eval}
\subsection{Experimental setup} \label{sec:exp_setup}

\noindent \textbf{Benchmarks.} We compare quantitatively and qualitatively with the two most commonly referenced baselines: CIAGAN~\cite{maximov2020ciagan} and FIT~\cite{gu2020password}, and with the state-of-the-art methods published recently for which public code is available: DP2~\cite{hukkelaas2023deepprivacy2}, RiDDLE~\cite{li2023riddle}, and FALCO~\cite{barattin2023attribute}. FALCO performs an adaptive normalization that can lead to washed out images, or to odd color artifacts if toggled off, we follow the authors' setting and leave it activated. 

\noindent \textbf{Settings and datasets.}
We use the same model in all experiments, which we train on data from CelebAMaskHQ~\cite{lee2020maskgan} for 600k iterations and continue with FFHQ training~\cite{karras2019style} for 400k iterations. For FFHQ, we use weak semantic labels predicted by a pretrained SegFormer~\cite{xie2021segformer}.
We use 2'000 held-out images from CelebAMaskHQ for evaluation.
On that set, we form 500 pairs using metadata such that they share the same identity, carefully excluding trivial duplicates or copies with only slightly different cropping. 
We show a paired example result from SiblingsDB~\cite{vieira2014detecting} for qualitative evaluation, where the image capture is standardized. The pairs from CelebAMaskHQ are more challenging than such a setting, or than any pairs that would be created out of a video sequence (due to temporal coherence). 

\subsection{Controllable generator} \label{sec:eval_generator}
We illustrate our generator's joint control over semantic and high-level attributes in Fig.~\ref{fig:control_joint}; we modify and accumulate changes across both semantic and high-level attributes. We provide extended evaluation in our supplementary, showing the preserved semantic control 
and the gained high-level control with our training and modified latents.

\subsection{Specialized inversion} \label{sec:eval_inversion}
With the inversion procedure of Sec.~\ref{sec:method_inversion}, we use the following loss weights: $\ell 1$ (1.0), $\ell 2$ (0.1), LPIPS (2.0), latent mean regularization (1.0), and segmentation loss (1.0), and Adam~\cite{kingma2014adam} with a 0.1 learning rate for 300 steps. For real images with no semantic masks, we obtain labels with a pretrained SegFormer~\cite{xie2021segformer} and map them to the 13-label subset~\cite{li2021semantic}. Fig.~\ref{fig:inversion_segmentation} shows inversion results with and without our semantic loss. We observe that this loss guides the inversion optimization, particularly for hats, earrings and clothes, and improves our results for clinical anonymization.

\begin{table}[t]
    \centering
    \resizebox{\columnwidth}{!}{%
    \begin{NiceTabular}{p{1em}l|c|c|c|c|c|c}
    \toprule
    & Face Recog. $\uparrow$ & \multicolumn{2}{c}{CASIA~\cite{yi2014learning}} & \multicolumn{2}{c}{VGGFace2~\cite{cao2018vggface2}} & \multicolumn{2}{c}{FaceNet~\cite{schroff2015facenet}} \\
    & Setting                              & Single & Paired & Single & Paired & Single & Paired \\
    \midrule
    \multirow{6}{*}{\rotatebox[origin=c]{90}{Standard}} 
    & CIAGAN~\cite{maximov2020ciagan}      & 98.6 & 97.8 & 98.4 & 97.4 & \textbf{99.9} & \textbf{100.0} \\
    & FIT~\cite{gu2020password}            & 99.1 & 98.2 & 98.3 & 98.8 & 99.1 & \underline{99.8} \\
    & DP2~\cite{hukkelaas2023deepprivacy2} & 99.1 & 99.2 & 98.5 & 98.8 & 83.2 & 85.0 \\
    & RiDDLE~\cite{li2023riddle}           & \textbf{99.9} & \textbf{100.0} & \textbf{99.7} & \textbf{99.6} & 96.7 & 96.8 \\
    & FALCO~\cite{barattin2023attribute}   & 98.7 & 98.6 & 98.7 & \textbf{99.6} & 90.6 & 90.4 \\
    & Ours                                 & 98.6 & 98.4 & 95.8 & 96.0 & 97.9 & 98.2 \\
    \hdashline
    \multirow{3}{*}{\rotatebox[origin=c]{90}{Clinical}} 
    & Ours (mouth)                         & \underline{99.8} & \underline{99.8} & \underline{98.9} & 99.0 & 98.2 & 98.0 \\
    & Ours (nose)                          & \underline{99.8} & 99.4 & 96.1 & 95.6 & 98.9 & 99.6 \\
    & Ours (eyes)                          & \underline{99.8} & \underline{99.8} & 98.8 & \underline{99.2} & \underline{99.6} & \textbf{100.0} \\
    \bottomrule
    \end{NiceTabular}
    }
    \caption{Benchmarking results on \textit{de-identification} rates, using multiple face recognition methods, for single and paired standard images, as well as their clinical counterparts. We achieve good de-identification comparable with the state-of-the-art methods.}
    \label{table:deID_comparison}
\end{table}

\subsection{Clinical image anonymization} \label{sec:eval_anon_clinical}

We evaluate VerA on clinical image anonymization, in the single-image and paired-image cases. For single images, illustrated in Fig.~\ref{fig:single_semantic_compact}, the objective is to preserve the semantic region with clinical intervention (such as mouth, nose, or eyes) during anonymization. While available anonymization methods typically destroy this information, VerA can preserve desired regions. We quantitatively evaluate our semantic preservation in Table~\ref{table:semantic_preservation} with $\ell_1$, PSNR, semantic mask IoU, and average landmark offset for the region of interest. VerA preserves well both the content and the area of the target component, while other methods perform poorly (even with our prior-based blending on their results, cf. supplementary). Furthermore, when preserving areas like the mouth, we still flexibly modify the nose and the eyes. We also provide de-identification evaluation using CASIA~\cite{yi2014learning}, VGGFace2~\cite{cao2018vggface2}, and FaceNet~\cite{schroff2015facenet} in Table~\ref{table:deID_comparison}, showing that VerA is competitive with state-of-the-art solutions. 
For paired images, such as before-and-after pairs, the objective is to anonymize them while keeping a consistent synthetic identity and  preserving the desired component. We illustrate this in Fig.~\ref{fig:paired_semantic_all}, on an example with less pair variation from SiblingsDB. Available methods do not preserve the semantic component in the single-image setting (Fig.~\ref{fig:single_semantic_compact}), nor preserve consistent identities within pairs (next section), therefore, we omit them in the paired clinical setting of Fig.~\ref{fig:paired_semantic_all}. Table~\ref{table:deID_comparison} shows that our paired clinical anonymization is also on par with the state of the art in de-identification. Lastly, we show in Table~\ref{table:pair_ID_match} a quantitative evaluation supporting that our paired anonymization yields consistent synthetic identities. Our results are close to the recognition rate of the ground-truth input pair itself when anonymizing pairs, while our non-paired anonymization yields diverse identities. 

\subsection{Standard image anonymization} \label{sec:eval_anon_regular}

Aside from clinical (single or paired) image anonymization, we also evaluate our proposed VerA on the standard image anonymization task. We follow the commonly used in-place setting~\cite{maximov2020ciagan,gu2020password,hukkelaas2023deepprivacy2,wen2023divide} that enables scene preservation. This setting anonymizes precisely the inside of the face, and is more constraining than methods that allow more modifications~\cite{li2023riddle,barattin2023attribute}. We illustrate standard anonymization on single images in the top rows of Fig.~\ref{fig:regular}. We also show good quantitative de-identification performance in Table~\ref{table:deID_comparison}, on par with state-of-the-art approaches. Lastly, VerA also enables the anonymization of paired standard images of the same person. We show sample results of this anonymization in the bottom of Fig.~\ref{fig:regular} compared with state-of-the-art methods. Our approach preserves the consistency within the pair in terms of identity far better than baselines. This is also supported by our quantitative evaluation in Table~\ref{table:pair_ID_match}. We achieve the highest consistency in terms of re-identification rate within each pair, with facial embedding distance that is very close to the distance within the input pair itself.

\begin{table}[t]
    \centering
    \resizebox{1.0\columnwidth}{!}{%
    \begin{NiceTabular}{p{1em}l|c|c|c|c|c|c}
    \toprule
    & Face Recog.   & \multicolumn{2}{c}{CASIA~\cite{yi2014learning}} & \multicolumn{2}{c}{VGGFace2~\cite{cao2018vggface2}} & \multicolumn{2}{c}{FaceNet~\cite{schroff2015facenet}} \\
    & Setting                              & Rate $\uparrow$ & Dist. $\downarrow$ & Rate $\uparrow$ & Dist. $\downarrow$ & Rate $\uparrow$ & Dist. $\downarrow$ \\
    \midrule
    & Input pair                           & 92.8 &   -   & 90.0 &   -   & 89.20 &   - \\    
    \hdashline
    \multirow{6}{*}{\rotatebox[origin=c]{90}{Single image}} 
    & CIAGAN~\cite{maximov2020ciagan}      & 65.4 & +0.161 & 74.1 & +0.136 & 74.5 & +1.694 \\
    & FIT~\cite{gu2020password}            & 15.6 & +0.355 & 10.8 & +0.443 & 24.4 & +4.771 \\
    & DP2~\cite{hukkelaas2023deepprivacy2} & 30.8 & +0.293 & 23.6 & +0.347 & 75.6 & +0.976 \\
    & RiDDLE~\cite{li2023riddle}           & 12.0 & +0.469 & 10.4 & +0.515 & 29.6 & +4.939 \\
    & FALCO~\cite{barattin2023attribute}   & 52.8 & +0.224 & 40.4 & +0.300 & 62.0 & +2.413 \\
    & Ours (standard)                       & 47.2 & +0.242 & 26.8 & +0.364 & 69.6 & +1.702 \\
    \hdashline
    \multirow{4}{*}{\rotatebox[origin=c]{90}{Paired}} 
    & Ours (standard)                        & 85.2 & +0.065 & 71.6 & +0.154 & 85.6 & +0.132 \\
    & Ours (mouth)                          & \underline{89.6} & \underline{+0.026} & 78.8 & \underline{+0.102} & \underline{87.2} & \underline{+0.109} \\ 
    & Ours (nose)                          & \textbf{90.0} & \textbf{-0.001}& \textbf{82.4} & \textbf{+0.074} & \textbf{89.2} & \textbf{+0.065} \\
    & Ours (eyes)                         & 88.8 & +0.030 & \underline{80.4} & \underline{+0.102} & 83.6 & +0.324 \\
    \bottomrule
    \end{NiceTabular}
    }
    \caption{Face recognition re-identification rates and face embedding distances, \textit{within pairs}. We show the distance change relative to the distance within the input pair itself. Our paired anonymization achieves higher rates, and lower distances, indicating that the identity is consistent between the two images inside each pair.}
    \label{table:pair_ID_match}
\end{table}

\begin{table}[t]
    \centering
    \resizebox{\columnwidth}{!}{%
    \begin{NiceTabular}{l|c|c||c|c|c|c}
    \toprule
    \multirow{2}{*}{Method}              & \multicolumn{2}{c}{FID $\downarrow$} & \multicolumn{2}{c}{Bounding box $\uparrow$} & \multicolumn{2}{c}{Face detection $\uparrow$} \\
                                         & FFHQ & CelebAHQ & MTCNN & Dlib  & MTCNN & Dlib  \\
    \midrule
    CIAGAN~\cite{maximov2020ciagan}      & 138.87 & 74.91 & 0.82  & 0.91  & 0.93  & 0.94  \\
    FIT~\cite{gu2020password}            & 125.67 & 79.80 & \textbf{0.92}  & \textbf{0.97}  & 0.97  & \underline{0.99}  \\
    DP2~\cite{hukkelaas2023deepprivacy2} & \underline{69.36} & \underline{19.34} & 0.88  & 0.88  & 0.92  & 0.96  \\
    RiDDLE~\cite{li2023riddle}           & 137.96 & 68.63 & 0.90  & \underline{0.95}  & \textbf{1.00}  & \textbf{1.00}  \\
    FALCO~\cite{barattin2023attribute}   & 87.81 & 37.62 & 0.90  & 0.94  & \underline{0.99}  & \textbf{1.00}  \\
    Ours                                 & \textbf{57.77} & \textbf{12.57} & \underline{0.91}  & \underline{0.95}  & 0.96  & \underline{0.99}  \\
    \bottomrule
    \end{NiceTabular}
    }
    \caption{Downstream utility evaluation for photorealism/diversity (FID~\cite{heusel2017gans}), bounding box IoU, and face detection rates (MTCNN~\cite{zhang2016joint}, Dlib~\cite{king2009dlib}). We achieve the best FID, and are on par with the best bounding box IoU and the best detection scores.}
    \label{table:utility}
\end{table}

\begingroup
\setlength{\tabcolsep}{-0.3pt} 
\renewcommand{\arraystretch}{0.65} 
\begin{figure*}[t]
\centering
\begin{tabular}{lccccccc}
\hspace{-0.0cm}
\rotatebox[origin=c]{90}{\makebox[1mm]{Single}} \hspace{-0.0cm} &
\begin{tabular}{c} \includegraphics[width=0.135\linewidth]{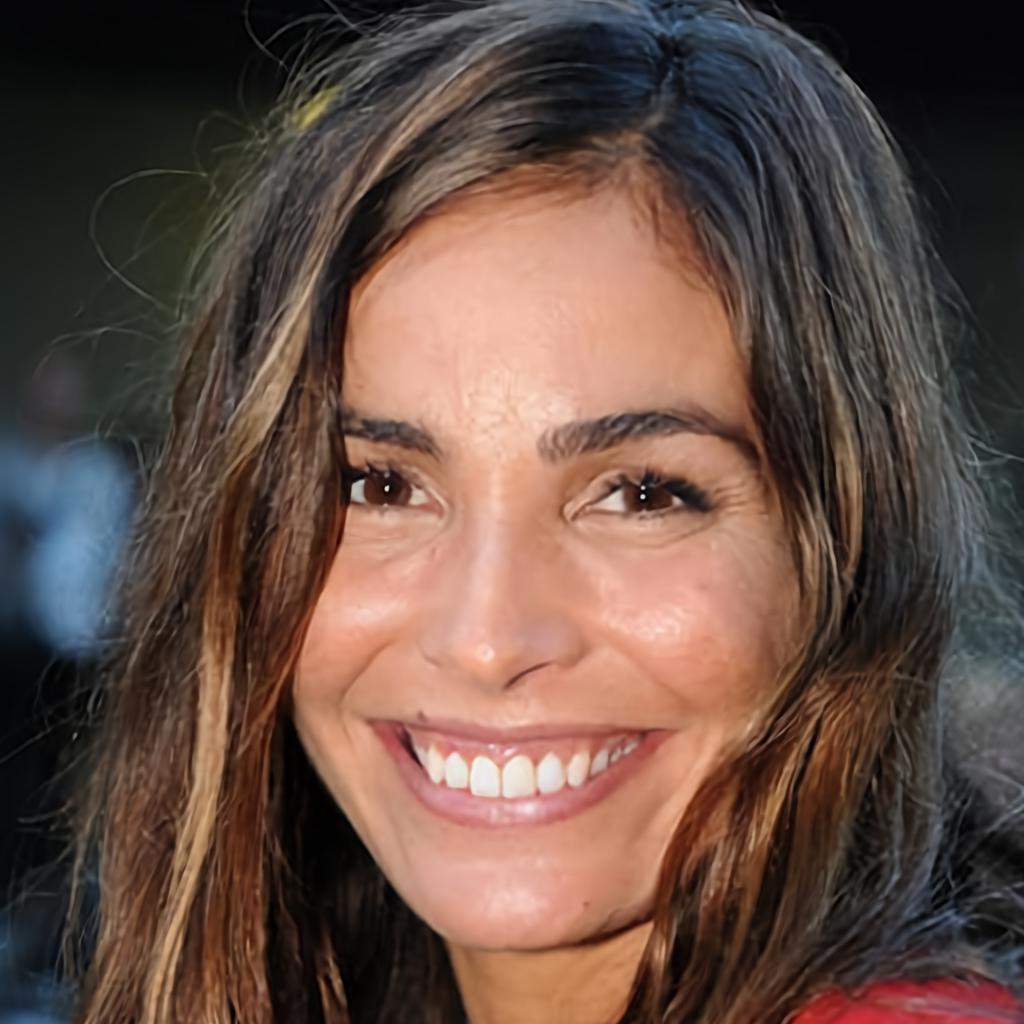} \end{tabular} \hspace{-0.0cm}
&
\begin{tabular}{c} \includegraphics[width=0.135\linewidth]{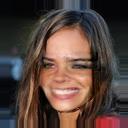} \end{tabular} \hspace{-0.0cm}
&
\begin{tabular}{c} \includegraphics[width=0.135\linewidth]{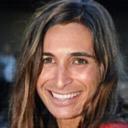} \end{tabular} \hspace{-0.0cm}
&
\begin{tabular}{c} \includegraphics[width=0.135\linewidth]{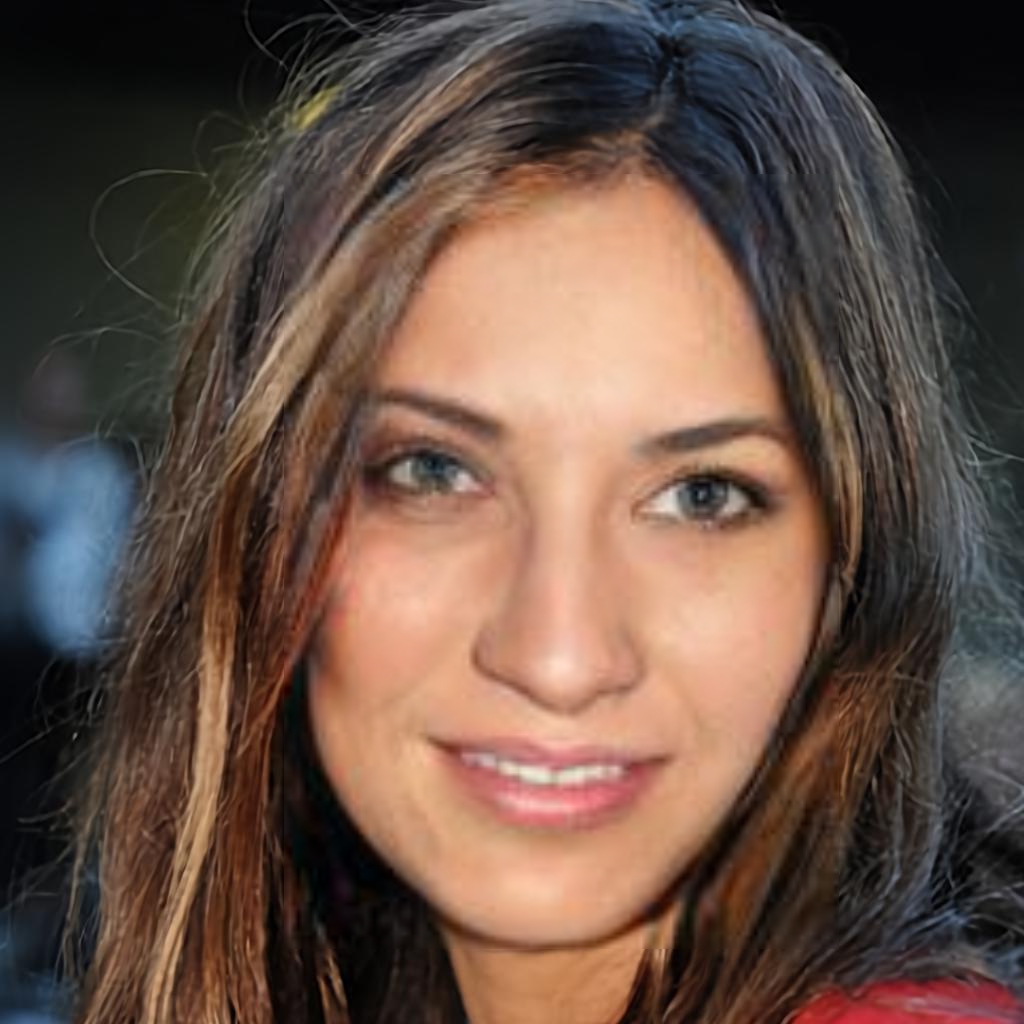} \end{tabular} \hspace{-0.0cm}
&
\begin{tabular}{c} \includegraphics[width=0.135\linewidth]{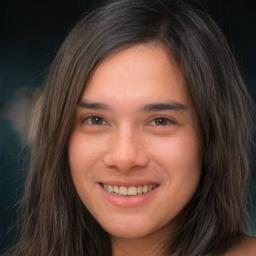} \end{tabular} \hspace{-0.0cm}
&
\begin{tabular}{c} \includegraphics[width=0.135\linewidth]{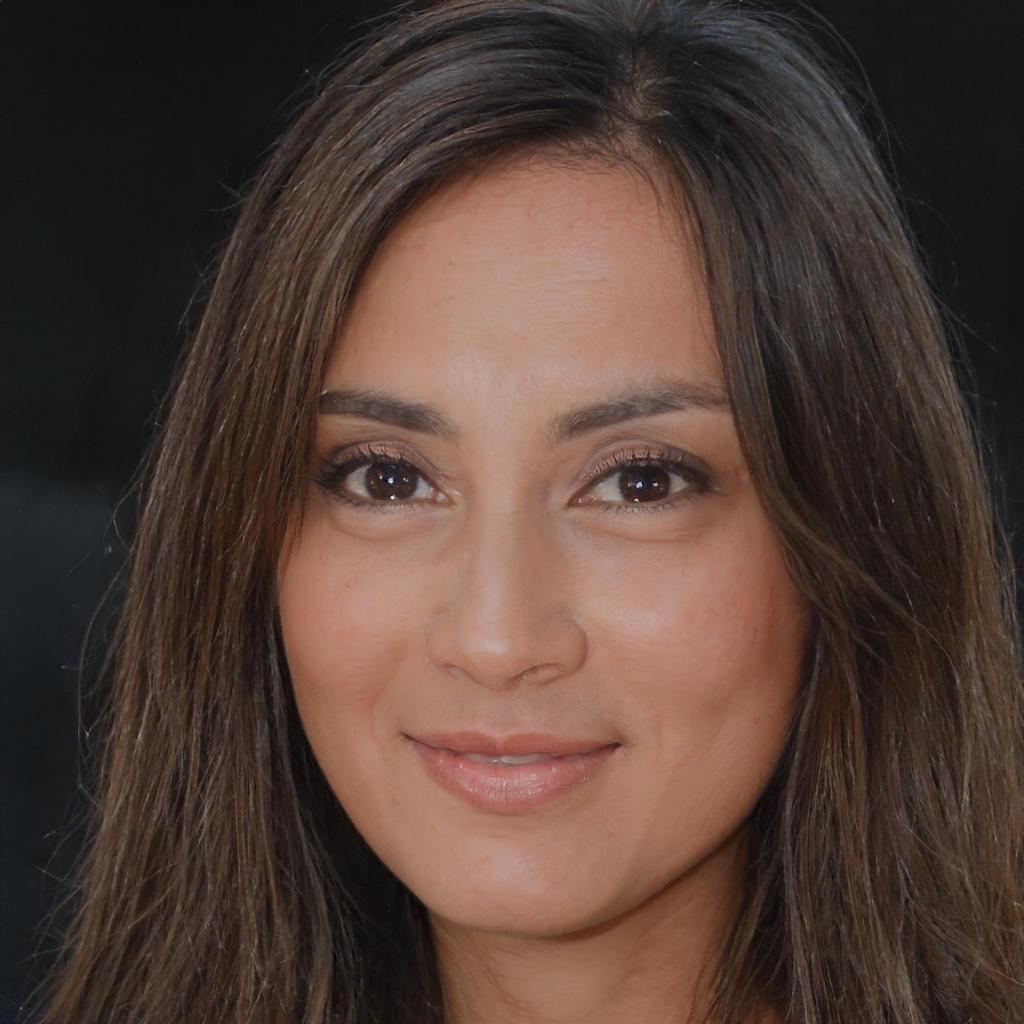} \end{tabular} \hspace{-0.0cm}
&
\begin{tabular}{c} \includegraphics[width=0.135\linewidth]{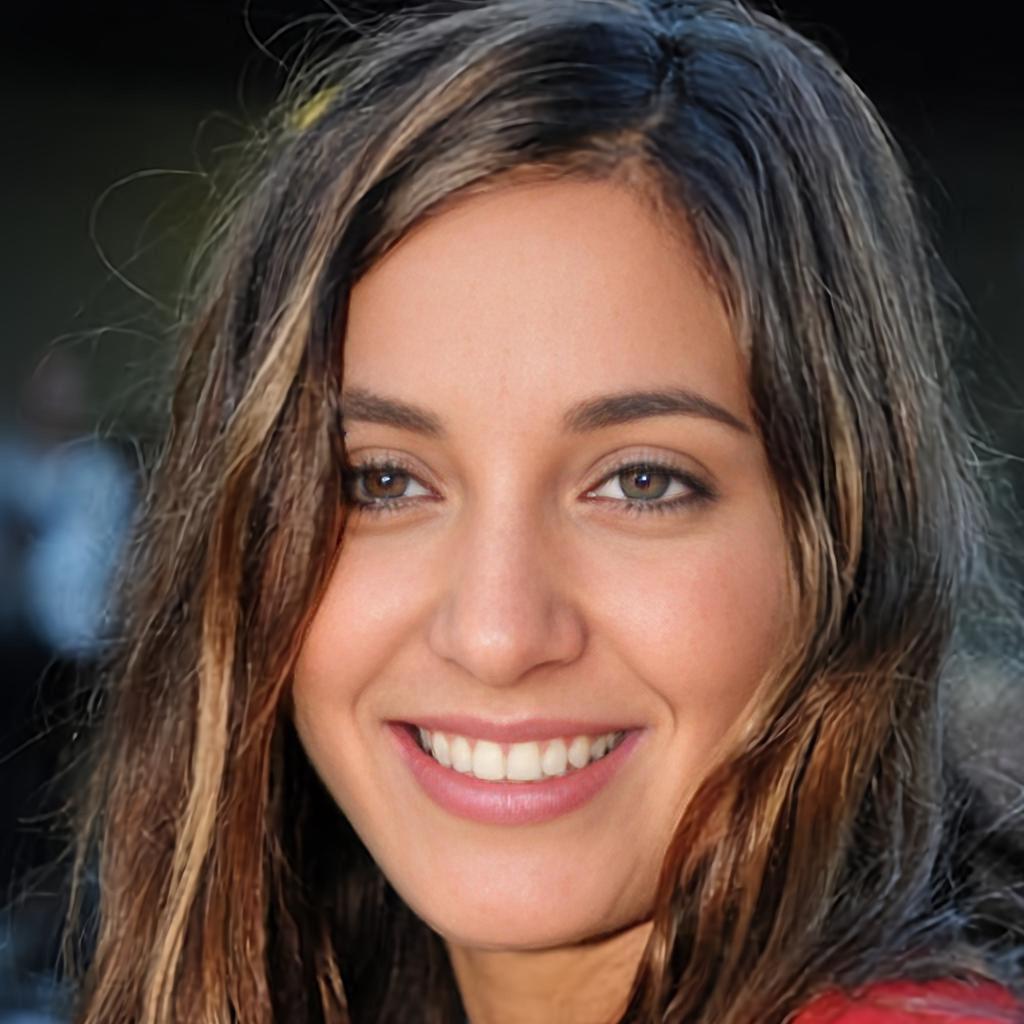} \end{tabular} \hspace{-0.0cm}
\\
\hspace{-0.0cm}
\rotatebox[origin=c]{90}{\makebox[1mm]{Single}} \hspace{-0.0cm} &
\begin{tabular}{c} \includegraphics[width=0.135\linewidth]{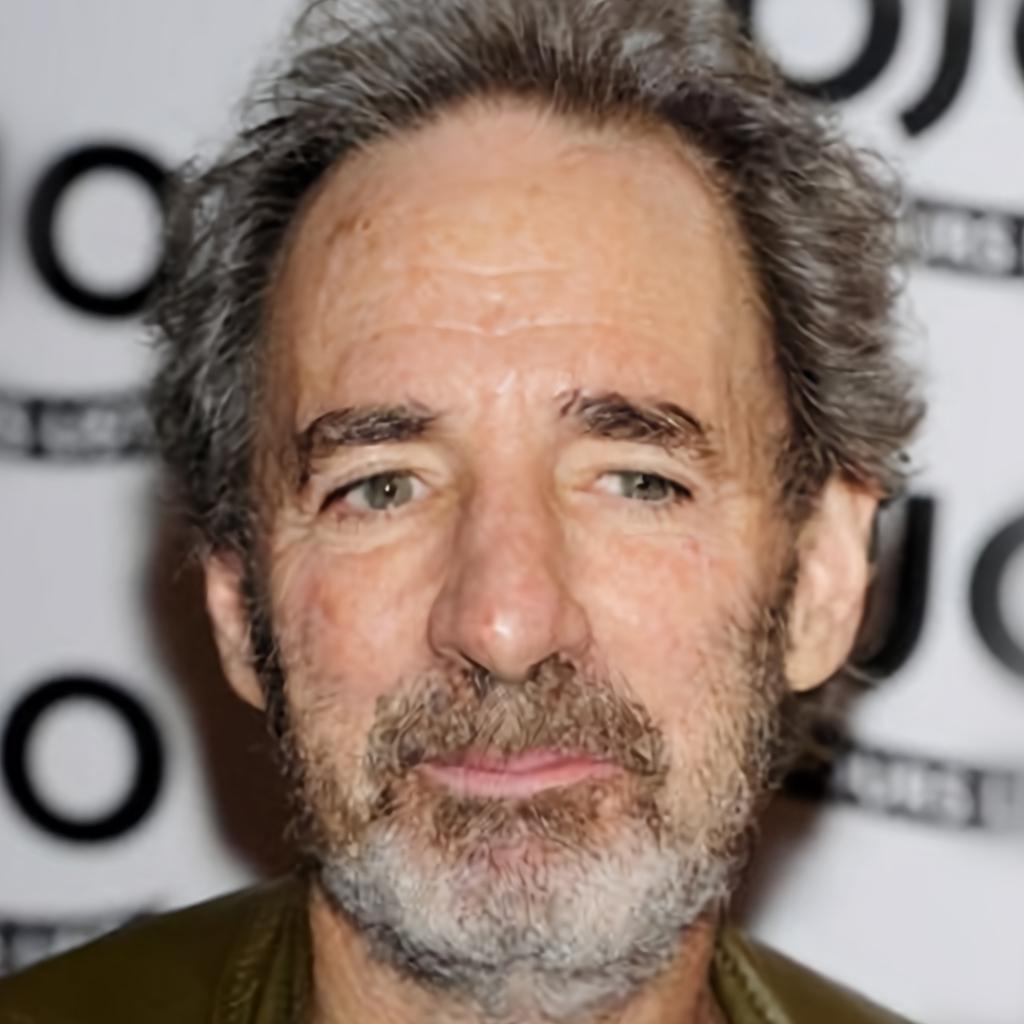} \end{tabular} \hspace{-0.0cm}
&
\begin{tabular}{c} \includegraphics[width=0.135\linewidth]{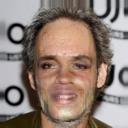} \end{tabular} \hspace{-0.0cm}
&
\begin{tabular}{c} \includegraphics[width=0.135\linewidth]{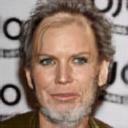} \end{tabular} \hspace{-0.0cm}
&
\begin{tabular}{c} \includegraphics[width=0.135\linewidth]{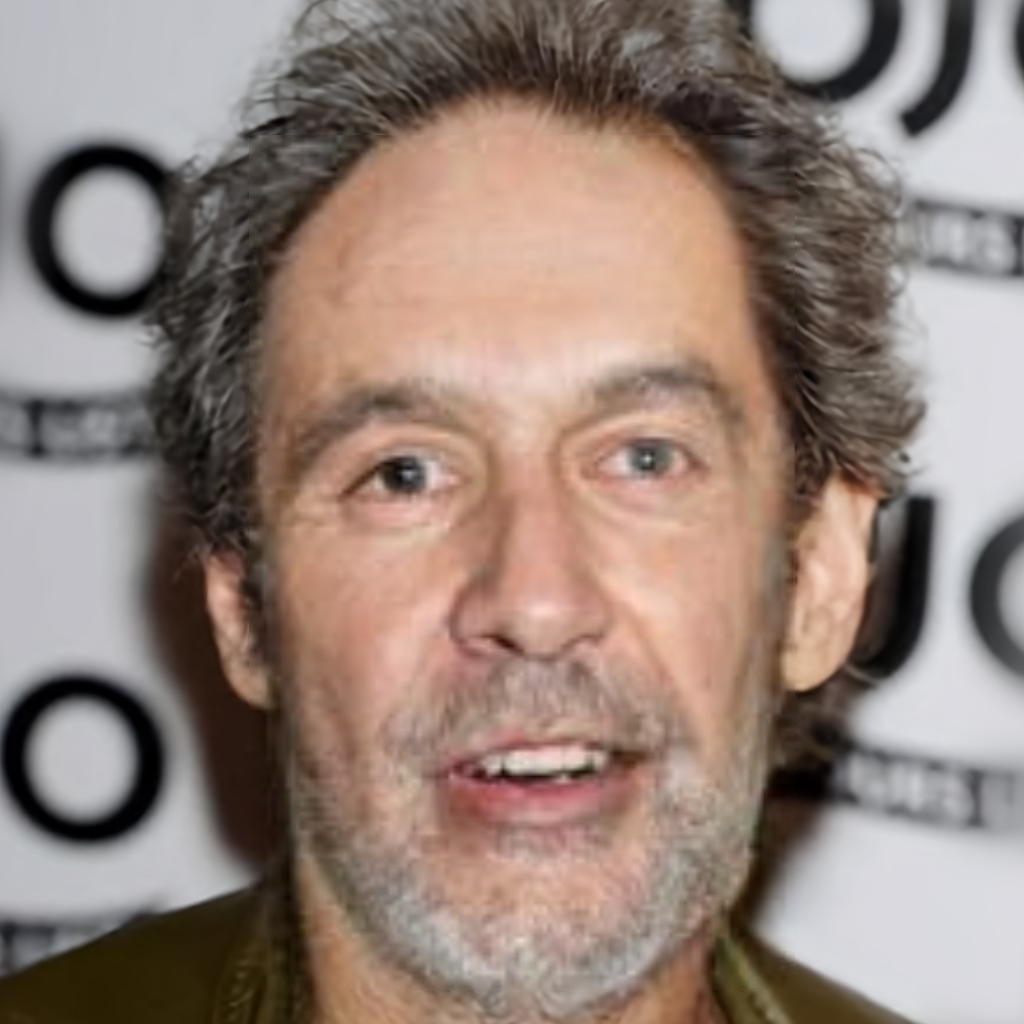} \end{tabular} \hspace{-0.0cm}
&
\begin{tabular}{c} \includegraphics[width=0.135\linewidth]{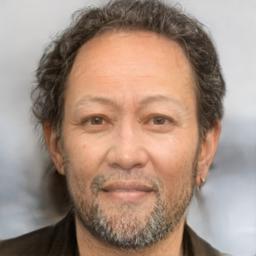} \end{tabular} \hspace{-0.0cm}
&
\begin{tabular}{c} \includegraphics[width=0.135\linewidth]{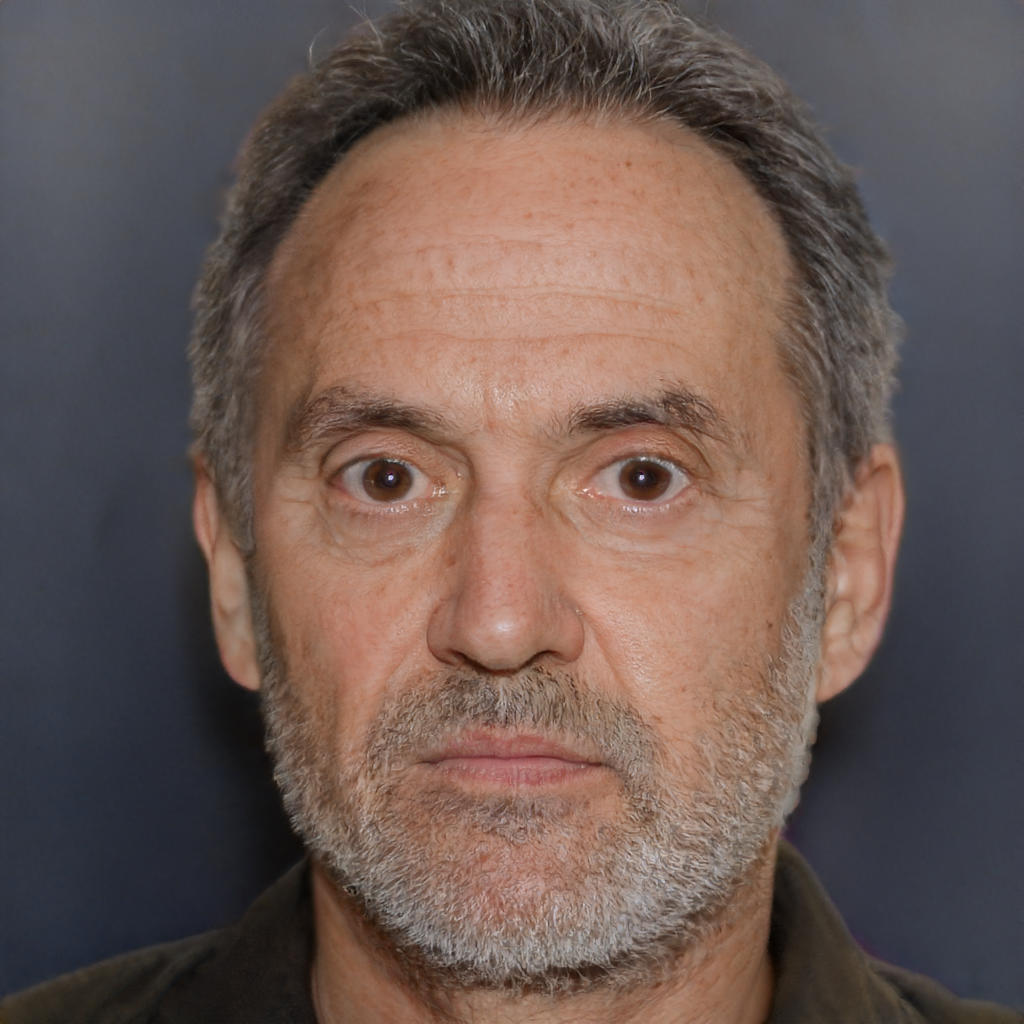} \end{tabular} \hspace{-0.0cm}
&
\begin{tabular}{c} \includegraphics[width=0.135\linewidth]{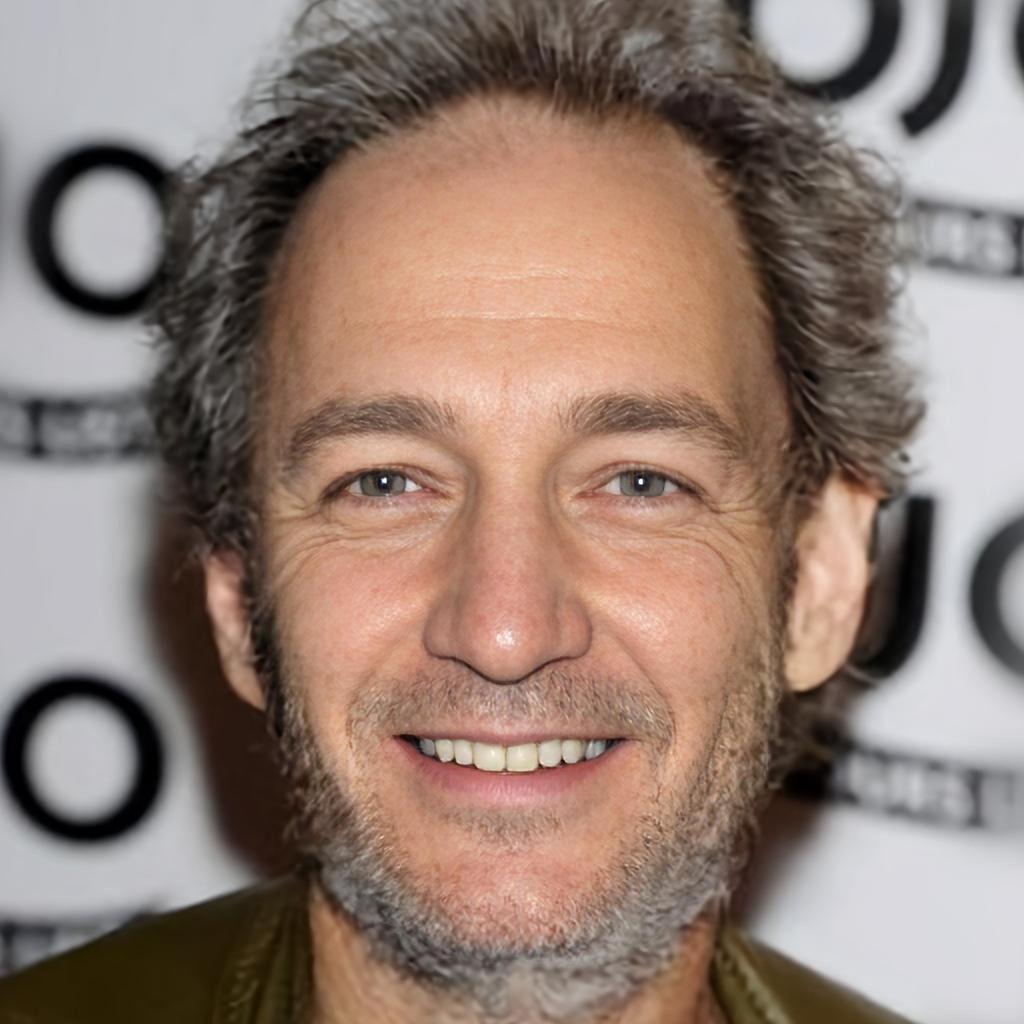} \end{tabular} \hspace{-0.0cm}
\\
\hspace{-0.0cm}
\multirow{2}{*}{\rotatebox[origin=c]{90}{\makebox[20mm]{Paired}}} \hspace{-0.0cm} &
\begin{tabular}{c} \includegraphics[width=0.135\linewidth]{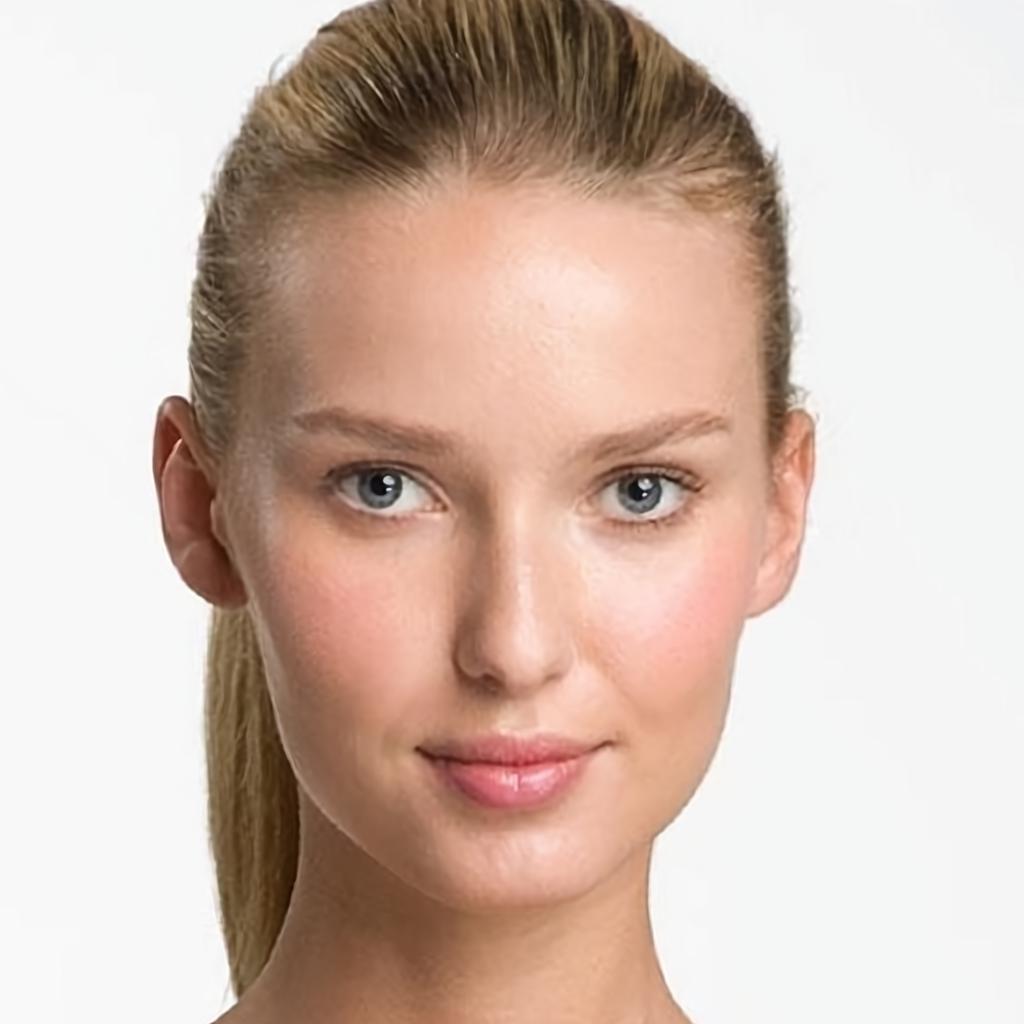} \end{tabular} \hspace{-0.0cm}
&
\begin{tabular}{c} \includegraphics[width=0.135\linewidth]{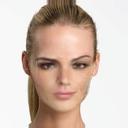} \end{tabular} \hspace{-0.0cm}
&
\begin{tabular}{c} \includegraphics[width=0.135\linewidth]{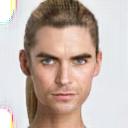} \end{tabular} \hspace{-0.0cm}
&
\begin{tabular}{c} \includegraphics[width=0.135\linewidth]{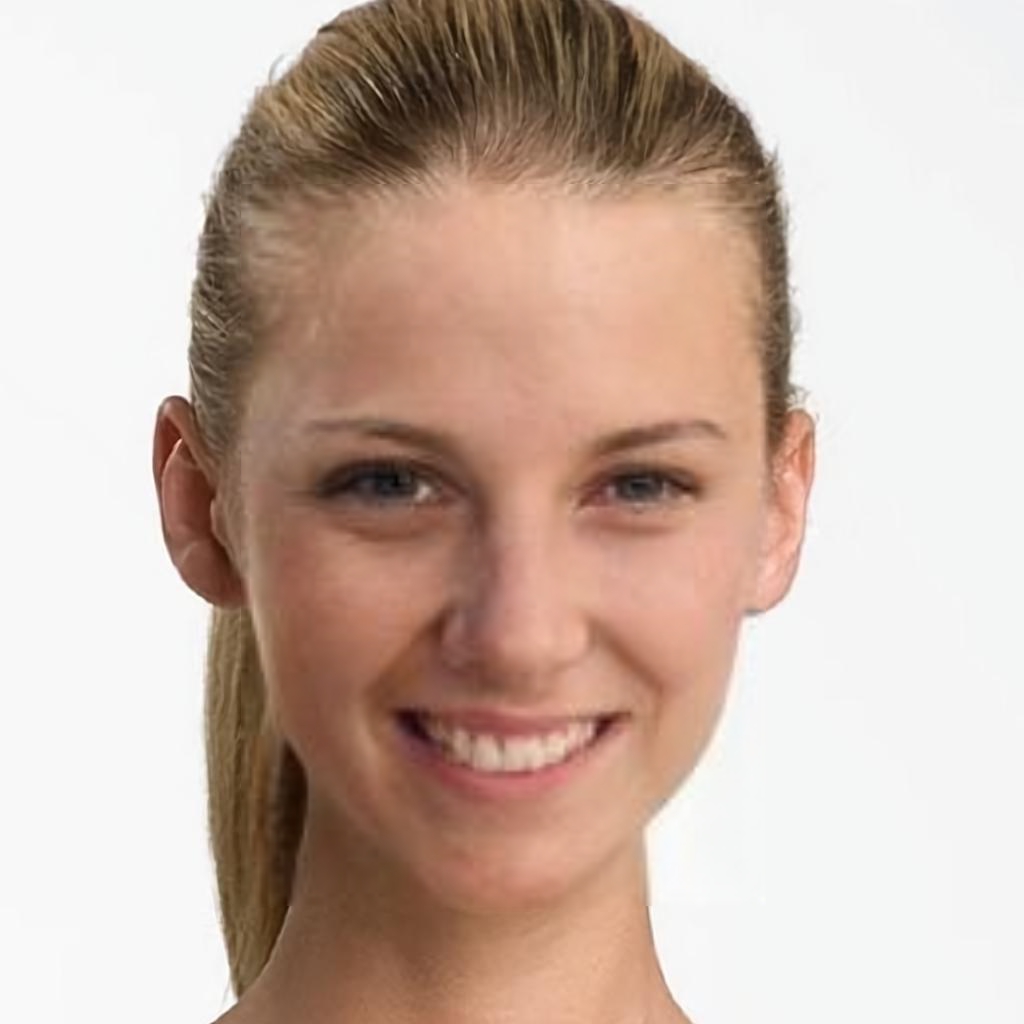} \end{tabular} \hspace{-0.0cm}
&
\begin{tabular}{c} \includegraphics[width=0.135\linewidth]{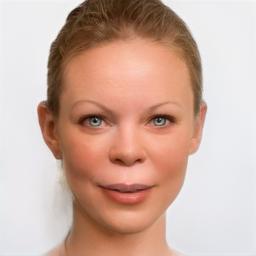} \end{tabular} \hspace{-0.0cm}
&
\begin{tabular}{c} \includegraphics[width=0.135\linewidth]{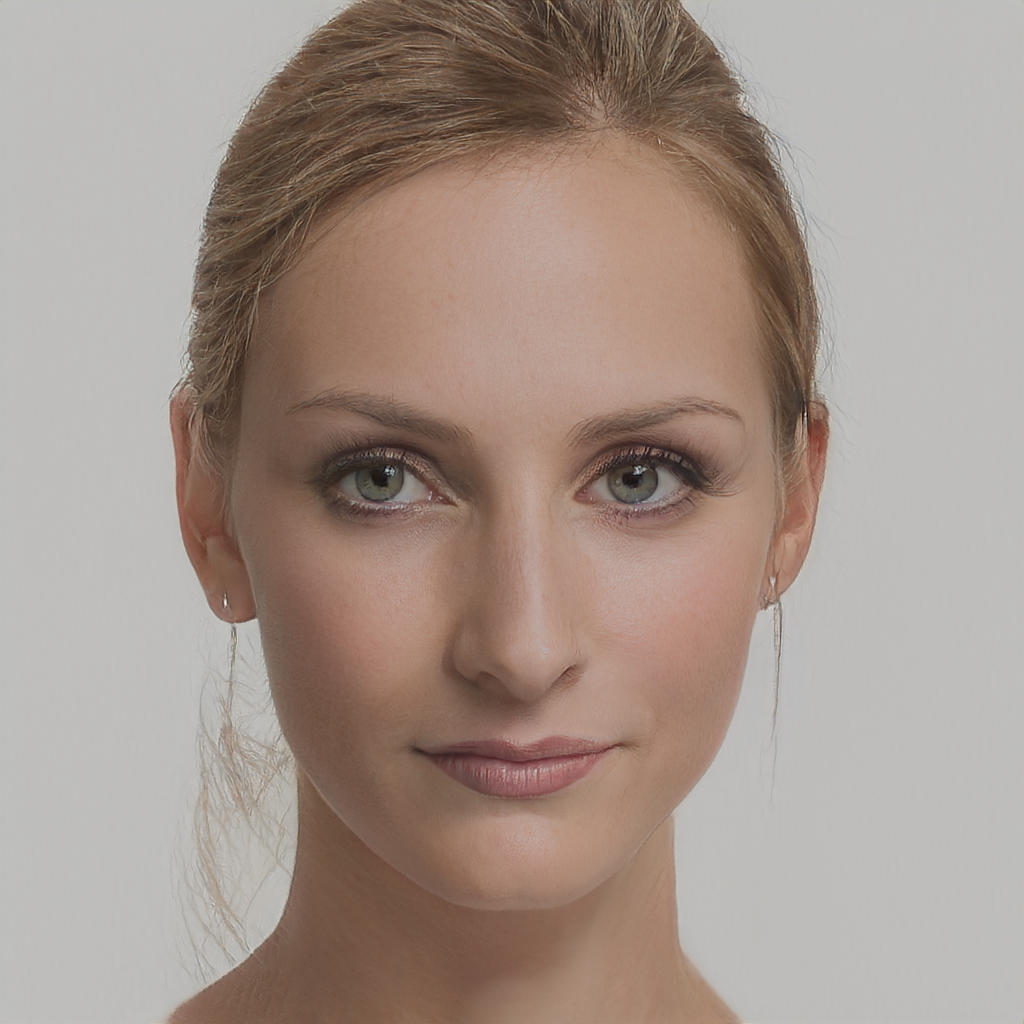} \end{tabular} \hspace{-0.0cm}
&
\begin{tabular}{c} \includegraphics[width=0.135\linewidth]{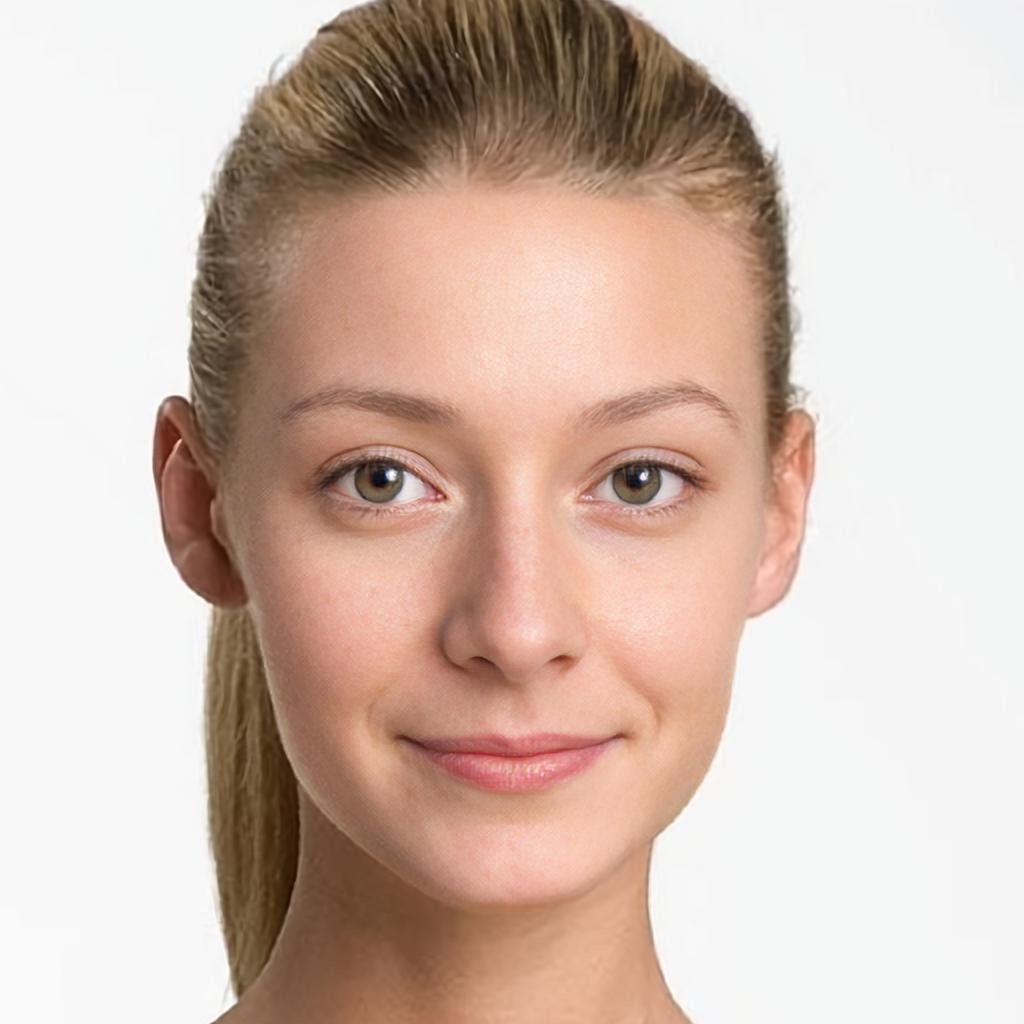} \end{tabular} \hspace{-0.0cm}
\\
\hspace{-0.0cm}
\hspace{-0.0cm} &
\begin{tabular}{c} \includegraphics[width=0.135\linewidth]{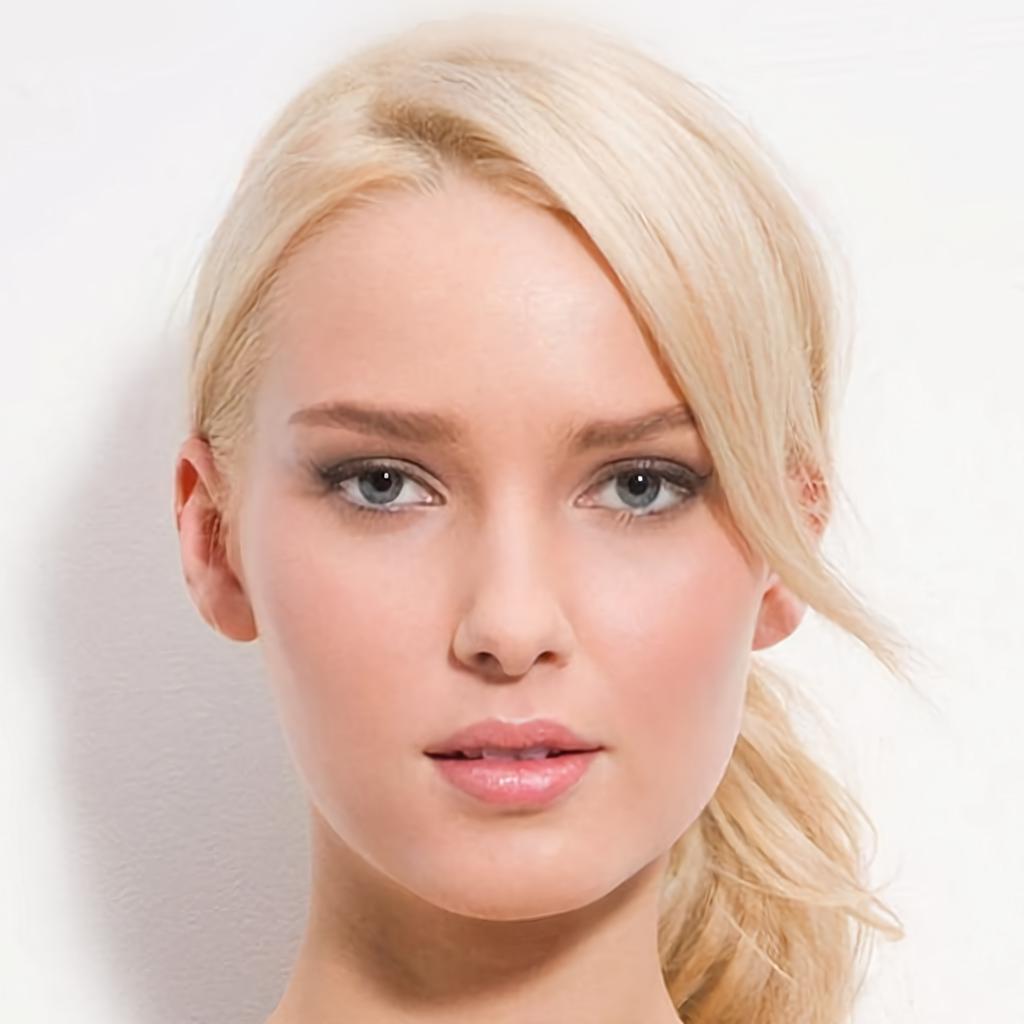} \end{tabular} \hspace{-0.0cm}
&
\begin{tabular}{c} \includegraphics[width=0.135\linewidth]{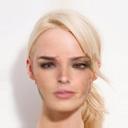} \end{tabular} \hspace{-0.0cm}
&
\begin{tabular}{c} \includegraphics[width=0.135\linewidth]{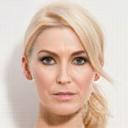} \end{tabular} \hspace{-0.0cm}
&
\begin{tabular}{c} \includegraphics[width=0.135\linewidth]{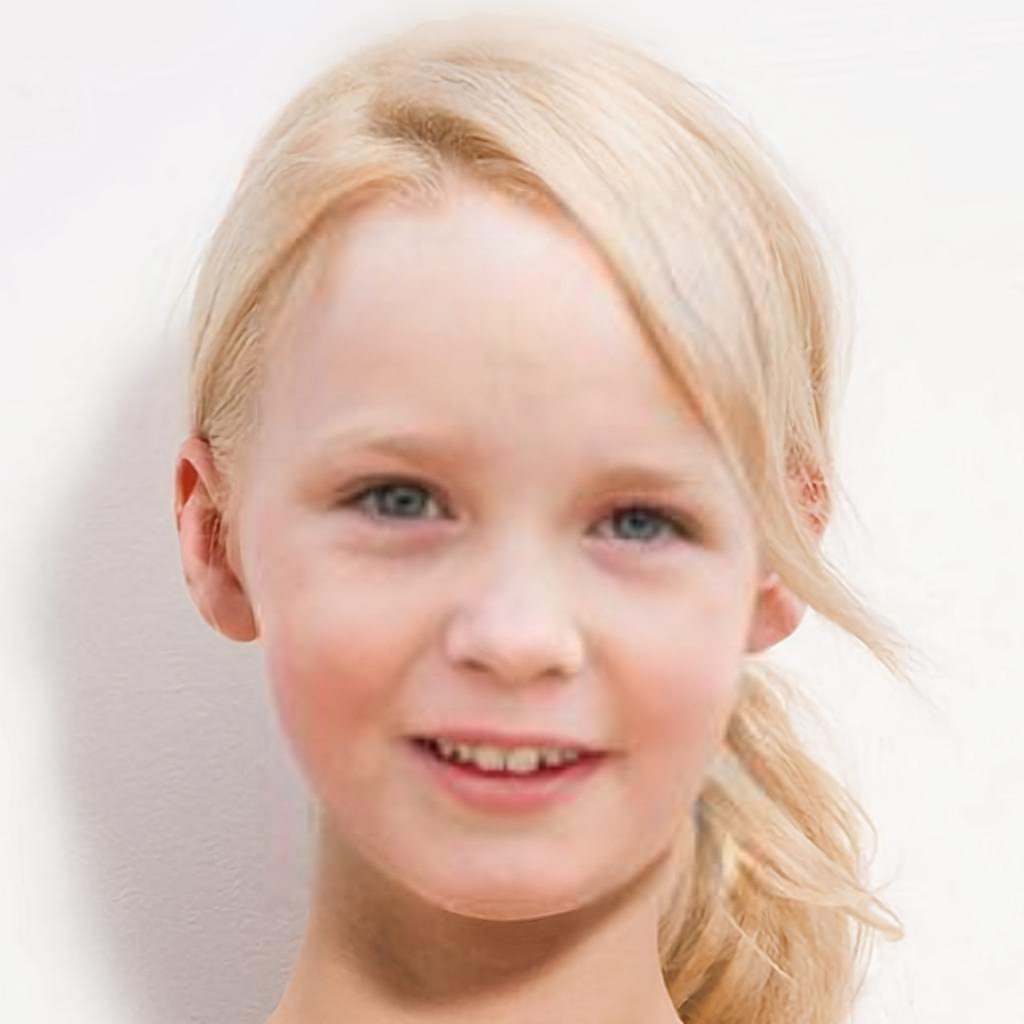} \end{tabular} \hspace{-0.0cm}
&
\begin{tabular}{c} \includegraphics[width=0.135\linewidth]{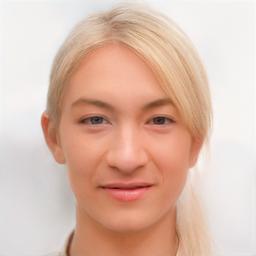} \end{tabular} \hspace{-0.0cm}
&
\begin{tabular}{c} \includegraphics[width=0.135\linewidth]{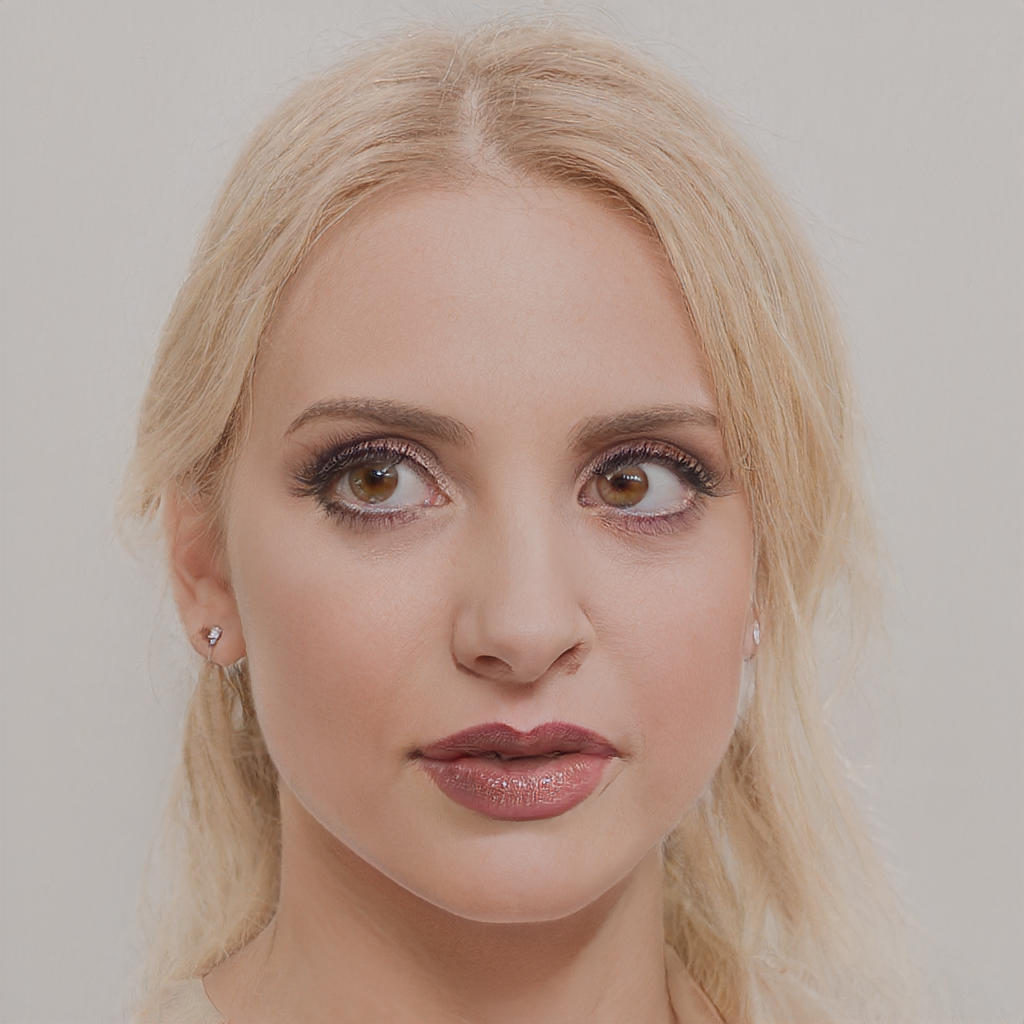} \end{tabular} \hspace{-0.0cm}
&
\begin{tabular}{c} \includegraphics[width=0.135\linewidth]{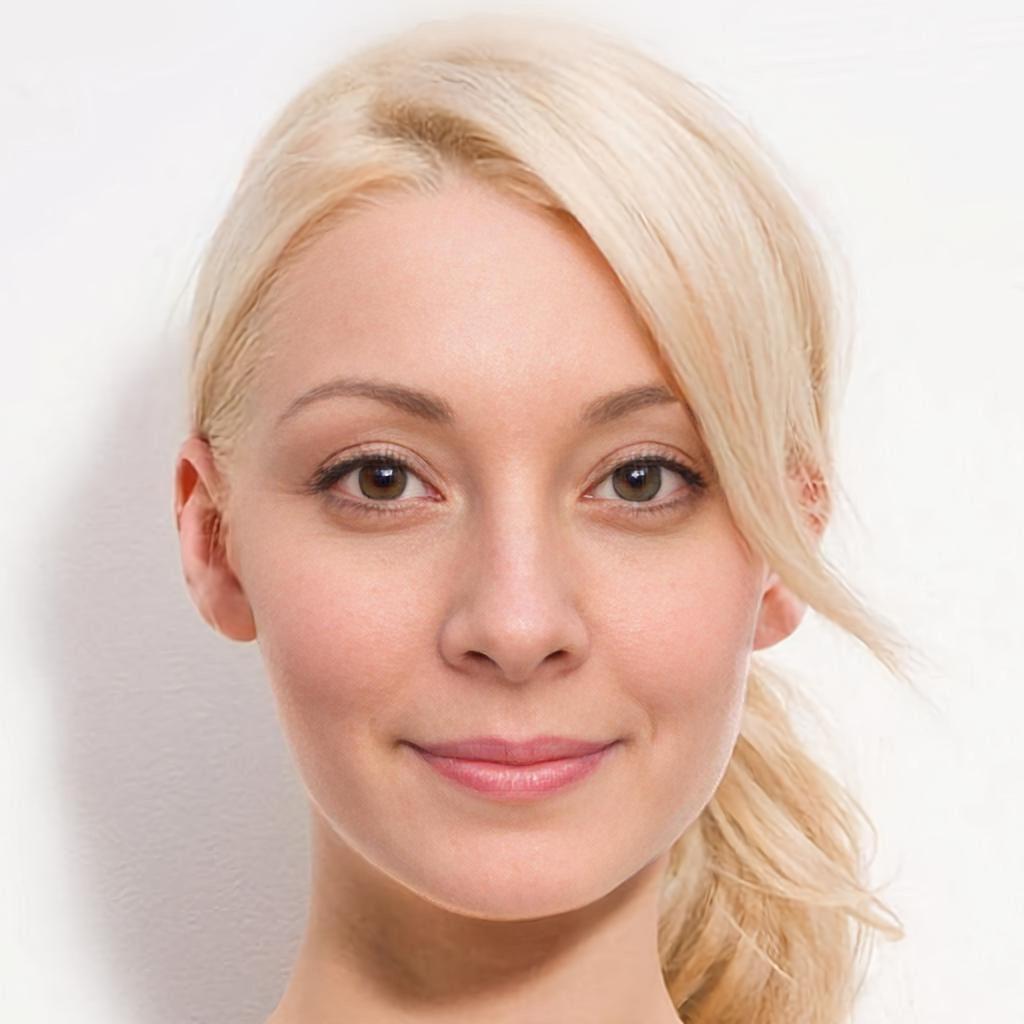} \end{tabular} \hspace{-0.0cm}
\\
\hspace{-0.0cm}
\rotatebox[origin=c]{90}{\makebox[1mm]{ }} \hspace{-0.0cm} &
\begin{tabular}{c} \makebox[0.135\linewidth]{Input} \end{tabular} \hspace{-0.0cm}
&
\begin{tabular}{c} \makebox[0.135\linewidth]{CIAGAN~\cite{maximov2020ciagan}} \end{tabular} \hspace{-0.0cm}
&
\begin{tabular}{c} \makebox[0.135\linewidth]{FIT~\cite{gu2020password}} \end{tabular} \hspace{-0.0cm}
&
\begin{tabular}{c} \makebox[0.135\linewidth]{DP2~\cite{hukkelaas2023deepprivacy2}} \end{tabular} \hspace{-0.0cm}
&
\begin{tabular}{c} \makebox[0.135\linewidth]{RiDDLE~\cite{li2023riddle}} \end{tabular} \hspace{-0.0cm}
&
\begin{tabular}{c} \makebox[0.135\linewidth]{FALCO~\cite{barattin2023attribute}} \end{tabular} \hspace{-0.0cm}
&
\begin{tabular}{c} \makebox[0.135\linewidth]{Ours} \end{tabular} \hspace{-0.0cm}
\\
\end{tabular}
\caption{Sample qualitative results of \textit{standard single-image} and \textit{standard paired-image} anonymization, compared with the state of the art. Best viewed zoomed in on screen.}
\label{fig:regular}
\end{figure*}
\endgroup

\subsection{Downstream utility} \label{sec:eval_downstream}
Beyond simple de-identification, it is important for the resulting images to be useful downstream. We conduct an evaluation of photorealism and diversity through FID~\cite{heusel2017gans}, which measures distribution matching, and report the results in Table~\ref{table:utility}. We outperform the state-of-the-art anonymization methods and the older baselines by a large margin. We also provide bounding box IoU results as well as face detection rates using MTCNN~\cite{zhang2016joint} and Dlib~\cite{king2009dlib} where we are consistently on par with the state-of-the-art. 

\section{Discussion} \label{sec:discussion}
\noindent \textbf{Limitations.} Our generator training requires semantic maps. We predict them for FFHQ images without noting any subsequent degradation, due to the good performance of the semantic segmentation model.
The final resolution of $512\times 512$ can be deemed low for clinical use.
Training larger models would require even more GPU resources, so we believe using off-the-shelf super-resolution models to upscale synthesized images is a promising direction.
\\
\noindent \textbf{Ethics considerations.} Although our anonymization is for privacy-preserving image dissemination, it may be used maliciously, for instance, to remove a person without consent. Also, training data may contain underlying distribution biases. For fairness, data could be balanced accordingly or results re-sampled.
For clinical anonymization, if a malicious third party knows that an image was anonymized and even knows which semantic component was preserved, they could try to re-identify the protected person by reverse searching this component. In practice however, reverse searching the internet based on only one semantic component is neither readily feasible nor likely to yield reliable results, particularly as the clinical input image is kept offline or under restricted access.
This risk, however, is inherent in our clinical image anonymization task and must be addressed through such existing laws and methods that restrict access and dissemination of clinical data, except authorized professionals.
It is also important to note that different images might have their exceptional features and different semantic regions might exacerbate such existing risks. This applies to subjects with identifying features on skin (such as tattoos) or facial jewelry. Similarly, the risk of reidentification would increase if the subject was a public figure.

\section{Conclusion} \label{sec:conclusion}
We present VerA, a versatile facial image anonymization framework. VerA enables clinical anonymization preserving desired semantic components, to show medical intervention results on the face. We also address both standard and clinical anonymization of paired images, enabling before-and-after clinical image anonymization. VerA can additionally perform standard image anonymization, on par with the state-of-the-art methods in de-identification, with improved photorealism and downstream utility. We hope our progress, coupled with the common privacy guarantees on the clinical input images, paves the way for safe dissemination of clinical results aiding future patients, practitioners in sharing and analyzing results, and ultimately enabling the training of downstream methods.

\clearpage

{\small
\bibliographystyle{ieee_fullname}
\bibliography{egbib}
}

\clearpage



\centerline{\large\bf Supplementary Material}%
\vspace*{12pt}%
\textit{
    We present extended results illustrating the control of our image generator, both in terms of semantic and high-level generative control. We additionally propose extended anonymization evaluation for the different problem settings. Namely, further results on standard single-image anonymization, clinical single-image anonymization, as well as the paired counterparts where two images of the same person need to be anonymized consistently. We also present illustrative examples of full-image anonymization and more evaluation on downstream utility. Lastly, we provide an ablation study of our proposed mirroring contrastive learning and the projection heads we learn on top of the pretrained high-level encoders. We add the table of contents below for more convenience in navigating between the sections. All the supplementary results support what we present in our main manuscript.
}
\vspace*{12pt}


\renewcommand{\contentsname}{} 
\section*{Table of Contents} 
\vspace{-0.8cm}

\begingroup
\let\clearpage\relax
\setcounter{tocdepth}{2}
\tableofcontents
\endgroup

\renewcommand\thesection{\Alph{section}}
\setcounter{section}{0}

\addtocontents{toc}{\protect\setcounter{tocdepth}{2}}

\section{Extended results of controllable synthesis} \label{sec:supp_synthesis}

\subsection{Semantic generative image control}
\begin{figure}[t]
    \centering
    \begin{subfigure}{\linewidth}
        \includegraphics[clip,width=\textwidth]{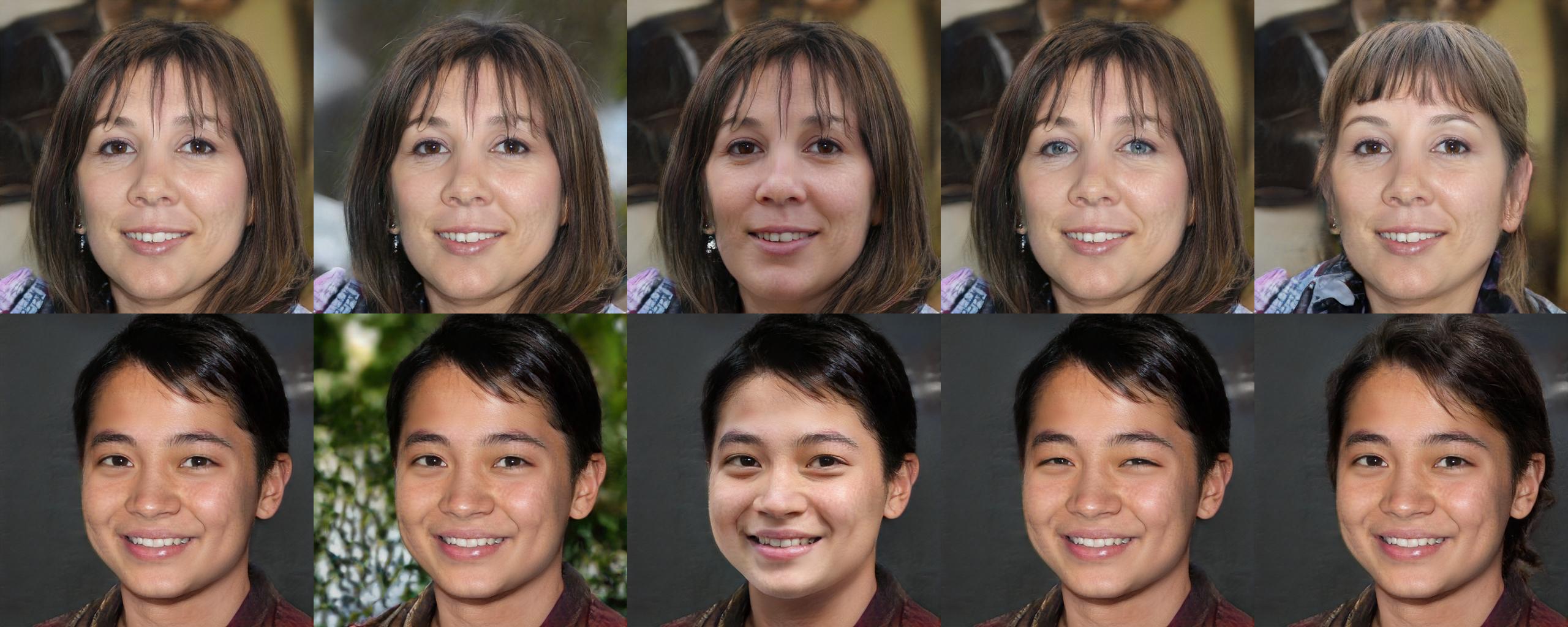}
        \caption{Independent change of semantic components}
        \label{fig:control_semantic_a}
    \end{subfigure}
    \begin{subfigure}{\linewidth}
        \includegraphics[clip,width=\textwidth]{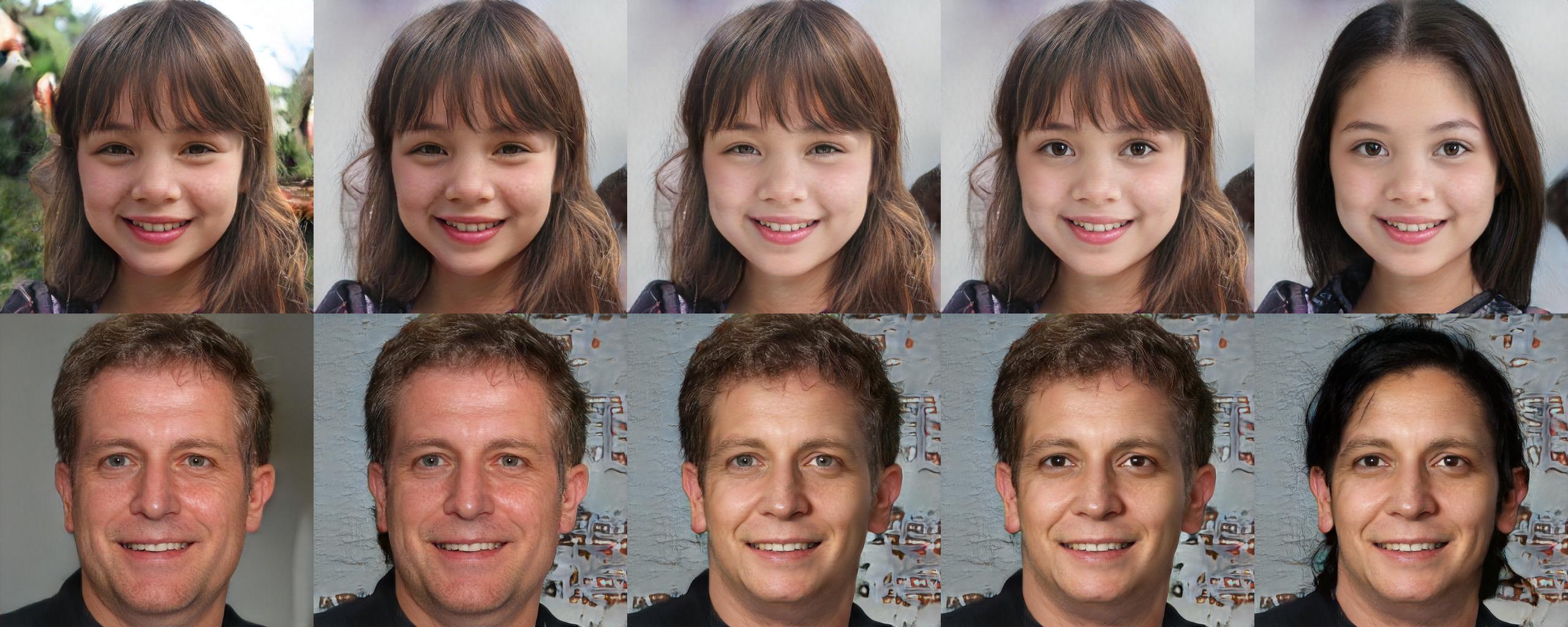}
        \caption{Same as (a), with accumulation of changes from left to right}
        \label{fig:control_semantic_b}
    \end{subfigure}
    \caption{Semantic control with changes applied (a) independently, and (b) cumulatively from left to right. The leftmost image is the original, and from left to right we change both the structure and texture of: \textit{background, face, eyes} and \textit{hair}.}
    \label{fig:control_semantic}
\end{figure}

We illustrate the disentangled semantic control capabilities of our generator in Fig.~\ref{fig:control_semantic}. We show two examples in Fig.~\ref{fig:control_semantic_a} where each column has a different semantic change relative to the leftmost column. In Fig.~\ref{fig:control_semantic_b}, the changes are accumulated from left to right, modifying in order the background, face, eyes and hair. This semantic control is due to the architecture components from SemanticStyleGAN~\cite{shi2022semanticstylegan} that extends on StyleGAN2~\cite{karras2020analyzing}, and we show here that the high-level control that we achieved with our training does not block the semantic control.

\begingroup
\setlength{\tabcolsep}{1pt} 
\renewcommand{\arraystretch}{0} 

\begin{figure}[!t]
    \centering
    \begin{tabular}{cc}
      \begin{tabular}{c} \rotatebox[origin=c]{90}{\makebox[1pt]{Pose}} \end{tabular} &
      \begin{tabular}{c} \includegraphics[width=0.9\linewidth]{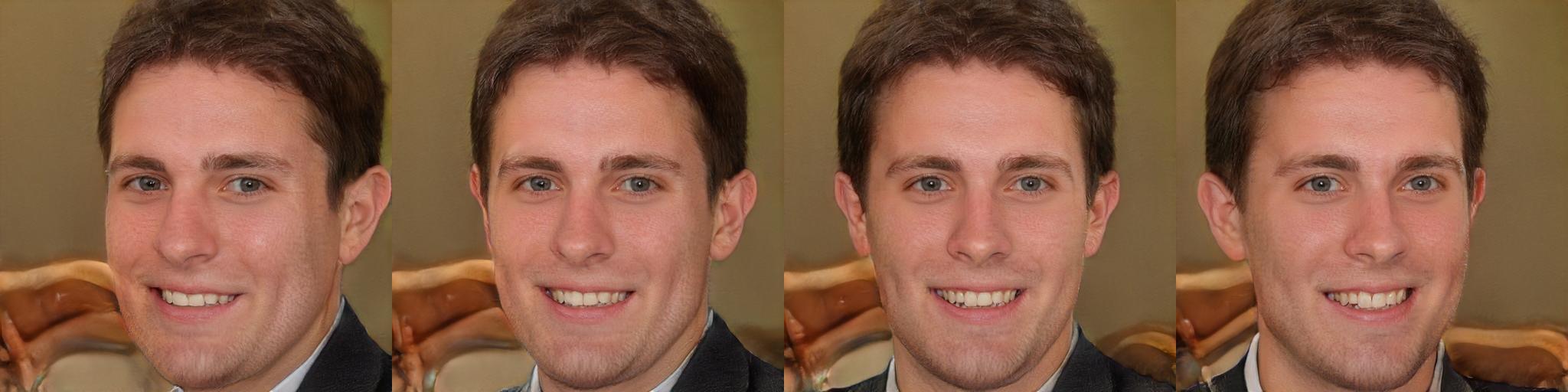} \end{tabular} \\
      \begin{tabular}{c} \rotatebox[origin=c]{90}{\makebox[1pt]{Age}} \end{tabular} &
      \begin{tabular}{c} \includegraphics[width=0.9\linewidth]{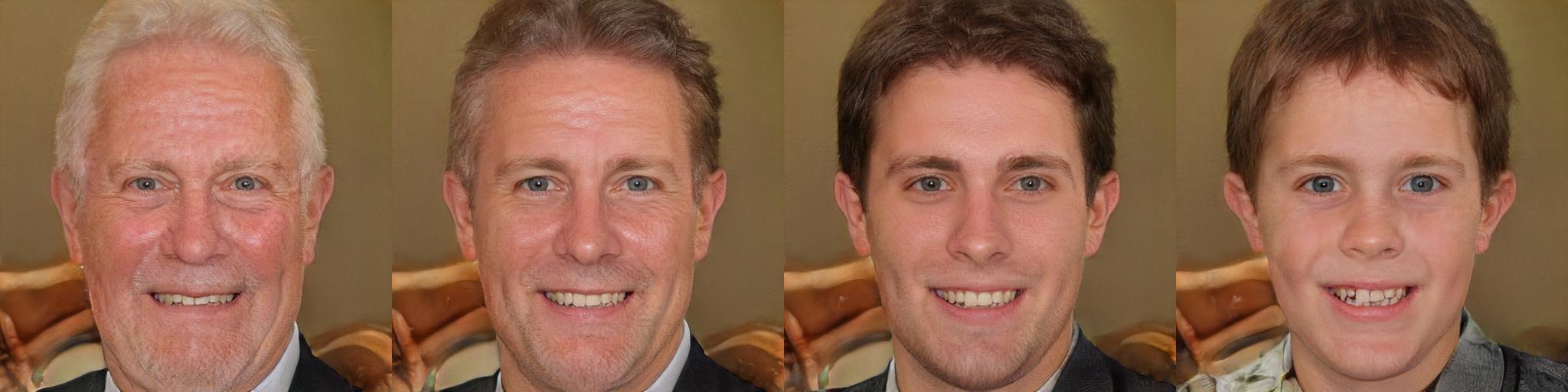} \end{tabular}  \\
    \end{tabular}
    \caption{Illustration of high-level changes, on the same person. The top row modifies the \textit{pose} gradually, and the bottom row modifies \textit{age} gradually. All other attributes remain unchanged.}
    \label{fig:control_high_sweep} 
\end{figure}

\endgroup

\begin{figure}[!t]
    \centering
    \begin{subfigure}{\linewidth}
        \includegraphics[clip,width=\textwidth]{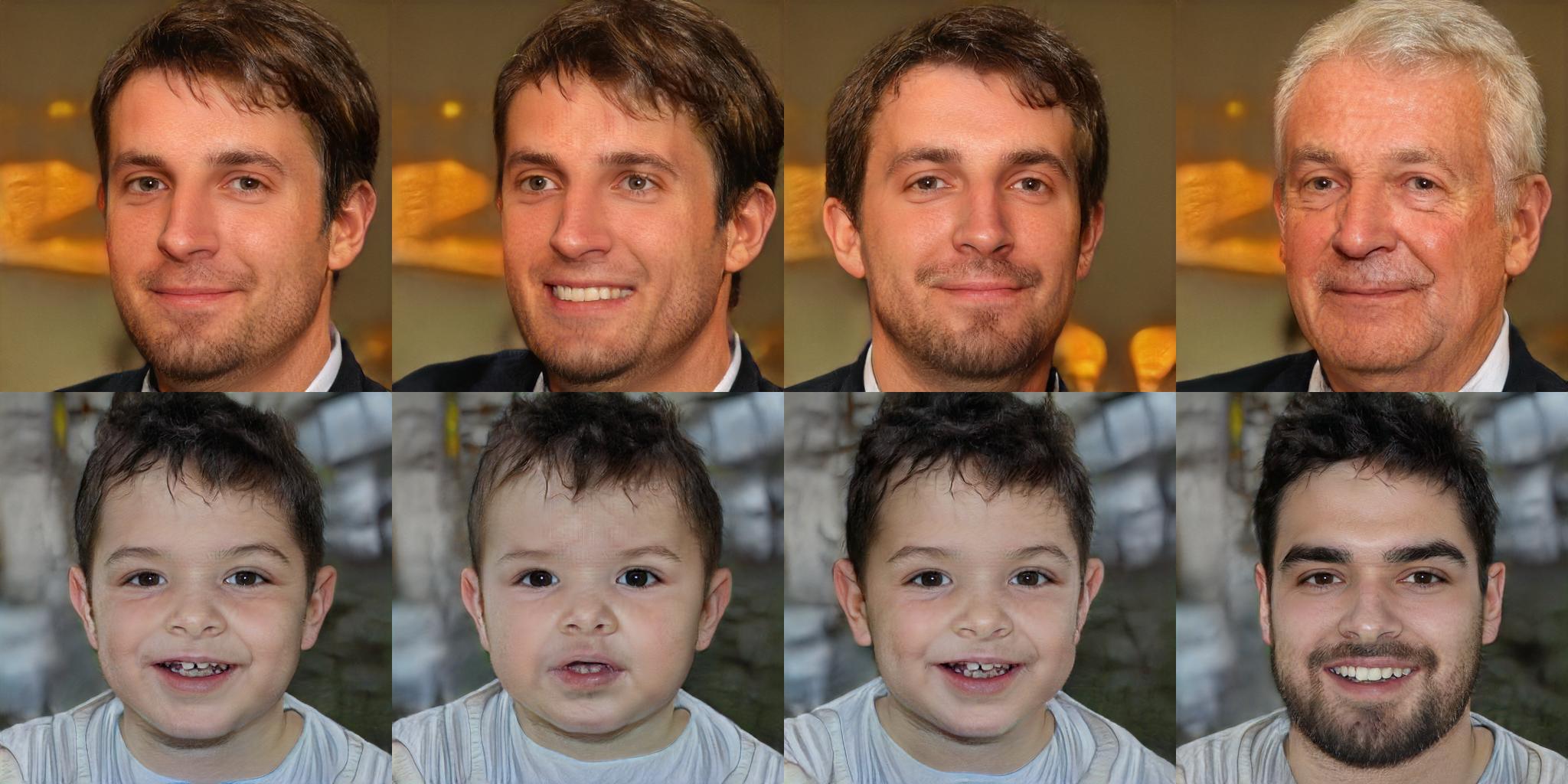}
        \caption{Independent change of high-level attributes}
        \label{fig:control_high_a}
    \end{subfigure}
    \begin{subfigure}{\linewidth}
        \includegraphics[clip,width=\textwidth]{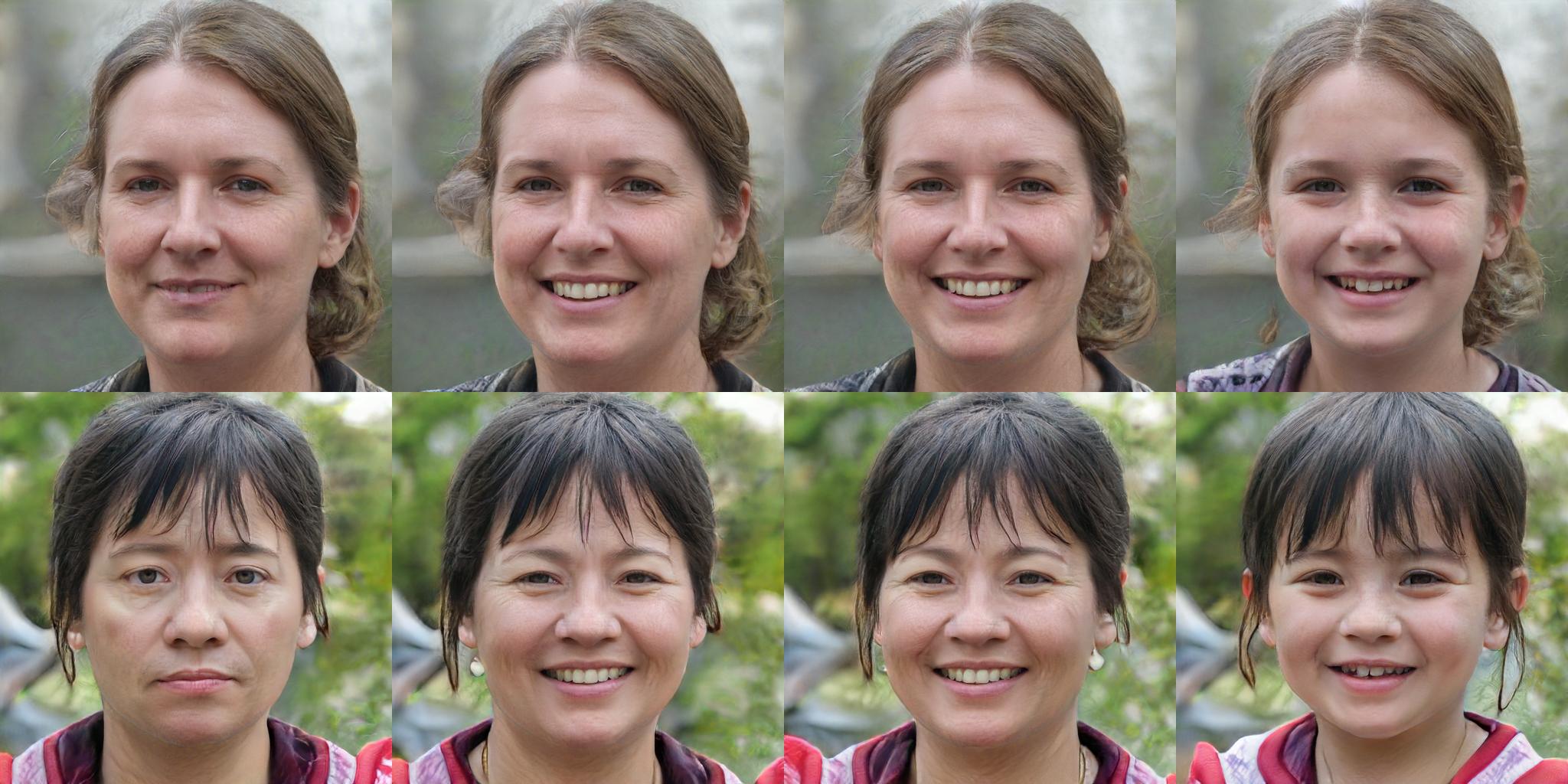}
        \caption{Same as (a), with accumulation of changes from left to right}
        \label{fig:control_high_b}
    \end{subfigure}%
    \caption{High-level attribute control, with changes applied (a) independently, and (b) cumulatively from left to right. The leftmost image is the original, and from left to right we change: \textit{expression, orientation}, and \textit{age}.}
    \label{fig:control_high}
\end{figure}


\begin{figure}[!t]
    \centering
    \begin{tabular}{cccc}
    \multicolumn{2}{l}{
    \includegraphics[width=0.45\linewidth, trim={8 0 5 0},clip]{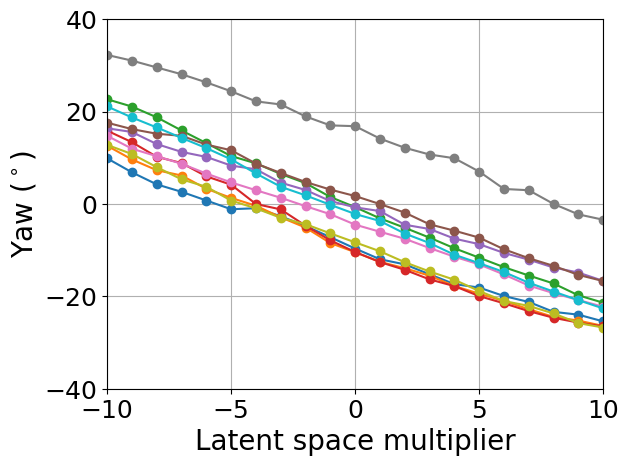}}
     &  
     \multicolumn{2}{l}{
    \includegraphics[width=0.45\linewidth, trim={8 0 5 0},clip]{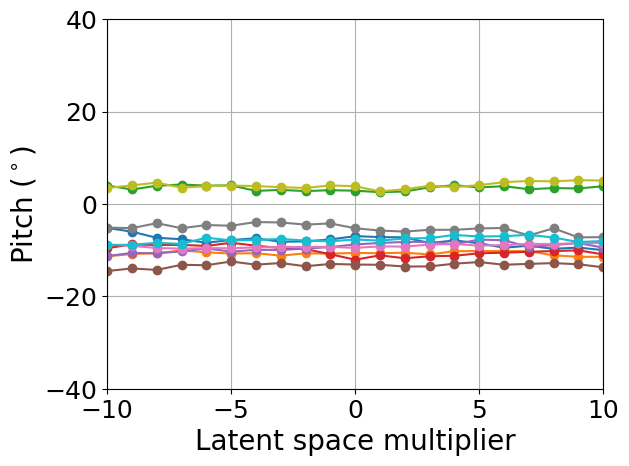}}\\
    \includegraphics[width=0.19\linewidth]{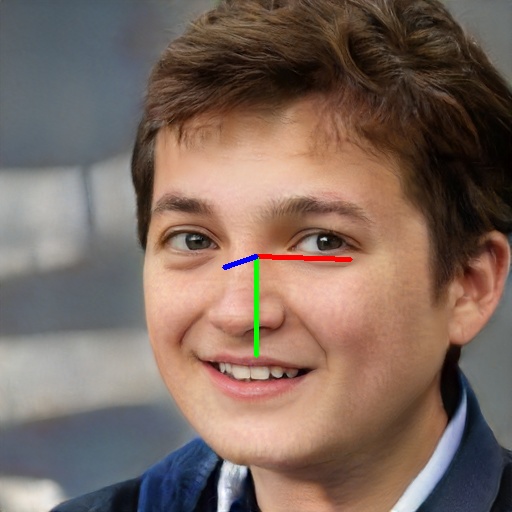} &
    \includegraphics[width=0.19\linewidth]{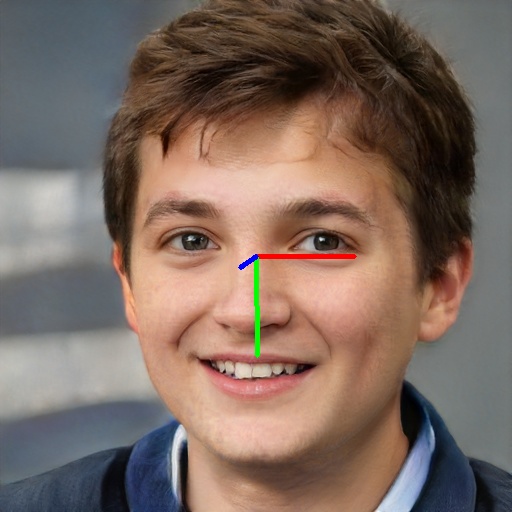} &
    \includegraphics[width=0.19\linewidth]{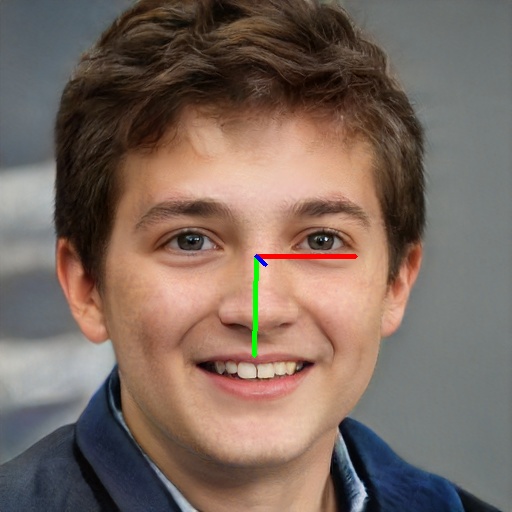} &
    \includegraphics[width=0.19\linewidth]{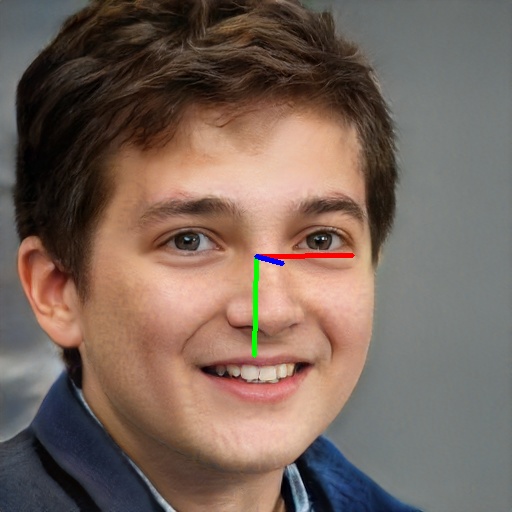}
    \end{tabular}
    \caption{We linearly vary the multiplier (x-axis) that shifts the latent vector in the pose latent space along the yaw direction obtained from PCA and observe with a 6DRepNet~\cite{hempel20226d} the resulting yaw (in the top left plot) and pitch (in the top right plot). We observe that the yaw linearly varies with our linear shift consistently across all images, while the pitch remains stable, proving the high-quality of our high-level attribute disentanglement. The bottom part shows a corresponding sample image with varying yaw.}
    \label{fig:control_yaw_pitch}    
\end{figure}  

\begingroup
\setlength{\tabcolsep}{0.5pt} 
\renewcommand{\arraystretch}{0.7} 
\begin{figure*}[ht]
\centering
\begin{tabular}{ccccccc}
\begin{tabular}{c} \includegraphics[width=0.136\linewidth]{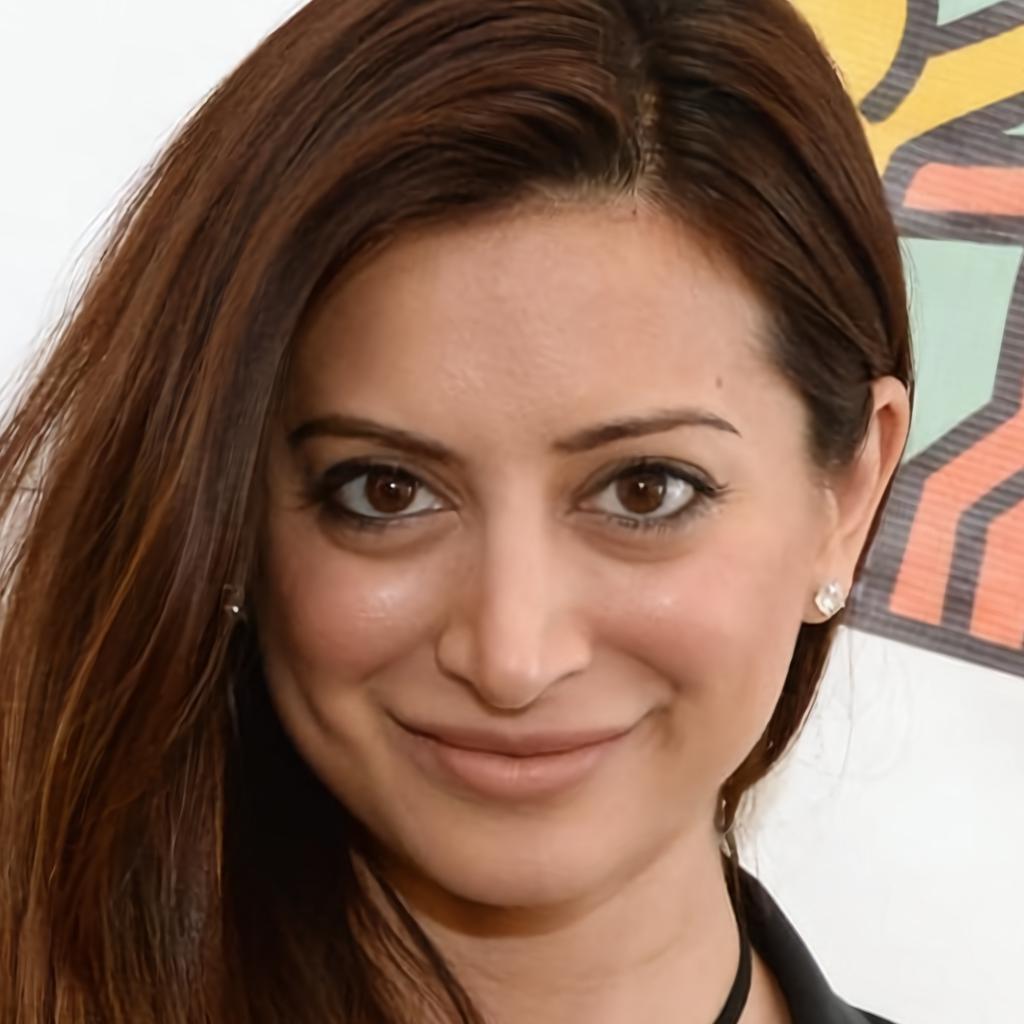} \end{tabular}
&
\begin{tabular}{c} \includegraphics[width=0.136\linewidth]{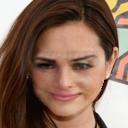} \end{tabular}
&
\begin{tabular}{c} \includegraphics[width=0.136\linewidth]{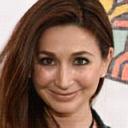} \end{tabular}
&
\begin{tabular}{c} \includegraphics[width=0.136\linewidth]{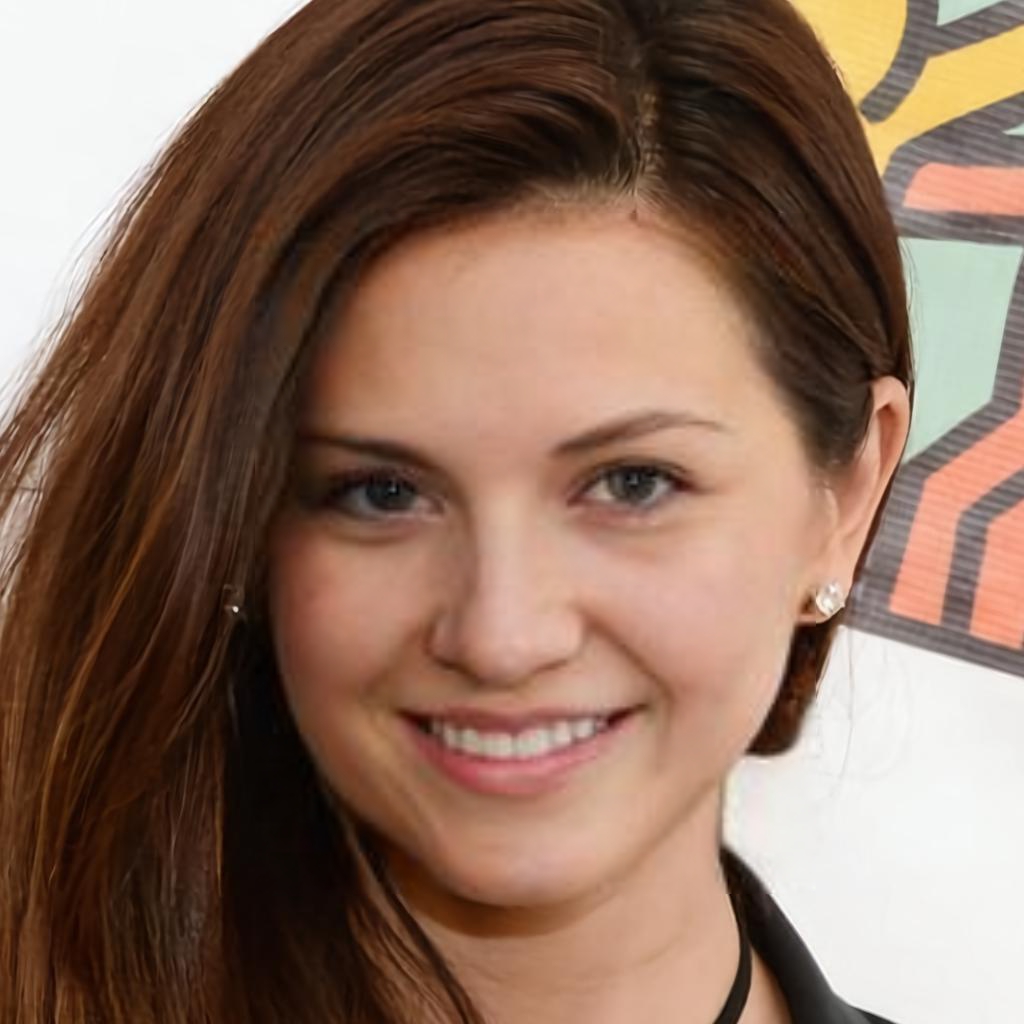} \end{tabular}
&
\begin{tabular}{c} \includegraphics[width=0.136\linewidth]{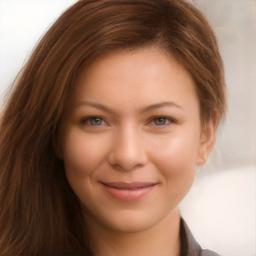} \end{tabular}
&
\begin{tabular}{c} \includegraphics[width=0.136\linewidth]{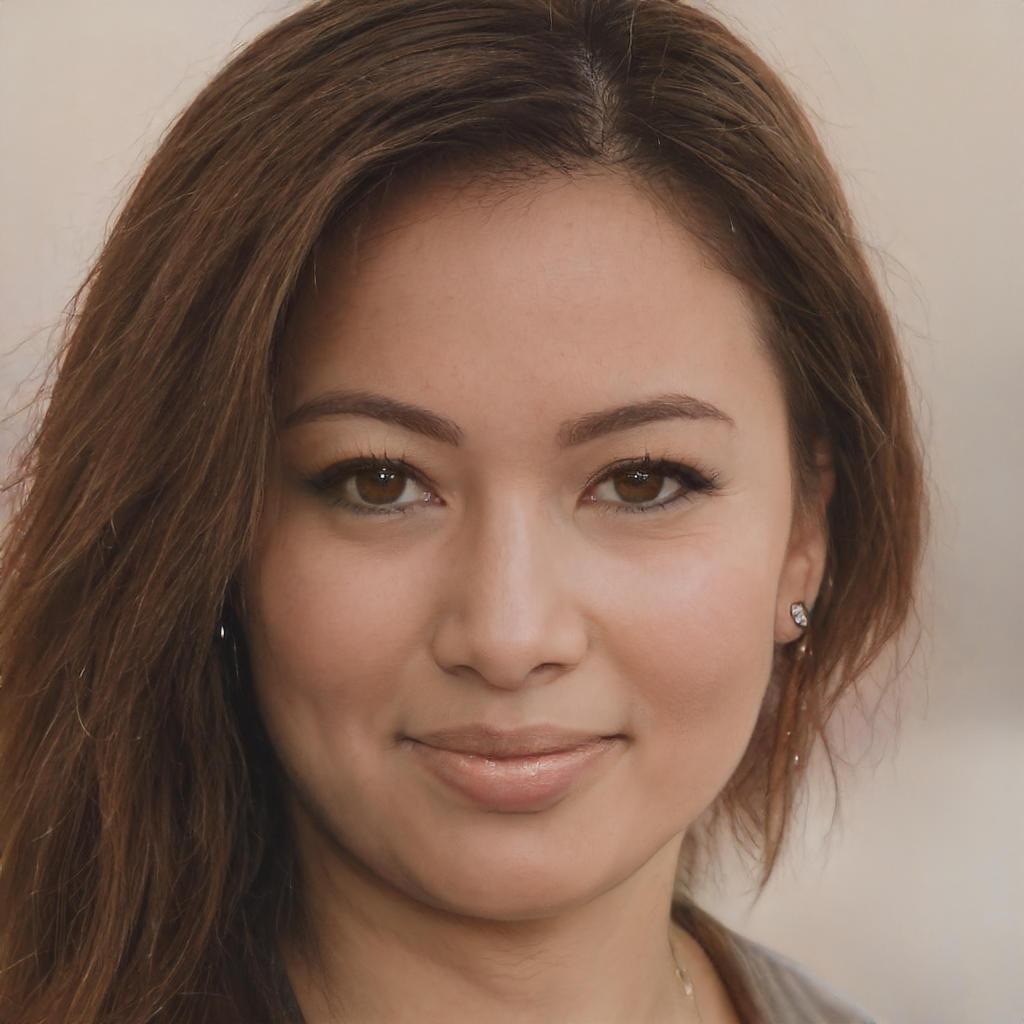} \end{tabular}
&
\begin{tabular}{c} \includegraphics[width=0.136\linewidth]{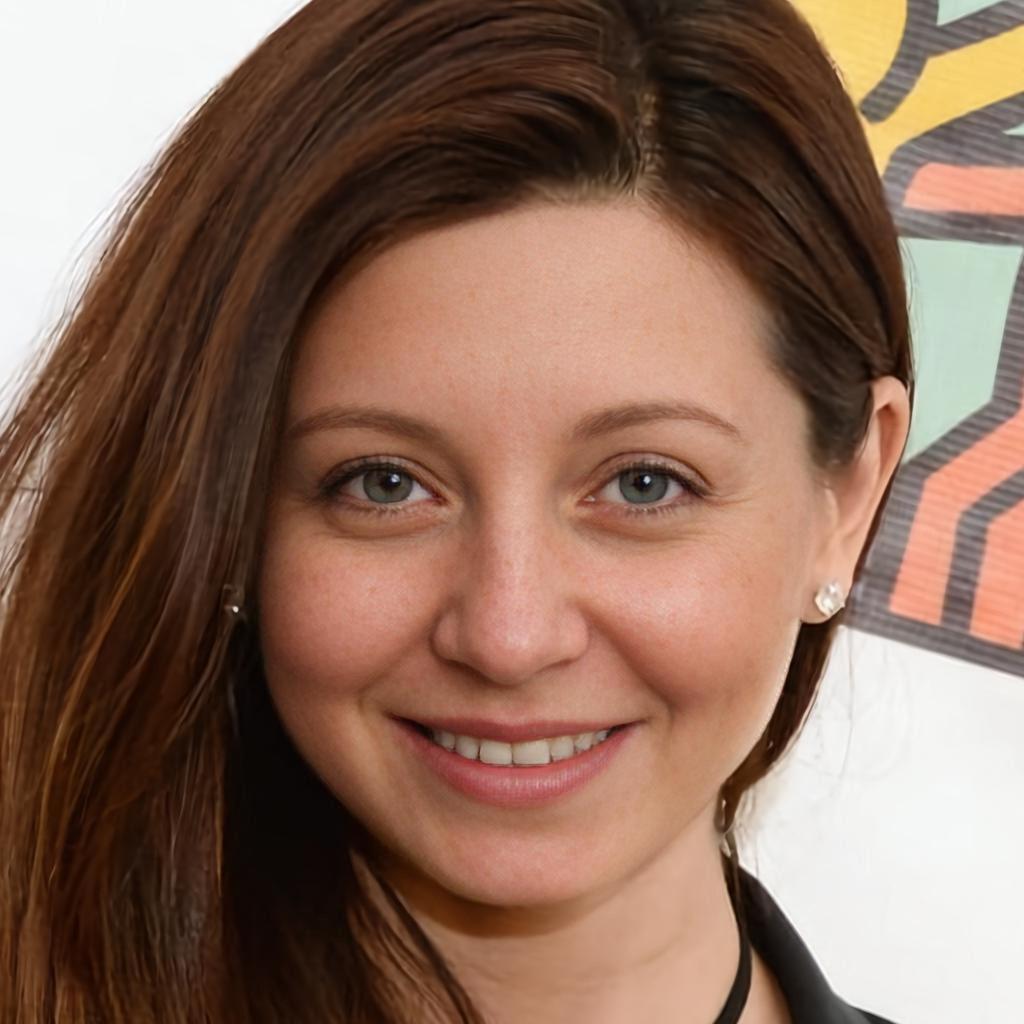} \end{tabular}
\\
\begin{tabular}{c} \includegraphics[width=0.136\linewidth]{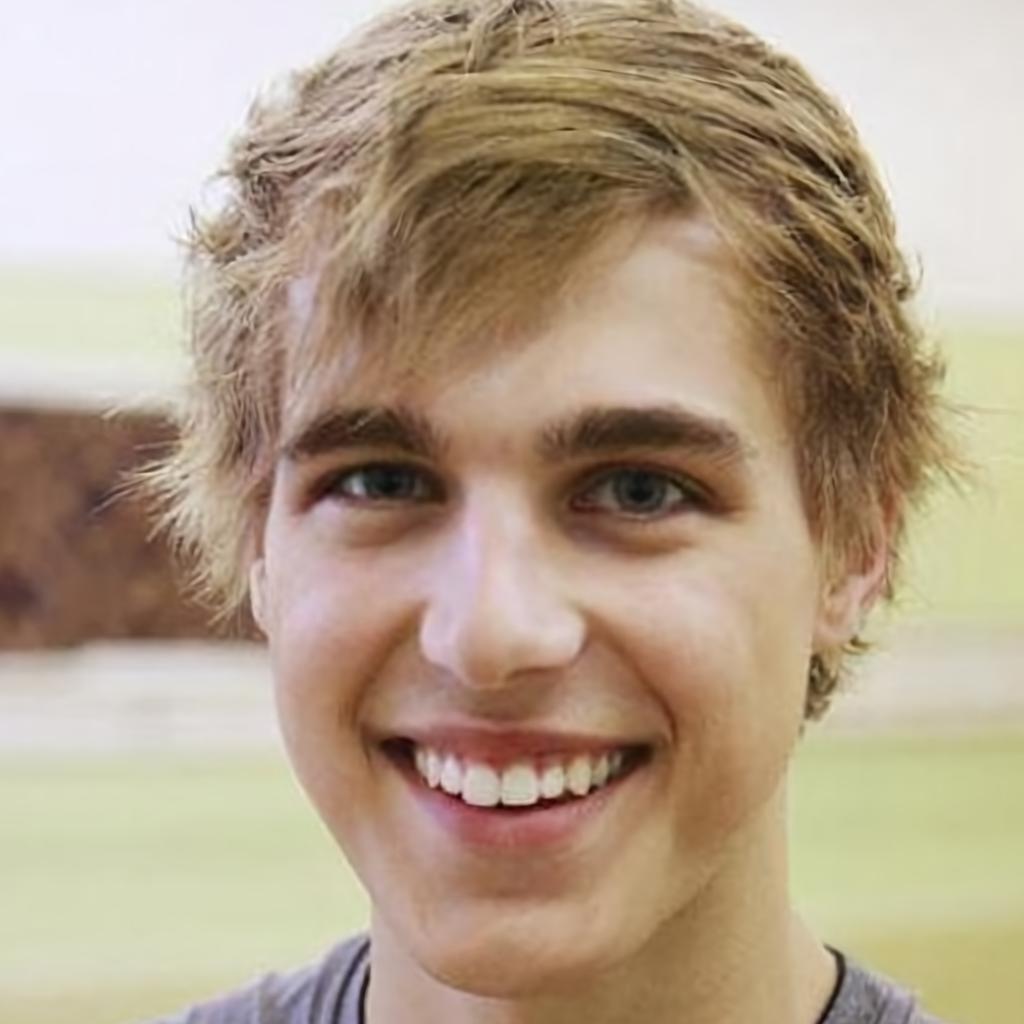} \end{tabular}
&
\begin{tabular}{c} \includegraphics[width=0.136\linewidth]{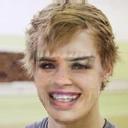} \end{tabular}
&
\begin{tabular}{c} \includegraphics[width=0.136\linewidth]{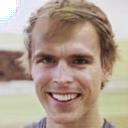} \end{tabular}
&
\begin{tabular}{c} \includegraphics[width=0.136\linewidth]{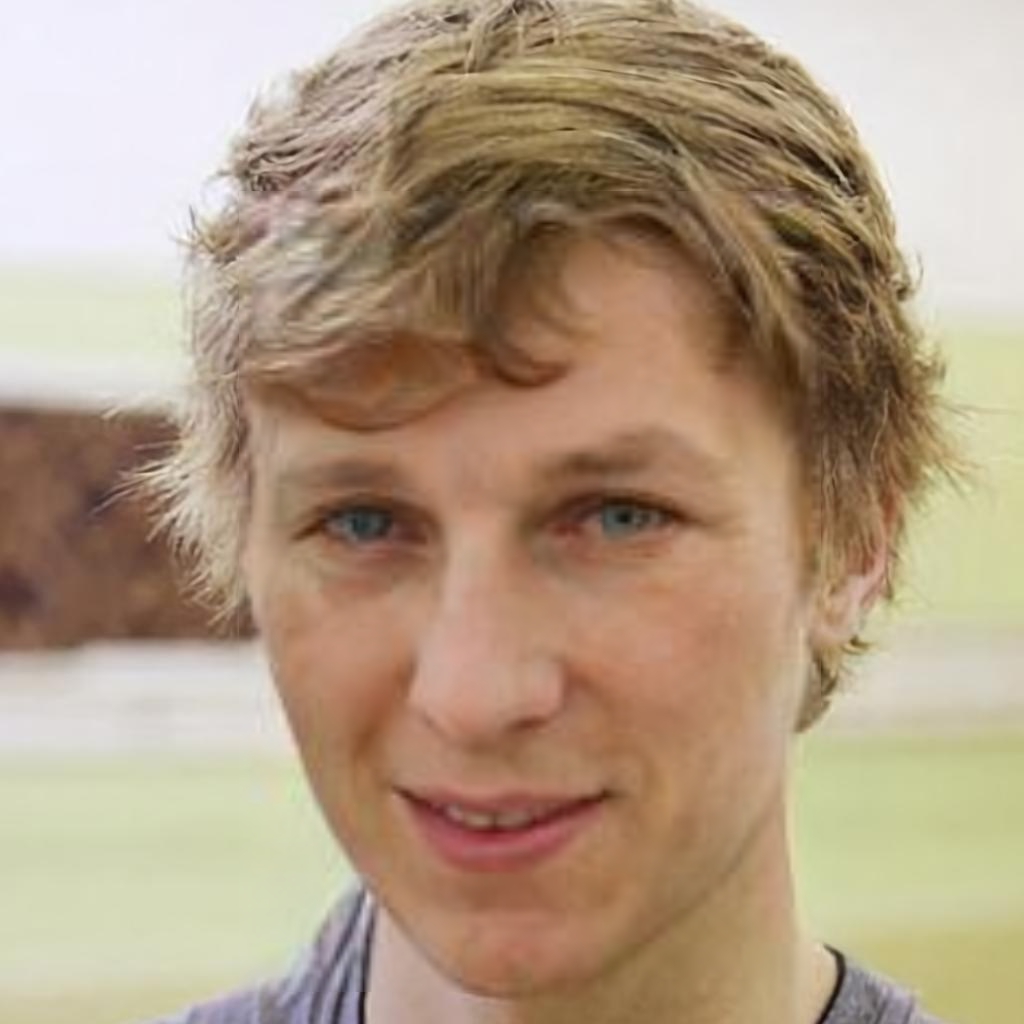} \end{tabular}
&
\begin{tabular}{c} \includegraphics[width=0.136\linewidth]{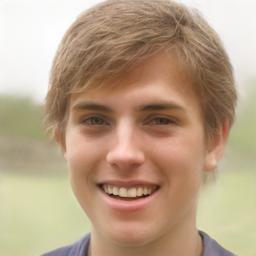} \end{tabular}
&
\begin{tabular}{c} \includegraphics[width=0.136\linewidth]{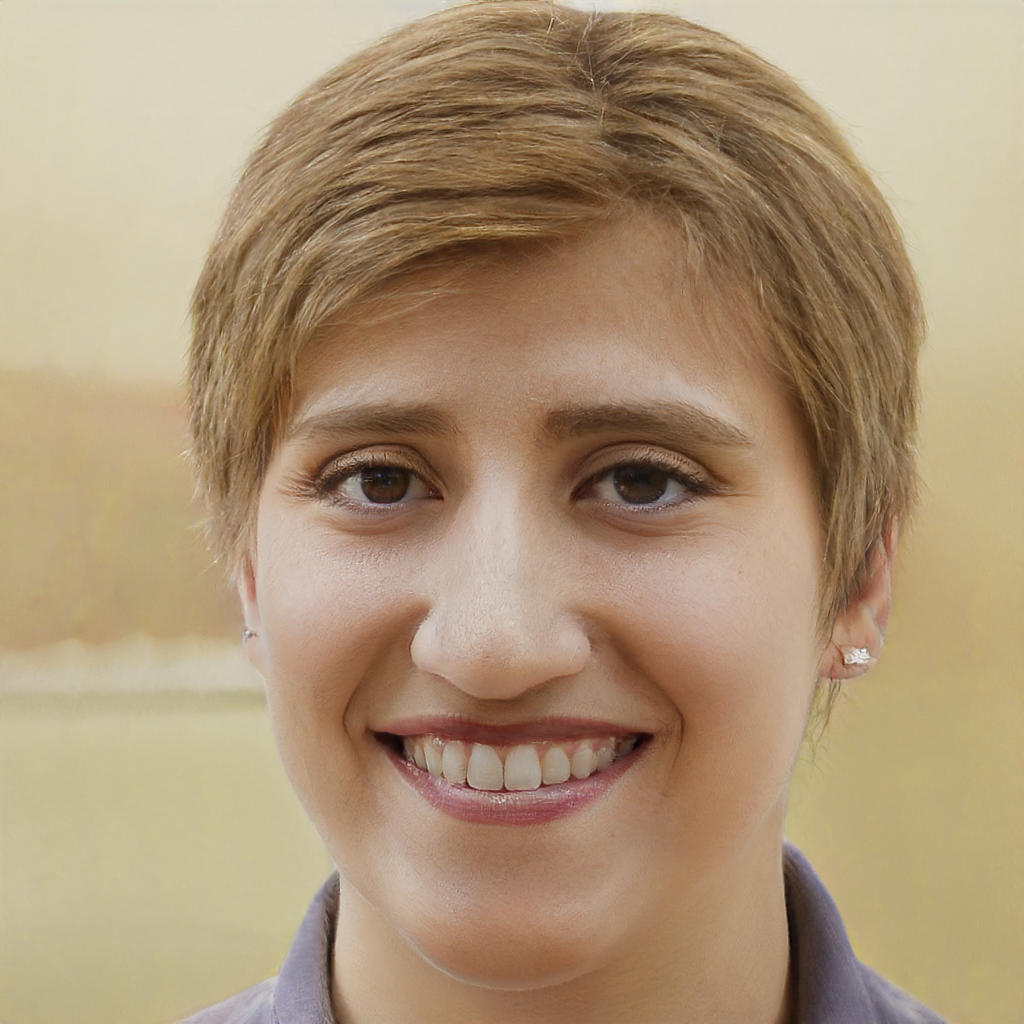} \end{tabular}
&
\begin{tabular}{c} \includegraphics[width=0.136\linewidth]{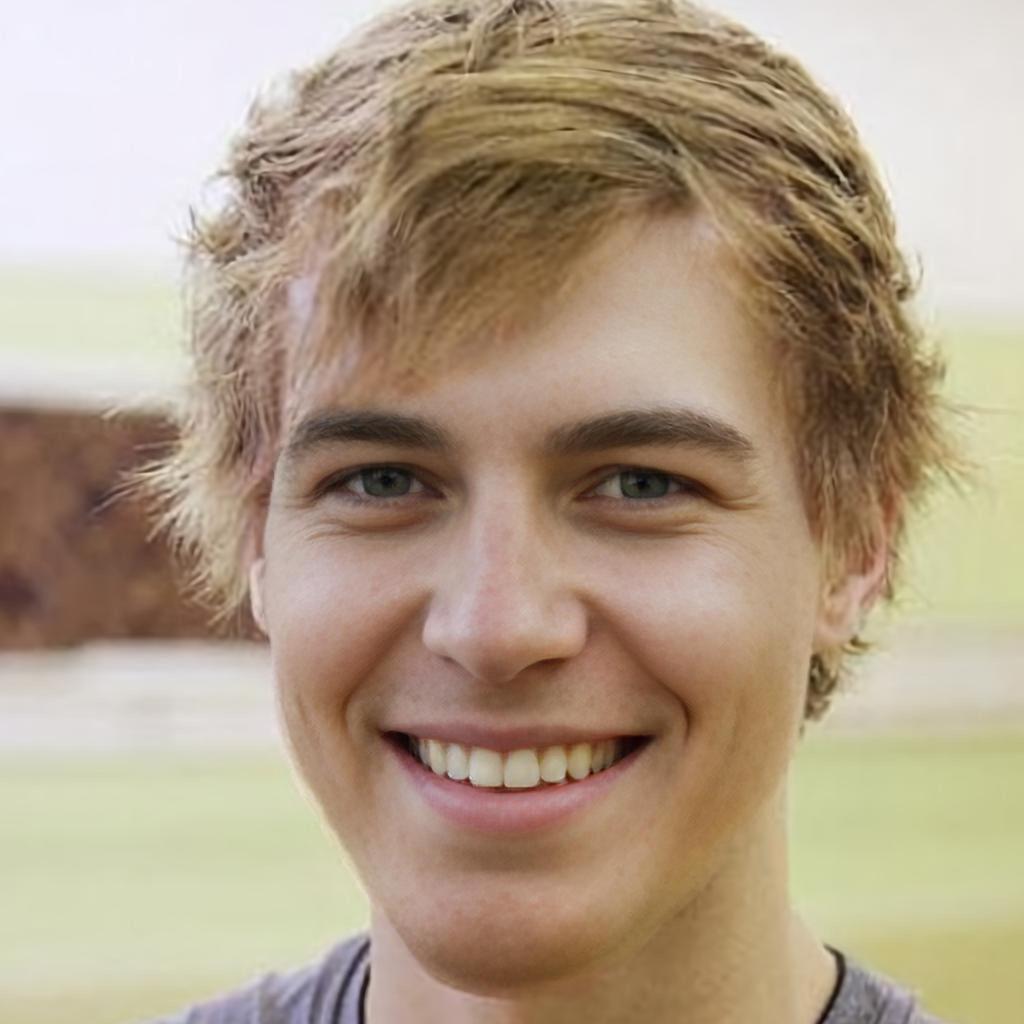} \end{tabular}
\\
\begin{tabular}{c} \includegraphics[width=0.136\linewidth]{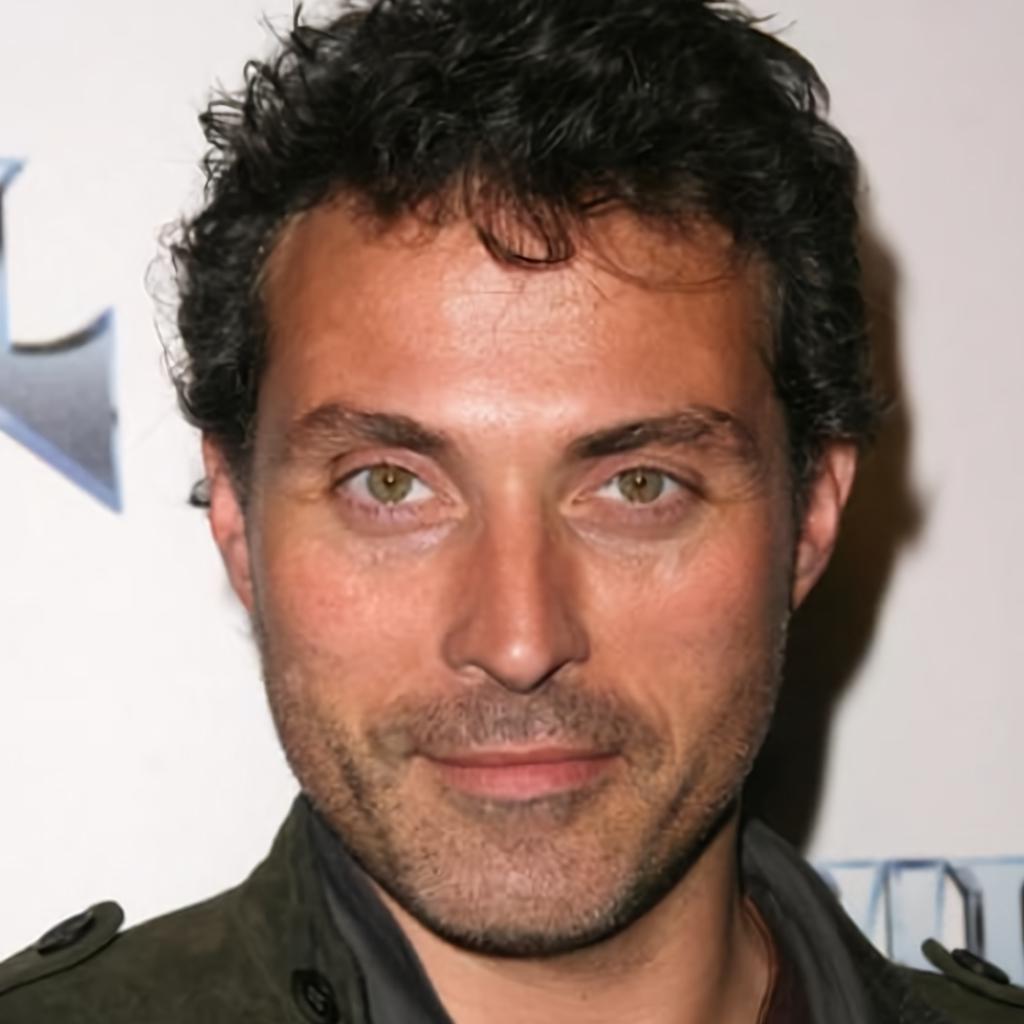} \end{tabular}
&
\begin{tabular}{c} \includegraphics[width=0.136\linewidth]{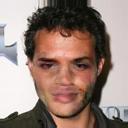} \end{tabular}
&
\begin{tabular}{c} \includegraphics[width=0.136\linewidth]{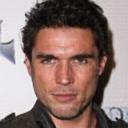} \end{tabular}
&
\begin{tabular}{c} \includegraphics[width=0.136\linewidth]{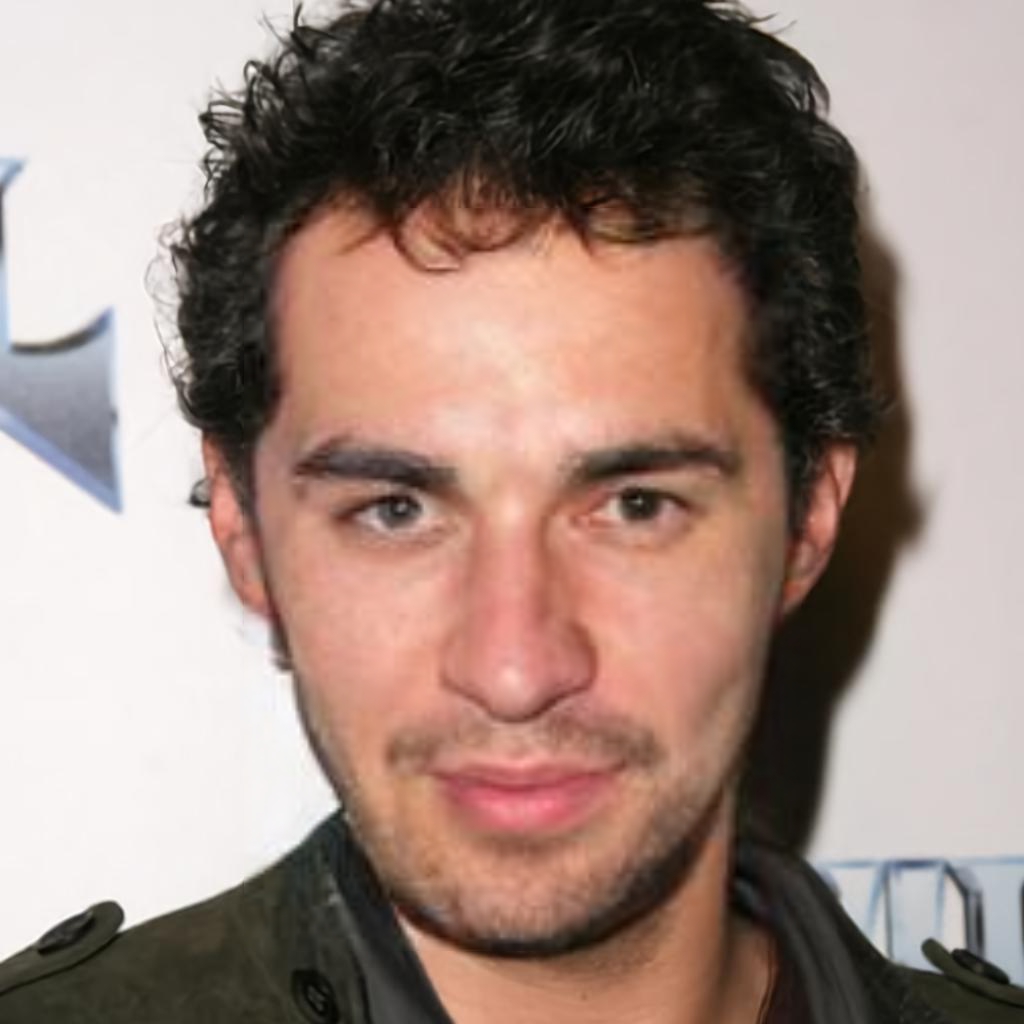} \end{tabular}
&
\begin{tabular}{c} \includegraphics[width=0.136\linewidth]{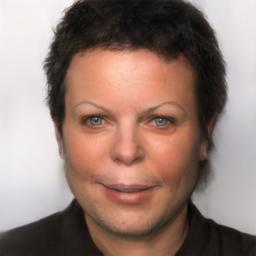} \end{tabular}
&
\begin{tabular}{c} \includegraphics[width=0.136\linewidth]{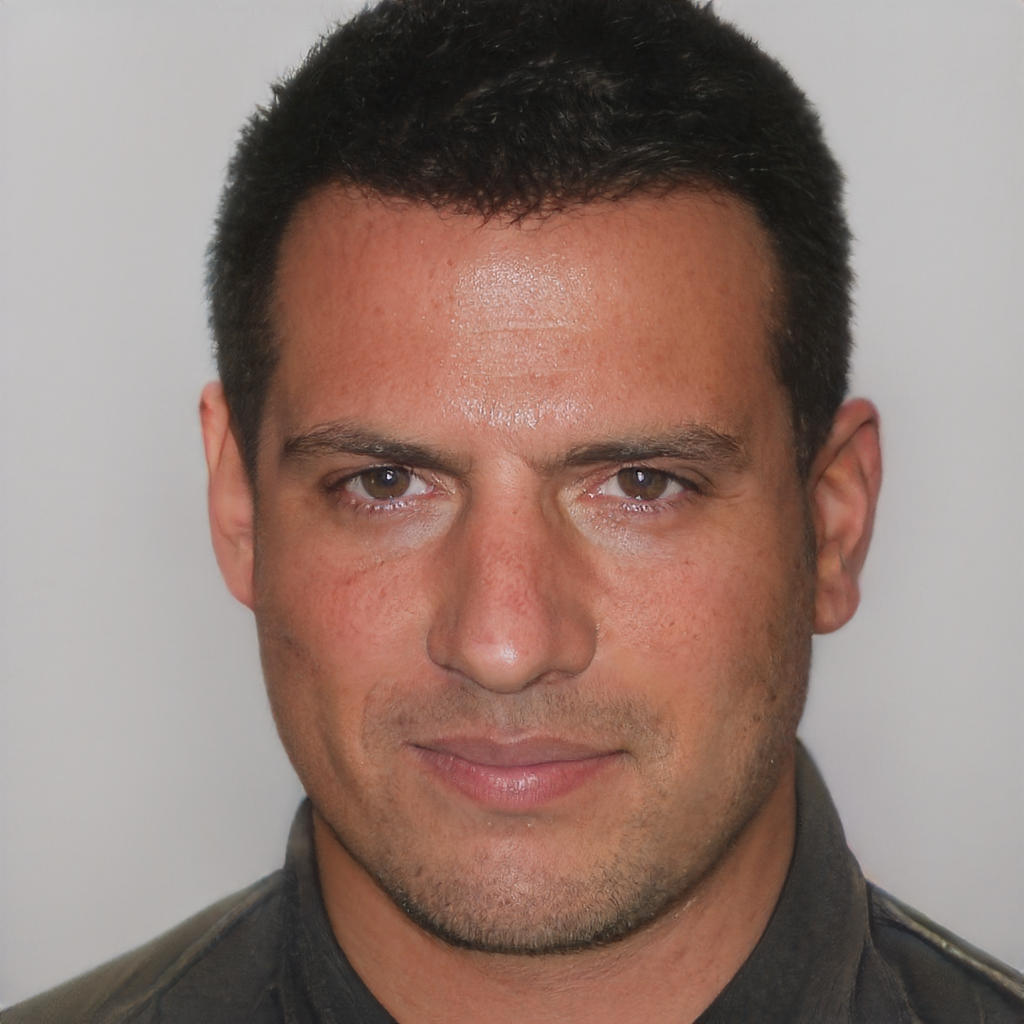} \end{tabular}
&
\begin{tabular}{c} \includegraphics[width=0.136\linewidth]{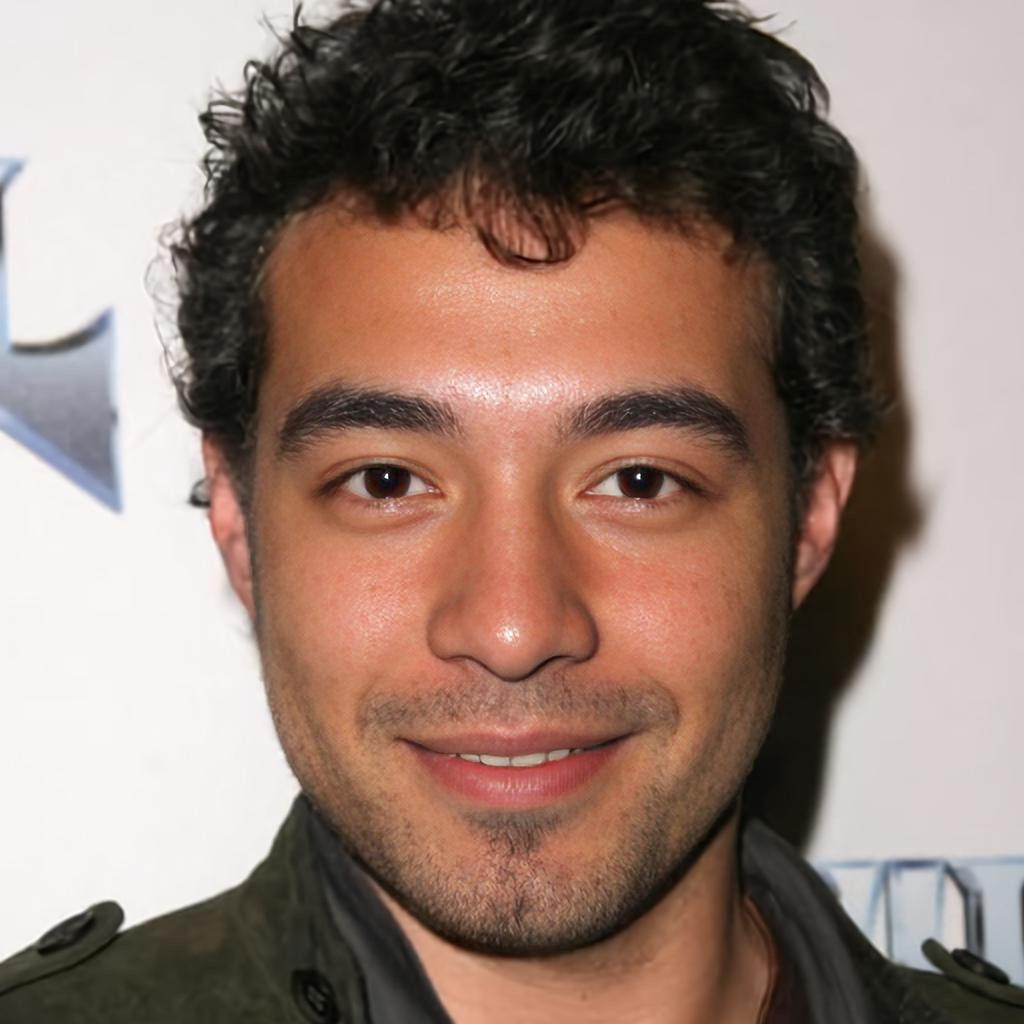} \end{tabular}
\\
\begin{tabular}{c} \includegraphics[width=0.136\linewidth]{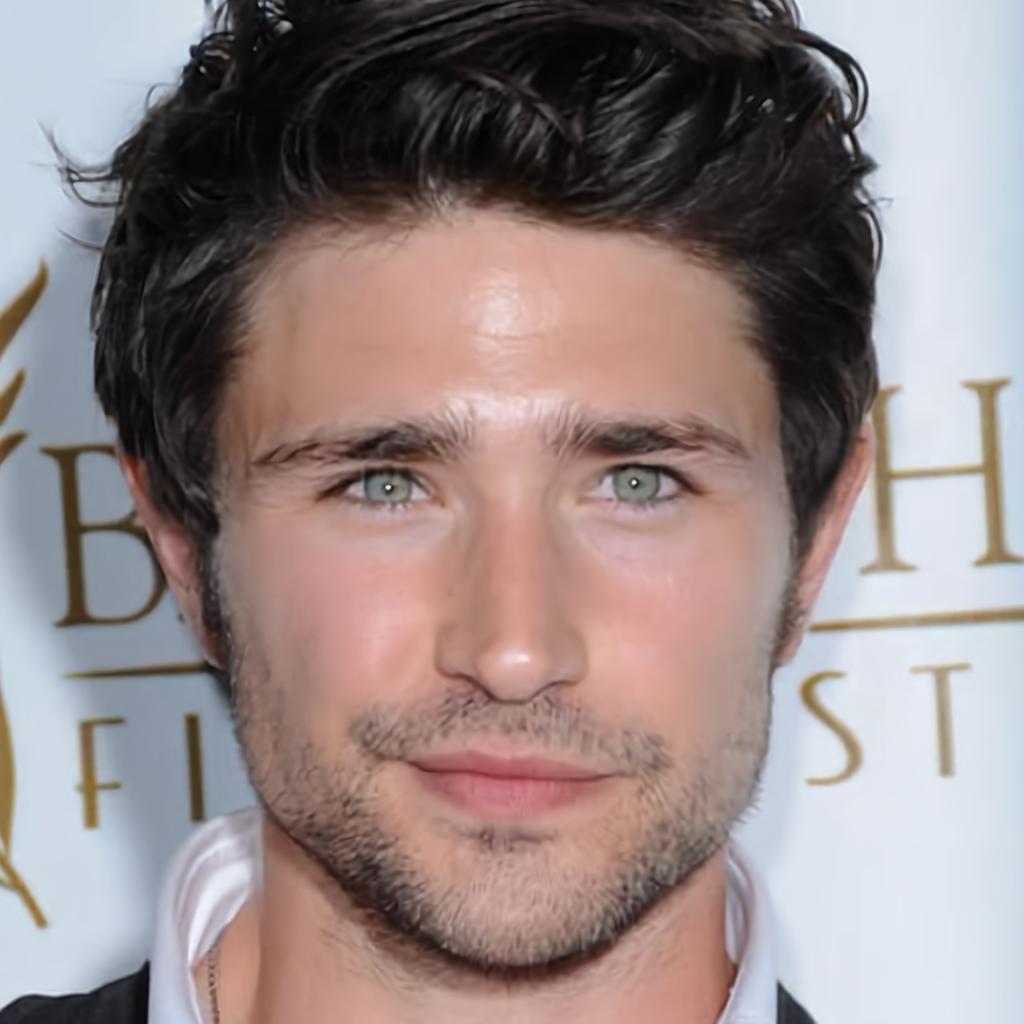} \end{tabular}
&
\begin{tabular}{c} \includegraphics[width=0.136\linewidth]{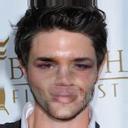} \end{tabular}
&
\begin{tabular}{c} \includegraphics[width=0.136\linewidth]{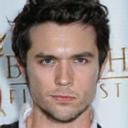} \end{tabular}
&
\begin{tabular}{c} \includegraphics[width=0.136\linewidth]{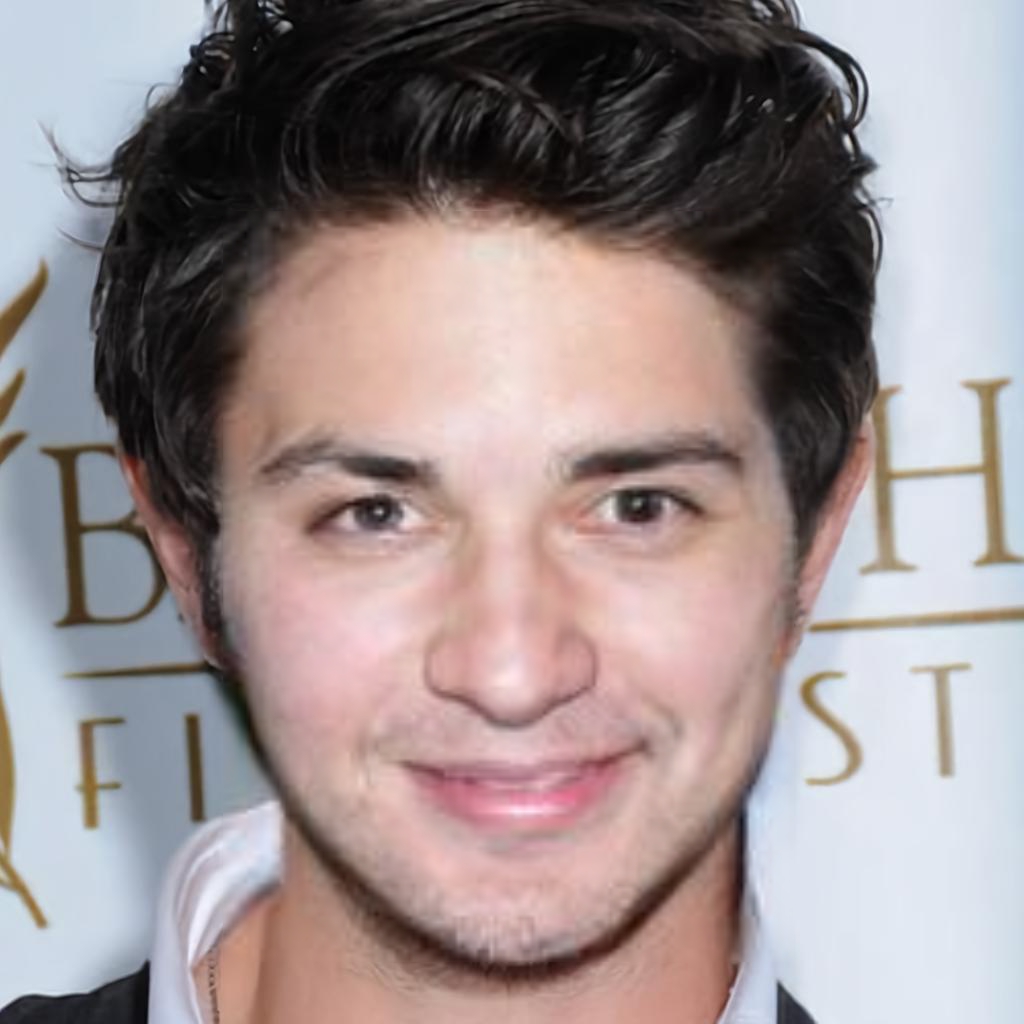} \end{tabular}
&
\begin{tabular}{c} \includegraphics[width=0.136\linewidth]{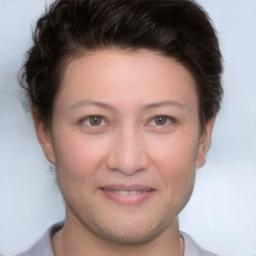} \end{tabular}
&
\begin{tabular}{c} \includegraphics[width=0.136\linewidth]{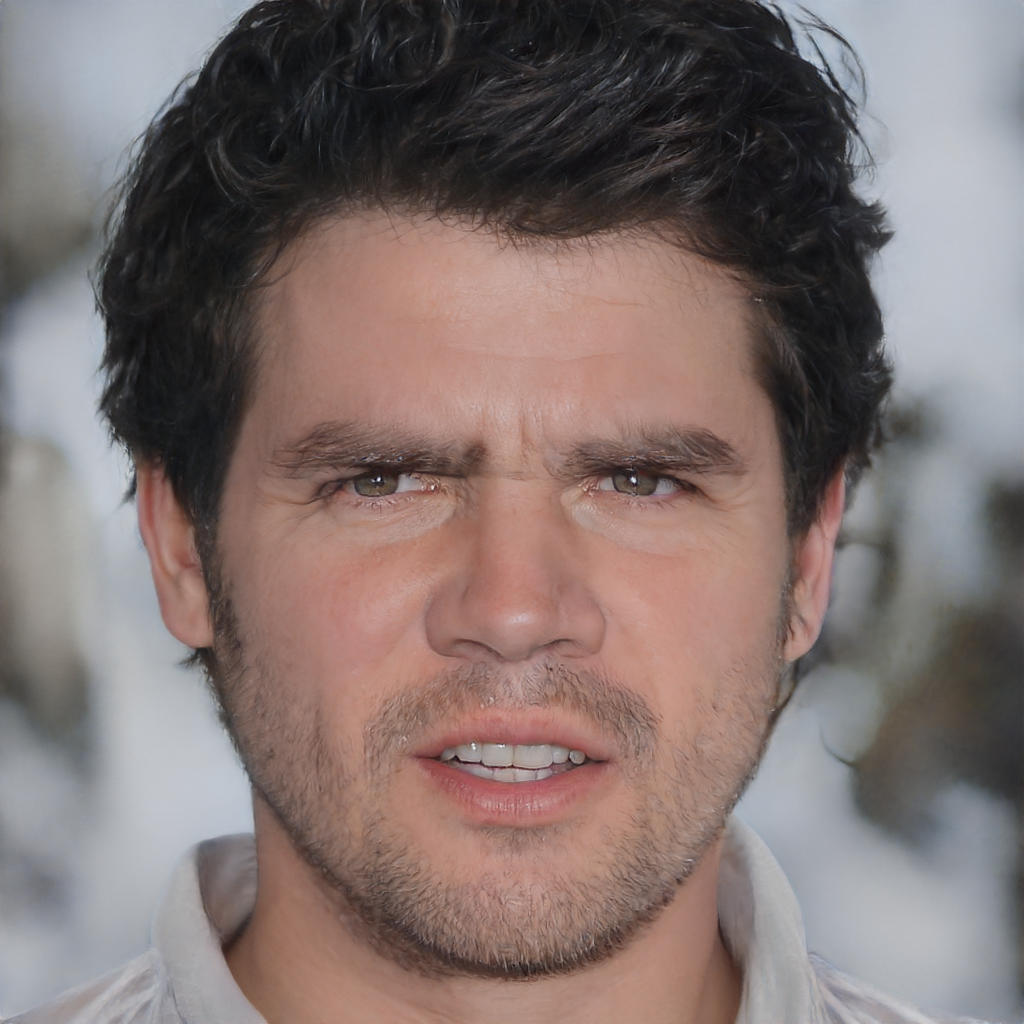} \end{tabular}
&
\begin{tabular}{c} \includegraphics[width=0.136\linewidth]{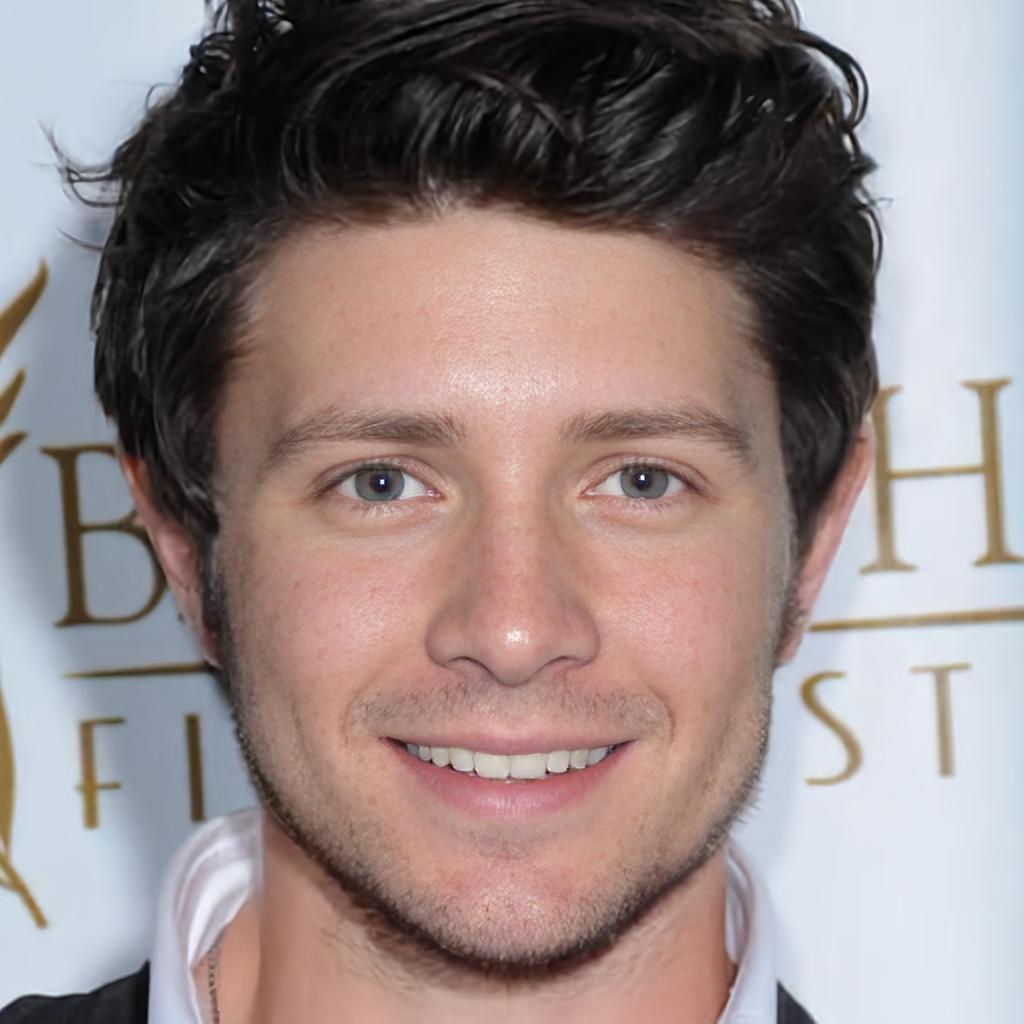} \end{tabular}
\\
\begin{tabular}{c} \includegraphics[width=0.136\linewidth]{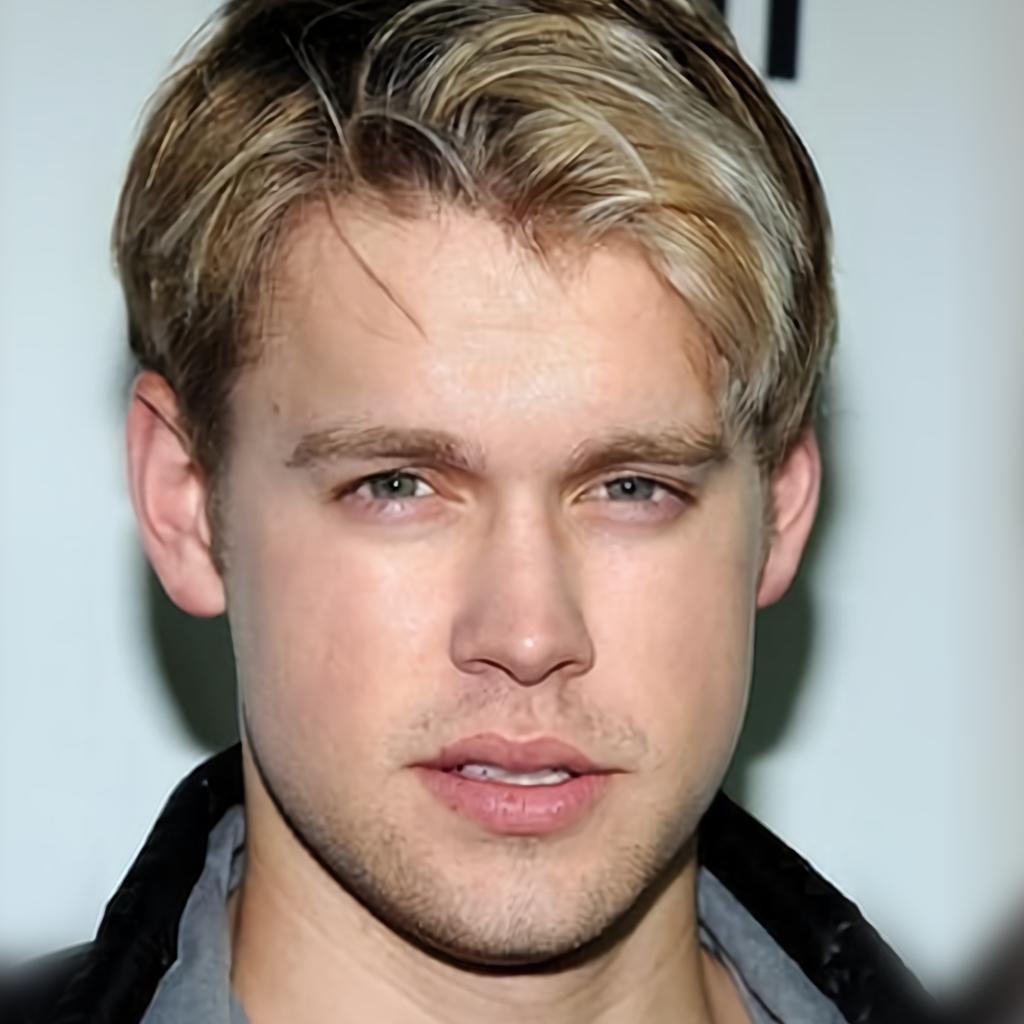} \end{tabular}
&
\begin{tabular}{c} \includegraphics[width=0.136\linewidth]{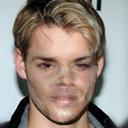} \end{tabular}
&
\begin{tabular}{c} \includegraphics[width=0.136\linewidth]{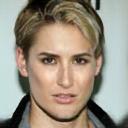} \end{tabular}
&
\begin{tabular}{c} \includegraphics[width=0.136\linewidth]{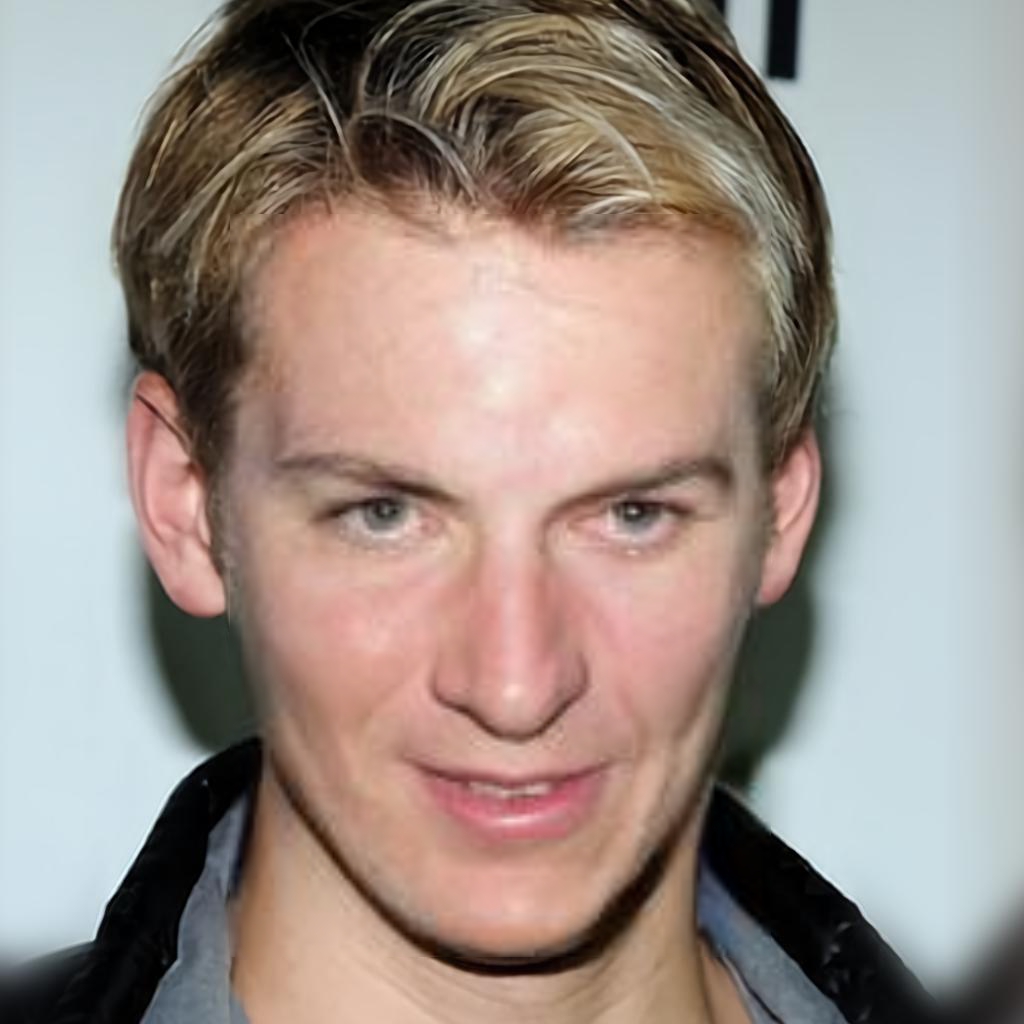} \end{tabular}
&
\begin{tabular}{c} \includegraphics[width=0.136\linewidth]{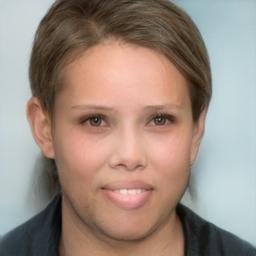} \end{tabular}
&
\begin{tabular}{c} \includegraphics[width=0.136\linewidth]{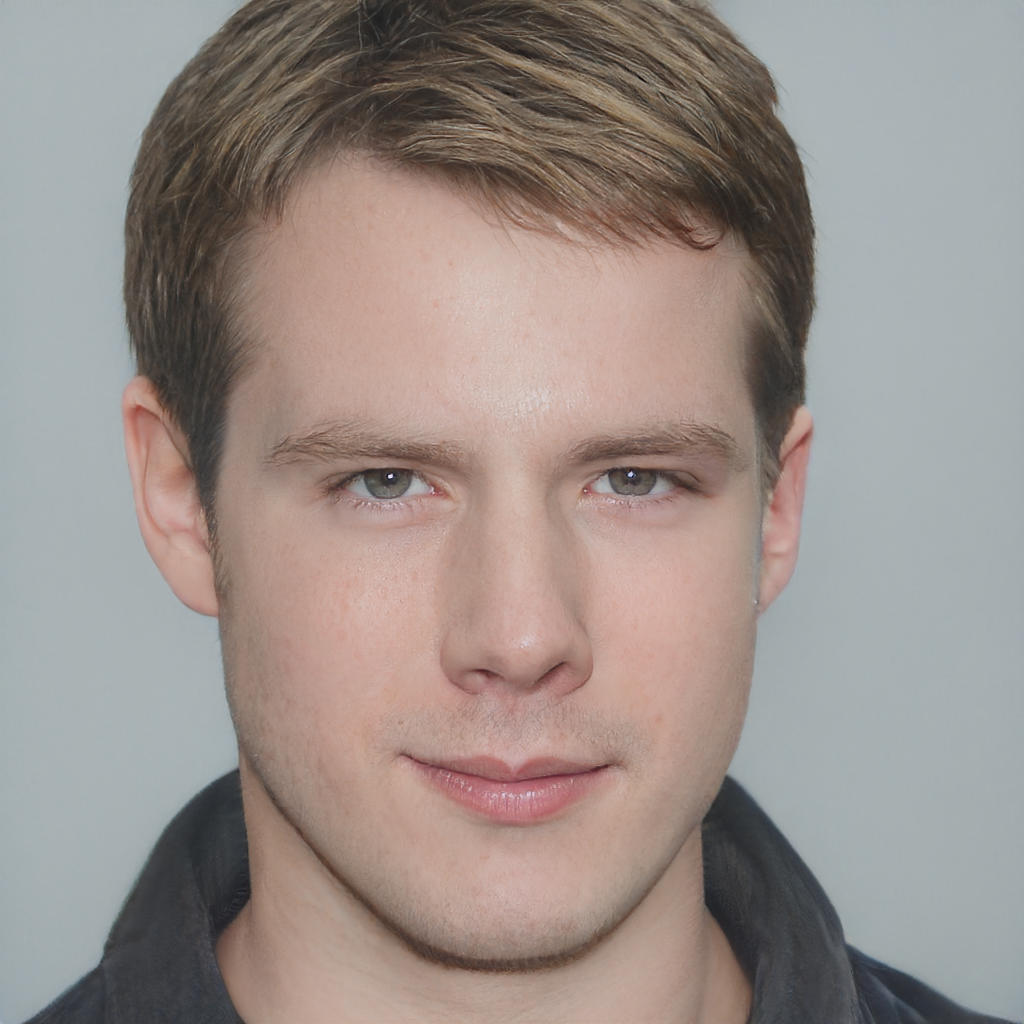} \end{tabular}
&
\begin{tabular}{c} \includegraphics[width=0.136\linewidth]{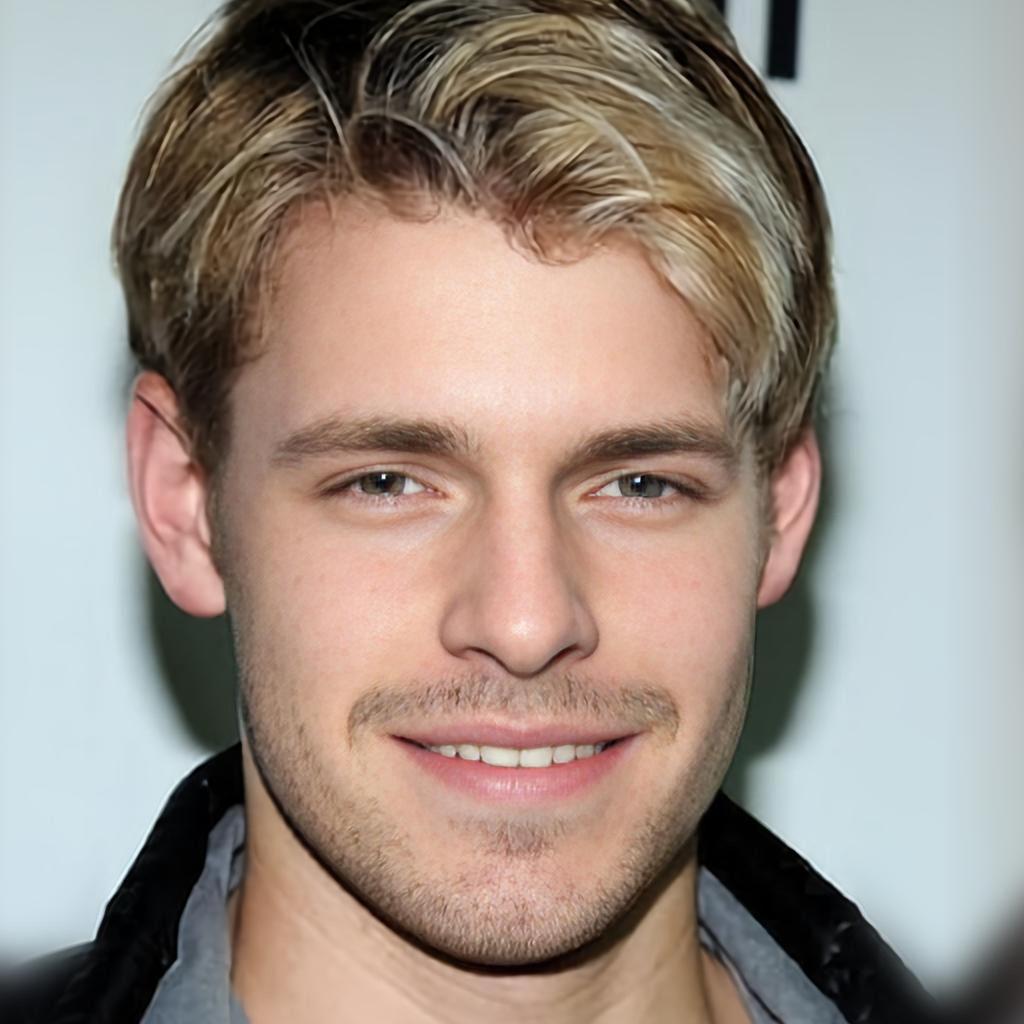} \end{tabular}
\\
\begin{tabular}{c} \includegraphics[width=0.136\linewidth]{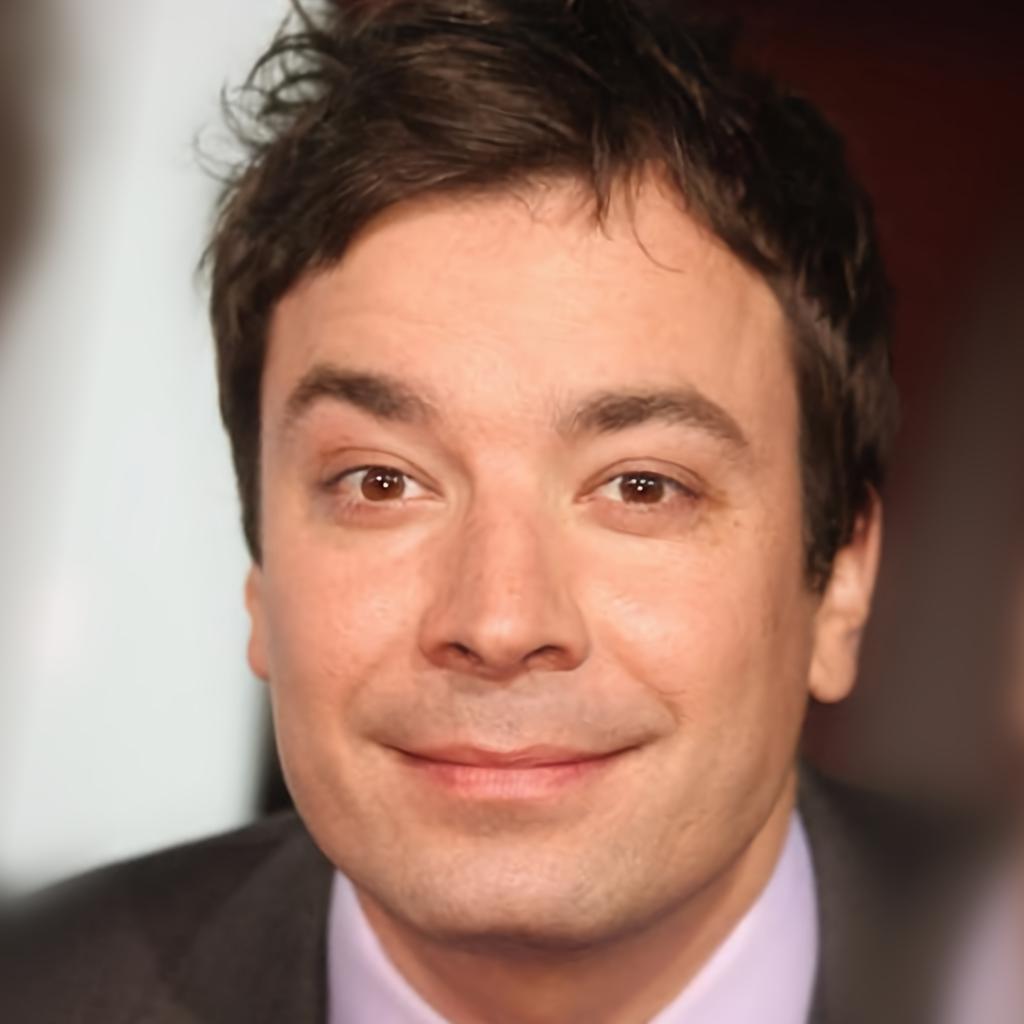} \end{tabular}
&
\begin{tabular}{c} \includegraphics[width=0.136\linewidth]{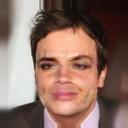} \end{tabular}
&
\begin{tabular}{c} \includegraphics[width=0.136\linewidth]{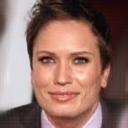} \end{tabular}
&
\begin{tabular}{c} \includegraphics[width=0.136\linewidth]{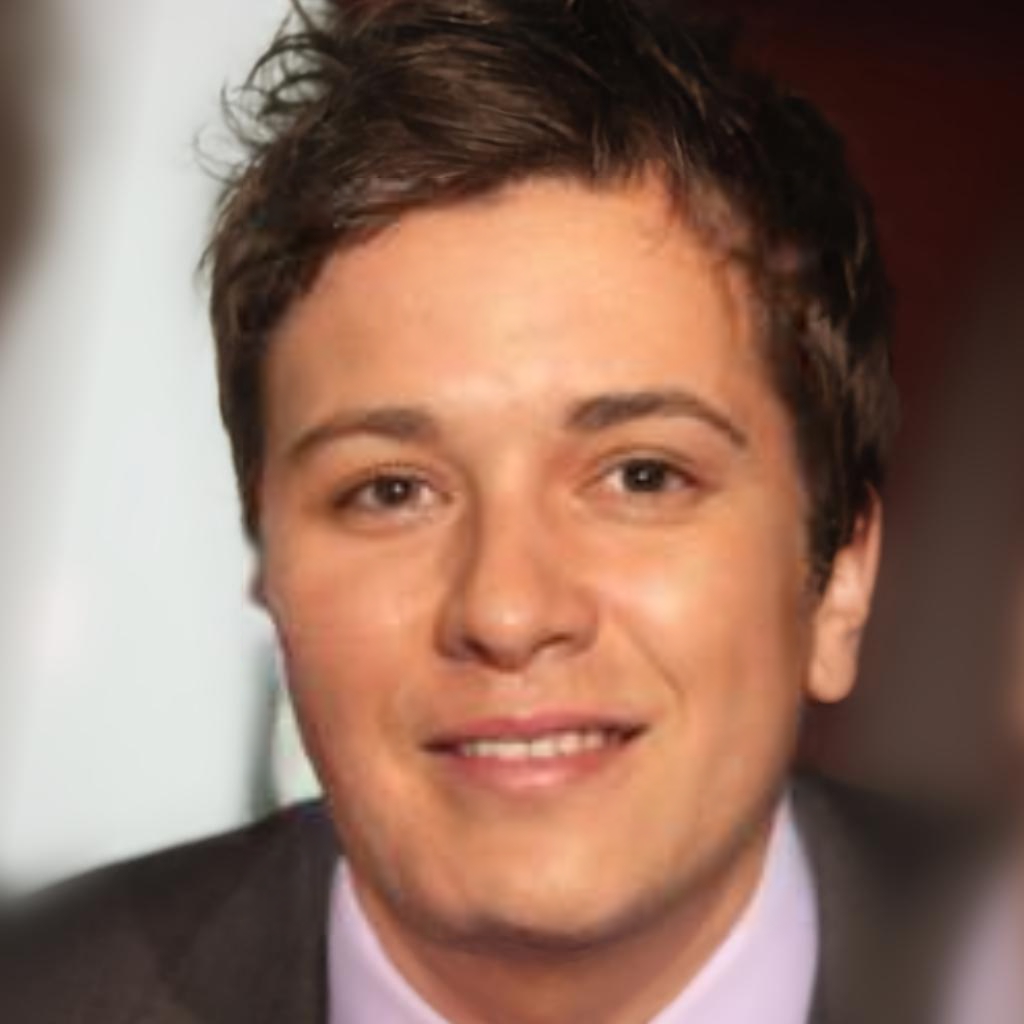} \end{tabular}
&
\begin{tabular}{c} \includegraphics[width=0.136\linewidth]{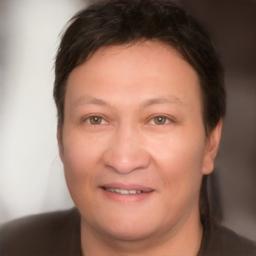} \end{tabular}
&
\begin{tabular}{c} \includegraphics[width=0.136\linewidth]{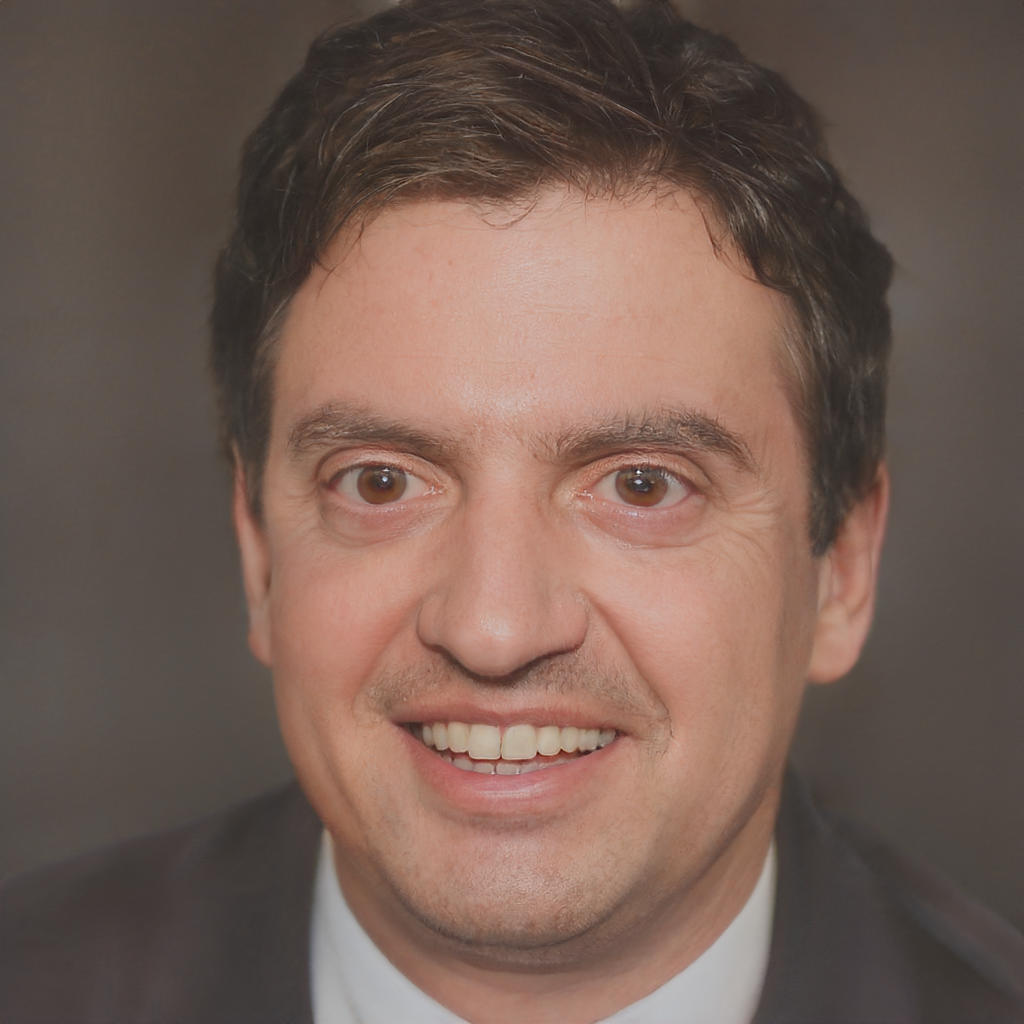} \end{tabular}
&
\begin{tabular}{c} \includegraphics[width=0.136\linewidth]{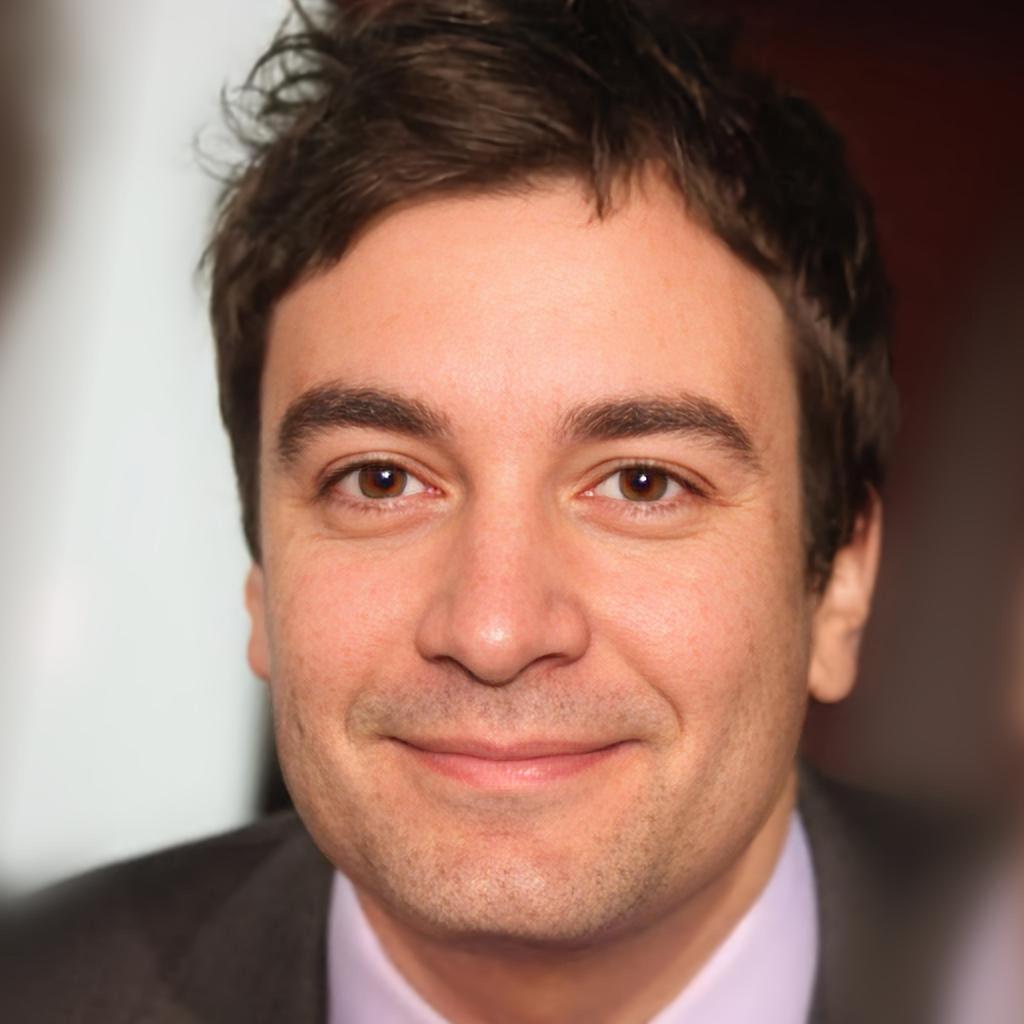} \end{tabular}
\\
\begin{tabular}{c} \includegraphics[width=0.136\linewidth]{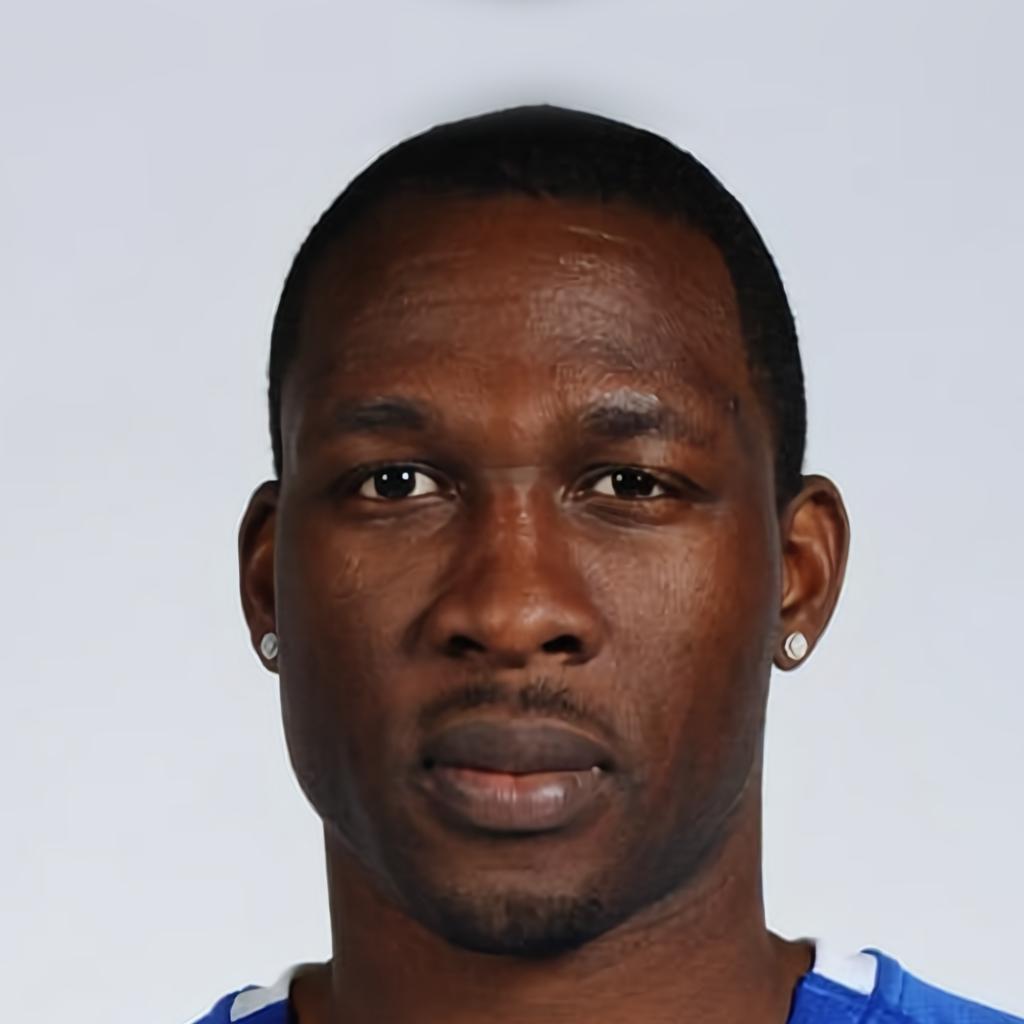} \end{tabular}
&
\begin{tabular}{c} \includegraphics[width=0.136\linewidth]{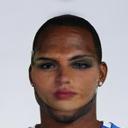} \end{tabular}
&
\begin{tabular}{c} \includegraphics[width=0.136\linewidth]{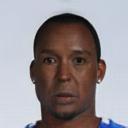} \end{tabular}
&
\begin{tabular}{c} \includegraphics[width=0.136\linewidth]{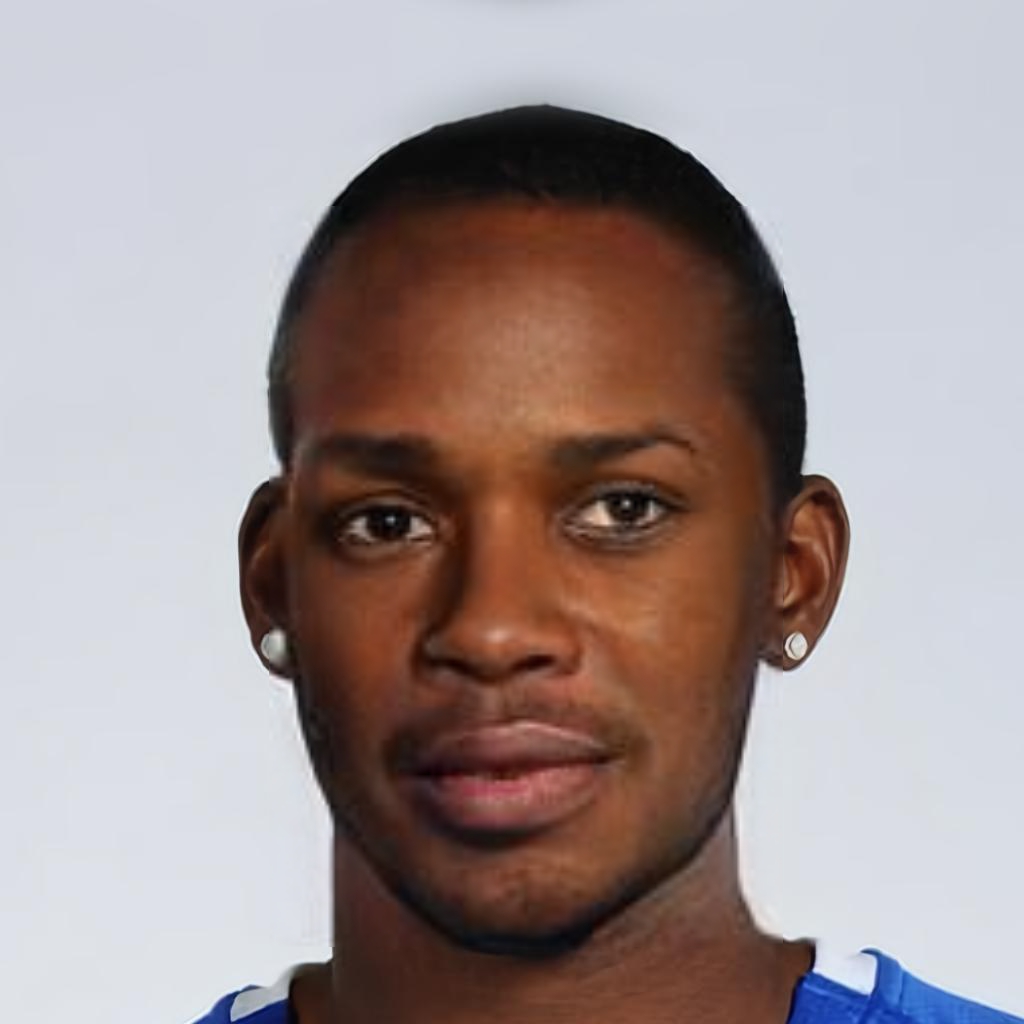} \end{tabular}
&
\begin{tabular}{c} \includegraphics[width=0.136\linewidth]{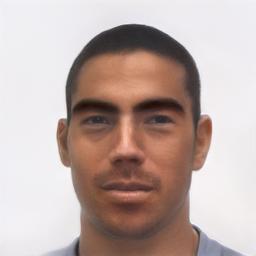} \end{tabular}
&
\begin{tabular}{c} \includegraphics[width=0.136\linewidth]{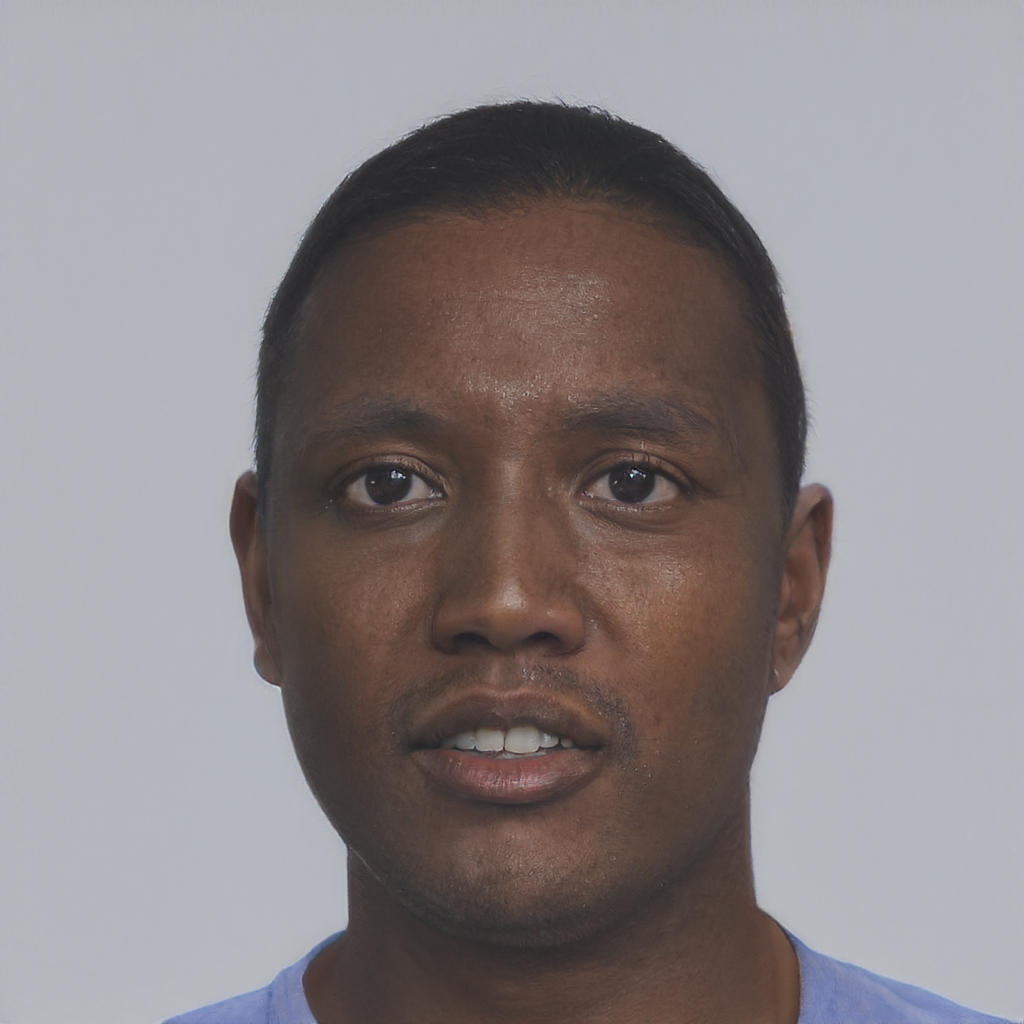} \end{tabular}
&
\begin{tabular}{c} \includegraphics[width=0.136\linewidth]{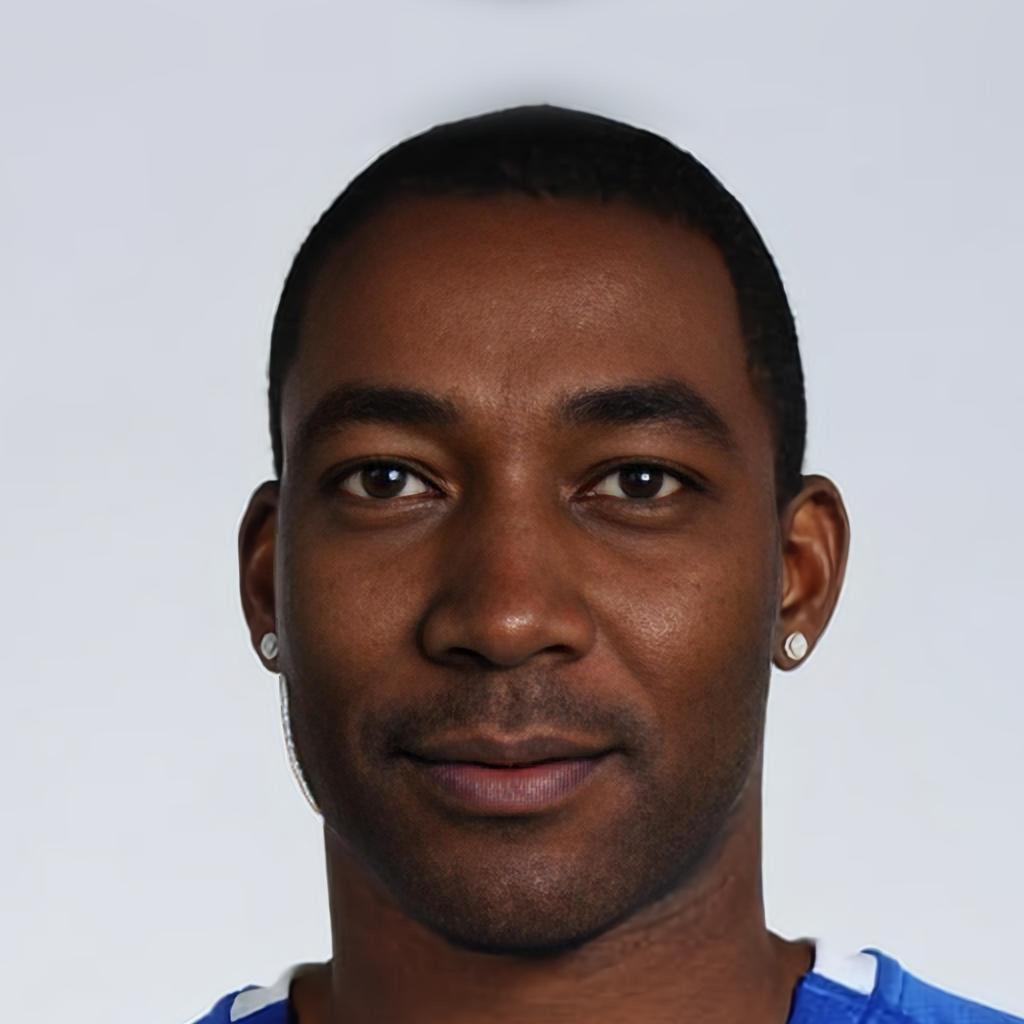} \end{tabular}
\\
\begin{tabular}{c} \includegraphics[width=0.136\linewidth]{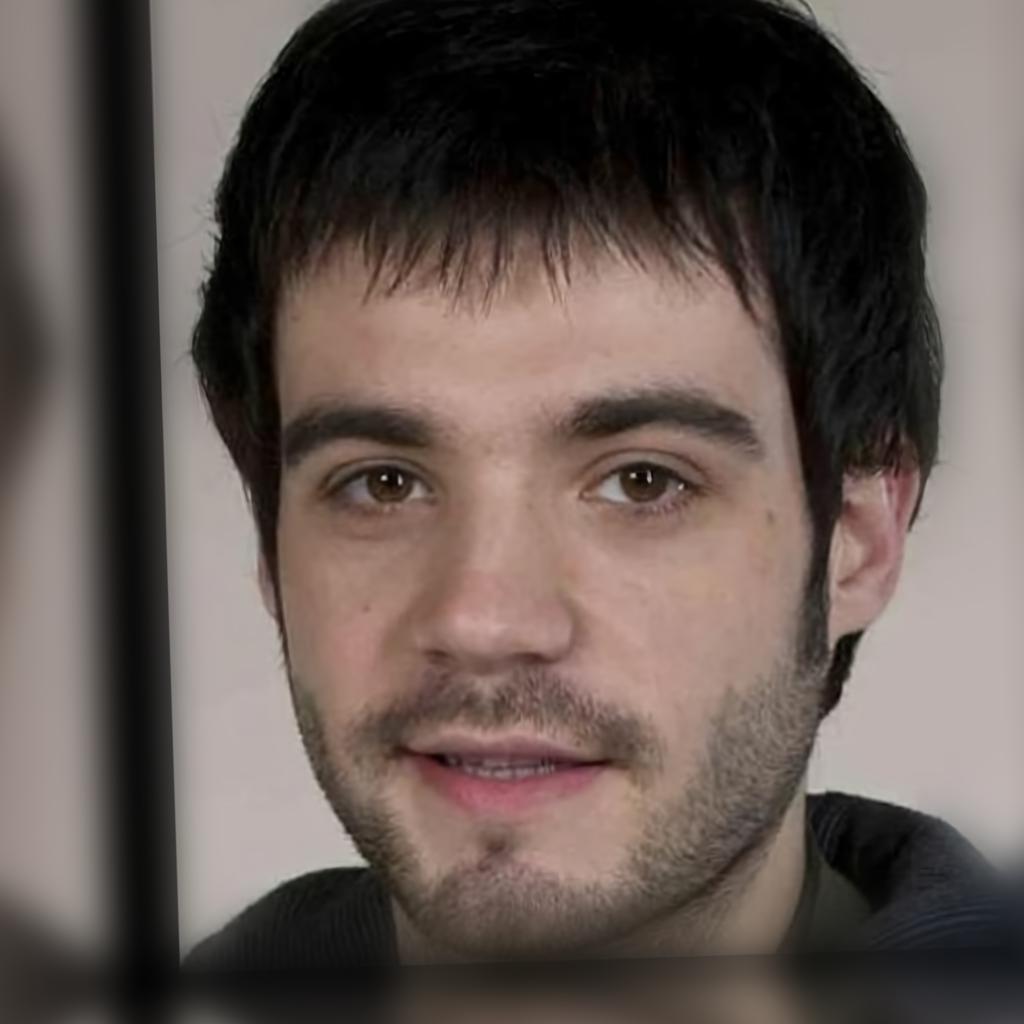} \end{tabular}
&
\begin{tabular}{c} \includegraphics[width=0.136\linewidth]{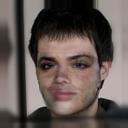} \end{tabular}
&
\begin{tabular}{c} \includegraphics[width=0.136\linewidth]{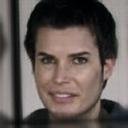} \end{tabular}
&
\begin{tabular}{c} \includegraphics[width=0.136\linewidth]{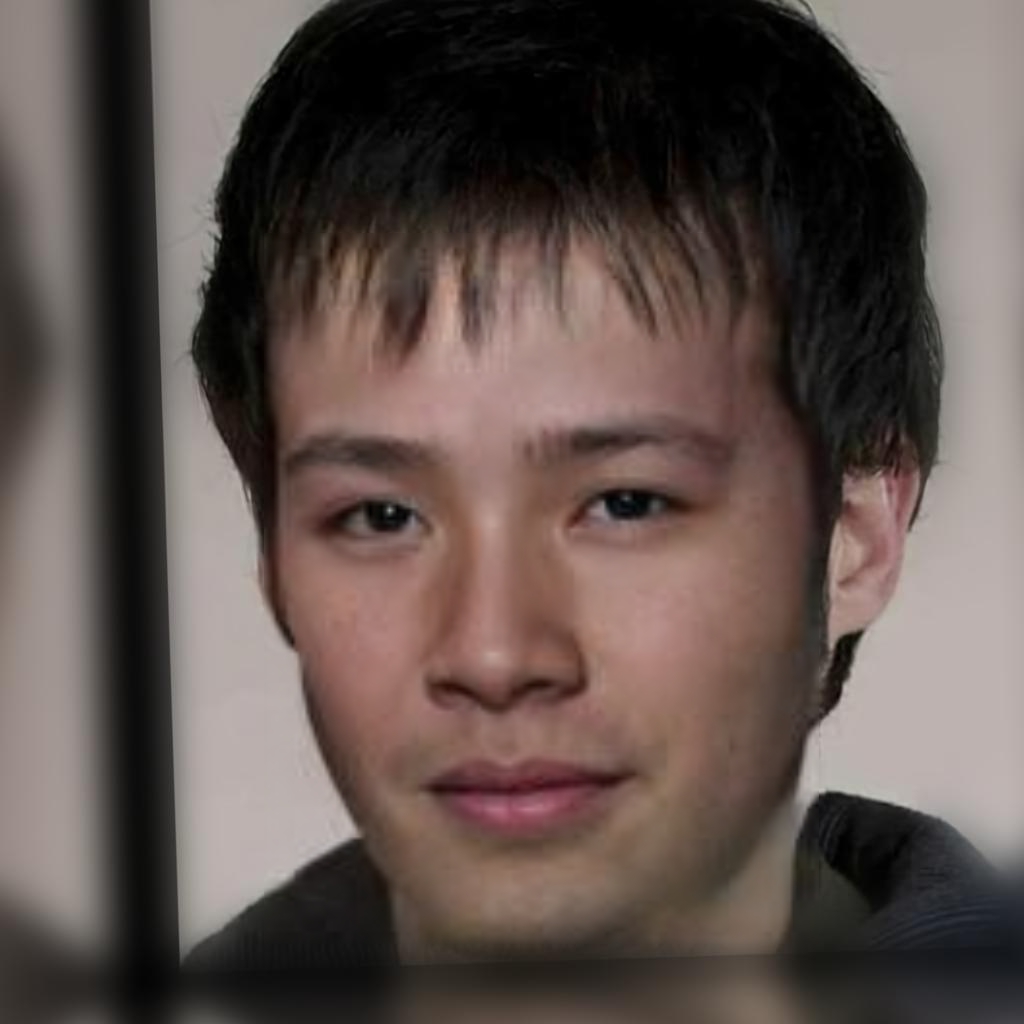} \end{tabular}
&
\begin{tabular}{c} \includegraphics[width=0.136\linewidth]{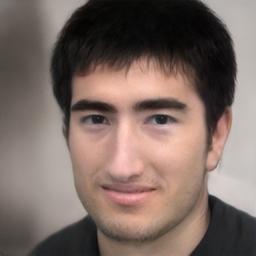} \end{tabular}
&
\begin{tabular}{c} \includegraphics[width=0.136\linewidth]{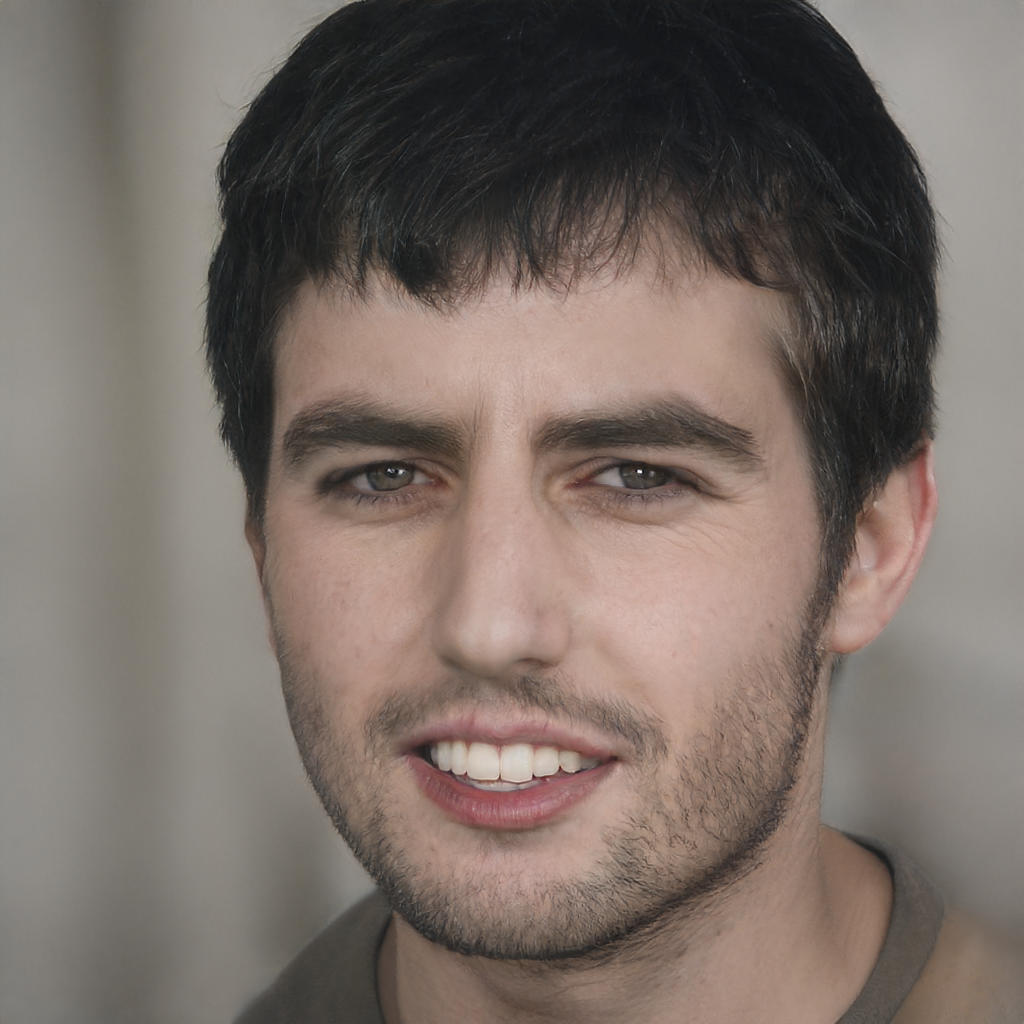} \end{tabular}
&
\begin{tabular}{c} \includegraphics[width=0.136\linewidth]{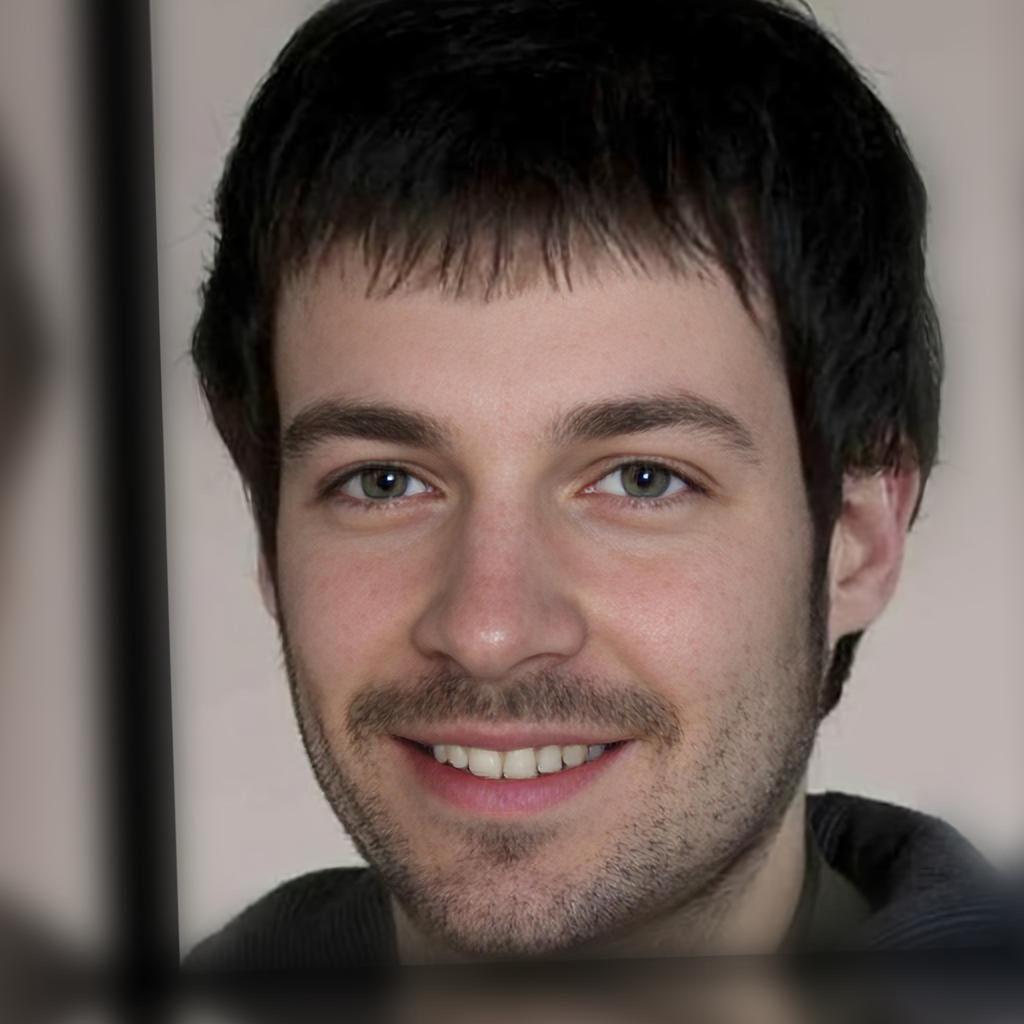} \end{tabular}
\\
\begin{tabular}{c} \makebox[0.136\linewidth]{Input} \end{tabular}
&
\begin{tabular}{c} \makebox[0.136\linewidth]{CIAGAN~\cite{maximov2020ciagan}} \end{tabular}
&
\begin{tabular}{c} \makebox[0.136\linewidth]{FIT~\cite{gu2020password}} \end{tabular}
&
\begin{tabular}{c} \makebox[0.136\linewidth]{DP2~\cite{hukkelaas2023deepprivacy2}} \end{tabular}
&
\begin{tabular}{c} \makebox[0.136\linewidth]{RiDDLE~\cite{li2023riddle}} \end{tabular}
&
\begin{tabular}{c} \makebox[0.136\linewidth]{FALCO~\cite{barattin2023attribute}} \end{tabular}
&
\begin{tabular}{c} \makebox[0.136\linewidth]{Ours} \end{tabular}
\\
\end{tabular}
\caption{Extensive qualitative evaluation results on the \textit{standard single-image} anonymization, benchmarking against the two most commonly referenced anonymization methods and the four most recent state-of-the-art anonymization approaches, on CelebAMaskHQ~\cite{lee2020maskgan} test samples. Our proposed VerA achieves good photorealistic results consistently while de-identifying the input image, outperforming the best baselines even on this standard (non clinical) single-image anonymization task.}
\label{fig:anon_inplace_celebahq}
\end{figure*}
\endgroup

\begingroup
\setlength{\tabcolsep}{0.5pt} 
\renewcommand{\arraystretch}{0.7} 
\begin{figure*}[ht]
\centering
\begin{tabular}{ccccccc}
\begin{tabular}{c} \includegraphics[width=0.136\linewidth]{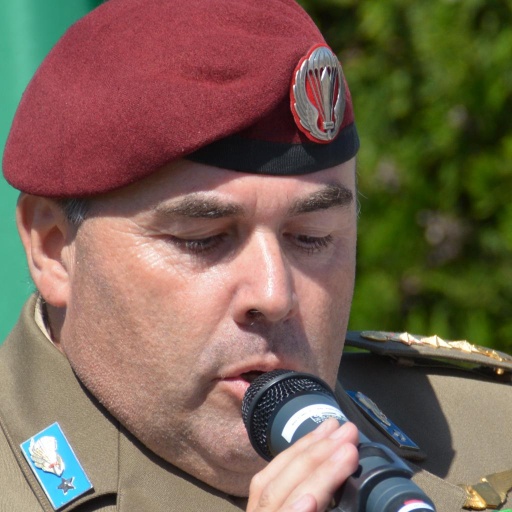} \end{tabular}
&
\begin{tabular}{c} \includegraphics[width=0.136\linewidth]{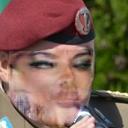} \end{tabular}
&
\begin{tabular}{c} \includegraphics[width=0.136\linewidth]{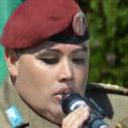} \end{tabular}
&
\begin{tabular}{c} \includegraphics[width=0.136\linewidth]{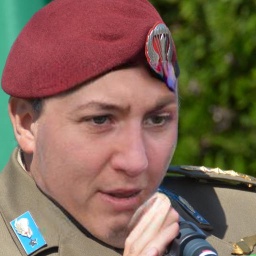} \end{tabular}
&
\begin{tabular}{c} \includegraphics[width=0.136\linewidth]{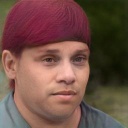} \end{tabular}
&
\begin{tabular}{c} \includegraphics[width=0.136\linewidth]{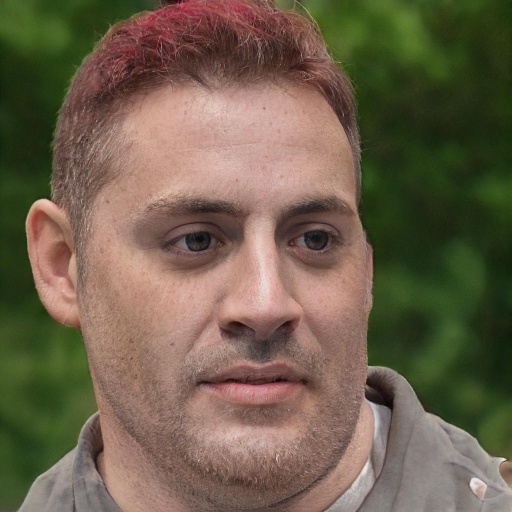} \end{tabular}
&
\begin{tabular}{c} \includegraphics[width=0.136\linewidth]{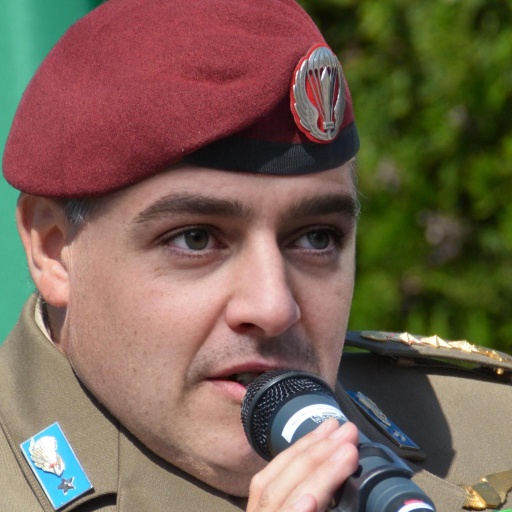} \end{tabular}
\\
\begin{tabular}{c} \includegraphics[width=0.136\linewidth]{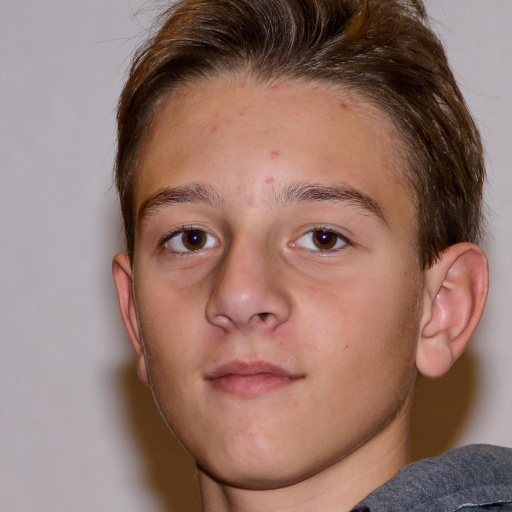} \end{tabular}
&
\begin{tabular}{c} \includegraphics[width=0.136\linewidth]{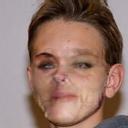} \end{tabular}
&
\begin{tabular}{c} \includegraphics[width=0.136\linewidth]{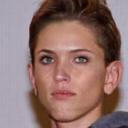} \end{tabular}
&
\begin{tabular}{c} \includegraphics[width=0.136\linewidth]{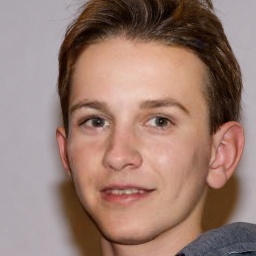} \end{tabular}
&
\begin{tabular}{c} \includegraphics[width=0.136\linewidth]{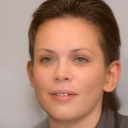} \end{tabular}
&
\begin{tabular}{c} \includegraphics[width=0.136\linewidth]{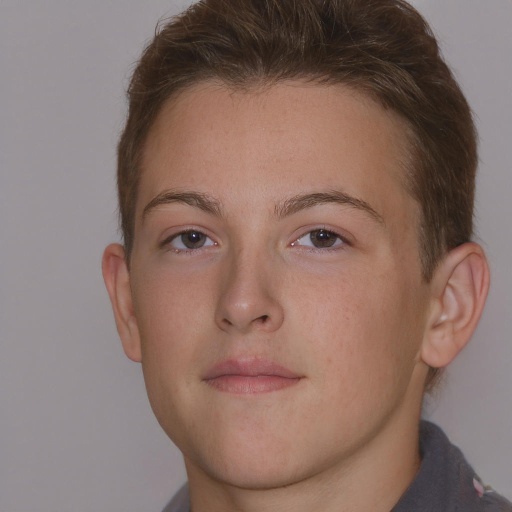} \end{tabular}
&
\begin{tabular}{c} \includegraphics[width=0.136\linewidth]{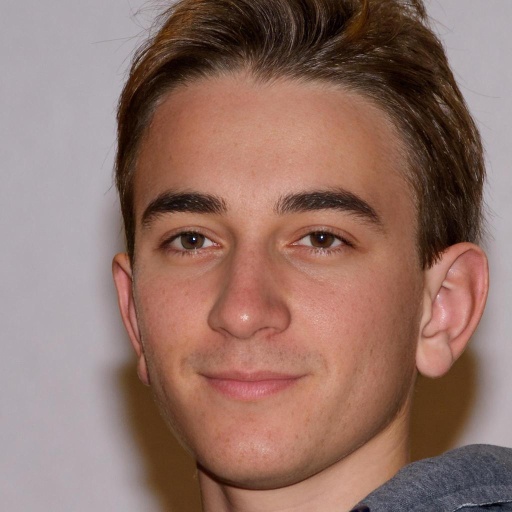} \end{tabular}
\\
\begin{tabular}{c} \includegraphics[width=0.136\linewidth]{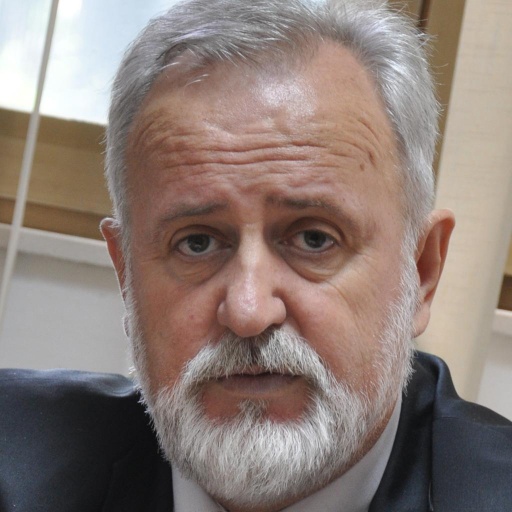} \end{tabular}
&
\begin{tabular}{c} \includegraphics[width=0.136\linewidth]{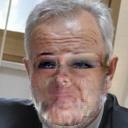} \end{tabular}
&
\begin{tabular}{c} \includegraphics[width=0.136\linewidth]{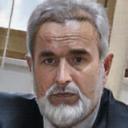} \end{tabular}
&
\begin{tabular}{c} \includegraphics[width=0.136\linewidth]{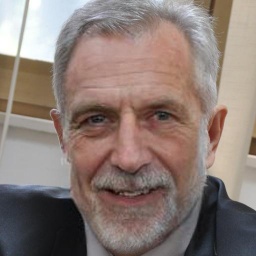} \end{tabular}
&
\begin{tabular}{c} \includegraphics[width=0.136\linewidth]{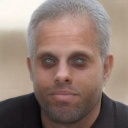} \end{tabular}
&
\begin{tabular}{c} \includegraphics[width=0.136\linewidth]{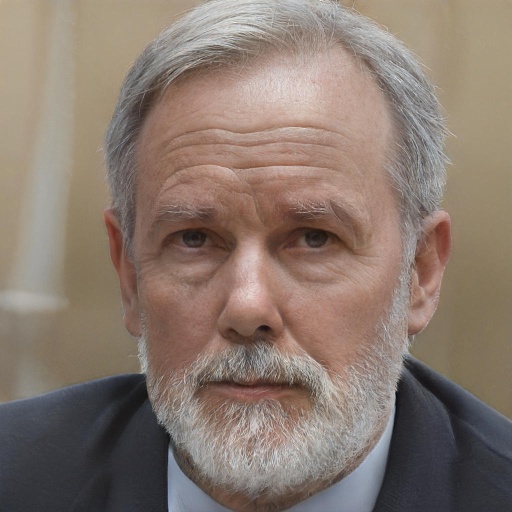} \end{tabular}
&
\begin{tabular}{c} \includegraphics[width=0.136\linewidth]{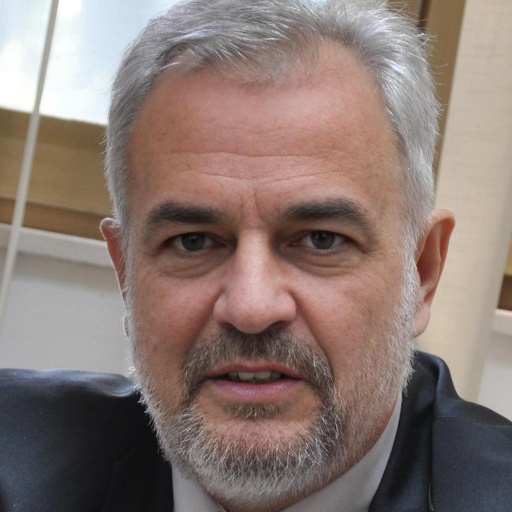} \end{tabular}
\\
\begin{tabular}{c} \includegraphics[width=0.136\linewidth]{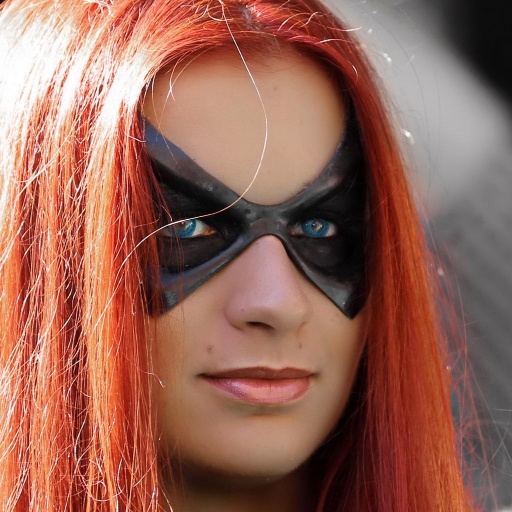} \end{tabular}
&
\begin{tabular}{c} \includegraphics[width=0.136\linewidth]{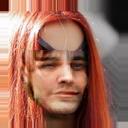} \end{tabular}
&
\begin{tabular}{c} \includegraphics[width=0.136\linewidth]{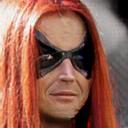} \end{tabular}
&
\begin{tabular}{c} \includegraphics[width=0.136\linewidth]{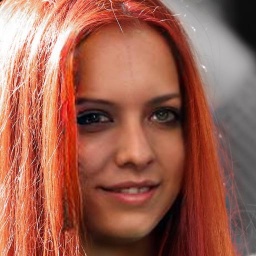} \end{tabular}
&
\begin{tabular}{c} \includegraphics[width=0.136\linewidth]{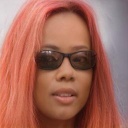} \end{tabular}
&
\begin{tabular}{c} \includegraphics[width=0.136\linewidth]{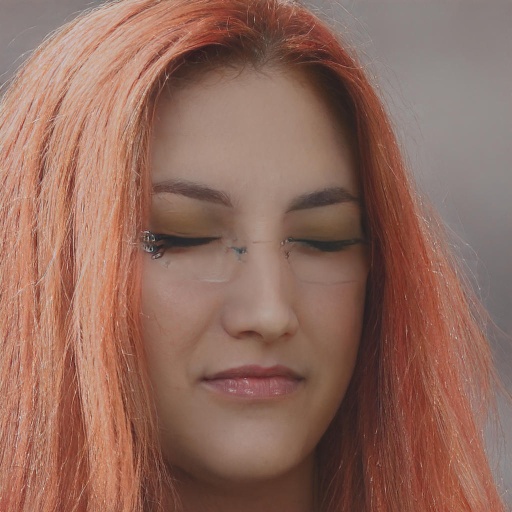} \end{tabular}
&
\begin{tabular}{c} \includegraphics[width=0.136\linewidth]{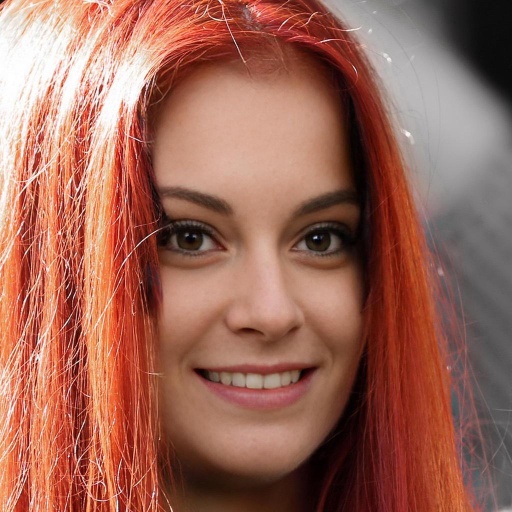} \end{tabular}
\\
\begin{tabular}{c} \includegraphics[width=0.136\linewidth]{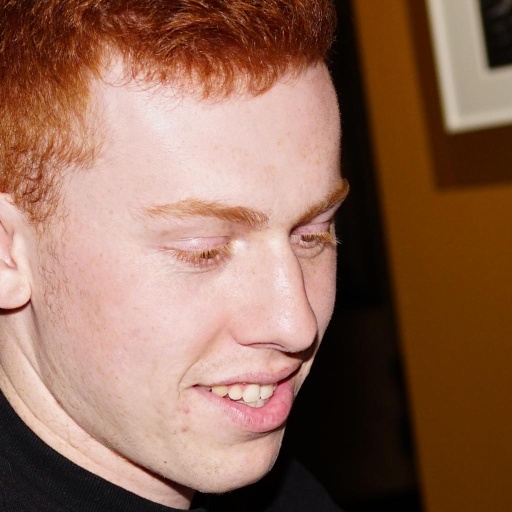} \end{tabular}
&
\begin{tabular}{c} \includegraphics[width=0.136\linewidth]{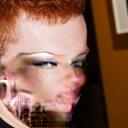} \end{tabular}
&
\begin{tabular}{c} \includegraphics[width=0.136\linewidth]{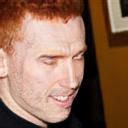} \end{tabular}
&
\begin{tabular}{c} \includegraphics[width=0.136\linewidth]{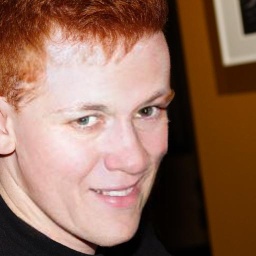} \end{tabular}
&
\begin{tabular}{c} \includegraphics[width=0.136\linewidth]{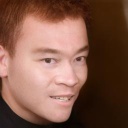} \end{tabular}
&
\begin{tabular}{c} \includegraphics[width=0.136\linewidth]{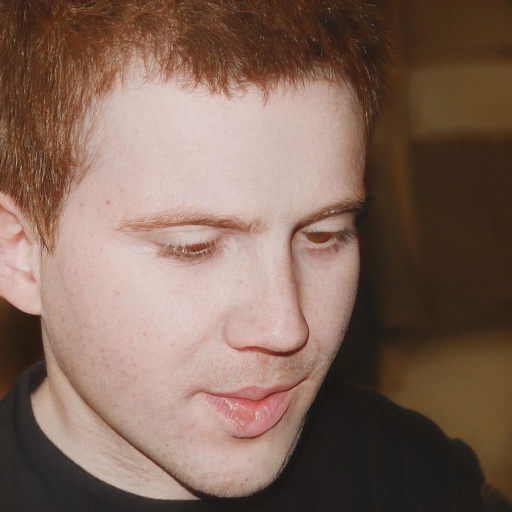} \end{tabular}
&
\begin{tabular}{c} \includegraphics[width=0.136\linewidth]{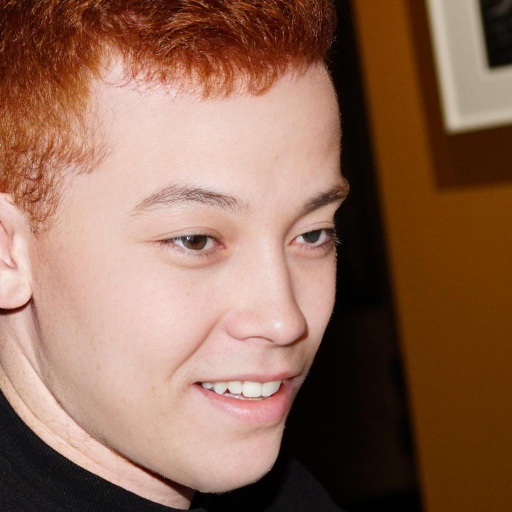} \end{tabular}
\\
\begin{tabular}{c} \includegraphics[width=0.136\linewidth]{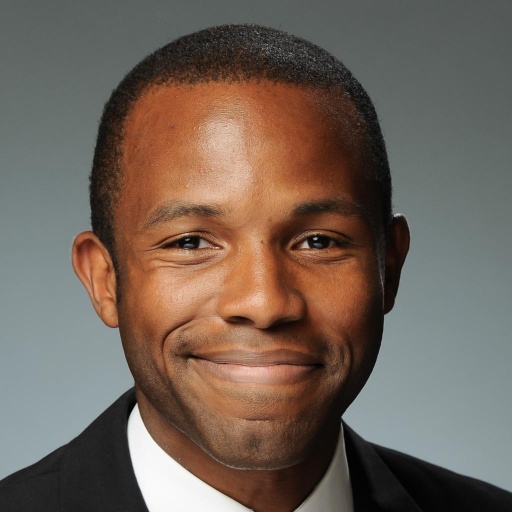} \end{tabular}
&
\begin{tabular}{c} \includegraphics[width=0.136\linewidth]{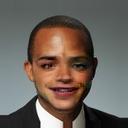} \end{tabular}
&
\begin{tabular}{c} \includegraphics[width=0.136\linewidth]{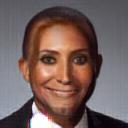} \end{tabular}
&
\begin{tabular}{c} \includegraphics[width=0.136\linewidth]{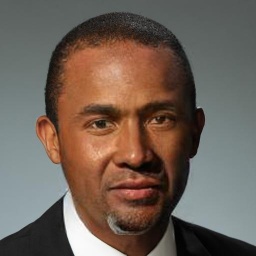} \end{tabular}
&
\begin{tabular}{c} \includegraphics[width=0.136\linewidth]{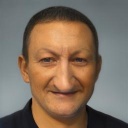} \end{tabular}
&
\begin{tabular}{c} \includegraphics[width=0.136\linewidth]{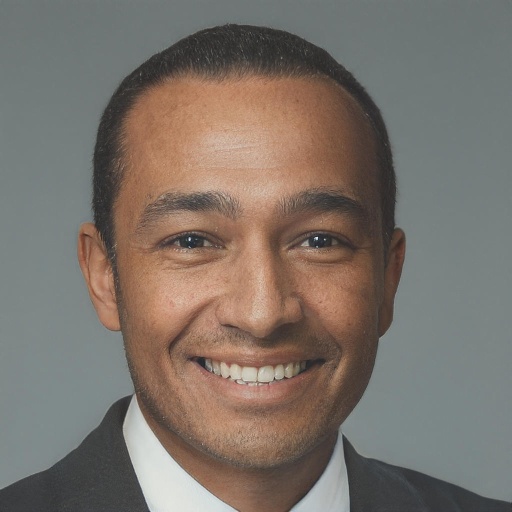} \end{tabular}
&
\begin{tabular}{c} \includegraphics[width=0.136\linewidth]{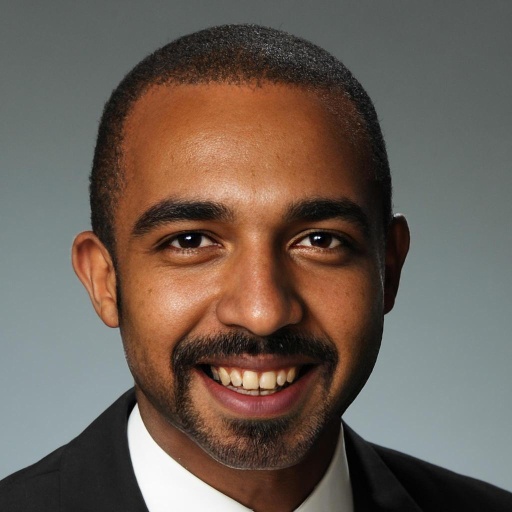} \end{tabular}
\\
\begin{tabular}{c} \includegraphics[width=0.136\linewidth]{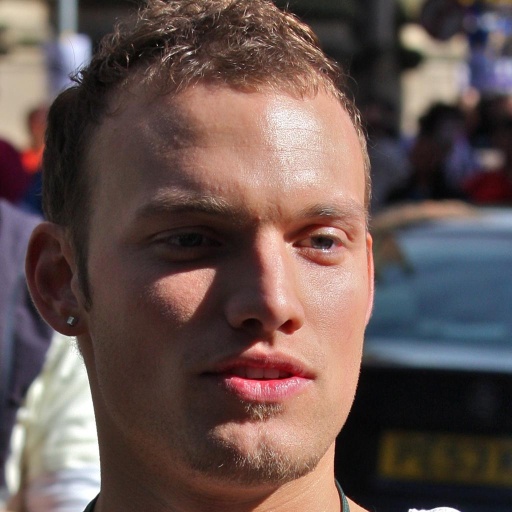} \end{tabular}
&
\begin{tabular}{c} \includegraphics[width=0.136\linewidth]{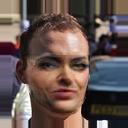} \end{tabular}
&
\begin{tabular}{c} \includegraphics[width=0.136\linewidth]{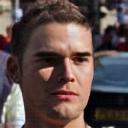} \end{tabular}
&
\begin{tabular}{c} \includegraphics[width=0.136\linewidth]{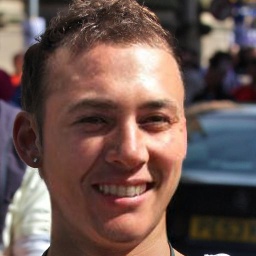} \end{tabular}
&
\begin{tabular}{c} \includegraphics[width=0.136\linewidth]{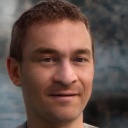} \end{tabular}
&
\begin{tabular}{c} \includegraphics[width=0.136\linewidth]{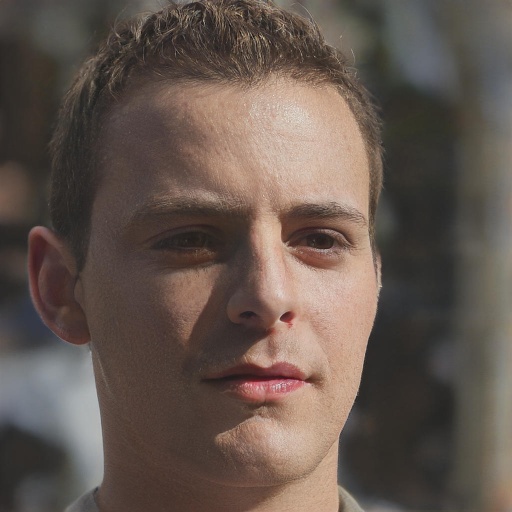} \end{tabular}
&
\begin{tabular}{c} \includegraphics[width=0.136\linewidth]{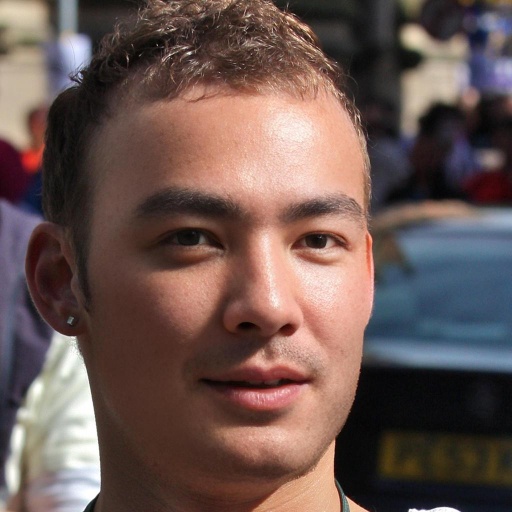} \end{tabular}
\\
\begin{tabular}{c} \includegraphics[width=0.136\linewidth]{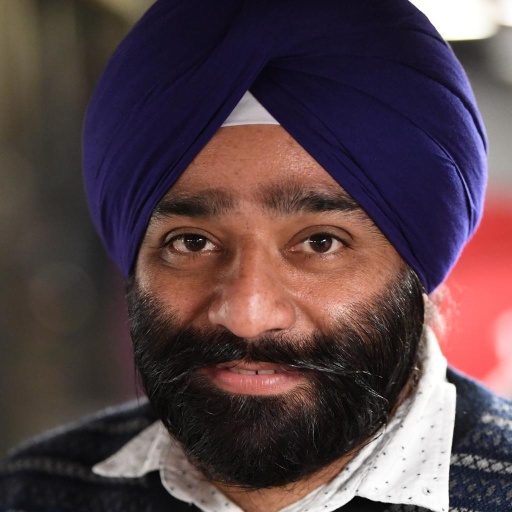} \end{tabular}
&
\begin{tabular}{c} \includegraphics[width=0.136\linewidth]{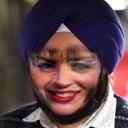} \end{tabular}
&
\begin{tabular}{c} \includegraphics[width=0.136\linewidth]{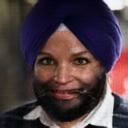} \end{tabular}
&
\begin{tabular}{c} \includegraphics[width=0.136\linewidth]{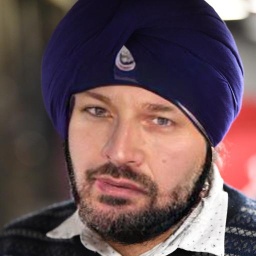} \end{tabular}
&
\begin{tabular}{c} \includegraphics[width=0.136\linewidth]{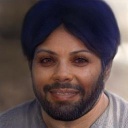} \end{tabular}
&
\begin{tabular}{c} \includegraphics[width=0.136\linewidth]{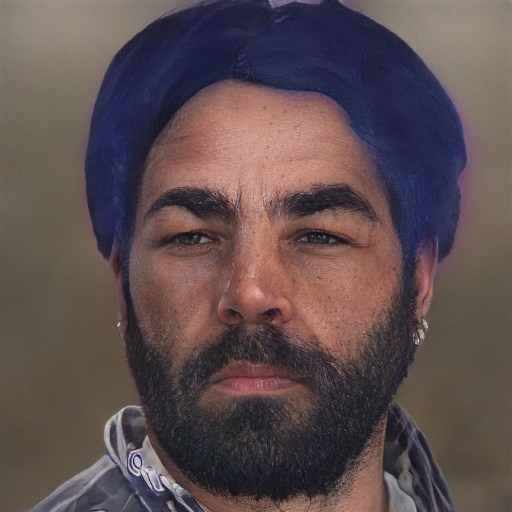} \end{tabular}
&
\begin{tabular}{c} \includegraphics[width=0.136\linewidth]{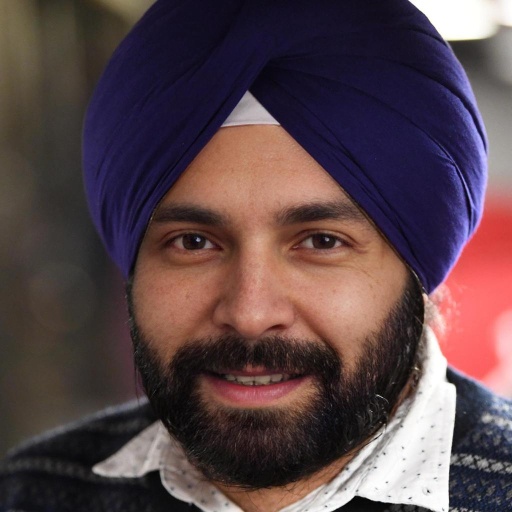} \end{tabular}
\\
\begin{tabular}{c} \makebox[0.136\linewidth]{Input} \end{tabular}
&
\begin{tabular}{c} \makebox[0.136\linewidth]{CIAGAN~\cite{maximov2020ciagan}} \end{tabular}
&
\begin{tabular}{c} \makebox[0.136\linewidth]{FIT~\cite{gu2020password}} \end{tabular}
&
\begin{tabular}{c} \makebox[0.136\linewidth]{DP2~\cite{hukkelaas2023deepprivacy2}} \end{tabular}
&
\begin{tabular}{c} \makebox[0.136\linewidth]{RiDDLE~\cite{li2023riddle}} \end{tabular}
&
\begin{tabular}{c} \makebox[0.136\linewidth]{FALCO~\cite{barattin2023attribute}} \end{tabular}
&
\begin{tabular}{c} \makebox[0.136\linewidth]{Ours} \end{tabular}
\\
\end{tabular}
\caption{Extensive qualitative evaluation results on the \textit{standard} single-image anonymization, similar to Fig.~\ref{fig:anon_inplace_celebahq}, but performed here on FFHQ~\cite{karras2019style} test samples. Thanks to semantic-aware inversion, VerA can robustly anonymize images with occluding objects and various accessories such as hats (first and last rows).}
\label{fig:anon_inplace_ffhq}
\end{figure*}
\endgroup

\subsection{High-level generative image control}
We show the high-level attribute control that our model achieves in Fig.~\ref{fig:control_high_sweep} flexibly on pose and age. We also show independent modifications that we apply in expression, pose, and age in Fig.~\ref{fig:control_high_a}. Fig.~\ref{fig:control_high_b} shows similar results but with the accumulation of high-level changes from left to right, sequentially altering the person's expression, orientation, and age. Lastly, we show in Fig.~\ref{fig:control_yaw_pitch} that even within pose we can disentangle yaw and pitch just with PCA. By performing a PCA decomposition over the pose latent, we can move in the direction of one component to modify yaw, and the other component for controlling pitch. As shown in the top part, we can linearly control the yaw (top left linear curves) without affecting the pitch (top right flat curves).

\begingroup
\setlength{\tabcolsep}{0.5pt} 
\renewcommand{\arraystretch}{0.7} 

\begin{figure}[!ht]
    \centering
    \begin{tabular}{ccccc}
      \multirow{2}{*}[+6ex]{\rotatebox[origin=lc]{90}{FALCO adaptive normalization}} &
      \begin{tabular}{c} \rotatebox[origin=c]{90}{\makebox[1mm]{On (default)}} \end{tabular} &
      \begin{tabular}{c} \includegraphics[width=0.29\linewidth]{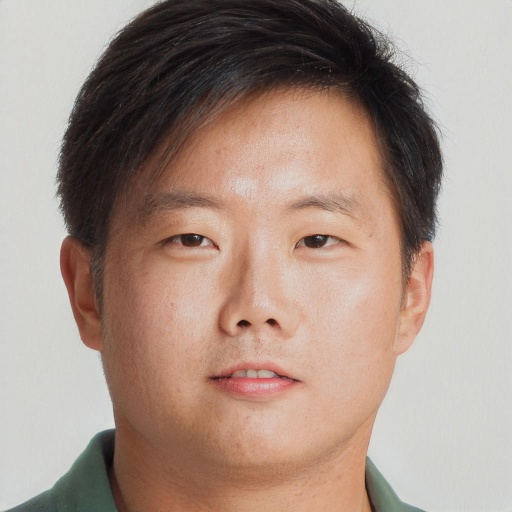} \end{tabular} &
      \begin{tabular}{c} \includegraphics[width=0.29\linewidth]{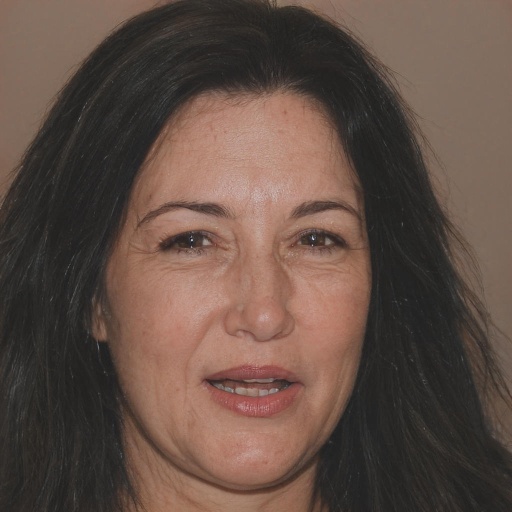} \end{tabular} &
      \begin{tabular}{c} \includegraphics[width=0.29\linewidth]{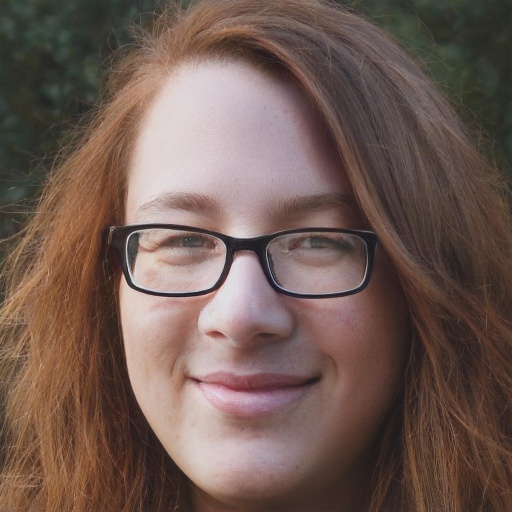} \end{tabular} \\
      & 
      \begin{tabular}{c} \rotatebox[origin=c]{90}{\makebox[1mm]{Off}} \end{tabular} &
      \begin{tabular}{c} \includegraphics[width=0.29\linewidth]{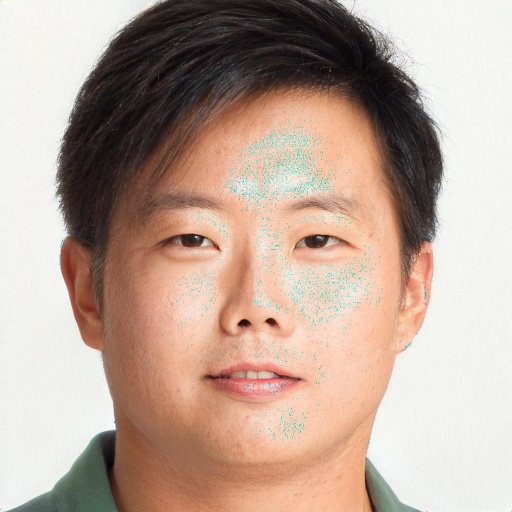} \end{tabular} &
      \begin{tabular}{c} \includegraphics[width=0.29\linewidth]{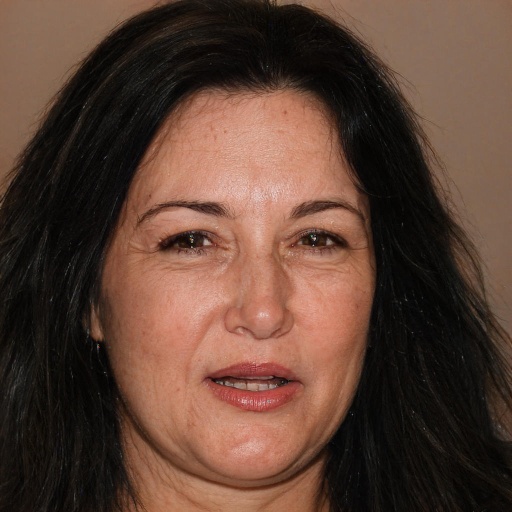} \end{tabular} &
      \begin{tabular}{c} \includegraphics[width=0.29\linewidth]{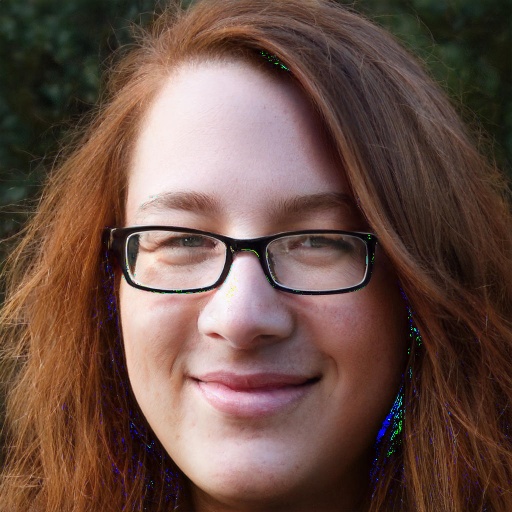} \end{tabular} \\
    \end{tabular}
    \caption{Illustration of the two settings of adaptive normalization in FALCO~\cite{barattin2023attribute}. The authors' default corresponds to the top row, and can lead to washed out final images. If toggled off, this setting can lead to odd color artifacts like the blue in the first and last column (bottom right corner between the face and the hair).}
    \label{fig:falco_norm} 
\end{figure}

\endgroup

\begingroup
\setlength{\tabcolsep}{0.5pt} 
\renewcommand{\arraystretch}{0.7} 

\begin{figure}[ht]
\centering
\begin{tabular}{cccc}
\begin{tabular}{c} \includegraphics[width=0.245\linewidth]{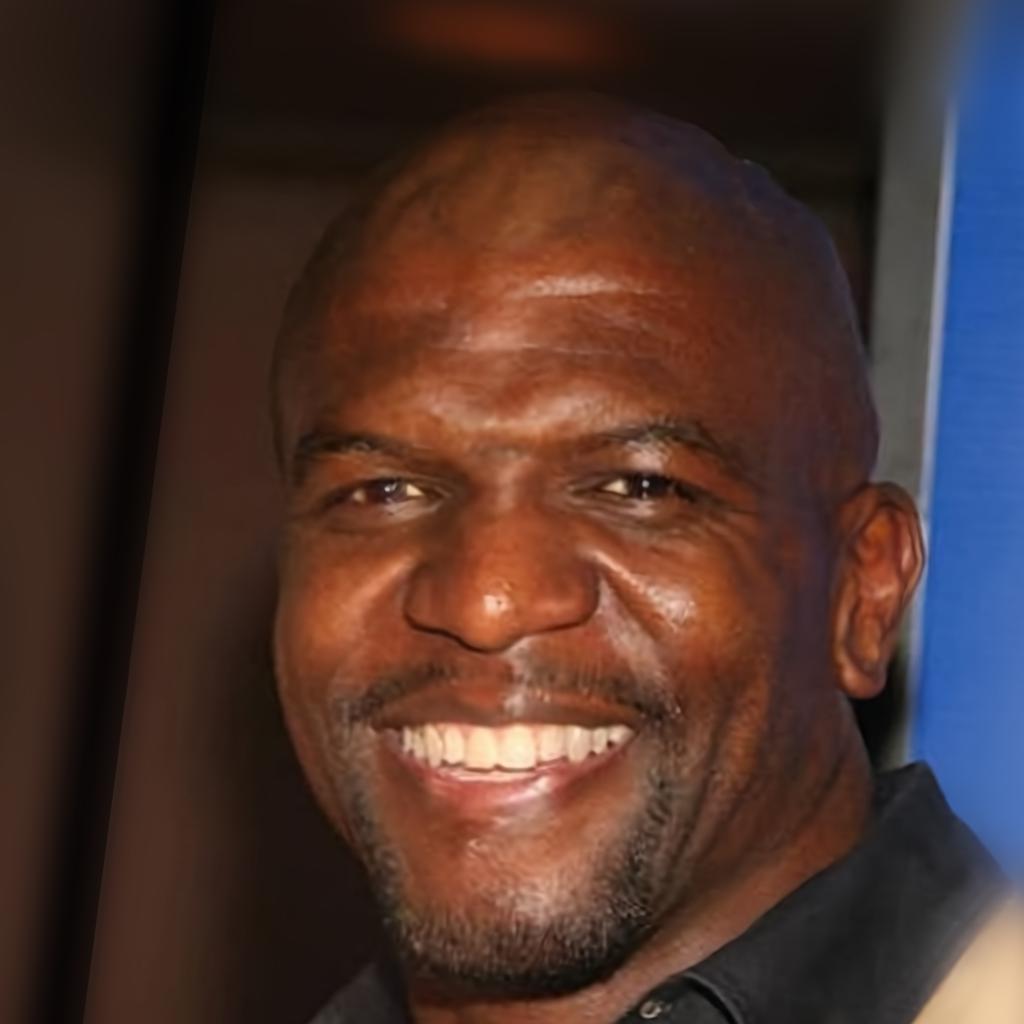} \end{tabular}
&
\begin{tabular}{c} \includegraphics[width=0.245\linewidth]{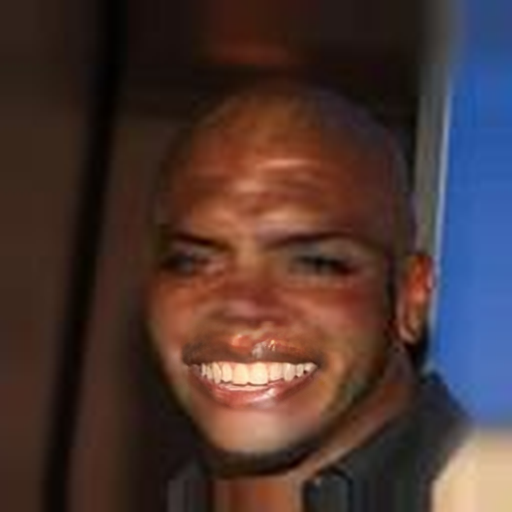} \end{tabular}
&
\begin{tabular}{c} \includegraphics[width=0.245\linewidth]{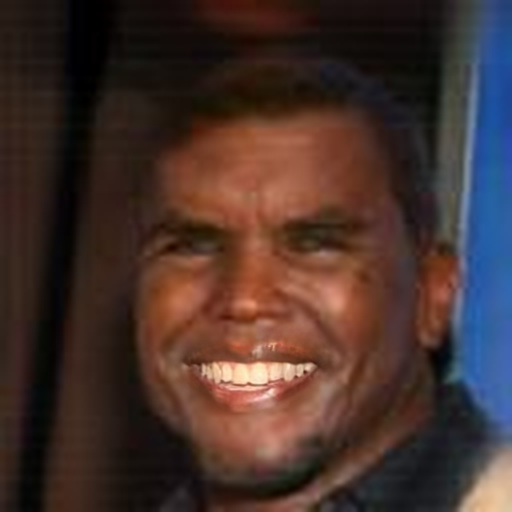} \end{tabular}
&
\begin{tabular}{c} \includegraphics[width=0.245\linewidth]{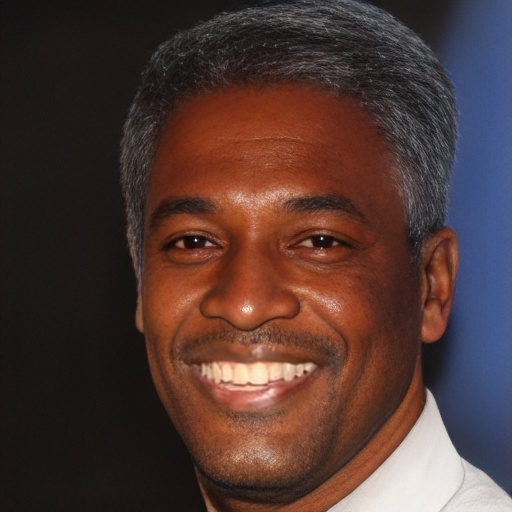} \end{tabular}
\\
\begin{tabular}{c} \makebox[0.245\linewidth]{Input} \end{tabular}
&
\begin{tabular}{c} \makebox[0.245\linewidth]{CIAGAN~\cite{maximov2020ciagan}} \end{tabular}
&
\begin{tabular}{c} \makebox[0.245\linewidth]{FIT~\cite{gu2020password}} \end{tabular}
&
\begin{tabular}{c} \makebox[0.245\linewidth]{\makecell{\textbf{Ours} \\ {blended}}} \end{tabular}
\\
\begin{tabular}{c} \includegraphics[width=0.245\linewidth]{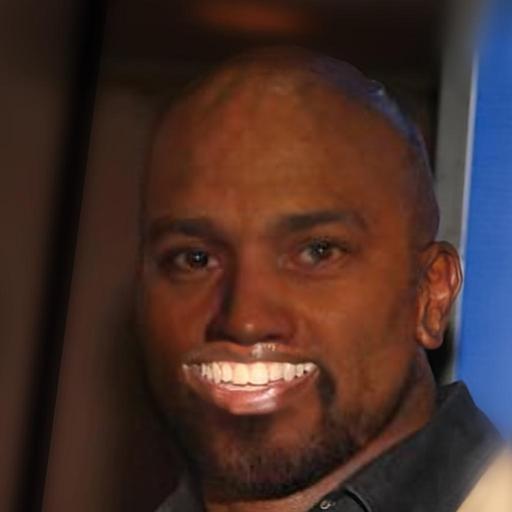} \end{tabular}
&
\begin{tabular}{c} \includegraphics[width=0.245\linewidth]{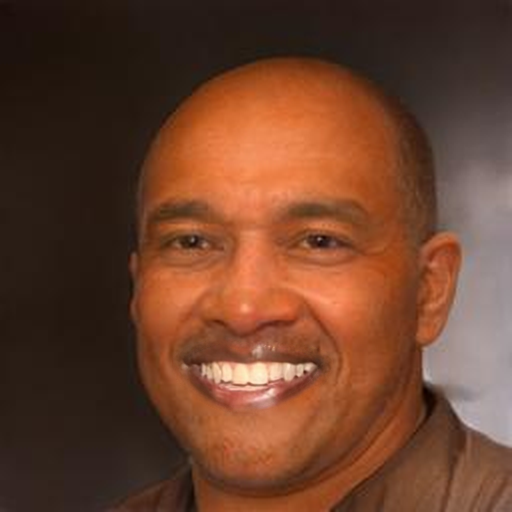} \end{tabular}
&
\begin{tabular}{c} \includegraphics[width=0.245\linewidth]{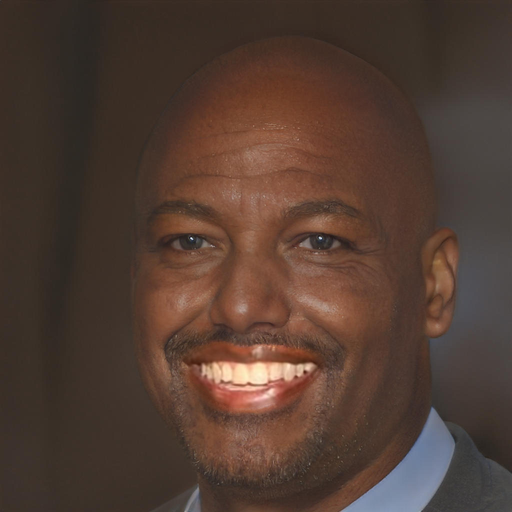} \end{tabular}
&
\begin{tabular}{c} \includegraphics[width=0.245\linewidth]{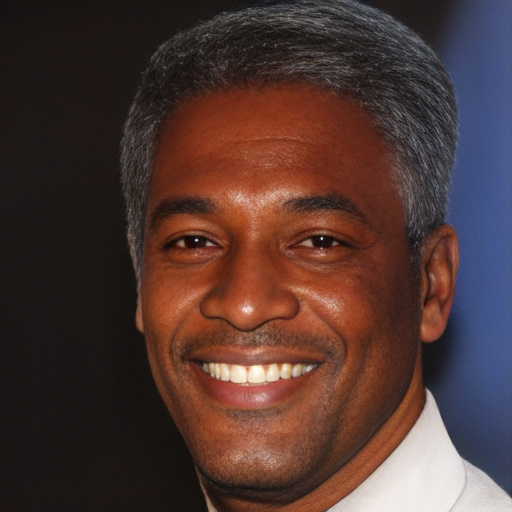} \end{tabular}
\\
\begin{tabular}{c} \makebox[0.245\linewidth]{DP2~\cite{hukkelaas2023deepprivacy2}} \end{tabular}
&
\begin{tabular}{c} \makebox[0.245\linewidth]{RiDDLE~\cite{li2023riddle}} \end{tabular}
&
\begin{tabular}{c} \makebox[0.245\linewidth]{FALCO~\cite{barattin2023attribute}} \end{tabular}
&
\begin{tabular}{c} \makebox[0.245\linewidth]{\makecell{\textbf{Ours} \\ \textcolor{red}{not blended}}} \end{tabular}
\\
\end{tabular}
\caption{Best viewed \textbf{zoomed in}. We blend the \textit{mouth} (of the same sample shown in Fig. 5 of the main manuscript) for competing methods, which yields less photorealistic results as competitors do not properly address semantic anonymization. We additionally show for illustration our own output without our region-of-interest blending.}
\label{fig:inpainted_baselines}
\end{figure}

\endgroup

\begin{table*}[t]
    \centering
    \resizebox{\linewidth}{!}{%
    \begin{NiceTabular}{ll|c|c|c|| c|c|c|| c|c|c|| c|c|c}
    \toprule
    & & \multicolumn{3}{c}{$\ell_1$ distance $\downarrow$} & \multicolumn{3}{c}{PSNR $\uparrow$} & \multicolumn{3}{c}{Semantic IoU $\uparrow$} & \multicolumn{3}{c}{Mean landmark offset $\downarrow$} \\
    & Method & Mouth & Nose & Eyes & Mouth & Nose & Eyes & Mouth & Nose & Eyes & Mouth & Nose & Eyes \\
    \midrule
    \multirow{6}{*}{\rotatebox[origin=c]{90}{Standard}}
    & CIAGAN~\cite{maximov2020ciagan}      & 38.53 & 31.73 & 54.00 & 14.20 & 15.76 & 11.60 & 0.53 & 0.53 & 0.01 & 17.50 & 21.26 & 43.88 \\
    & FIT~\cite{gu2020password}            & \underline{21.19} & \underline{17.04} & \underline{24.70} & \underline{19.12} & \underline{20.85} & \underline{17.53} & \underline{0.75} & \underline{0.81} & 0.57 & \underline{9.75}  & \underline{9.38}  & 8.82  \\
    & DP2~\cite{hukkelaas2023deepprivacy2} & 40.72 & 32.95 & 49.16 & 13.58 & 15.05 & 12.05 & 0.52 & 0.62 & 0.23 & 29.82 & 32.45 & 28.13 \\
    & RiDDLE~\cite{li2023riddle}           & 35.97 & 30.48 & 36.05 & 14.71 & 15.98 & 14.48 & 0.69 & 0.77 & 0.59 & 14.30 & 18.98 & 8.58  \\
    & FALCO~\cite{barattin2023attribute}   & 33.63 & 27.01 & 34.93 & 15.25 & 17.07 & 14.70 & 0.63 & 0.76 & 0.54 & 18.22 & 17.40 & 9.08  \\
    & Ours                                 & 34.59 & 21.89 & 34.82 & 14.69 & 18.09 & 14.38 & 0.64 & 0.78 & \underline{0.61} & 17.39 & 16.04 & \underline{7.62}  \\
    \hdashline
    \multirow{3}{*}{\rotatebox[origin=c]{90}{Clinical}}
    & Ours (mouth)  & \textbf{0.22}  & 22.32 & 36.20 & \textbf{54.42} & 18.01 & 14.15 & \textbf{0.90} & 0.79 & 0.60 & \textbf{8.13}  & 15.56 & 7.85 \\
    & Ours (nose)   & 34.63 & \textbf{0.21}  & 36.05 & 14.67 & \textbf{54.59} & 14.20 & 0.65 & \textbf{0.93} & 0.60 & 16.29 & \textbf{5.68}  & 7.68 \\
    & Ours (eyes)   & 34.91 & 22.17 & \textbf{0.35}  & 14.61 & 18.03 & \textbf{50.72} & 0.64 & 0.78 & \textbf{0.76} & 17.84 & 16.58 & \textbf{6.25} \\
    \bottomrule
    \end{NiceTabular}
    }
    \caption{Semantic preservation results, in terms of content ($\ell_1$, PSNR) and area (IoU, landmarks), evaluated on FFHQ~\cite{karras2019style}. We note two main observations: standard anonymization approaches destroy all semantic components that may need to be preserved in clinical images, and our clinical anonymization successfully preserves the desired component while also flexibly modifying non-blocked components as much as the baselines. Note: eye landmarks are key components in the alignment algorithm of FFHQ~\cite{karras2019style}, which results in similar eye landmarks across images, thus the generally lower average landmark offset.}
    \label{table:semantic_preservation_ffhq}
\end{table*}

\begin{table}[!t]
    \centering
    \resizebox{\columnwidth}{!}{%
    \begin{NiceTabular}{l|c|c||c|c|c|c}
    \toprule
    \multirow{2}{*}{Method}              & \multicolumn{2}{c}{FID $\downarrow$} & \multicolumn{2}{c}{Bounding box $\uparrow$} & \multicolumn{2}{c}{Face detection $\uparrow$} \\
                                         & FFHQ   & CelebAHQ & MTCNN & Dlib  & MTCNN & Dlib  \\
    \midrule
    CIAGAN~\cite{maximov2020ciagan}      & 109.92 & 93.46    & 0.80  & 0.88  & 0.90  & 0.90  \\
    FIT~\cite{gu2020password}            & 89.47  & 95.98    & \textbf{0.91}  & \textbf{0.94}  & 0.98  & \underline{0.99}  \\
    DP2~\cite{hukkelaas2023deepprivacy2} & \underline{23.41}  & \underline{51.89}    & 0.87  & 0.88  & 0.96  & 0.97  \\
    RiDDLE~\cite{li2023riddle}           & 69.93  & 66.95    & \underline{0.90}  & 0.91  & \textbf{1.00}  & \textbf{1.00}  \\
    FALCO~\cite{barattin2023attribute}   & 48.03  & 53.35    & 0.89  & 0.91  & \underline{0.99}  & \textbf{1.00}  \\
    Ours                                 & \textbf{13.79}  & \textbf{51.60}    & \textbf{0.91}  & \underline{0.92}  & 0.97  & \textbf{1.00}  \\
    \bottomrule
    \end{NiceTabular}
    }
    \caption{Downstream utility evaluation for photorealism/diversity (FID~\cite{heusel2017gans}), bounding box IoU, and face detection rates (MTCNN~\cite{zhang2016joint}, Dlib~\cite{king2009dlib}), which we compute over FFHQ~\cite{karras2019style} test data. We achieve the best FID, and are on par with the best bounding box IoU and the best detection scores.}
    \label{table:utility_ffhq}
\end{table}  

\section{Extended anonymization evaluation} \label{sec:eval_anonymization}

\begingroup
\setlength{\tabcolsep}{0.5pt} 
\renewcommand{\arraystretch}{0.7} 
\begin{figure*}[t]
\centering
\begin{tabular}{ccccccccc}
\begin{tabular}{c} \includegraphics[width=0.105\linewidth]{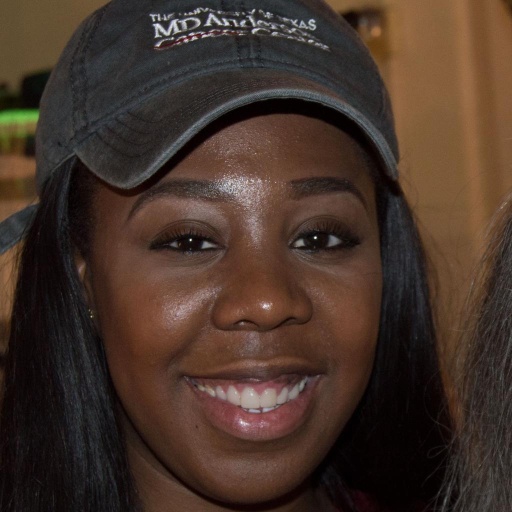} \end{tabular}
&
\begin{tabular}{c} \includegraphics[width=0.105\linewidth]{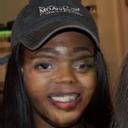} \end{tabular}
&
\begin{tabular}{c} \includegraphics[width=0.105\linewidth]{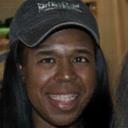} \end{tabular}
&
\begin{tabular}{c} \includegraphics[width=0.105\linewidth]{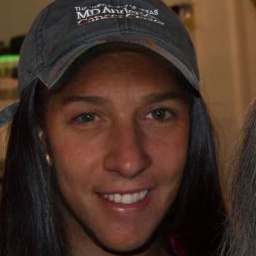} \end{tabular}
&
\begin{tabular}{c} \includegraphics[width=0.105\linewidth]{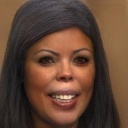} \end{tabular}
&
\begin{tabular}{c} \includegraphics[width=0.105\linewidth]{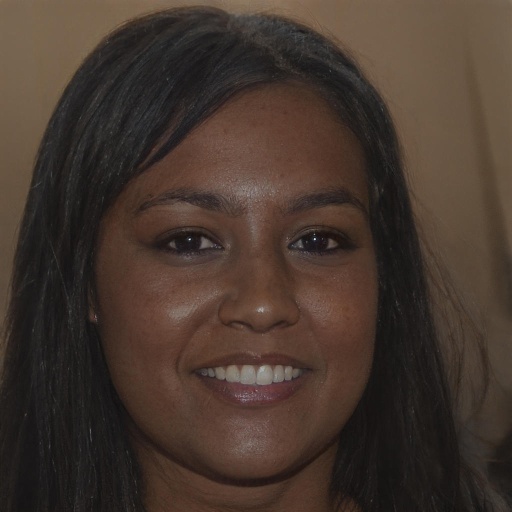} \end{tabular}
&
\begin{tabular}{c} \includegraphics[width=0.105\linewidth]{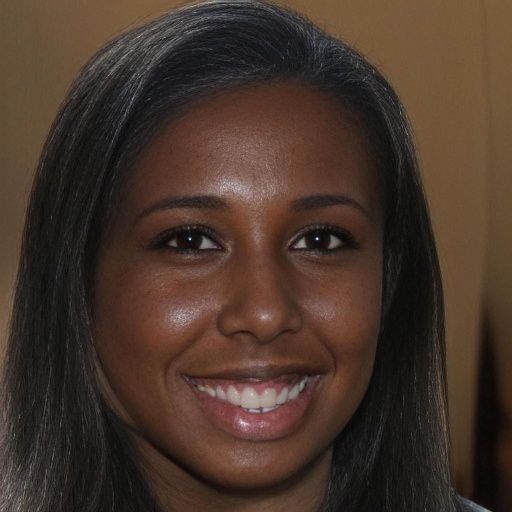} \end{tabular}
&
\begin{tabular}{c} \includegraphics[width=0.105\linewidth]{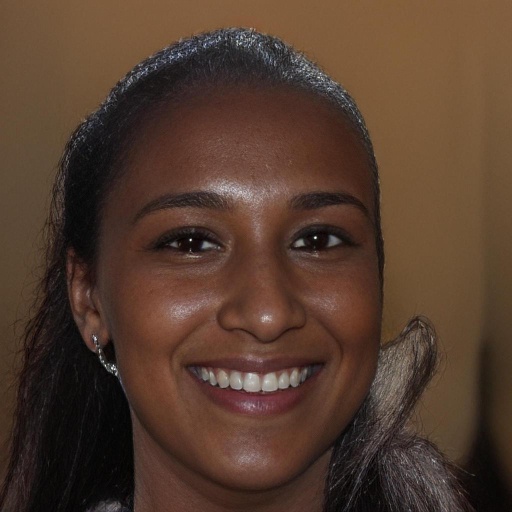} \end{tabular}
&
\begin{tabular}{c} \includegraphics[width=0.105\linewidth]{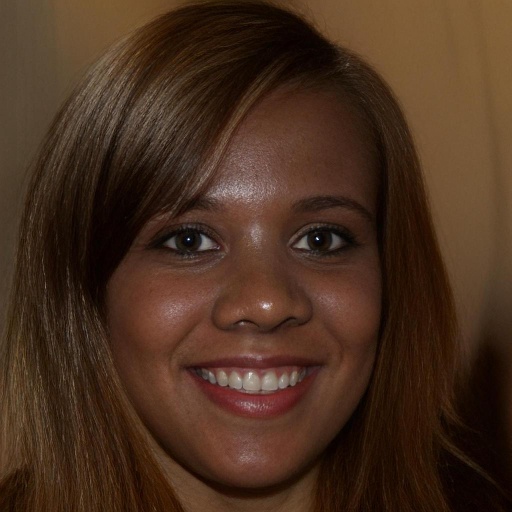} \end{tabular}
\\
\begin{tabular}{c} \includegraphics[width=0.105\linewidth]{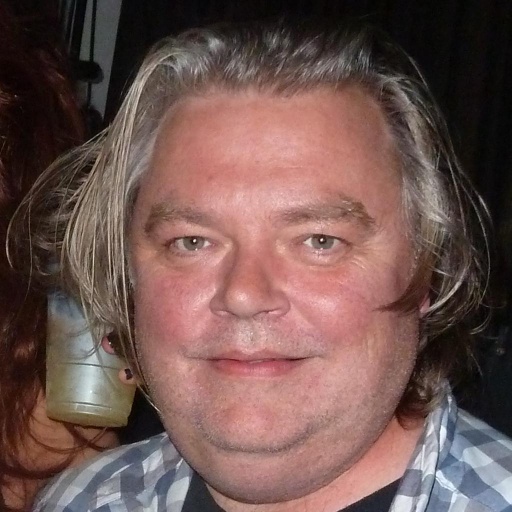} \end{tabular}
&
\begin{tabular}{c} \includegraphics[width=0.105\linewidth]{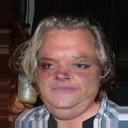} \end{tabular}
&
\begin{tabular}{c} \includegraphics[width=0.105\linewidth]{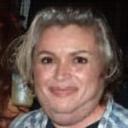} \end{tabular}
&
\begin{tabular}{c} \includegraphics[width=0.105\linewidth]{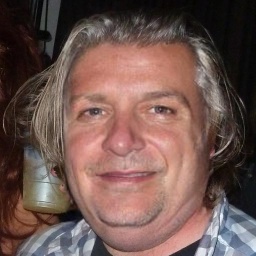} \end{tabular}
&
\begin{tabular}{c} \includegraphics[width=0.105\linewidth]{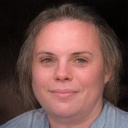} \end{tabular}
&
\begin{tabular}{c} \includegraphics[width=0.105\linewidth]{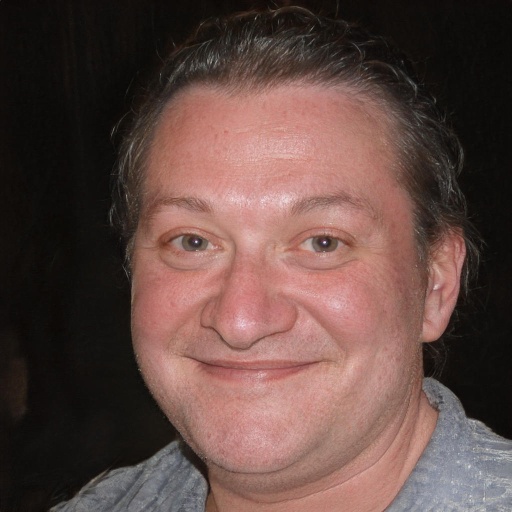} \end{tabular}
&
\begin{tabular}{c} \includegraphics[width=0.105\linewidth]{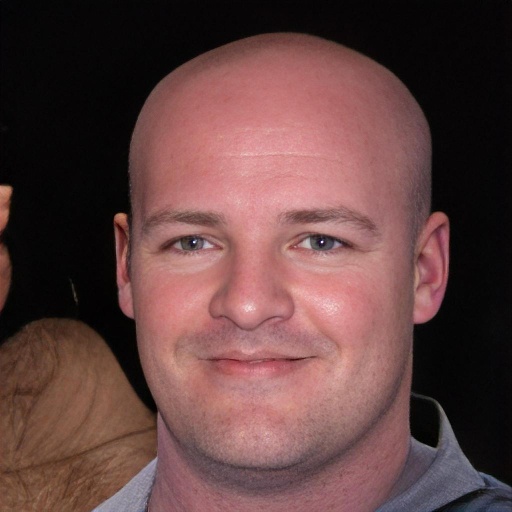} \end{tabular}
&
\begin{tabular}{c} \includegraphics[width=0.105\linewidth]{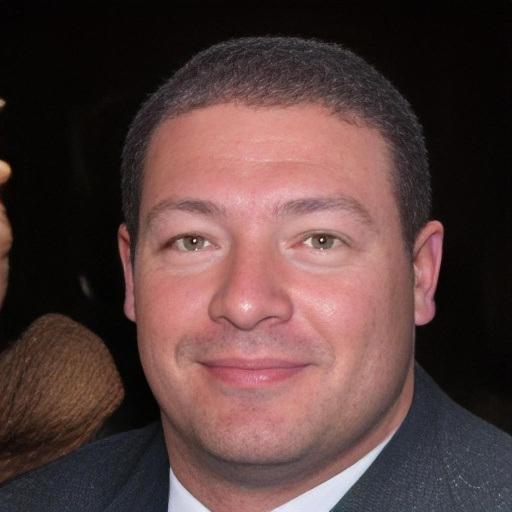} \end{tabular}
&
\begin{tabular}{c} \includegraphics[width=0.105\linewidth]{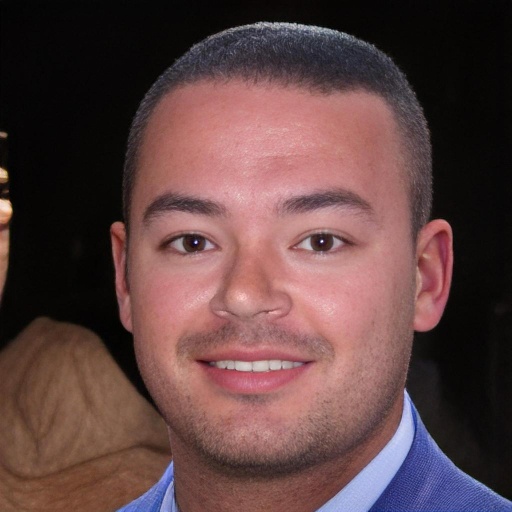} \end{tabular}
\\
\begin{tabular}{c} \includegraphics[width=0.105\linewidth]{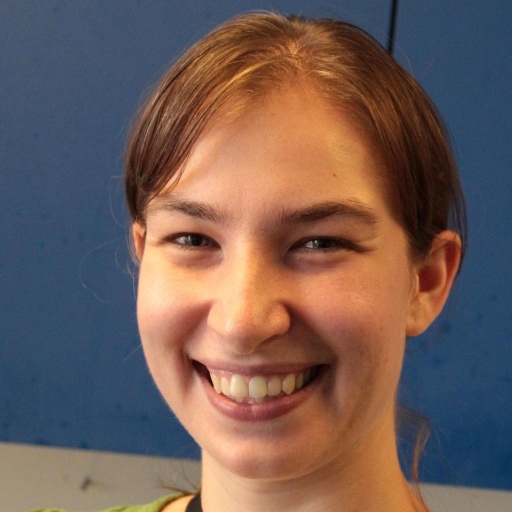} \end{tabular}
&
\begin{tabular}{c} \includegraphics[width=0.105\linewidth]{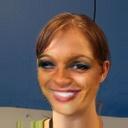} \end{tabular}
&
\begin{tabular}{c} \includegraphics[width=0.105\linewidth]{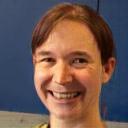} \end{tabular}
&
\begin{tabular}{c} \includegraphics[width=0.105\linewidth]{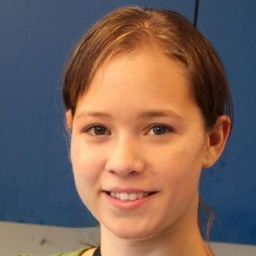} \end{tabular}
&
\begin{tabular}{c} \includegraphics[width=0.105\linewidth]{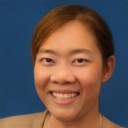} \end{tabular}
&
\begin{tabular}{c} \includegraphics[width=0.105\linewidth]{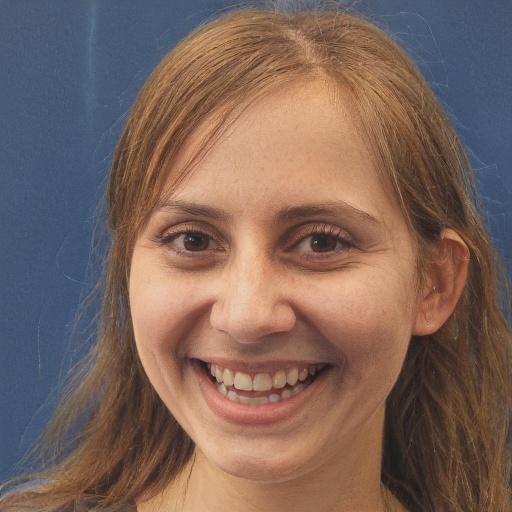} \end{tabular}
&
\begin{tabular}{c} \includegraphics[width=0.105\linewidth]{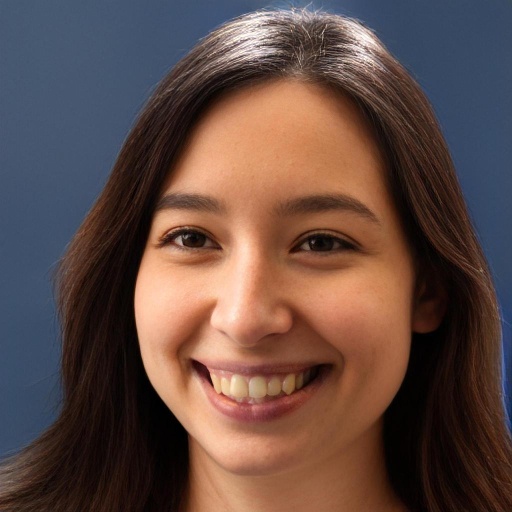} \end{tabular}
&
\begin{tabular}{c} \includegraphics[width=0.105\linewidth]{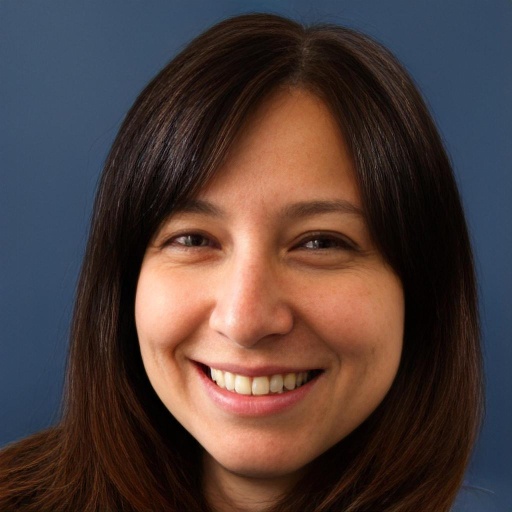} \end{tabular}
&
\begin{tabular}{c} \includegraphics[width=0.105\linewidth]{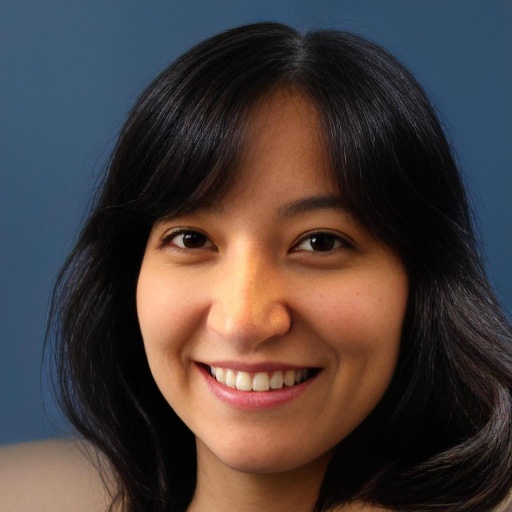} \end{tabular}
\\
\begin{tabular}{c} \includegraphics[width=0.105\linewidth]{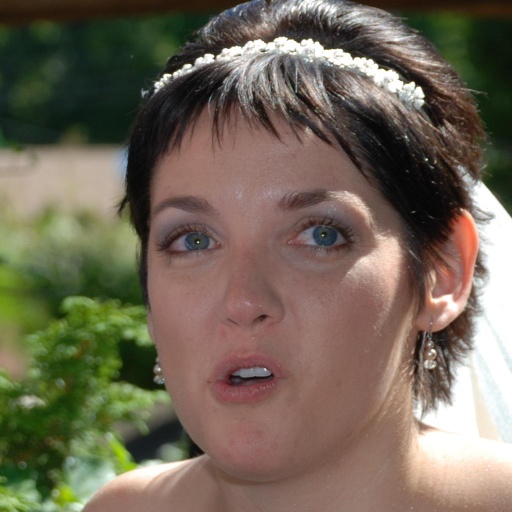} \end{tabular}
&
\begin{tabular}{c} \includegraphics[width=0.105\linewidth]{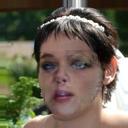} \end{tabular}
&
\begin{tabular}{c} \includegraphics[width=0.105\linewidth]{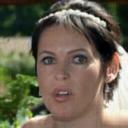} \end{tabular}
&
\begin{tabular}{c} \includegraphics[width=0.105\linewidth]{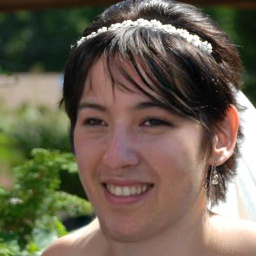} \end{tabular}
&
\begin{tabular}{c} \includegraphics[width=0.105\linewidth]{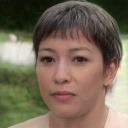} \end{tabular}
&
\begin{tabular}{c} \includegraphics[width=0.105\linewidth]{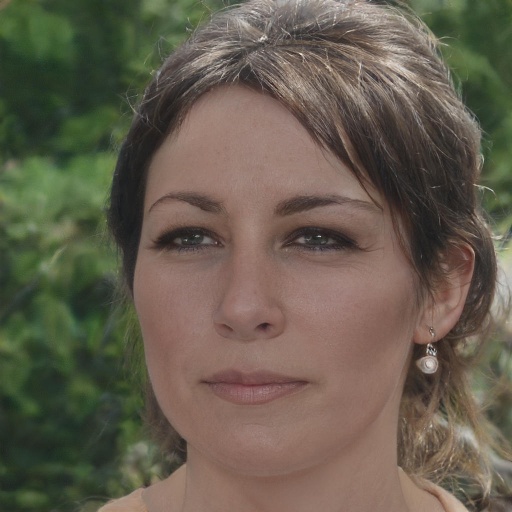} \end{tabular}
&
\begin{tabular}{c} \includegraphics[width=0.105\linewidth]{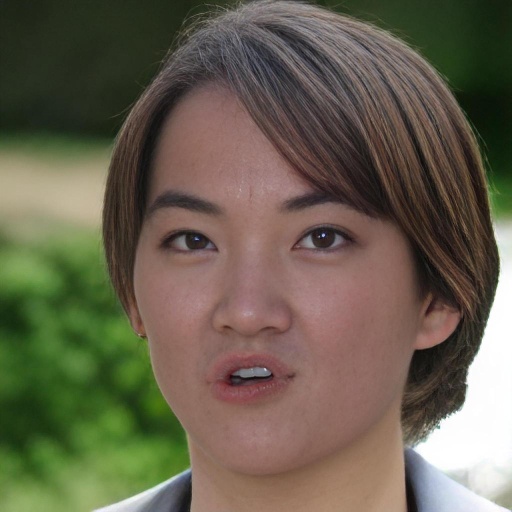} \end{tabular}
&
\begin{tabular}{c} \includegraphics[width=0.105\linewidth]{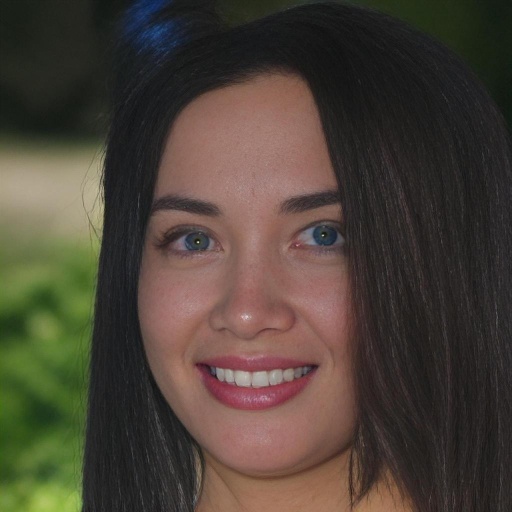} \end{tabular}
&
\begin{tabular}{c} \includegraphics[width=0.105\linewidth]{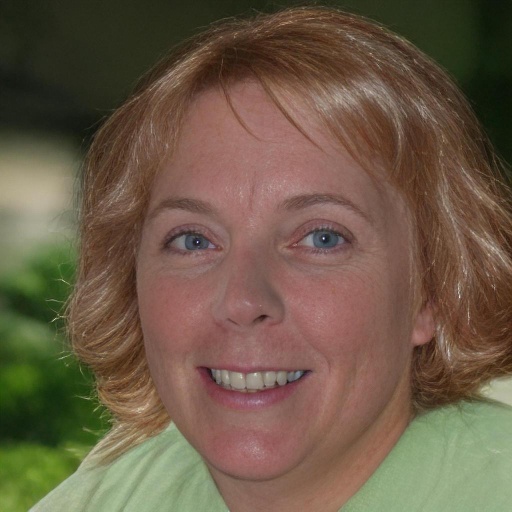} \end{tabular}
\\
\begin{tabular}{c} \includegraphics[width=0.105\linewidth]{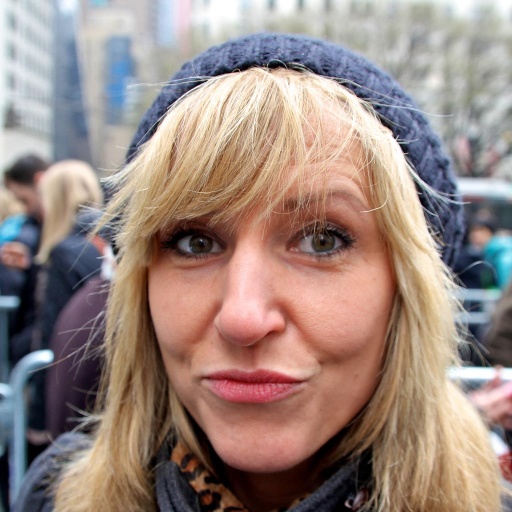} \end{tabular}
&
\begin{tabular}{c} \includegraphics[width=0.105\linewidth]{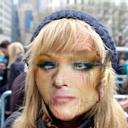} \end{tabular}
&
\begin{tabular}{c} \includegraphics[width=0.105\linewidth]{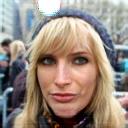} \end{tabular}
&
\begin{tabular}{c} \includegraphics[width=0.105\linewidth]{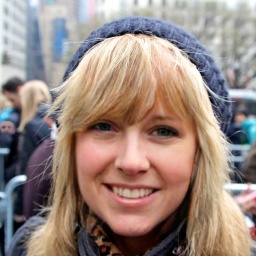} \end{tabular}
&
\begin{tabular}{c} \includegraphics[width=0.105\linewidth]{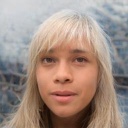} \end{tabular}
&
\begin{tabular}{c} \includegraphics[width=0.105\linewidth]{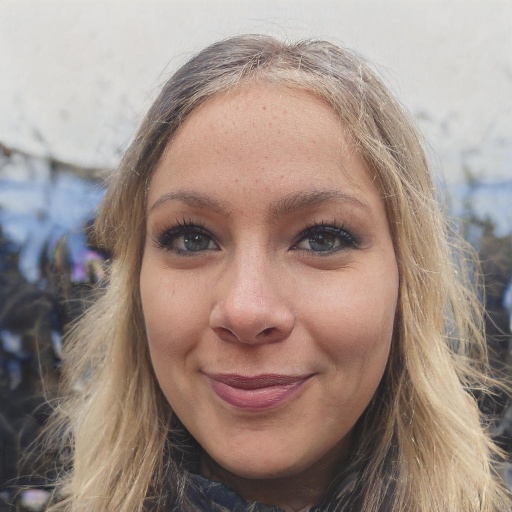} \end{tabular}
&
\begin{tabular}{c} \includegraphics[width=0.105\linewidth]{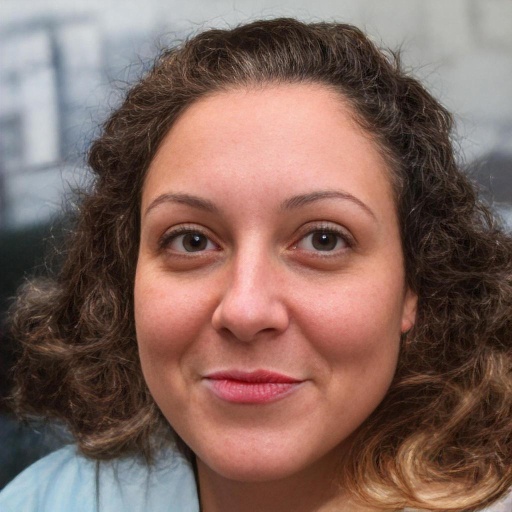} \end{tabular}
&
\begin{tabular}{c} \includegraphics[width=0.105\linewidth]{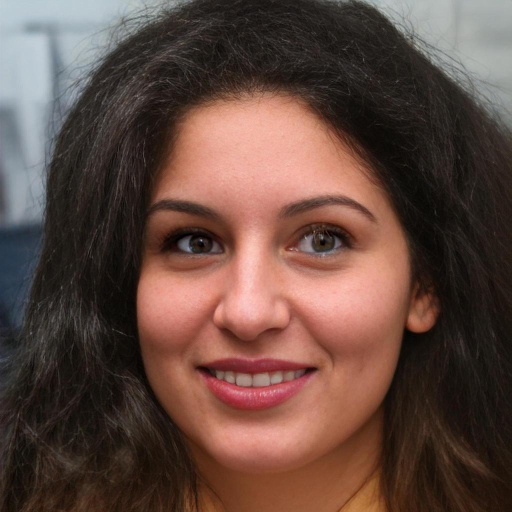} \end{tabular}
&
\begin{tabular}{c} \includegraphics[width=0.105\linewidth]{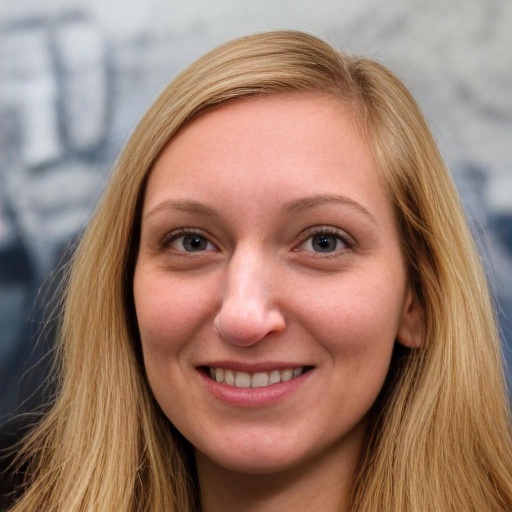} \end{tabular}
\\
\begin{tabular}{c} \includegraphics[width=0.105\linewidth]{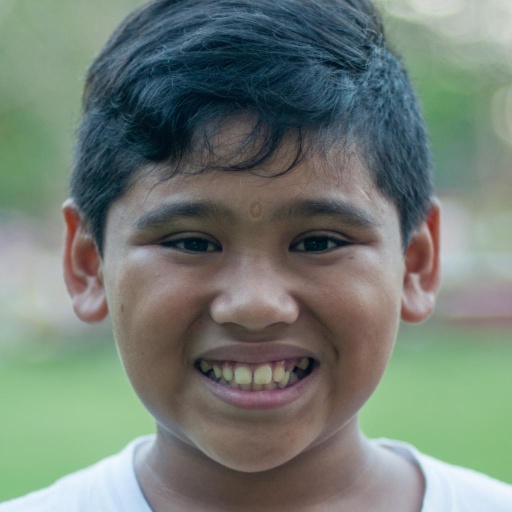} \end{tabular}
&
\begin{tabular}{c} \includegraphics[width=0.105\linewidth]{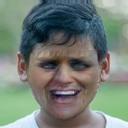} \end{tabular}
&
\begin{tabular}{c} \includegraphics[width=0.105\linewidth]{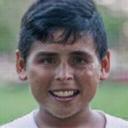} \end{tabular}
&
\begin{tabular}{c} \includegraphics[width=0.105\linewidth]{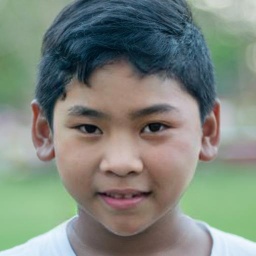} \end{tabular}
&
\begin{tabular}{c} \includegraphics[width=0.105\linewidth]{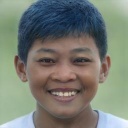} \end{tabular}
&
\begin{tabular}{c} \includegraphics[width=0.105\linewidth]{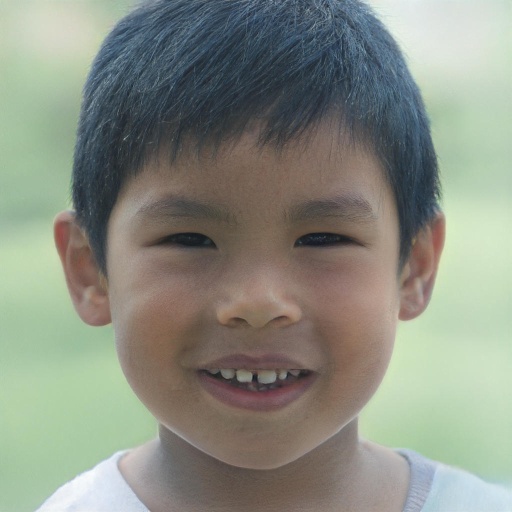} \end{tabular}
&
\begin{tabular}{c} \includegraphics[width=0.105\linewidth]{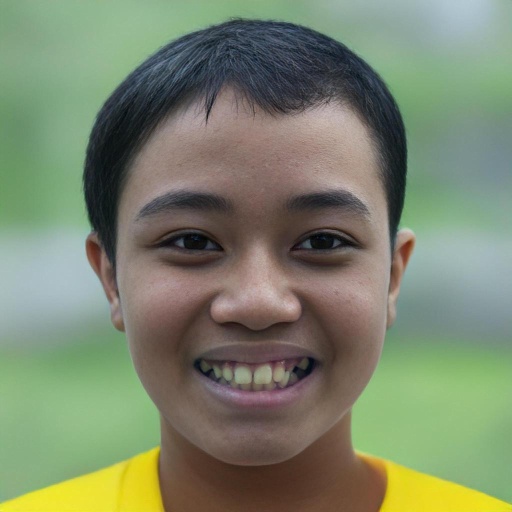} \end{tabular}
&
\begin{tabular}{c} \includegraphics[width=0.105\linewidth]{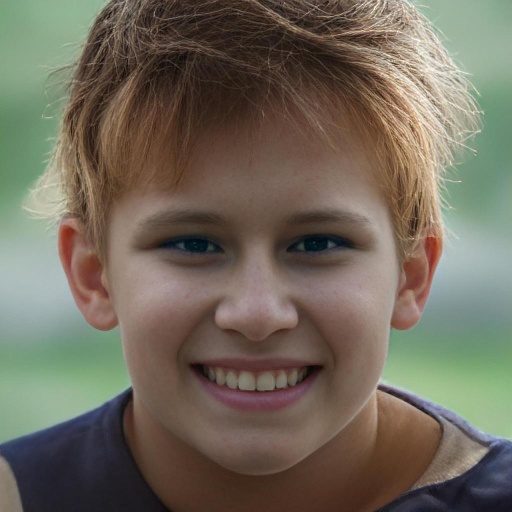} \end{tabular}
&
\begin{tabular}{c} \includegraphics[width=0.105\linewidth]{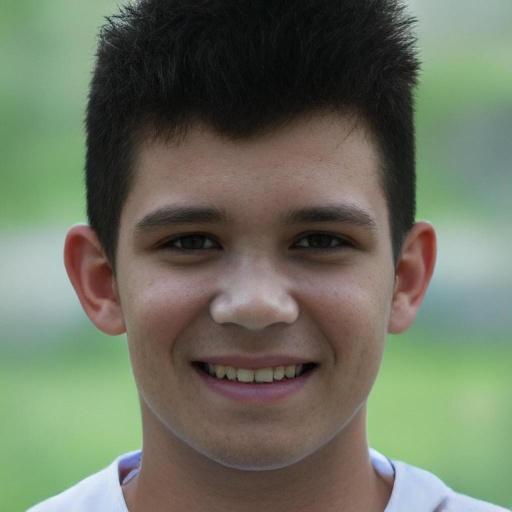} \end{tabular}
\\
\begin{tabular}{c} \includegraphics[width=0.105\linewidth]{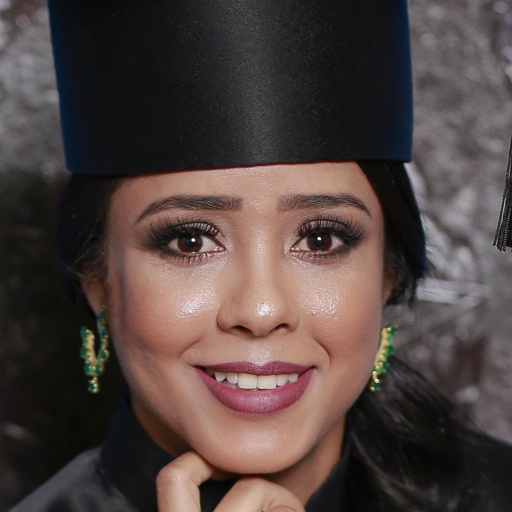} \end{tabular}
&
\begin{tabular}{c} \includegraphics[width=0.105\linewidth]{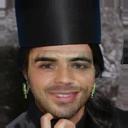} \end{tabular}
&
\begin{tabular}{c} \includegraphics[width=0.105\linewidth]{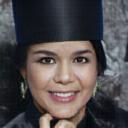} \end{tabular}
&
\begin{tabular}{c} \includegraphics[width=0.105\linewidth]{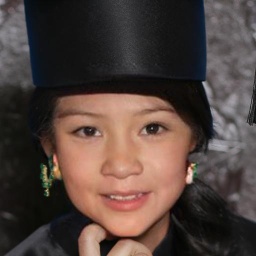} \end{tabular}
&
\begin{tabular}{c} \includegraphics[width=0.105\linewidth]{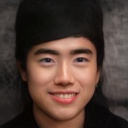} \end{tabular}
&
\begin{tabular}{c} \includegraphics[width=0.105\linewidth]{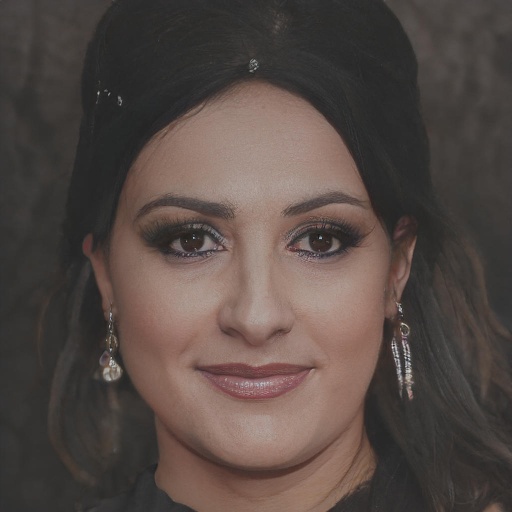} \end{tabular}
&
\begin{tabular}{c} \includegraphics[width=0.105\linewidth]{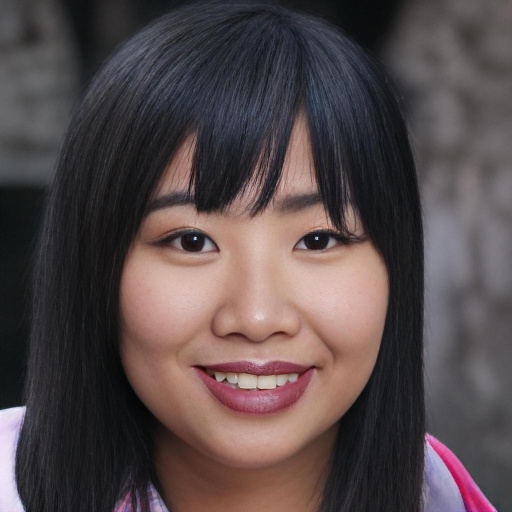} \end{tabular}
&
\begin{tabular}{c} \includegraphics[width=0.105\linewidth]{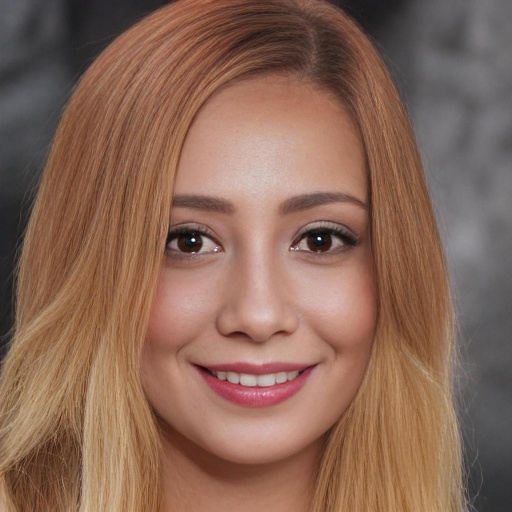} \end{tabular}
&
\begin{tabular}{c} \includegraphics[width=0.105\linewidth]{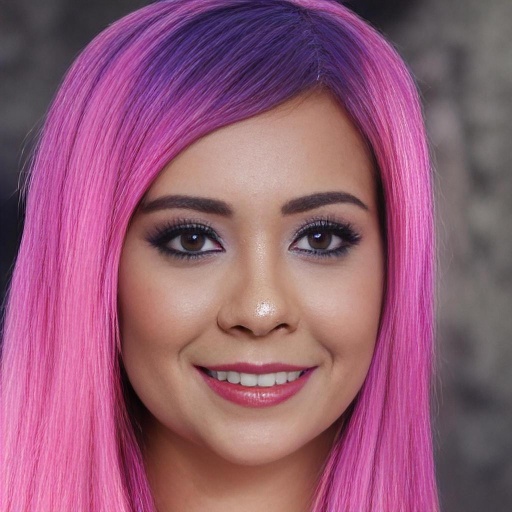} \end{tabular}
\\
\begin{tabular}{c} \makebox[0.105\linewidth]{Input} \end{tabular}
&
\begin{tabular}{c} \makebox[0.105\linewidth]{\makecell{CIAGAN \\ ~\cite{maximov2020ciagan}}} \end{tabular}
&
\begin{tabular}{c} \makebox[0.105\linewidth]{\makecell{FIT \\ ~\cite{gu2020password}}} \end{tabular}
&
\begin{tabular}{c} \makebox[0.105\linewidth]{\makecell{DP2 \\ ~\cite{hukkelaas2023deepprivacy2}}} \end{tabular}
&
\begin{tabular}{c} \makebox[0.105\linewidth]{\makecell{RiDDLE \\ ~\cite{li2023riddle}}} \end{tabular}
&
\begin{tabular}{c} \makebox[0.105\linewidth]{\makecell{FALCO \\ ~\cite{barattin2023attribute}}} \end{tabular}
&
\begin{tabular}{c} \makebox[0.105\linewidth]{\makecell{Ours \\ (mouth)}} \end{tabular}
&
\begin{tabular}{c} \makebox[0.105\linewidth]{\makecell{Ours \\ (eyes)}} \end{tabular}
&
\begin{tabular}{c} \makebox[0.105\linewidth]{\makecell{Ours \\ (nose)}} \end{tabular}
\\
\end{tabular}
\caption{Extensive qualitative evaluation results on the \textit{clinical} single-image anonymization, benchmarking against the two most commonly referenced anonymization methods and the four most recent state-of-the-art anonymization approaches, on FFHQ~\cite{karras2019style} test samples.}
\label{fig:single_semantic_ffhq}
\end{figure*}
\endgroup

\subsection{Standard single-image anonymization}
We provide extended benchmarking results of standard single-image anonymization in Fig.~\ref{fig:anon_inplace_celebahq} on the CelebAMaskHQ~\cite{lee2020maskgan} test set and in Fig.~\ref{fig:anon_inplace_ffhq} on the FFHQ~\cite{karras2019style} test set. We compare against the two most commonly referenced baselines CIAGAN~\cite{maximov2020ciagan} and FIT~\cite{gu2020password}, and the three most recent state-of-the-art methods that have public code available; DP2~\cite{hukkelaas2023deepprivacy2}, RiDDLE~\cite{li2023riddle} and FALCO~\cite{barattin2023attribute}. Our VerA results are the most photorealistic, consistently de-identifying the person, even on this setting of standard single-image anonymization. 

We make a note regarding the results of FALCO~\cite{barattin2023attribute}. As mentioned in our main text, FALCO performs an adaptive normalization that can lead to washed out images, or to odd color artifacts if toggled off. We follow the authors’ default setting and leave it activated in all our experiments. We illustrate this normalization's effect in Fig.~\ref{fig:falco_norm}.

\subsection{Clinical single-image anonymization}
We provide further benchmarking results on clinical single-image anonymization in Fig.~\ref{fig:single_semantic_ffhq}, using images from our test set in FFHQ~\cite{karras2019style}. We compare against the same set of methods, and include our clinical anonymization results that preserve the mouth, eyes, and nose, respectively. We also provide extended quantitative evaluation on semantic preservation, conducted on FFHQ~\cite{karras2019style}, in Table~\ref{table:semantic_preservation_ffhq}. All results support the same claims we make in our main manuscript. We further provide example results of competing methods, to which we add our own blending procedure in Fig.~\ref{fig:inpainted_baselines}. Other methods do not directly tackle clinical anonymization with a semantic region-of-interest, and their results with our added blending are less photorealistic. We also show our own output when we do not perform our blending for illustration purposes.

\subsection{Paired standard and clinical anonymization}
We provide further examples of paired anonymization, in both the standard and clinical setting, and comparing to all benchmarks in Fig.~\ref{fig:anon_siblings_paired} on a pair from the SiblingsDB dataset~\cite{vieira2014detecting}. We additionally provide images at higher resolution for illustrating a paired clinical anonymization example in Fig.~\ref{fig:paired_semantic_big}.

\subsection{Full-image in-place anonymization}
Fig.~\ref{fig:anon_inplace_full} shows numerous examples of full images that we anonymize using VerA. These examples serve as an illustration of the application to full-scene images, aside from the clinical use cases that our main manuscript focuses on.

\subsection{Downstream utility evaluation}
We repeat the downstream utility evaluation presented in our main text on the FFHQ~\cite{karras2019style} set, and compile the results in Table~\ref{table:utility_ffhq}. We achieve the best photorealism and diverse distribution measured by FID, followed by DP2. As for bounding boxes and face detection, we are on par with other state-of-the-art methods, all achieving significantly high performance. The results echo what we present and the conclusions in our main manuscript.

\begingroup
\setlength{\tabcolsep}{0.5pt} 
\renewcommand{\arraystretch}{0.7} 

\begin{figure*}[t]
\centering
\begin{tabular}{ccccc}

\begin{tabular}{c} \includegraphics[width=0.195\linewidth]{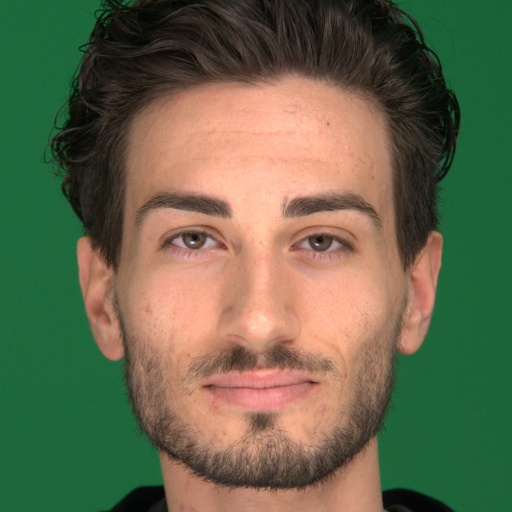} \end{tabular}
&
\begin{tabular}{c} \includegraphics[width=0.195\linewidth]{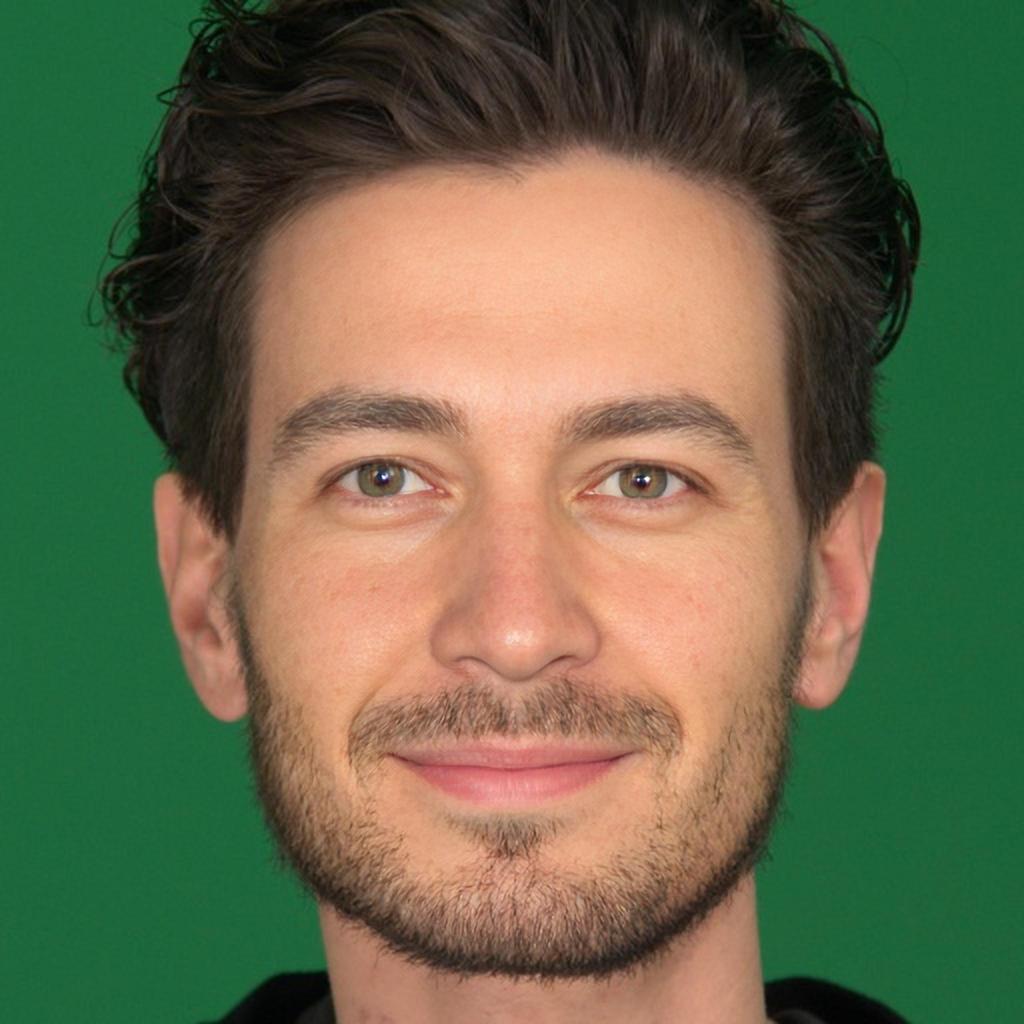} \end{tabular}
&
\begin{tabular}{c} \includegraphics[width=0.195\linewidth]{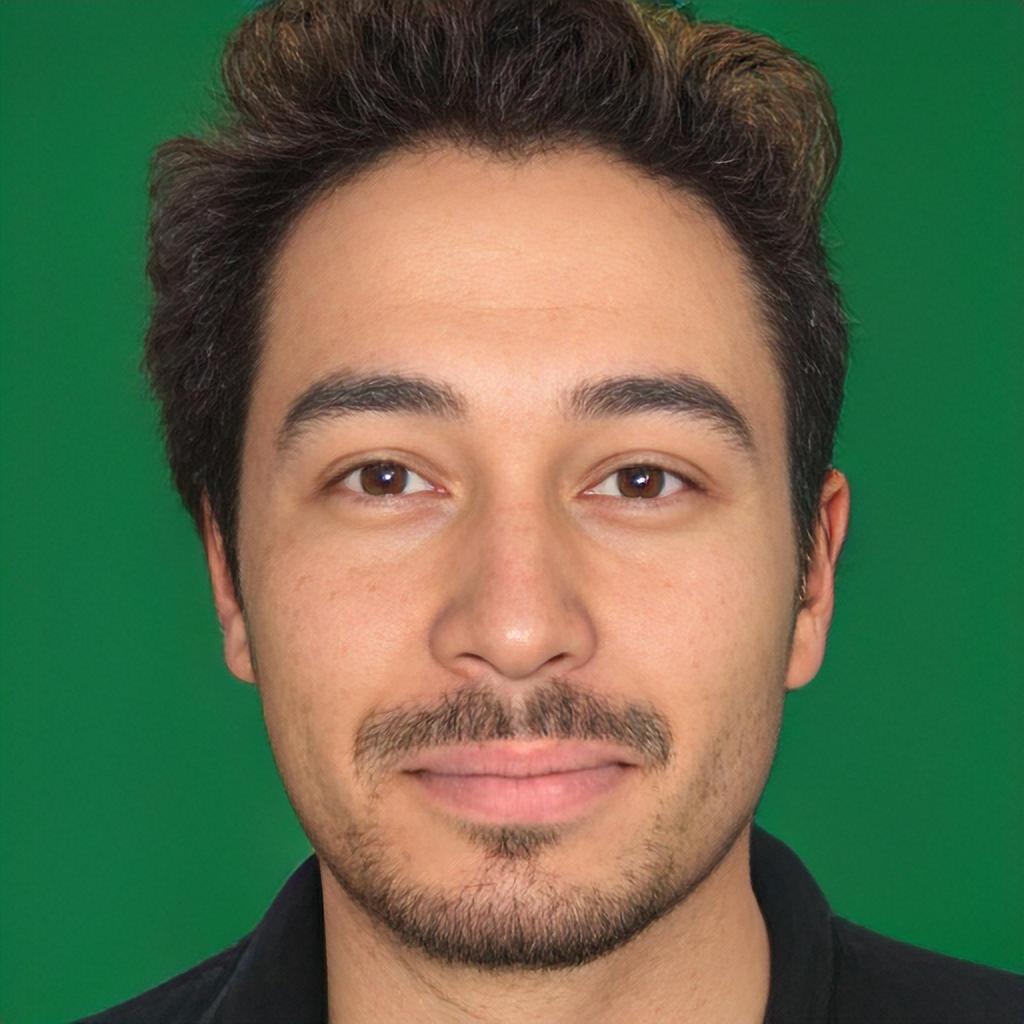} \end{tabular}
&
\begin{tabular}{c} \includegraphics[width=0.195\linewidth]{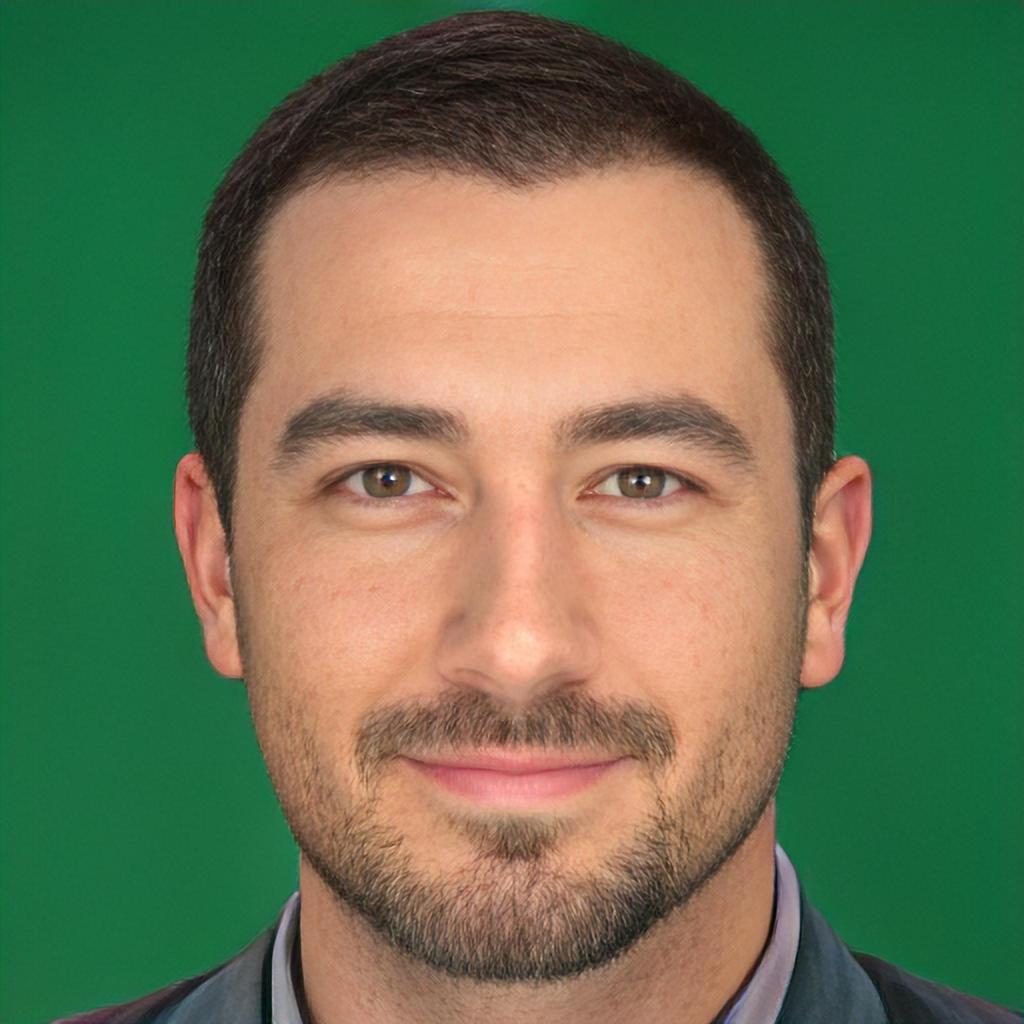} \end{tabular}
&
\begin{tabular}{c} \includegraphics[width=0.195\linewidth]{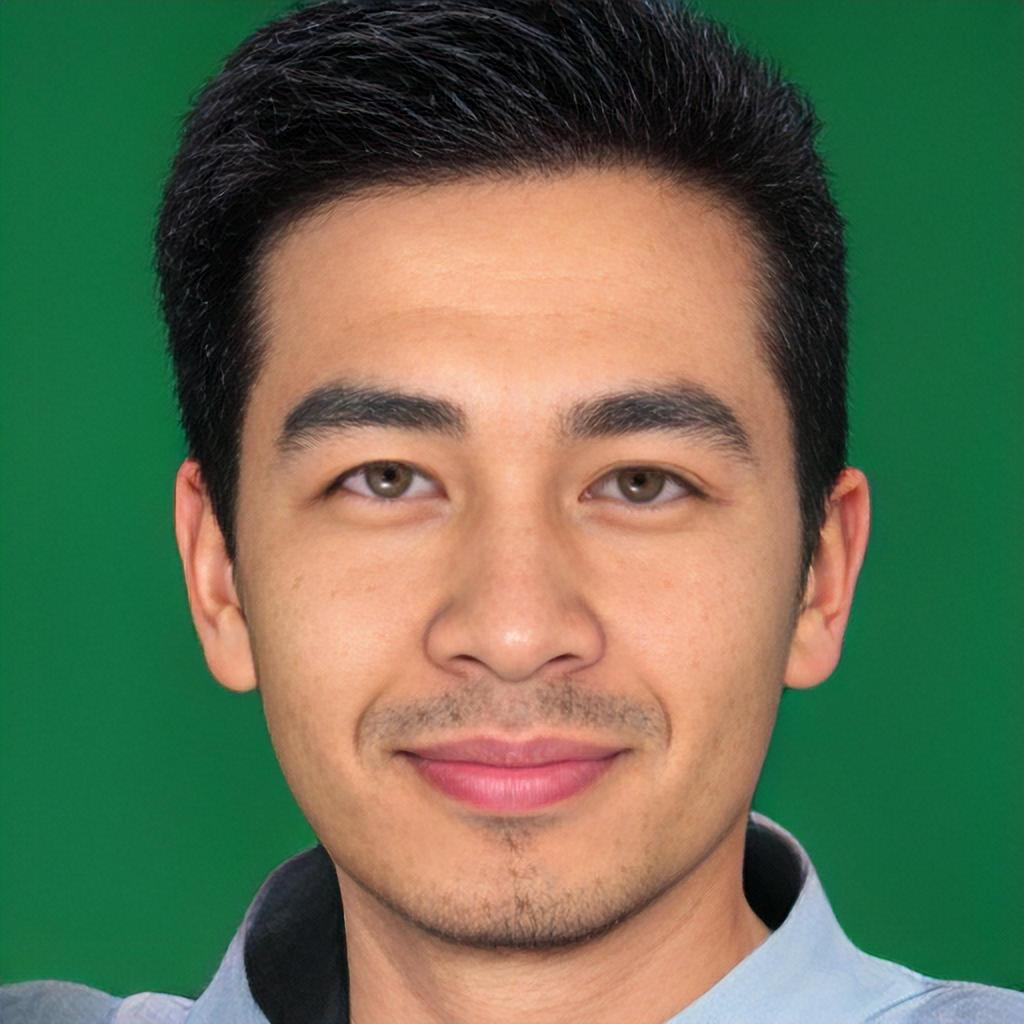} \end{tabular}
\\
\begin{tabular}{c} \includegraphics[width=0.195\linewidth]{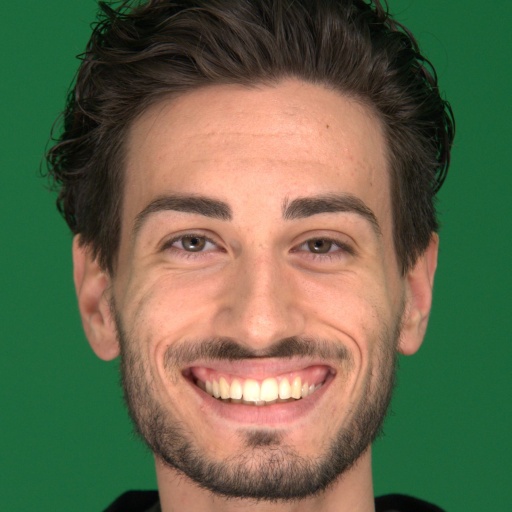} \end{tabular}
&
\begin{tabular}{c} \includegraphics[width=0.195\linewidth]{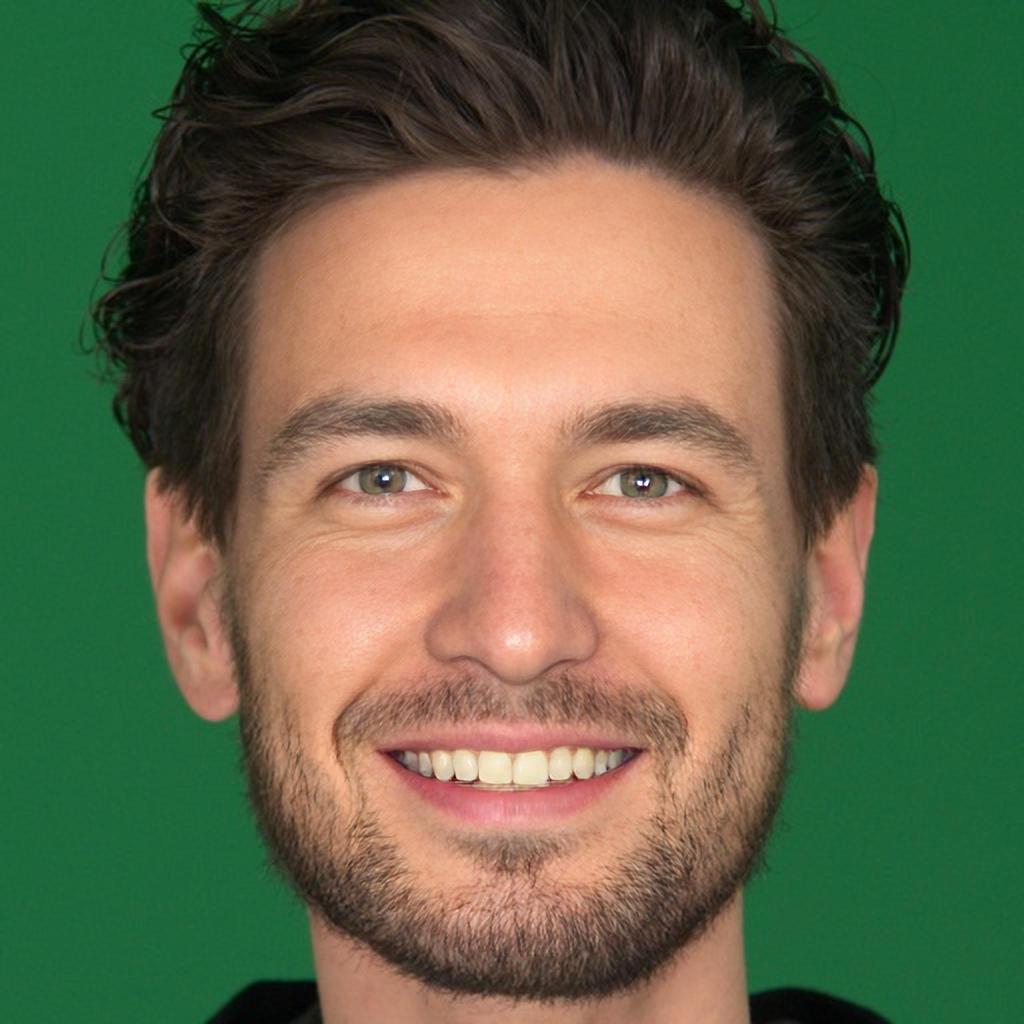} \end{tabular}
&
\begin{tabular}{c} \includegraphics[width=0.195\linewidth]{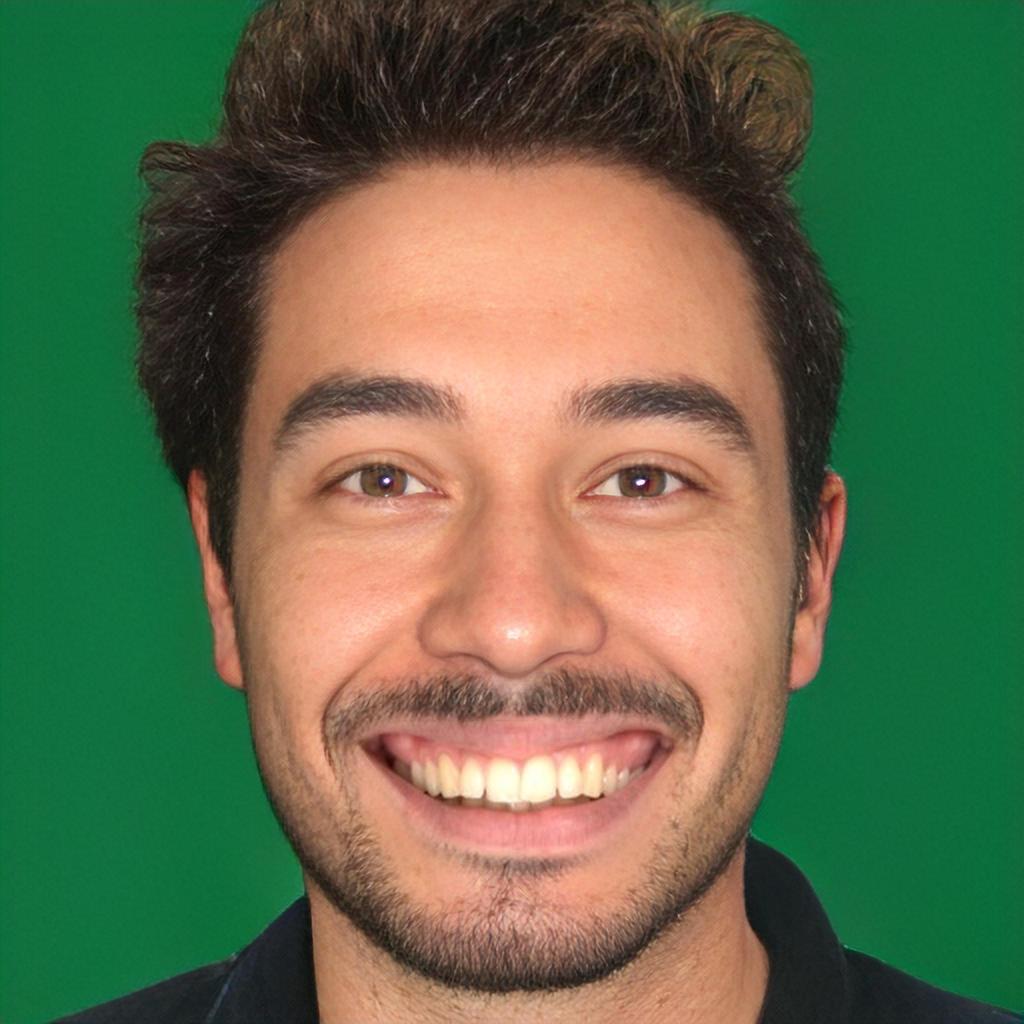} \end{tabular}
&
\begin{tabular}{c} \includegraphics[width=0.195\linewidth]{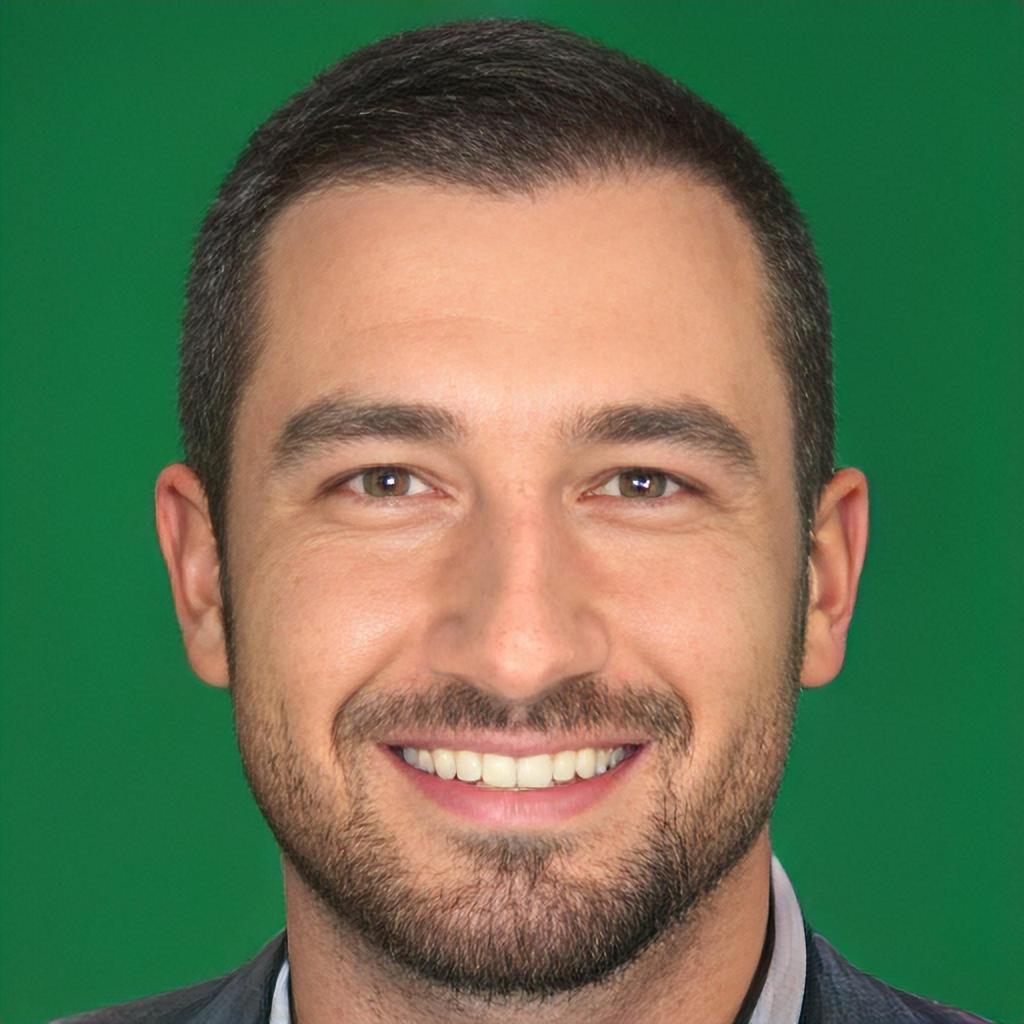} \end{tabular}
&
\begin{tabular}{c} \includegraphics[width=0.195\linewidth]{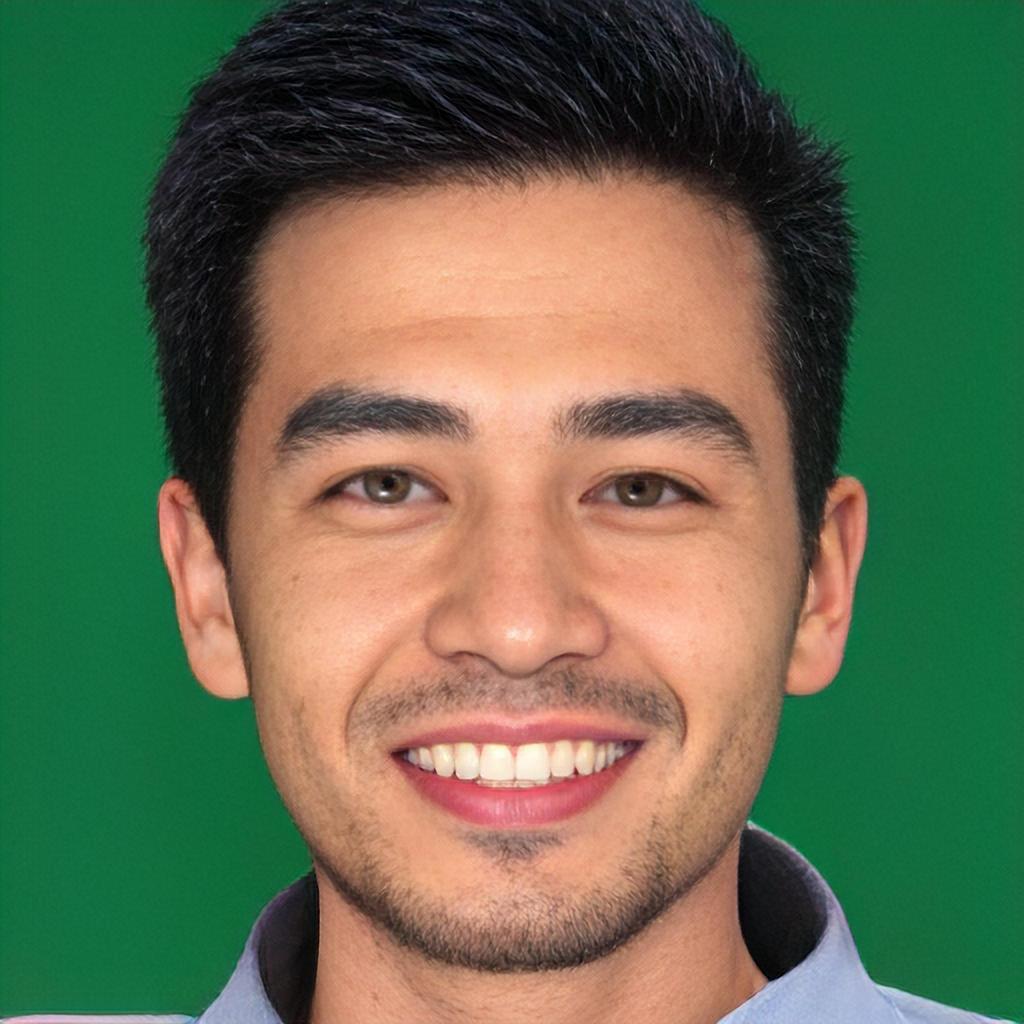} \end{tabular}
\\
\begin{tabular}{c} \makebox[0.195\linewidth]{Input} \end{tabular}
&
\begin{tabular}{c} \makebox[0.195\linewidth]{Ours (standard)} \end{tabular}
&
\begin{tabular}{c} \makebox[0.195\linewidth]{Ours (mouth)} \end{tabular}
&
\begin{tabular}{c} \makebox[0.195\linewidth]{Ours (nose)} \end{tabular}
&
\begin{tabular}{c} \makebox[0.195\linewidth]{Ours (eyes)} \end{tabular}
\\

\begin{tabular}{c} \includegraphics[width=0.195\linewidth]{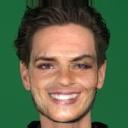} \end{tabular}
&
\begin{tabular}{c} \includegraphics[width=0.195\linewidth]{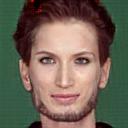} \end{tabular}
&
\begin{tabular}{c} \includegraphics[width=0.195\linewidth]{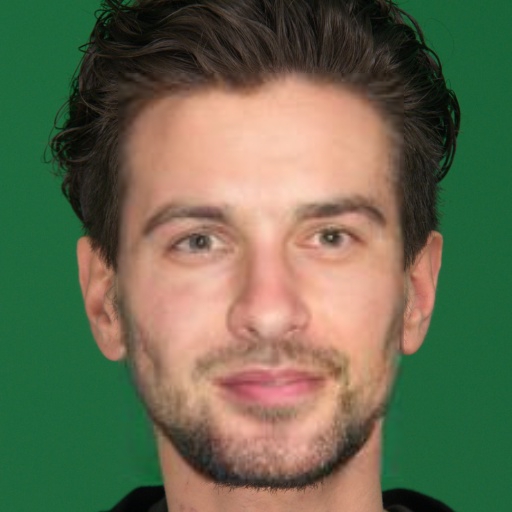} \end{tabular}
&
\begin{tabular}{c} \includegraphics[width=0.195\linewidth]{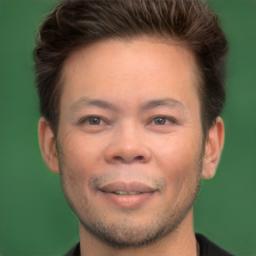} \end{tabular}
&
\begin{tabular}{c} \includegraphics[width=0.195\linewidth]{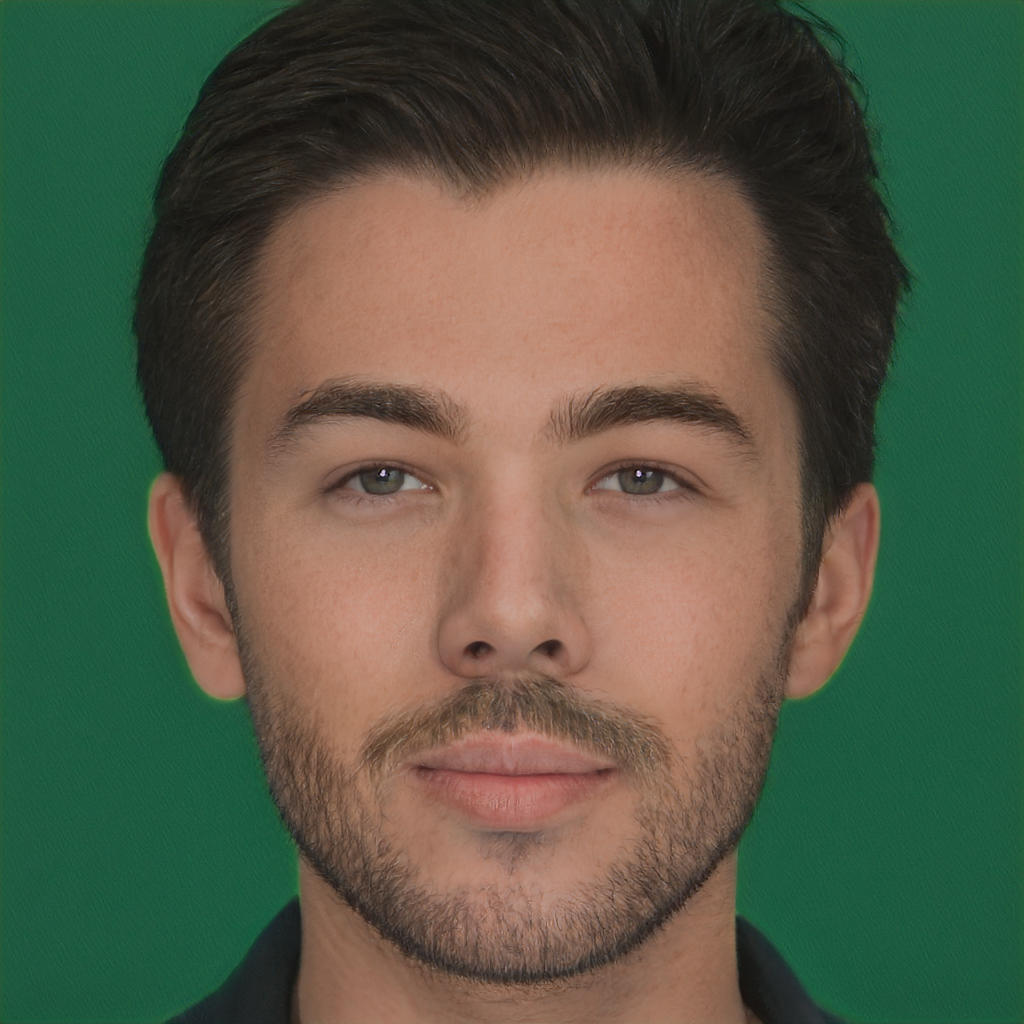} \end{tabular}
\\
\begin{tabular}{c} \includegraphics[width=0.195\linewidth]{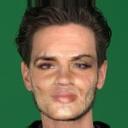} \end{tabular}
&
\begin{tabular}{c} \includegraphics[width=0.195\linewidth]{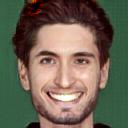} \end{tabular}
&
\begin{tabular}{c} \includegraphics[width=0.195\linewidth]{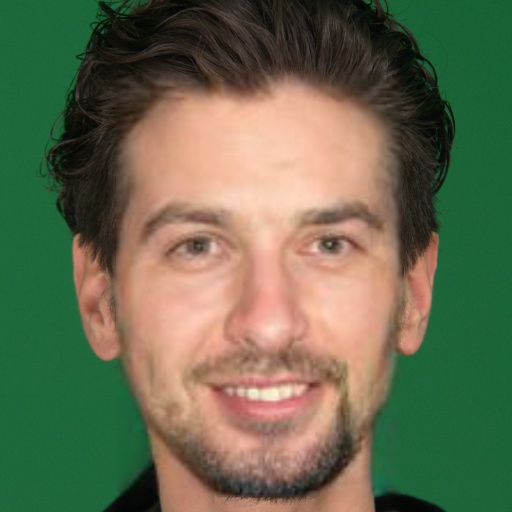} \end{tabular}
&
\begin{tabular}{c} \includegraphics[width=0.195\linewidth]{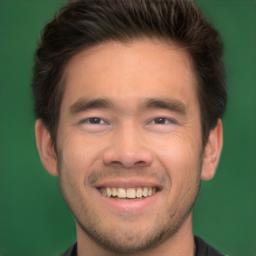} \end{tabular}
&
\begin{tabular}{c} \includegraphics[width=0.195\linewidth]{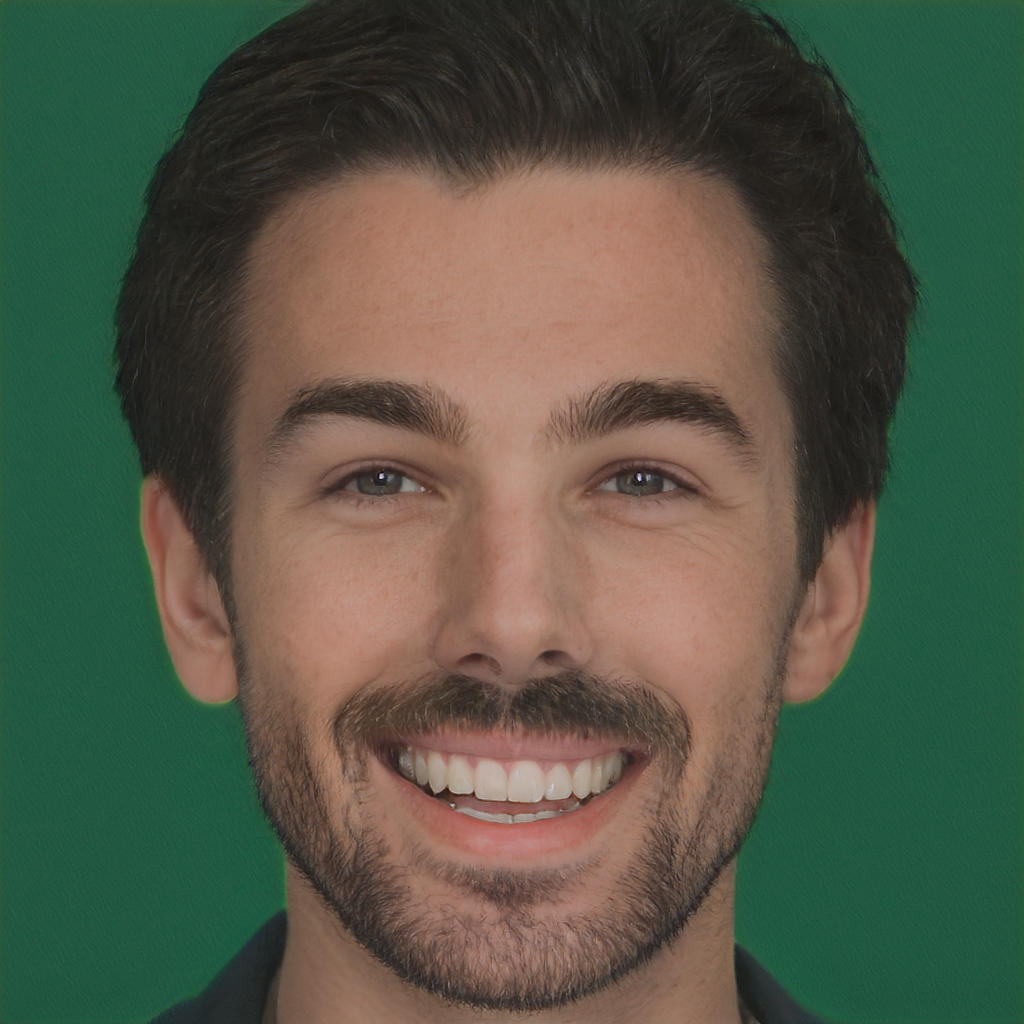} \end{tabular}
\\
\begin{tabular}{c} \makebox[0.195\linewidth]{CIAGAN~\cite{maximov2020ciagan}} \end{tabular}
&
\begin{tabular}{c} \makebox[0.195\linewidth]{FIT~\cite{gu2020password}} \end{tabular}
&
\begin{tabular}{c} \makebox[0.195\linewidth]{DP2~\cite{hukkelaas2023deepprivacy2}} \end{tabular}
&
\begin{tabular}{c} \makebox[0.195\linewidth]{RiDDLE~\cite{li2023riddle}} \end{tabular}
&
\begin{tabular}{c} \makebox[0.195\linewidth]{FALCO~\cite{barattin2023attribute}} \end{tabular}
\\
\end{tabular}
\caption{An example pair taken from the SiblingsDB dataset~\cite{vieira2014detecting} that we anonymize under the \textit{standard} setting and the \textit{clinical} setting preserving the \textit{mouth, nose}, and \textit{eyes} respectively. We compare our results to those of the five benchmarks. Note that in this setting DP2 achieves good identity consistency within the pair. This is because of the highly standardized capture, and the fact that DP2 inpaints the inside of the face conditioned on the outside, which in this setting is almost unchanged.}
\label{fig:anon_siblings_paired}
\end{figure*}

\endgroup

\section{Ablation experiments}

\subsection{Mirroring and projection heads ablation}
We perform a simple ablation study over our proposed contrastive mirroring strategy and over our projection heads that are learned on top of each pretrained high-level encoder. We provide the results in Fig.~\ref{fig:ablation}. The training contrastive loss for the pose high-level attribute is shown in the top plot. Each curve corresponds to training with our mirroring strategy and projection heads, as well as with the ablation of each. We obtain the best convergence when both components are included. This improved convergence results in improved high-level control, as illustrated visually in the bottom part of the figure. In every row, we sample multiple identities that have a fixed pose latent. Only the last row, training with both proposed strategies, consistently achieves the same pose.

\subsection{Prior-based blending and correction ablation}
The effects of prior-based blending and correction are hard to quantify, since evaluating the photorealism of images (or parts of images) is an open task, tackled only through proxy metrics. We once again use FID~\cite{heusel2017gans}, accompanied with the downstream utility metrics we compute using face detection models. We perform the ablation study over the postprocessing on the standard anonymization task and provide the results in Tab.~\ref{table:postproc_ablation_utility}. We also illustrate the separate and cumulative effects of both steps in Fig. \ref{fig:ablation_postproc}, on a sample from our validation set. Face detection models get minimally affected by the changes since only local and low-level features change through the postprocessing and image structures stay mostly similar. We can observe the effects of the postprocessing on FID values on CelebAHQ, but the FID computation with respect to an external dataset (FFHQ, in this example) fails to illustrate the changes in the images. Such small changes are not significant enough to alter the feature distributions with respect to another dataset.

\begingroup
\setlength{\tabcolsep}{0.5pt} 
\renewcommand{\arraystretch}{0.7} 

\begin{figure*}[t]
\centering
\begin{tabular}{ccccc}

\begin{tabular}{c} \includegraphics[width=0.195\linewidth]{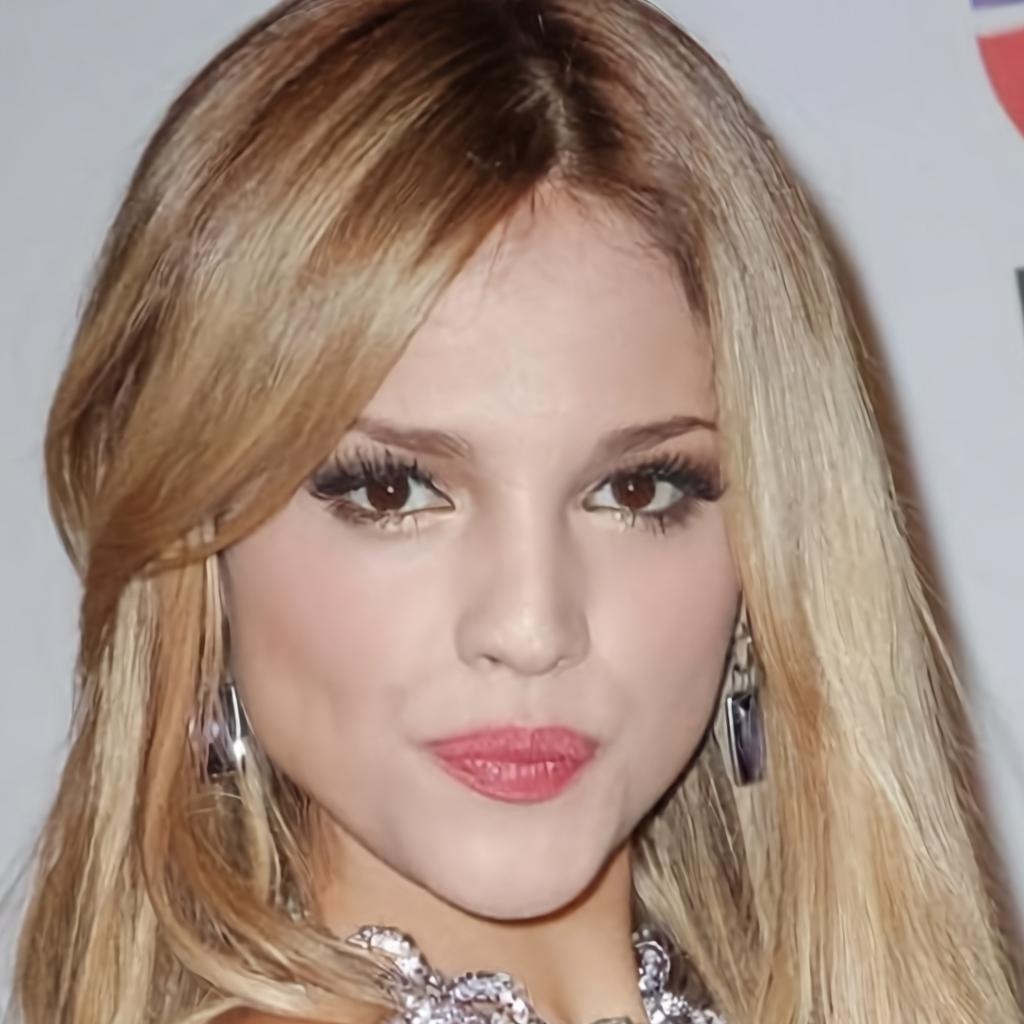} \end{tabular}
&
\begin{tabular}{c} \includegraphics[width=0.195\linewidth]{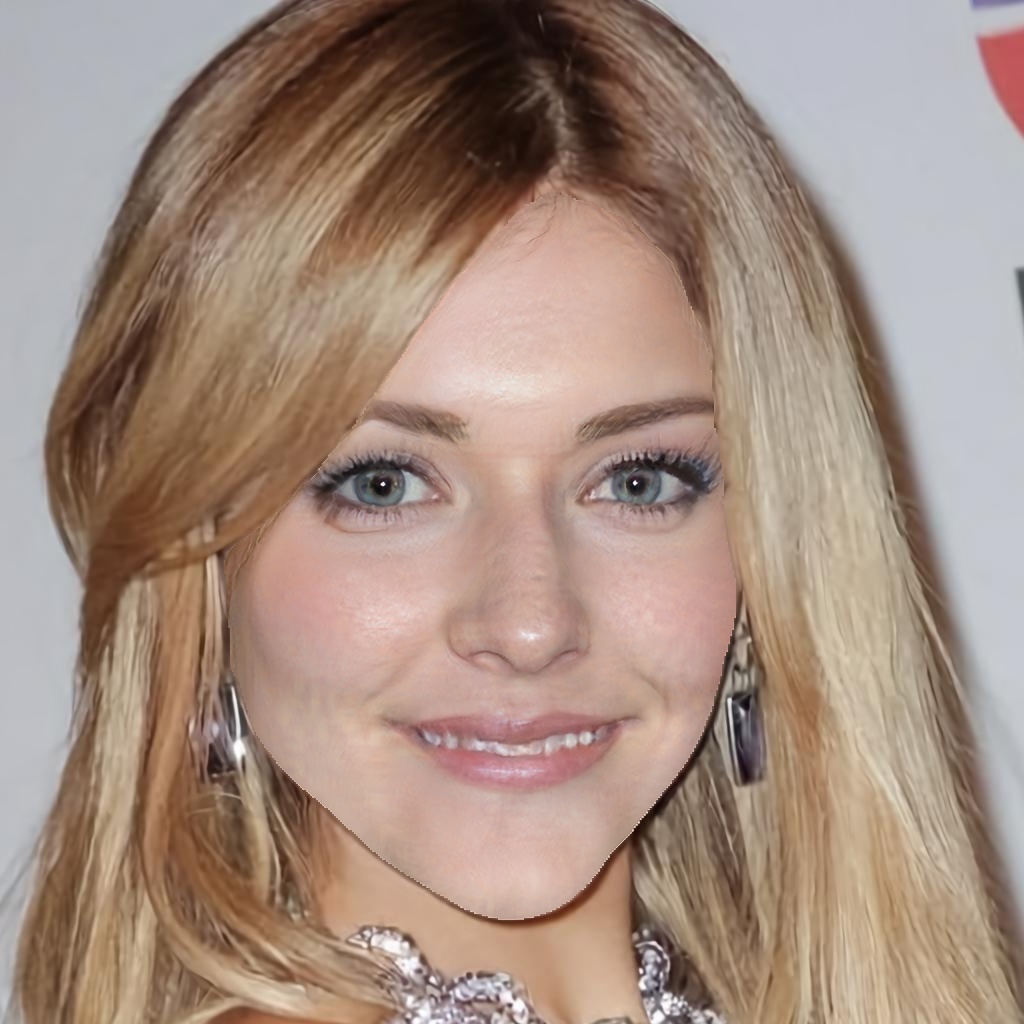} \end{tabular}
&
\begin{tabular}{c} \includegraphics[width=0.195\linewidth]{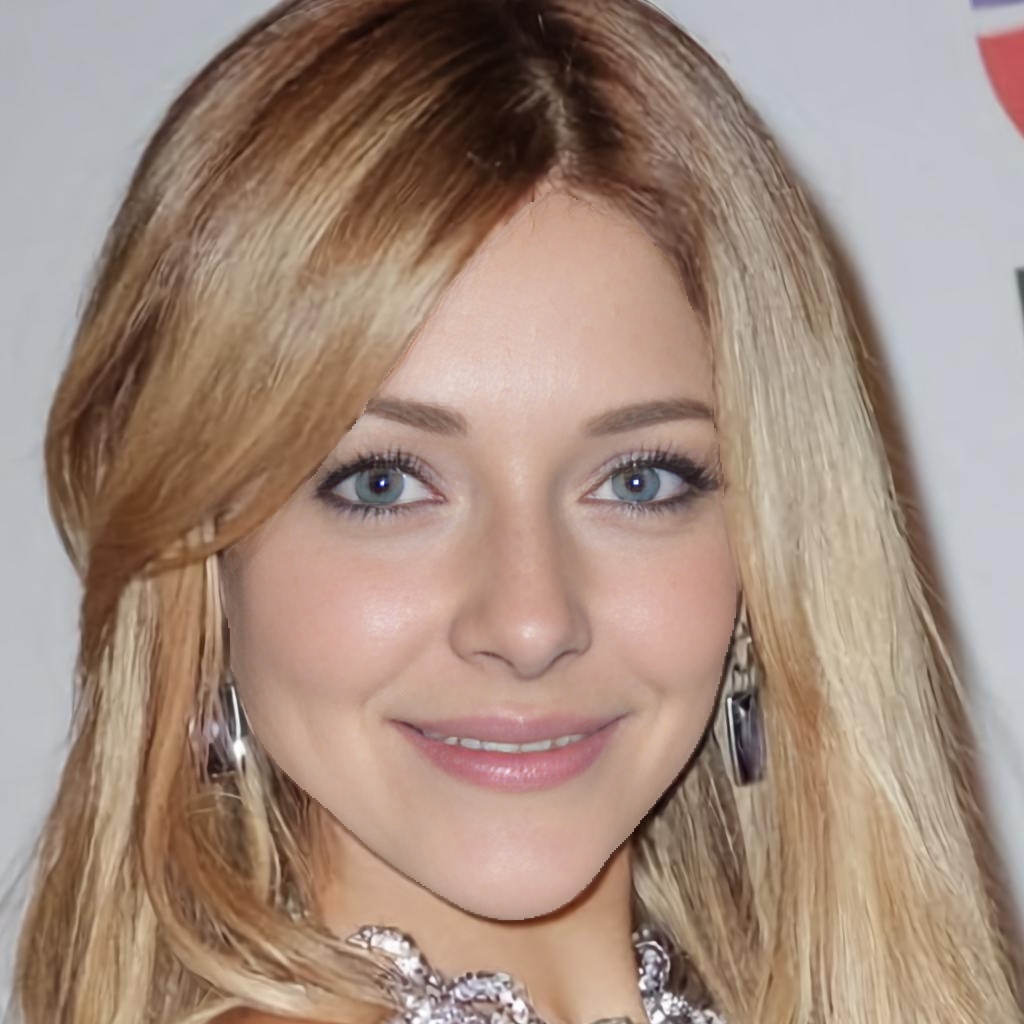} \end{tabular}
&
\begin{tabular}{c} \includegraphics[width=0.195\linewidth]{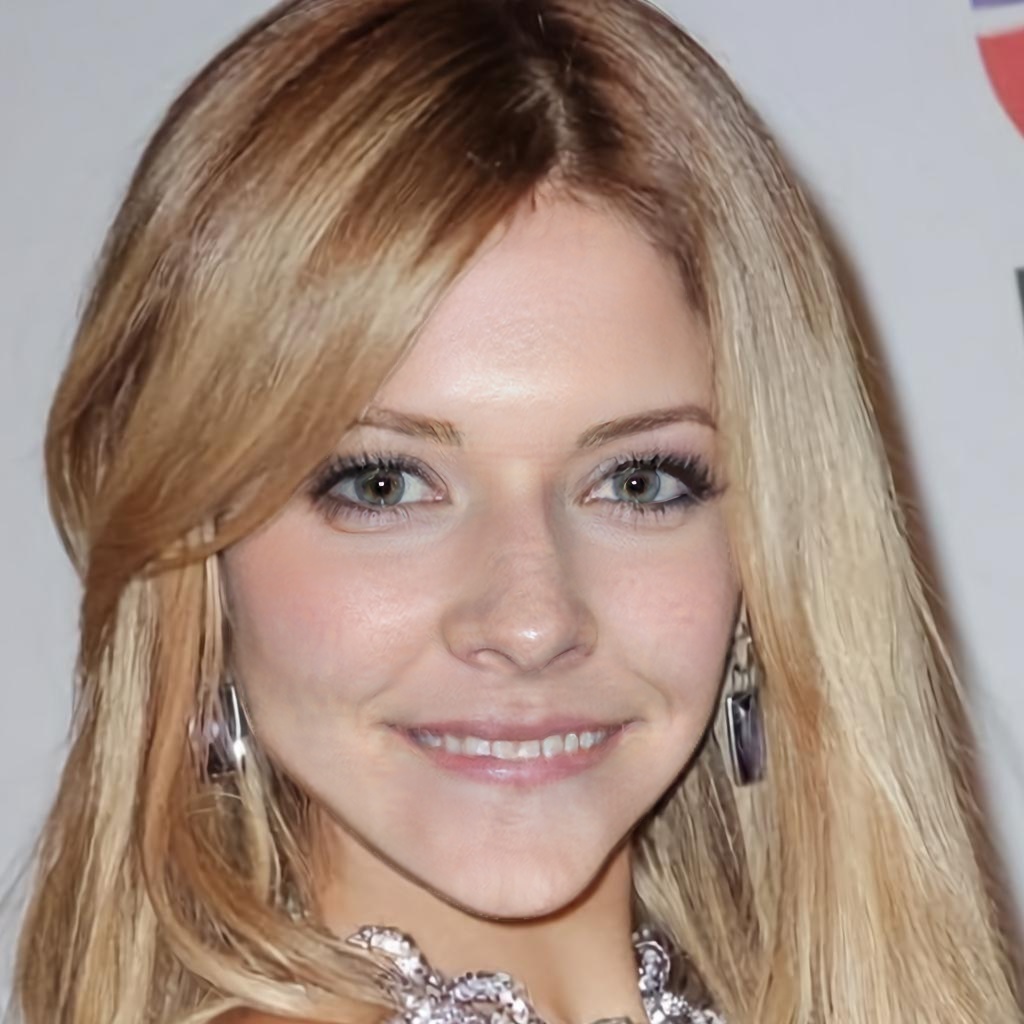} \end{tabular}
&
\begin{tabular}{c} \includegraphics[width=0.195\linewidth]{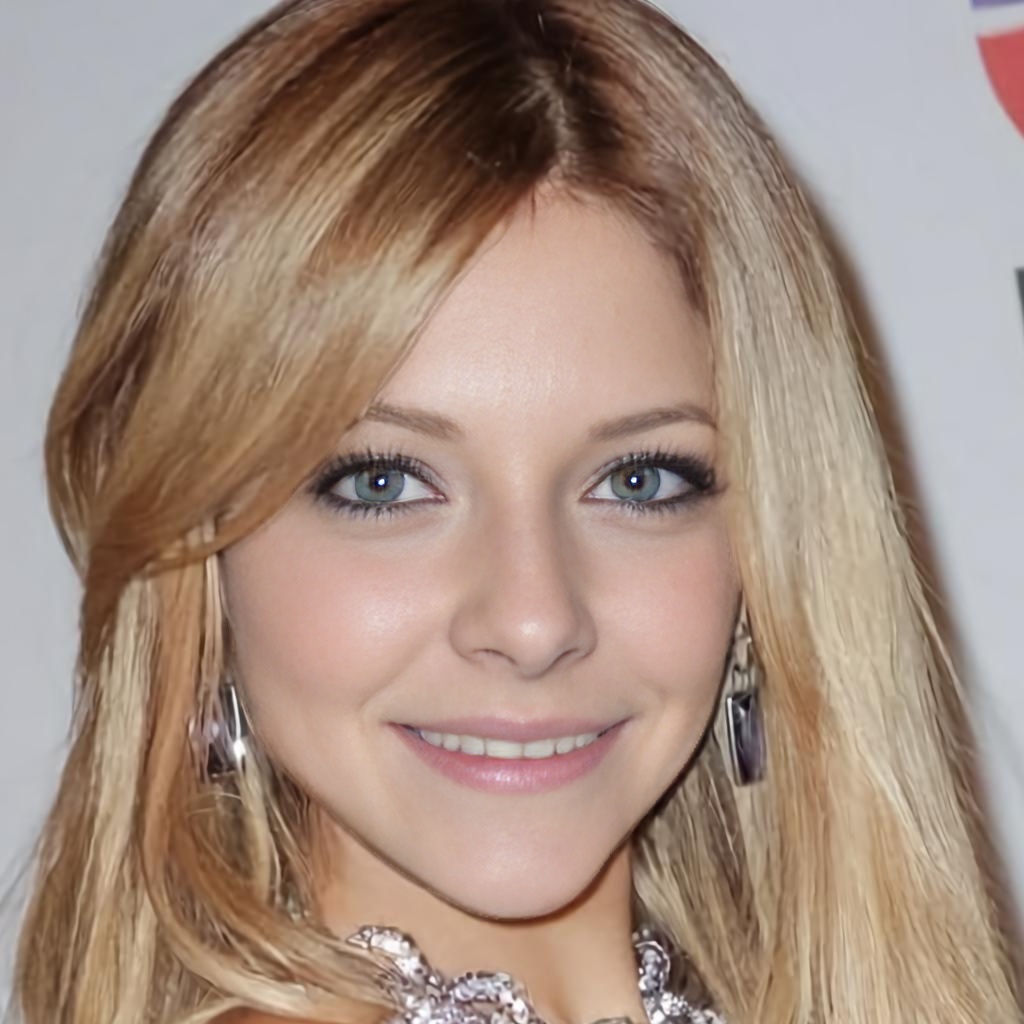} \end{tabular}
\\
\begin{tabular}{c} \makebox[0.195\linewidth]{Input} \end{tabular}
&
\begin{tabular}{c} \makebox[0.195\linewidth]{No postprocessing} \end{tabular}
&
\begin{tabular}{c} \makebox[0.195\linewidth]{Only correction} \end{tabular}
&
\begin{tabular}{c} \makebox[0.195\linewidth]{Only blending} \end{tabular}
&
\begin{tabular}{c} \makebox[0.195\linewidth]{Correction \& blending} \end{tabular}
\\
\end{tabular}
\caption{A sample from our ablation experiment over prior-based blending and correction. Prior-based blending fixes the border issues between preserved and non-preserved regions of the image, whereas prior-based correction fixes occasional GAN artifacts and improves texture. Neither step changes the overall structure of the image.}
\label{fig:ablation_postproc}
\end{figure*}

\endgroup

\begingroup
\setlength{\tabcolsep}{0.5pt} 
\renewcommand{\arraystretch}{0.7} 

\begin{figure}[t]
\centering
\begin{tabular}{cc}
\begin{tabular}{c} \includegraphics[width=0.49\linewidth]{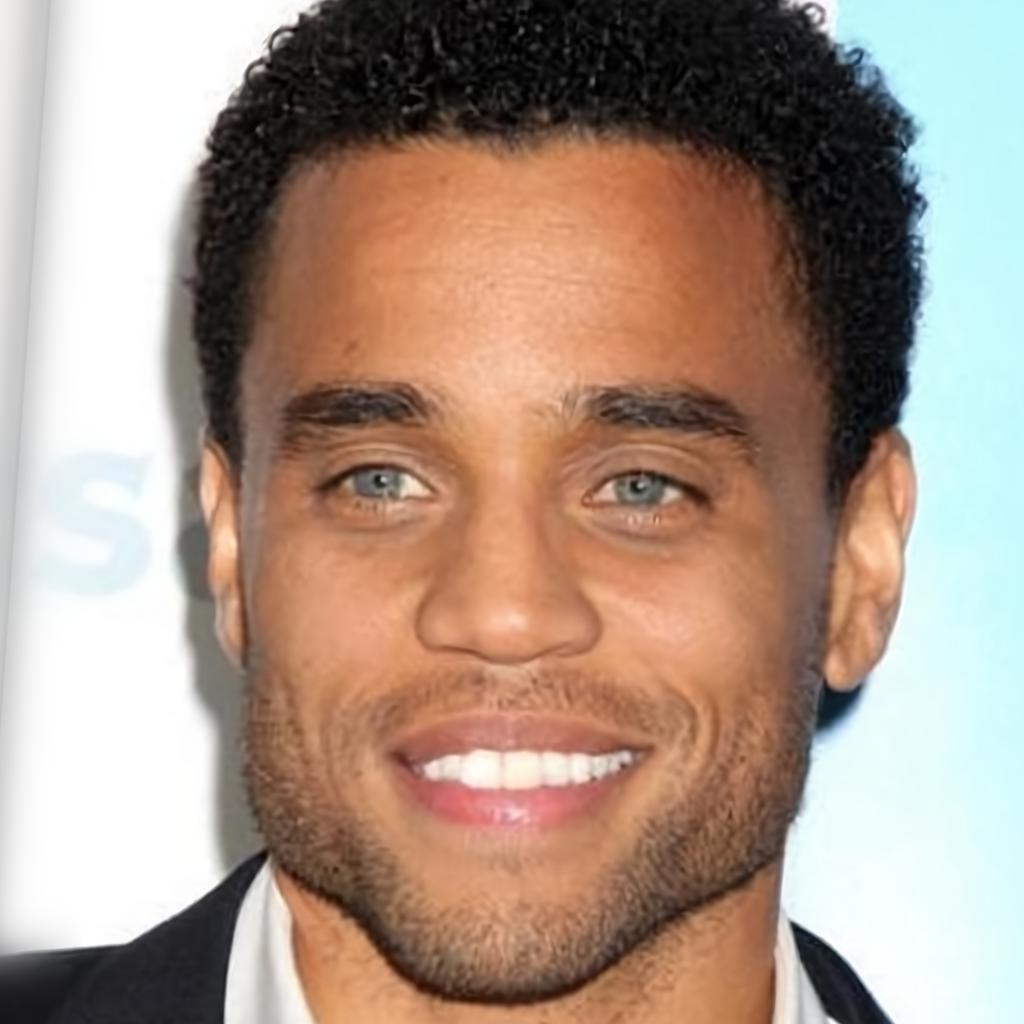} \end{tabular}
&
\begin{tabular}{c} \includegraphics[width=0.49\linewidth]{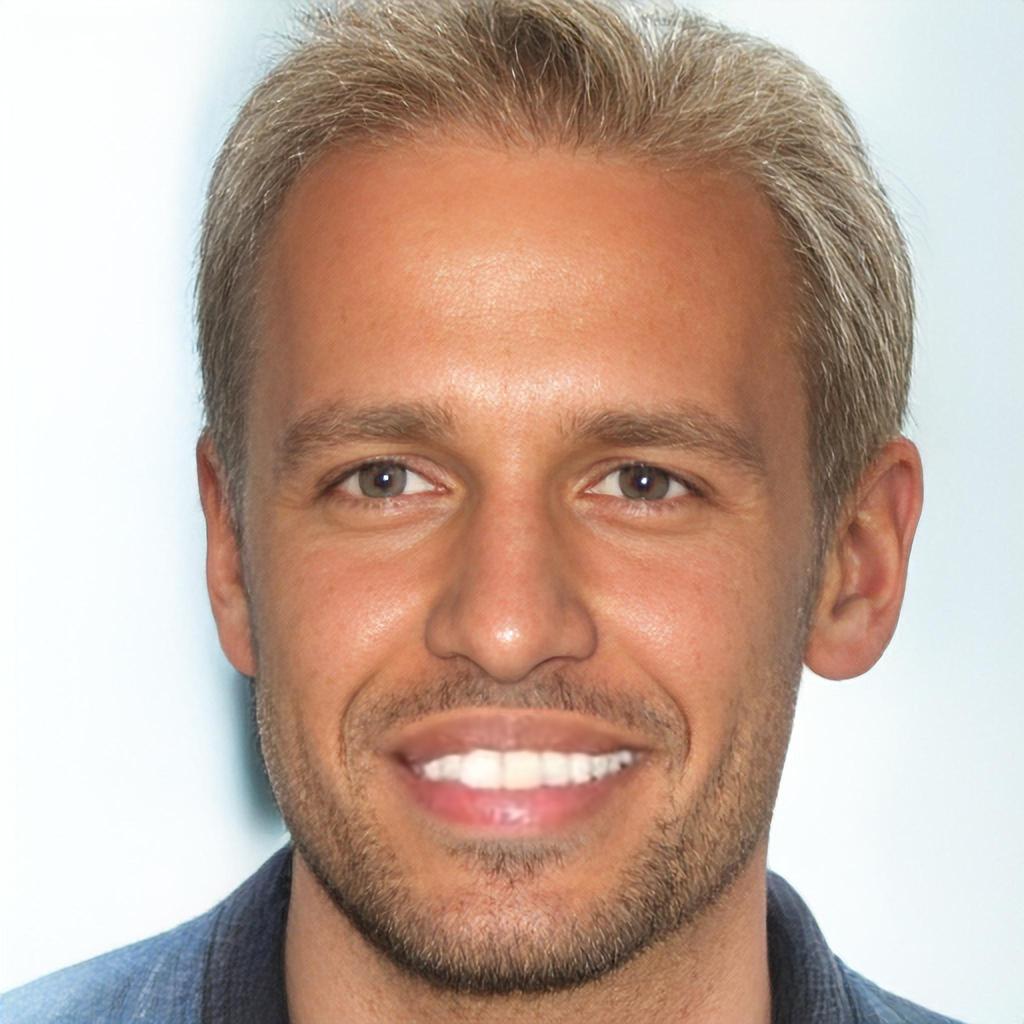} \end{tabular}
\\
\begin{tabular}{c} \includegraphics[width=0.49\linewidth]{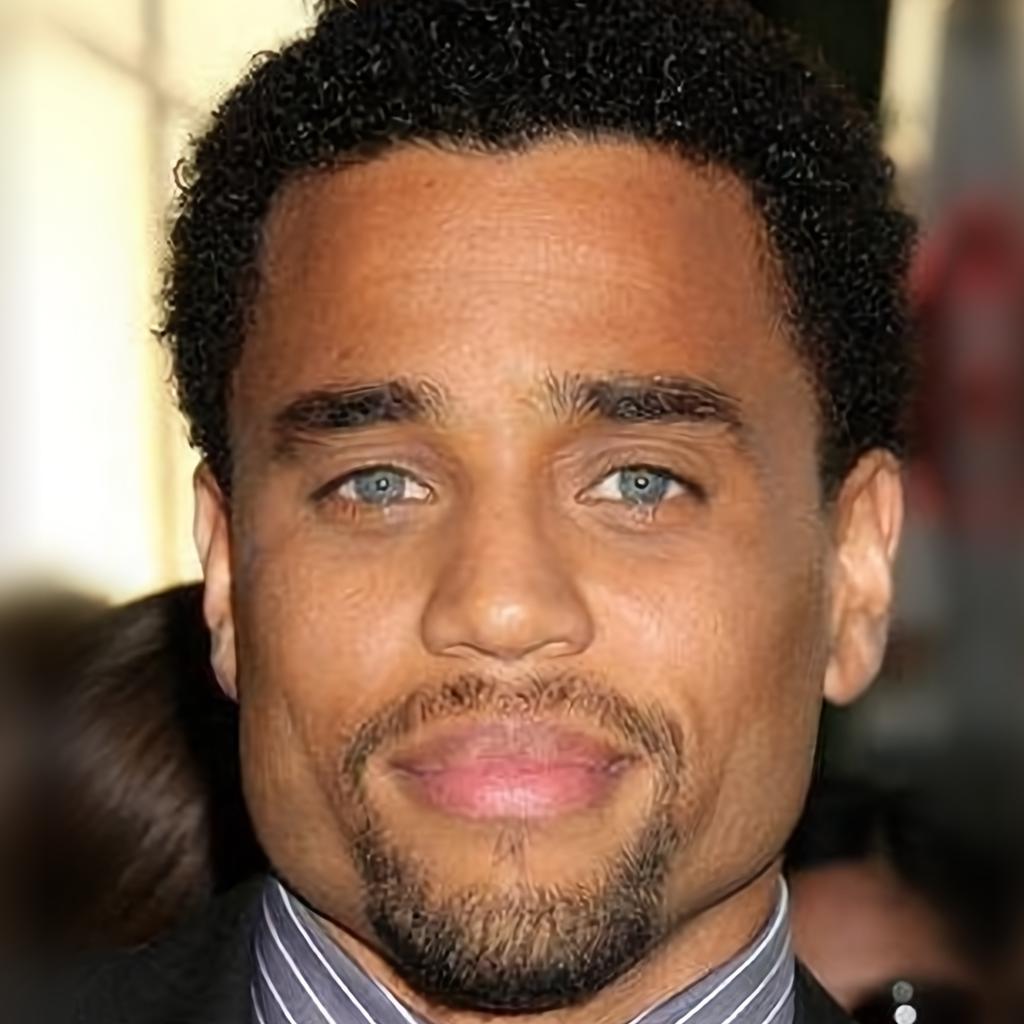} \end{tabular}
&
\begin{tabular}{c} \includegraphics[width=0.49\linewidth]{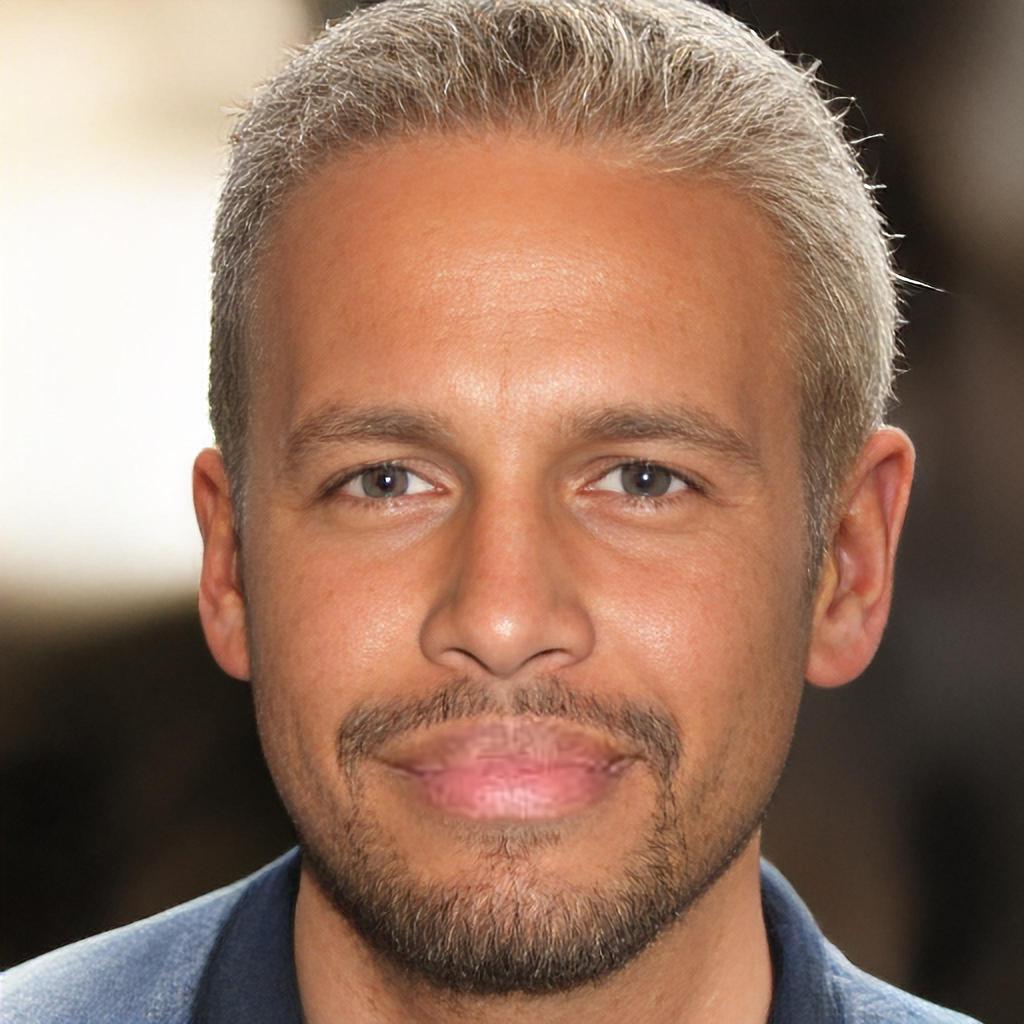} \end{tabular}
\\
\begin{tabular}{c} \makebox[0.49\linewidth]{Input \textbf{pair}} \end{tabular}
&
\begin{tabular}{c} \makebox[0.49\linewidth]{Ours (mouth preserved)} \end{tabular}
\\
\end{tabular}
\caption{Our \textit{clinical paired-image} anonymization preserving the mouth of the input pair, shown in larger resolution than the main manuscript for better illustration.}
\label{fig:paired_semantic_big}
\end{figure}

\endgroup
\begingroup
\setlength{\tabcolsep}{0.5pt} 
\renewcommand{\arraystretch}{0.7} 

\begin{figure}
    \centering
    \begin{tabular}{cccc}
    \multicolumn{4}{c}{
    \includegraphics[width=0.9\linewidth]{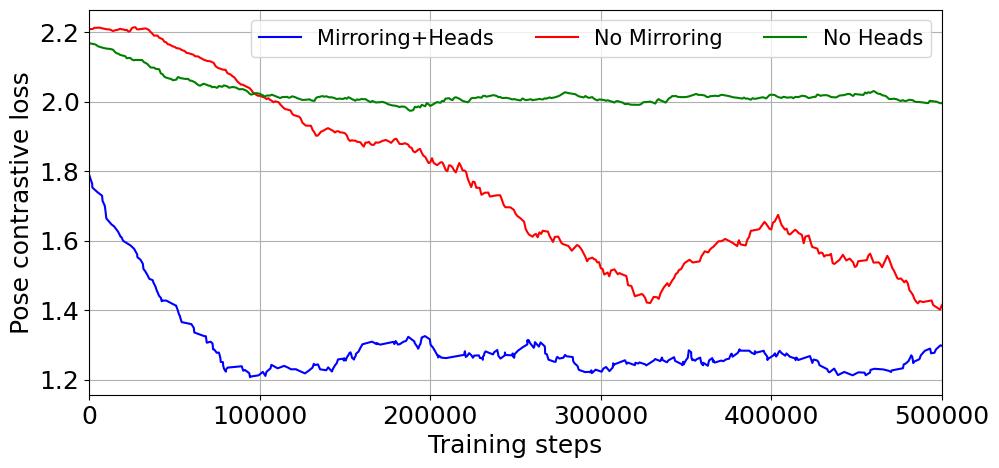}}\\
      \begin{tabular}{c} \rotatebox[origin=c]{90}{\makebox[1mm]{No mirroring}} \end{tabular} &
      \begin{tabular}{c} \includegraphics[width=0.3\linewidth]{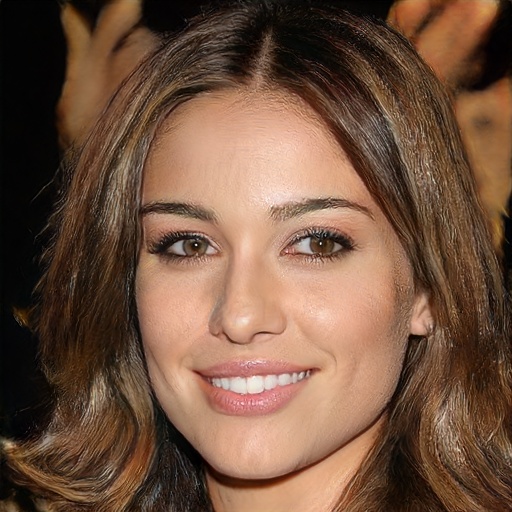} \end{tabular} &
      \begin{tabular}{c} \includegraphics[width=0.3\linewidth]{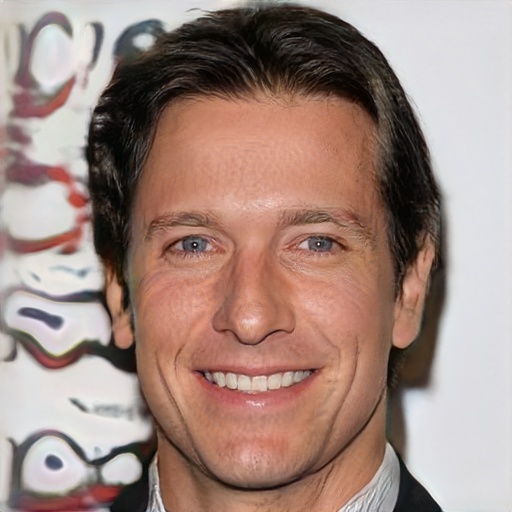} \end{tabular} &
      \begin{tabular}{c} \includegraphics[width=0.3\linewidth]{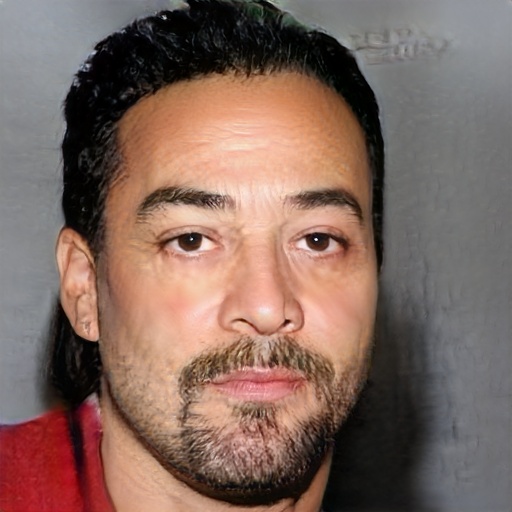} \end{tabular} \\
      \begin{tabular}{c} \rotatebox[origin=c]{90}{\makebox[1mm]{No heads}} \end{tabular} &
      \begin{tabular}{c} \includegraphics[width=0.3\linewidth]{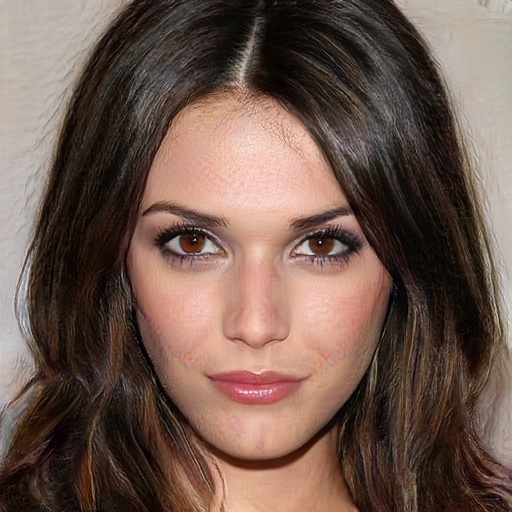} \end{tabular} &
      \begin{tabular}{c} \includegraphics[width=0.3\linewidth]{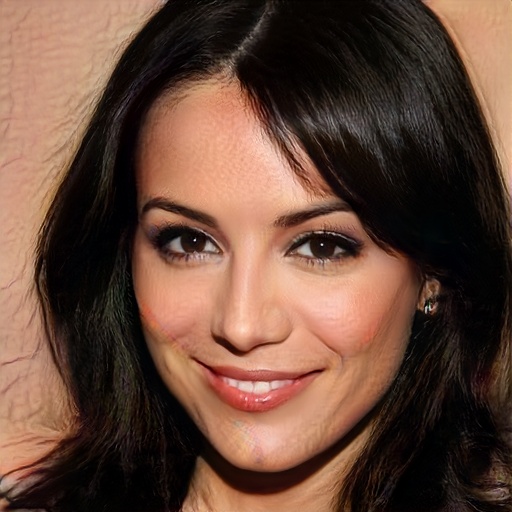} \end{tabular} &
      \begin{tabular}{c} \includegraphics[width=0.3\linewidth]{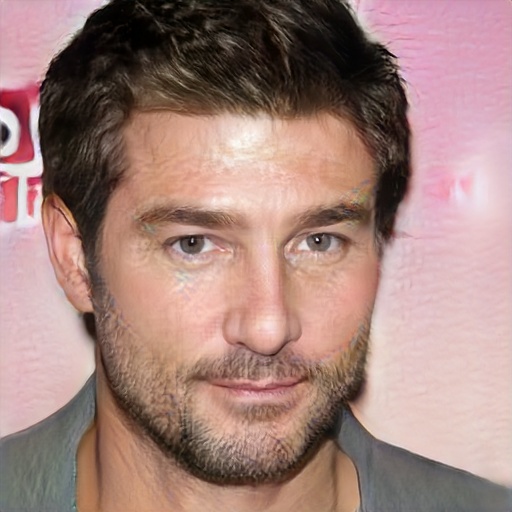} \end{tabular} \\
      \begin{tabular}{c} \rotatebox[origin=c]{90}{\makebox[1mm]{Mirroring+heads}} \end{tabular} &
      \begin{tabular}{c} \includegraphics[width=0.3\linewidth]{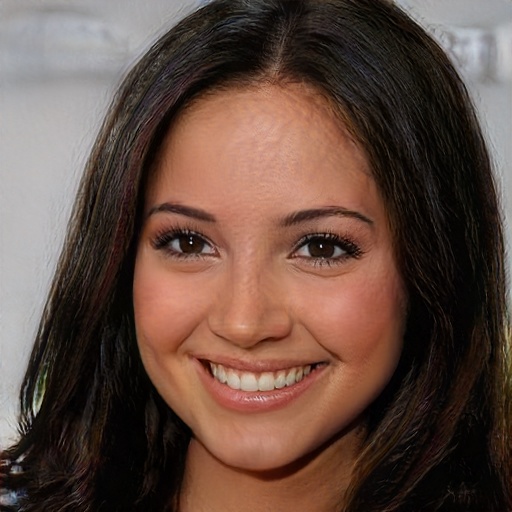} \end{tabular} &
      \begin{tabular}{c} \includegraphics[width=0.3\linewidth]{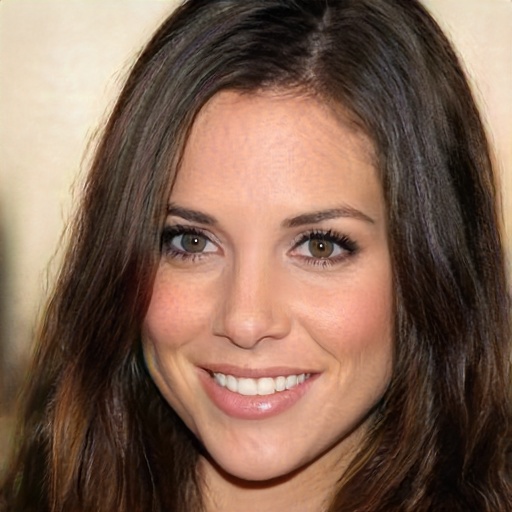} \end{tabular} &
      \begin{tabular}{c} \includegraphics[width=0.3\linewidth]{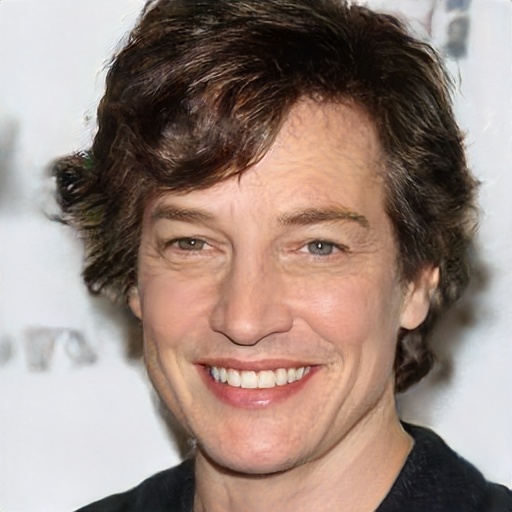} \end{tabular} \\
    \end{tabular}
    \caption{The top plot shows the training contrastive loss for the pose high-level attribute, with our mirroring strategy and projection heads, as well as with the ablation of each component individually. The best convergence is achieved with both of our components. The bottom part illustrates the resulting effect qualitatively. Each row samples multiple identities with a fixed target pose, however, only the last row successfully achieves the same consistent pose across all identities.}
    \label{fig:ablation}    
\end{figure}

\endgroup
\begin{table}[!t]
    \centering
    \resizebox{\columnwidth}{!}{%
    \begin{NiceTabular}{l|c|c||c|c|c|c}
    \toprule
    \multirow{2}{*}{Postprocessing method}              & \multicolumn{2}{c}{FID $\downarrow$} & \multicolumn{2}{c}{Bounding box $\uparrow$} & \multicolumn{2}{c}{Face detection $\uparrow$} \\
                                         & FFHQ   & CelebAHQ & MTCNN & Dlib  & MTCNN & Dlib  \\
    \midrule
    No postprocessing           & 56.92  & 15.81    & 0.908  & 0.955  & 0.963 & 0.990 \\
    Only correction         & 57.86  & 14.03    & 0.908  & 0.956  & 0.966 & 0.994 \\
    Only blending         & 56.93  & 13.28    & 0.907  & 0.949  & 0.959 & 0.990 \\
    Correction \& blending              & 57.89  & 12.49    & 0.908  & 0.952  & 0.962 & 0.990 \\
    \bottomrule
    \end{NiceTabular}
    }
    \caption{Effects of prior-based blending and prior-based correction, illustrated through a downstream utility evaluation for photorealism/diversity (FID~\cite{heusel2017gans}), bounding box IoU, and face detection rates (MTCNN~\cite{zhang2016joint}, Dlib~\cite{king2009dlib}) computed over held-out test set from CelebAMaskHQ. Results are comparable with Table 5 from the main manuscript, although there exists a slight difference in the metric for the complete pipeline, accounted by the hardware differences in these two experiments.}
    \label{table:postproc_ablation_utility}
\end{table}

\begingroup
\setlength{\tabcolsep}{0.5pt} 
\renewcommand{\arraystretch}{0.7} 

\begin{figure*}[t]
\centering
\begin{tabular}{cccc}
\begin{tabular}{c} \includegraphics[width=0.24\linewidth]{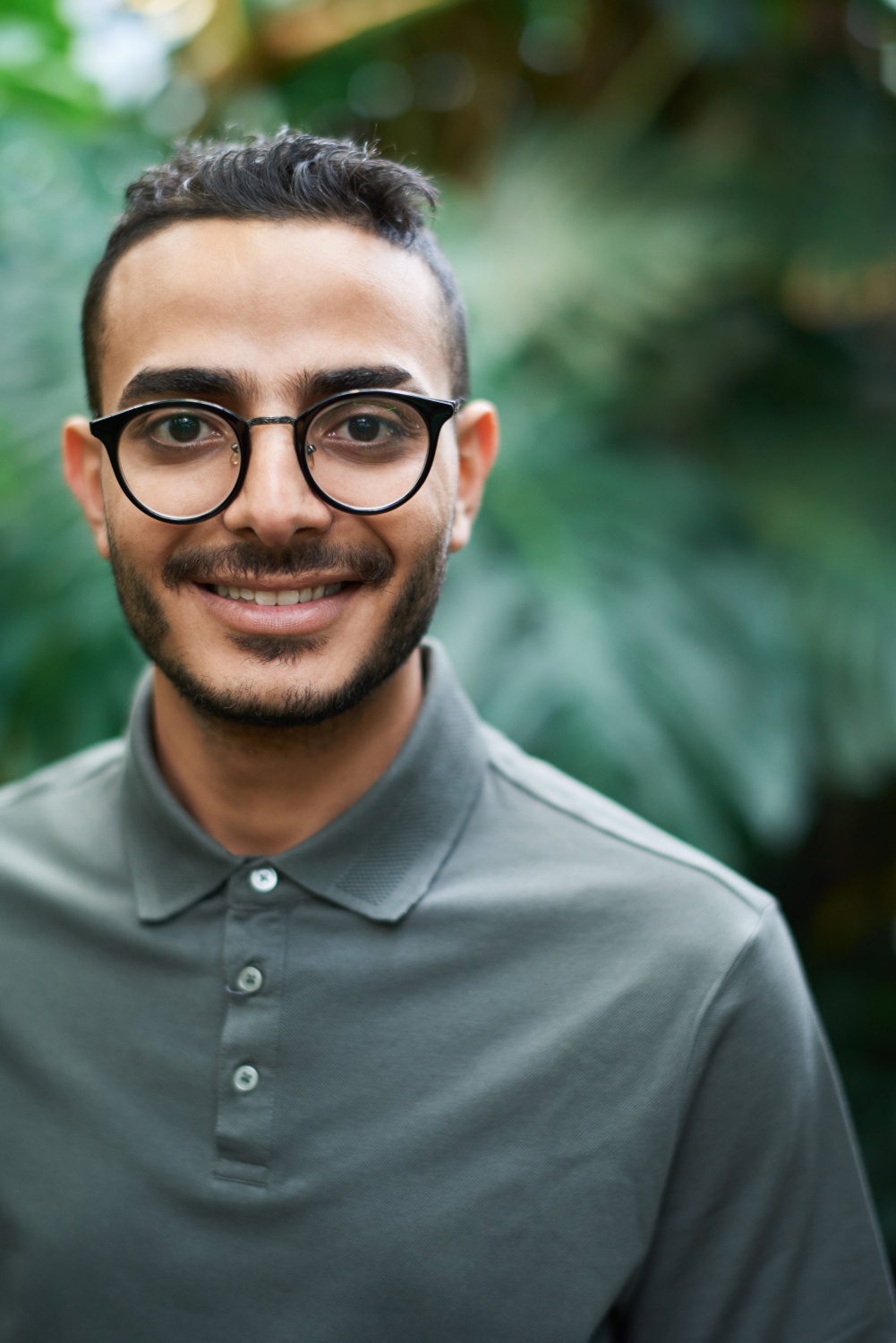} \end{tabular}
&
\begin{tabular}{c} \includegraphics[width=0.24\linewidth]{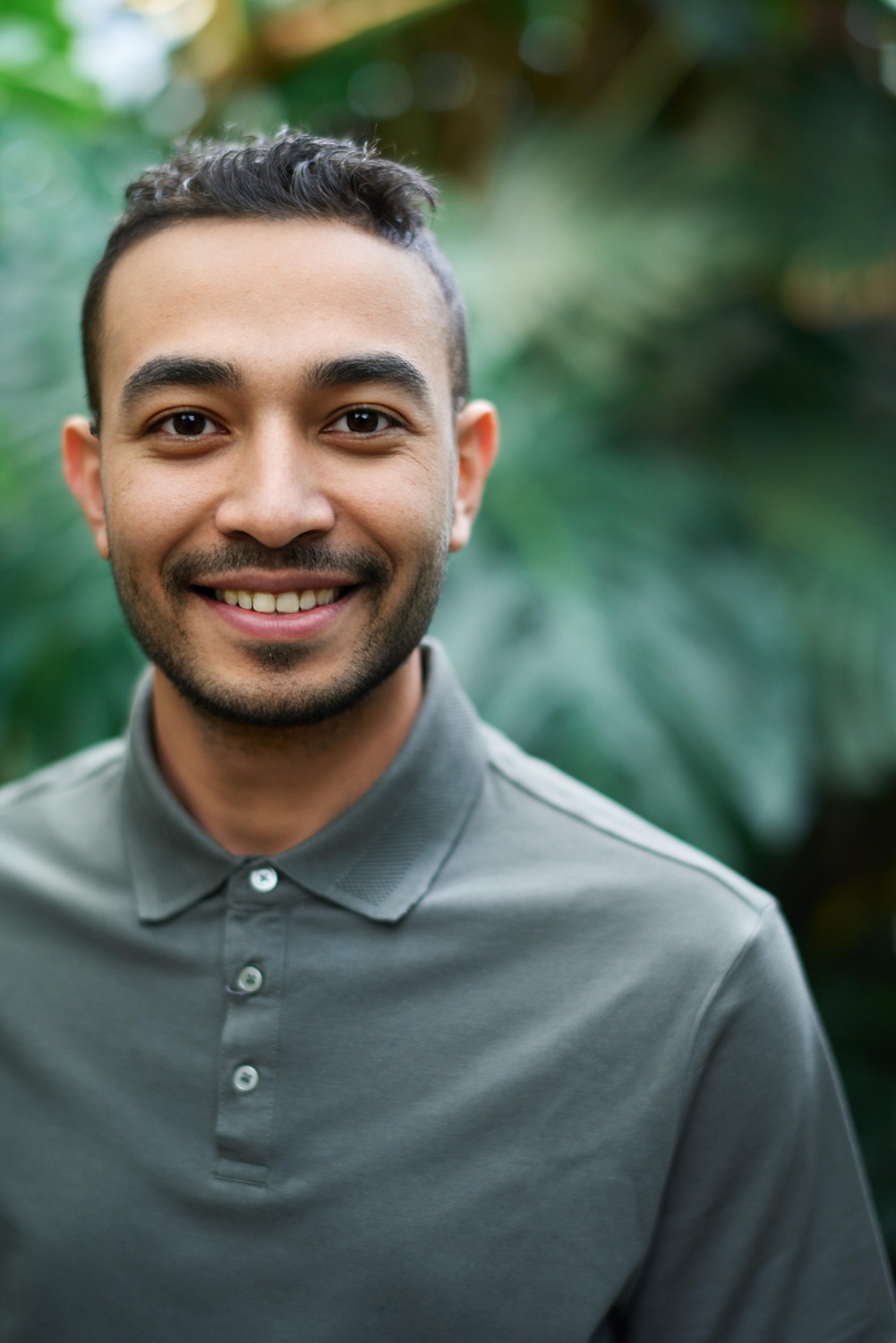} \end{tabular}
&
\begin{tabular}{c} \includegraphics[width=0.24\linewidth]{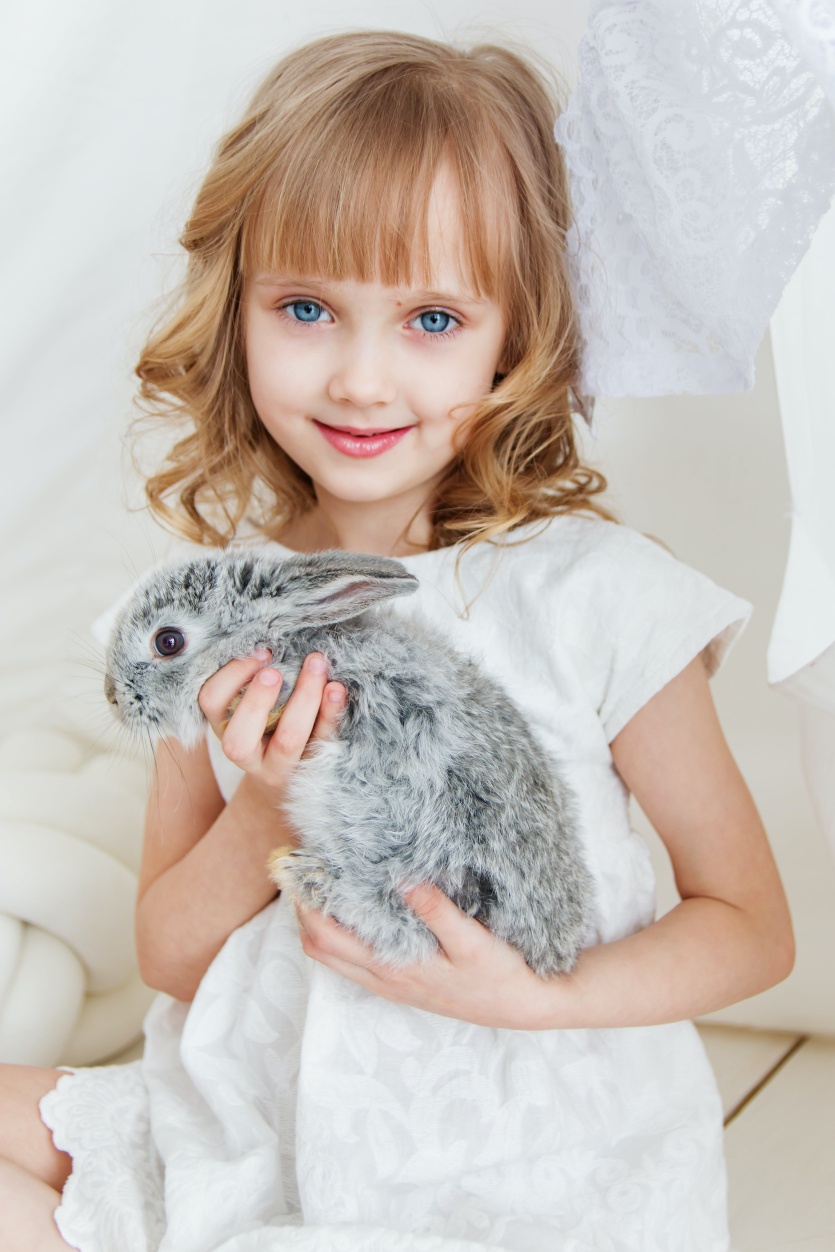} \end{tabular}
&
\begin{tabular}{c} \includegraphics[width=0.24\linewidth]{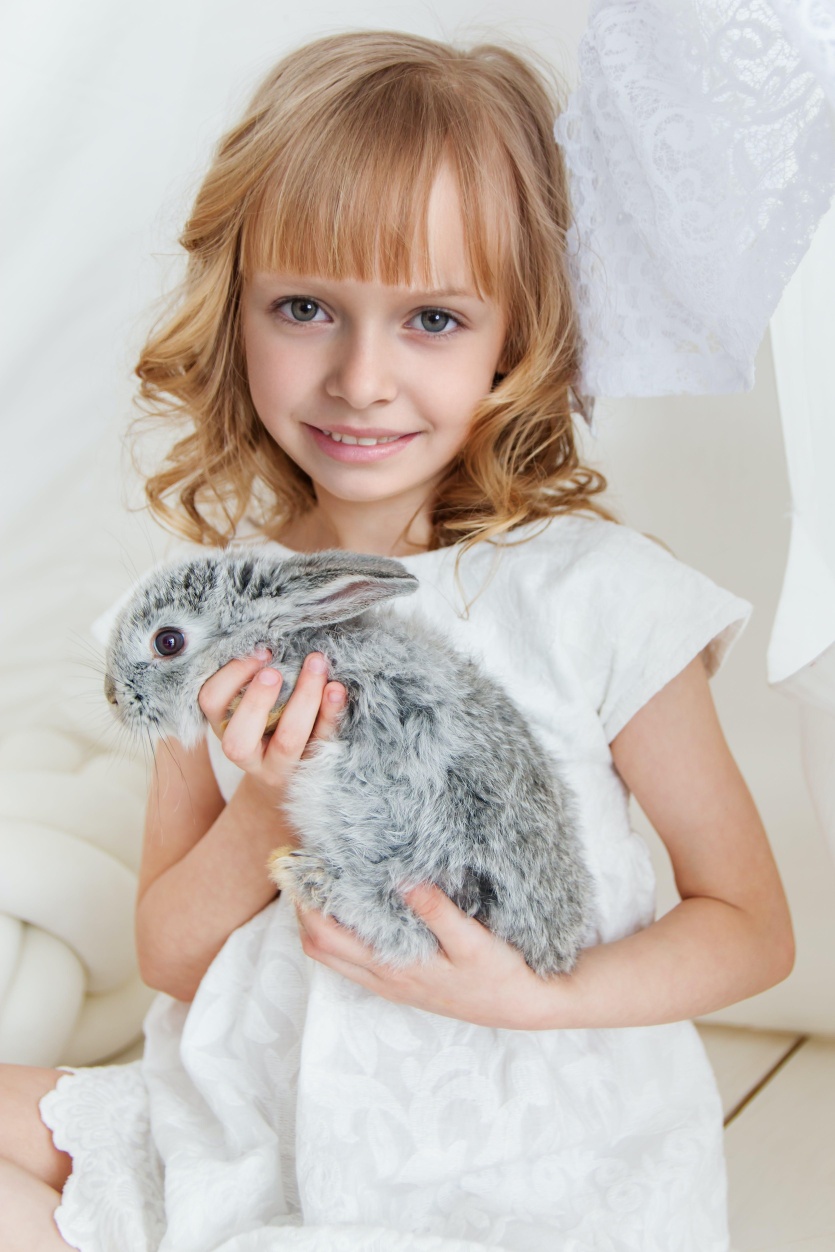} \end{tabular}
\\
\begin{tabular}{c} \includegraphics[width=0.24\linewidth]{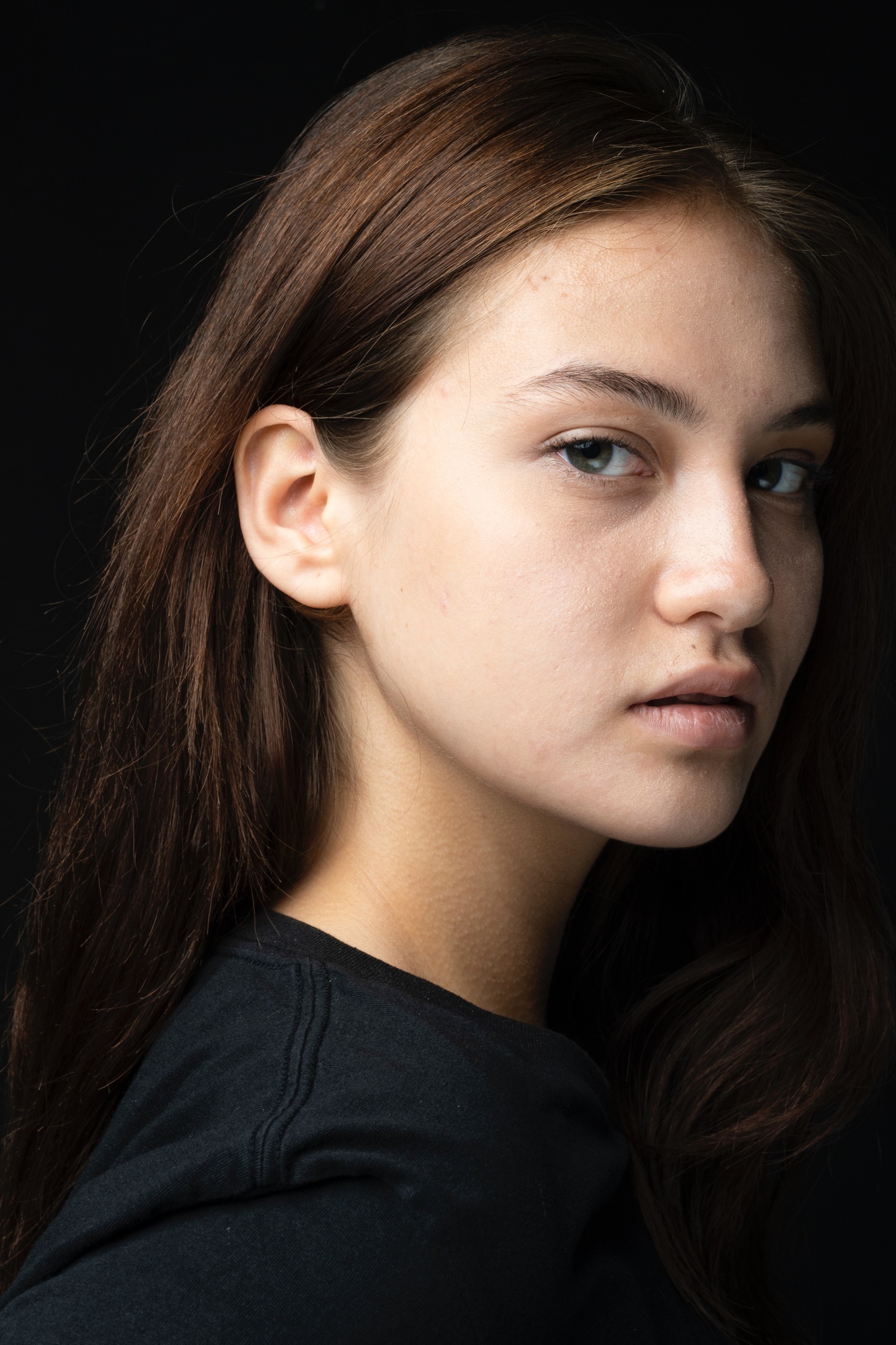} \end{tabular}
&
\begin{tabular}{c} \includegraphics[width=0.24\linewidth]{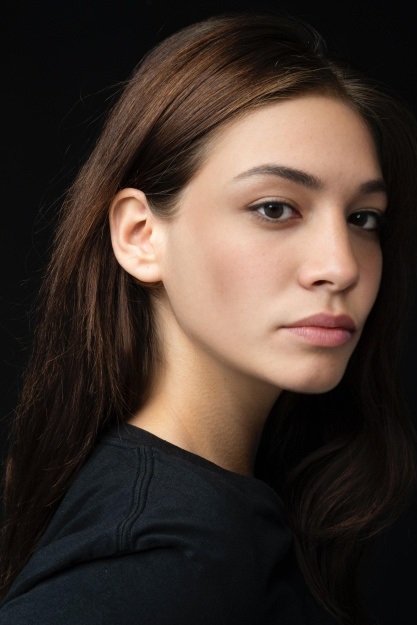} \end{tabular}
&
\begin{tabular}{c} \includegraphics[width=0.24\linewidth]{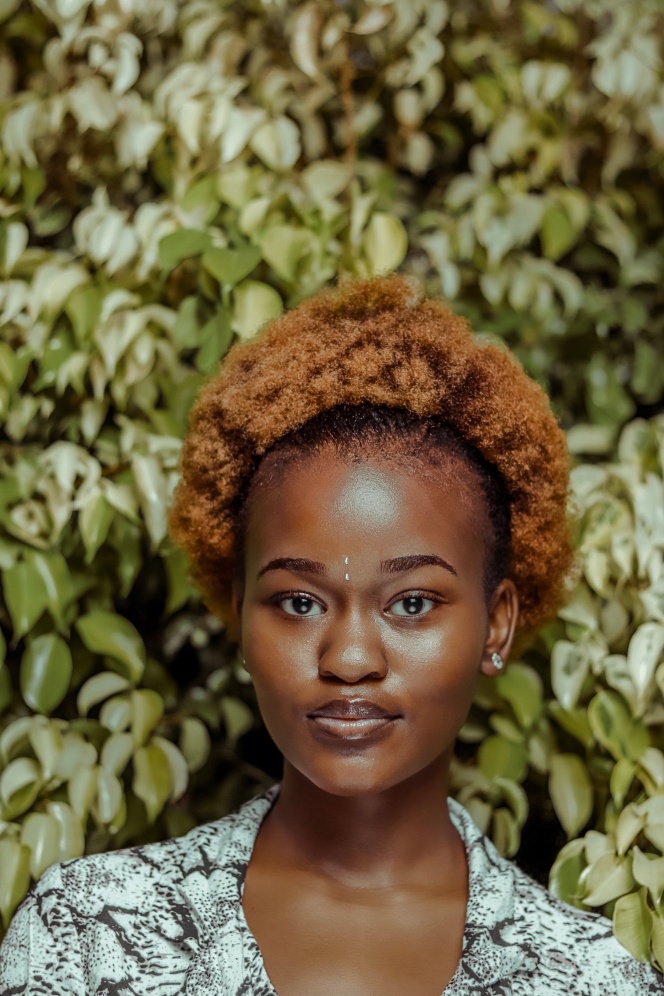} \end{tabular}
&
\begin{tabular}{c} \includegraphics[width=0.24\linewidth]{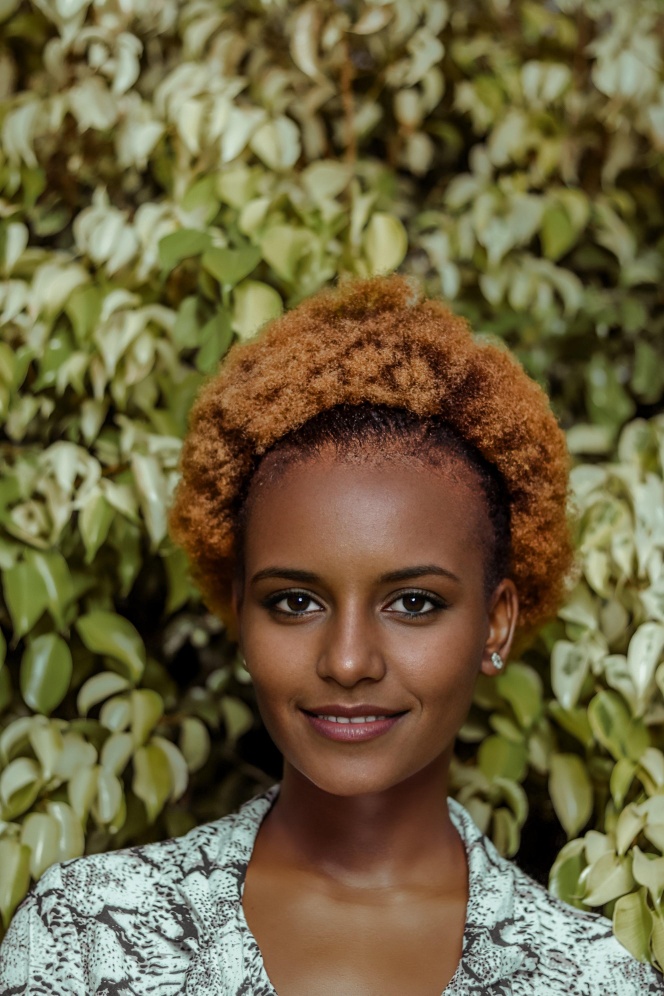} \end{tabular}
\\
\begin{tabular}{c} \includegraphics[width=0.24\linewidth]{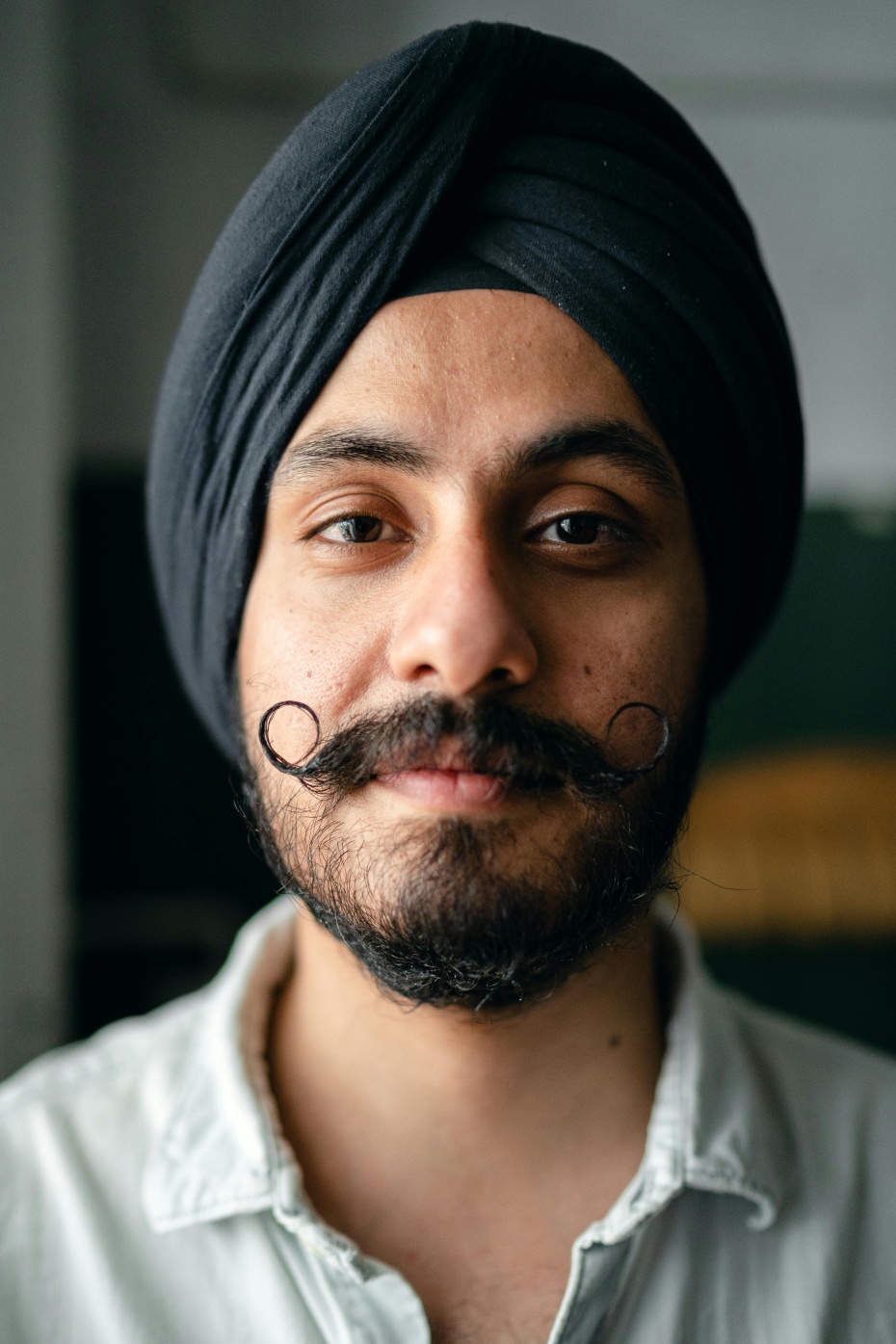} \end{tabular}
&
\begin{tabular}{c} \includegraphics[width=0.24\linewidth]{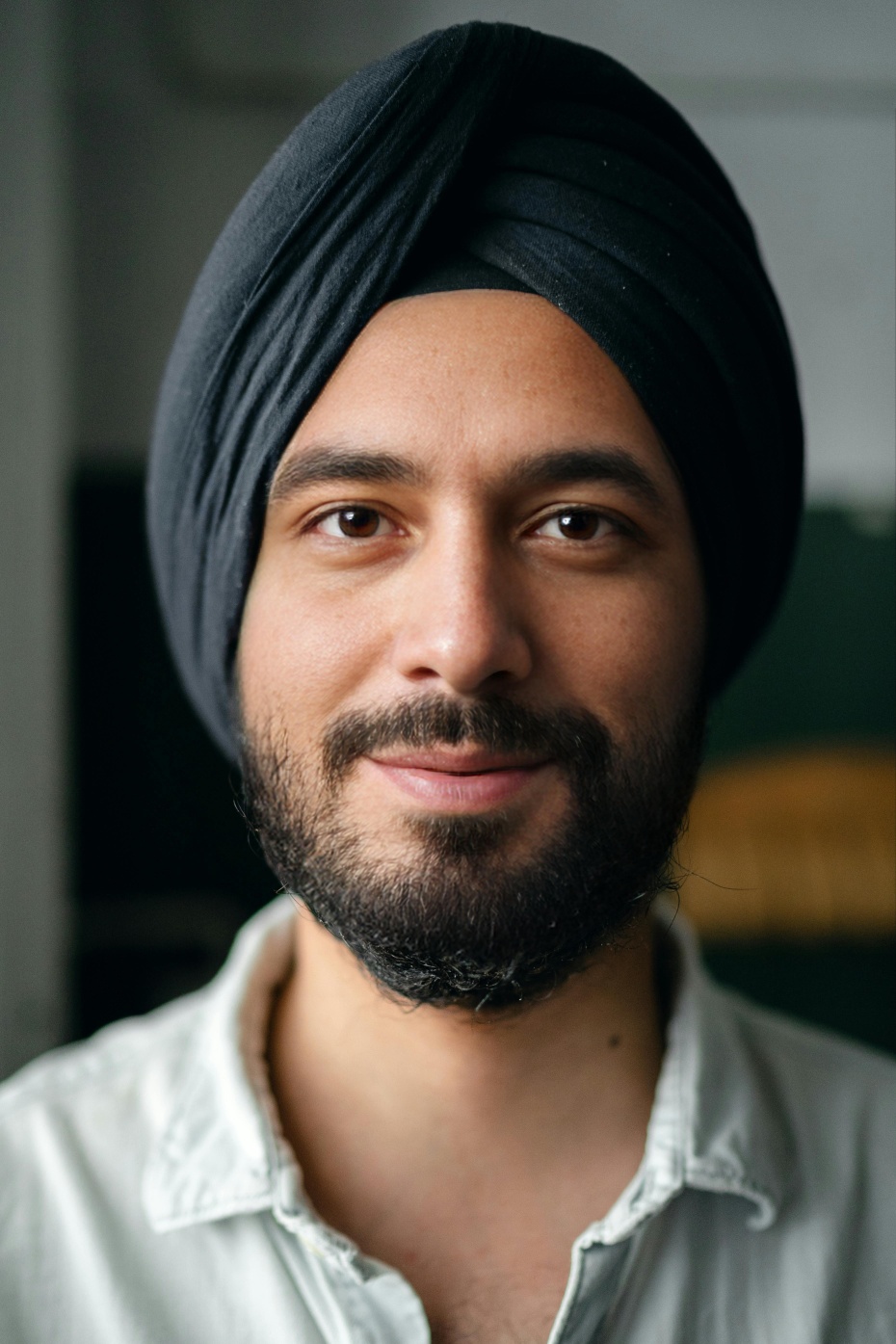} \end{tabular}
&
\begin{tabular}{c} \includegraphics[width=0.24\linewidth]{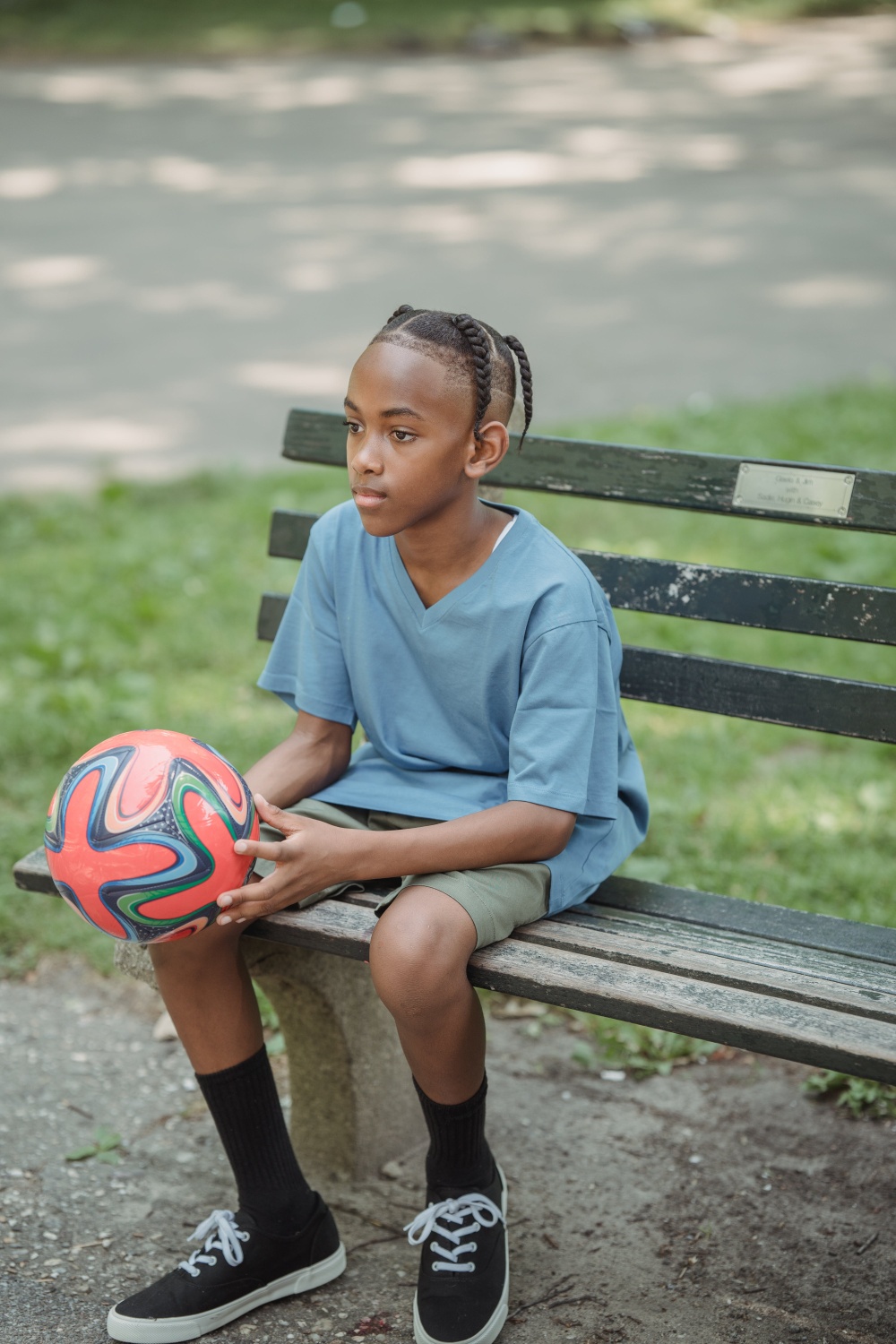} \end{tabular}
&
\begin{tabular}{c} \includegraphics[width=0.24\linewidth]{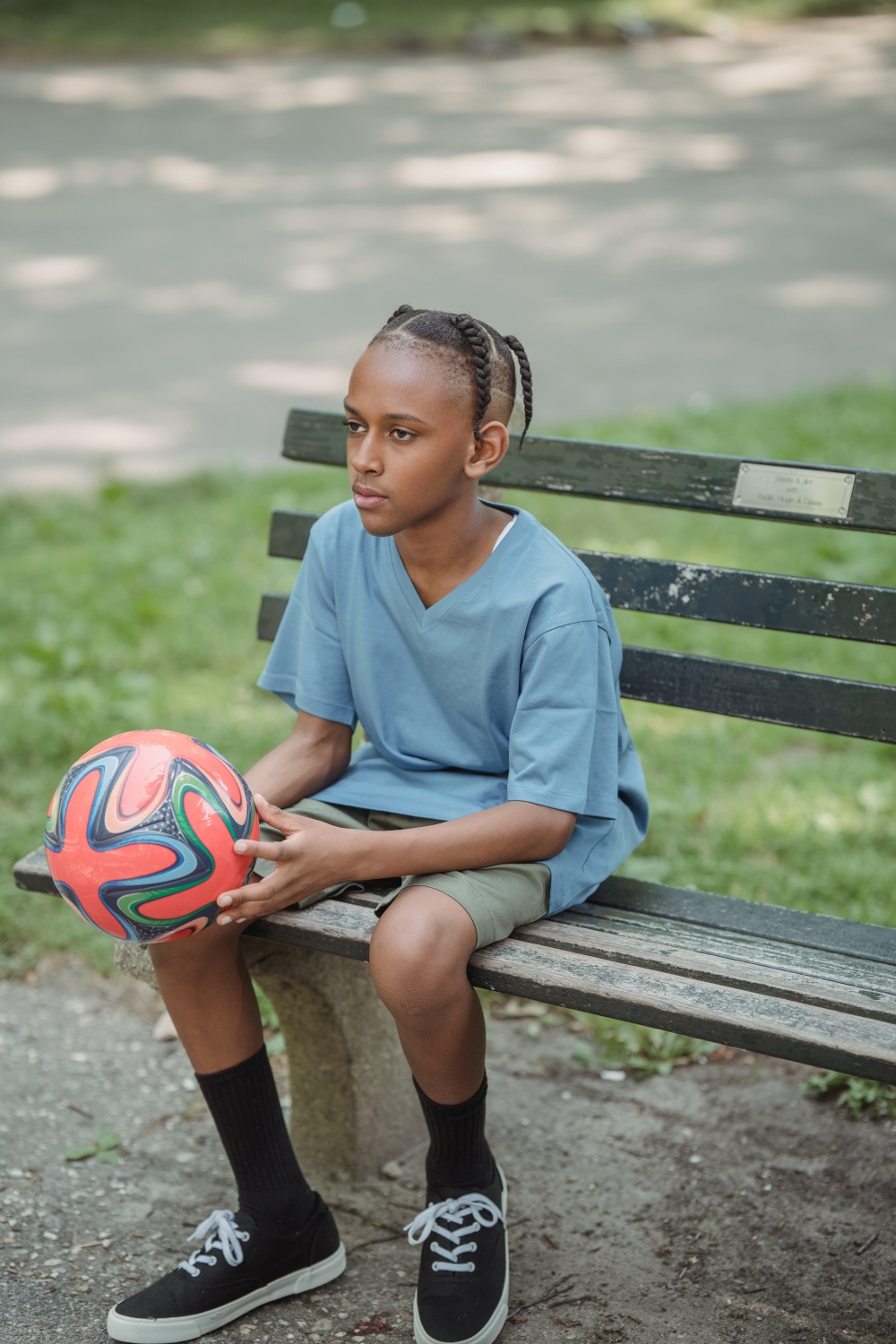} \end{tabular}
\\
\begin{tabular}{c} \includegraphics[width=0.24\linewidth]{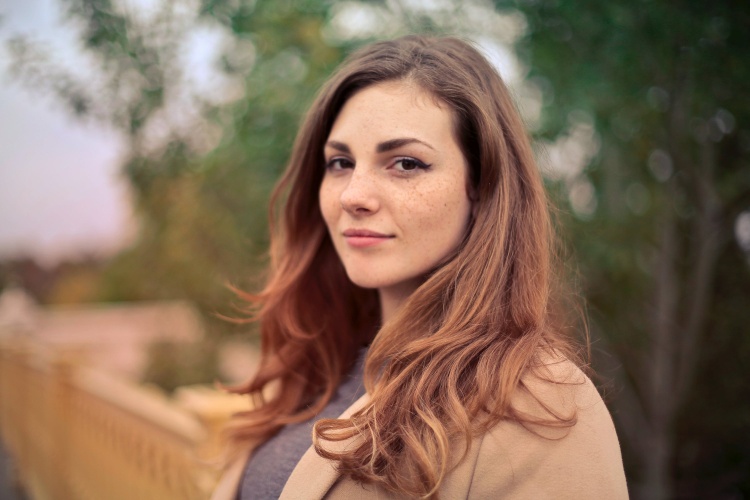} \end{tabular}
&
\begin{tabular}{c} \includegraphics[width=0.24\linewidth]{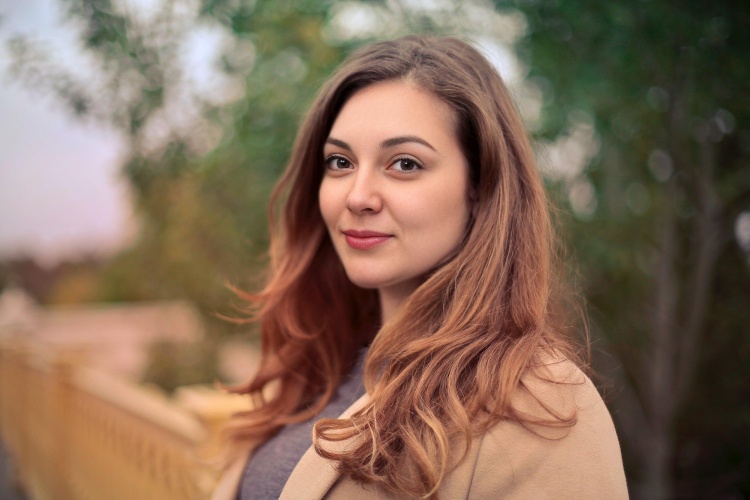} \end{tabular}
&
\begin{tabular}{c} \includegraphics[width=0.24\linewidth]{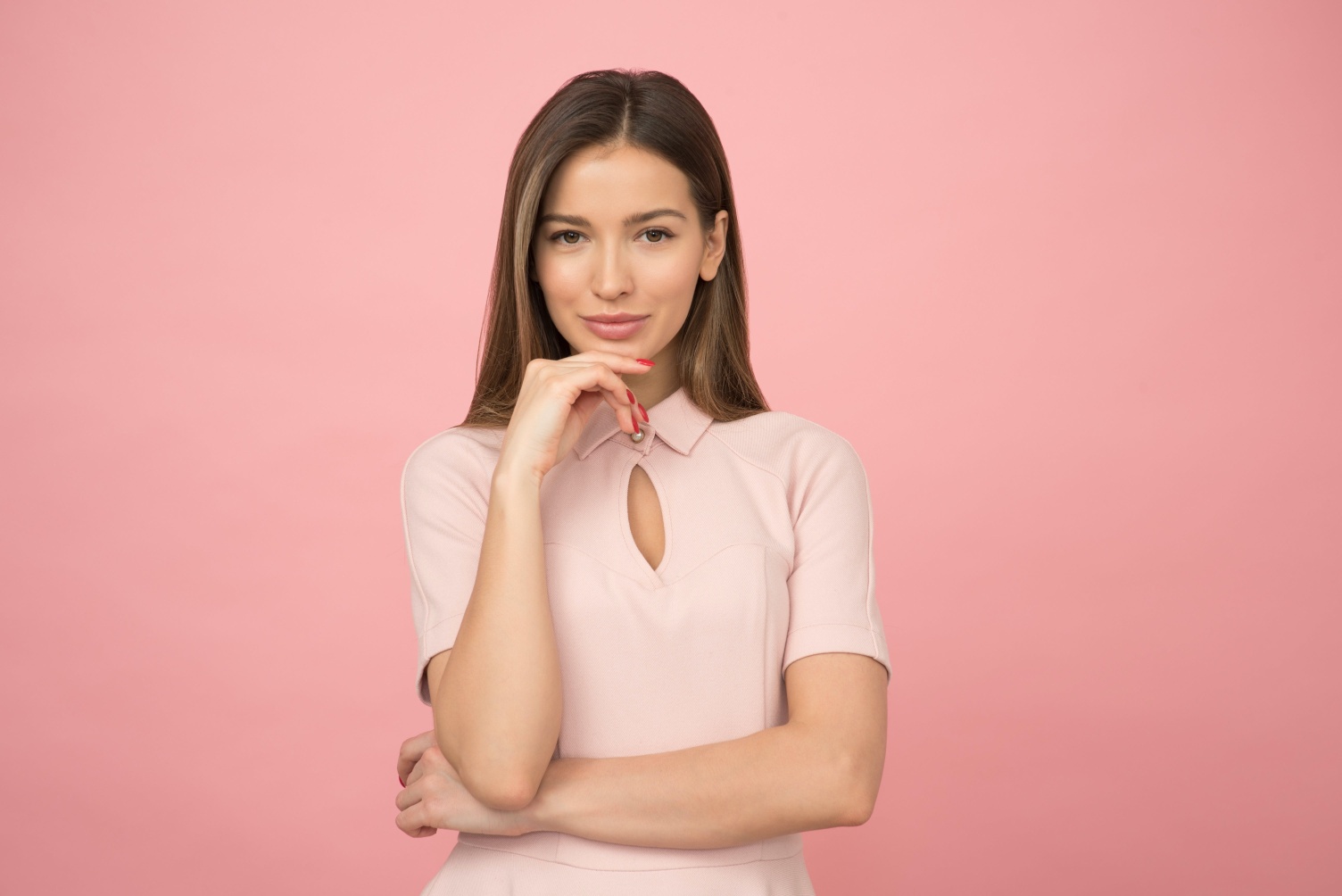} \end{tabular}
&
\begin{tabular}{c} \includegraphics[width=0.24\linewidth]{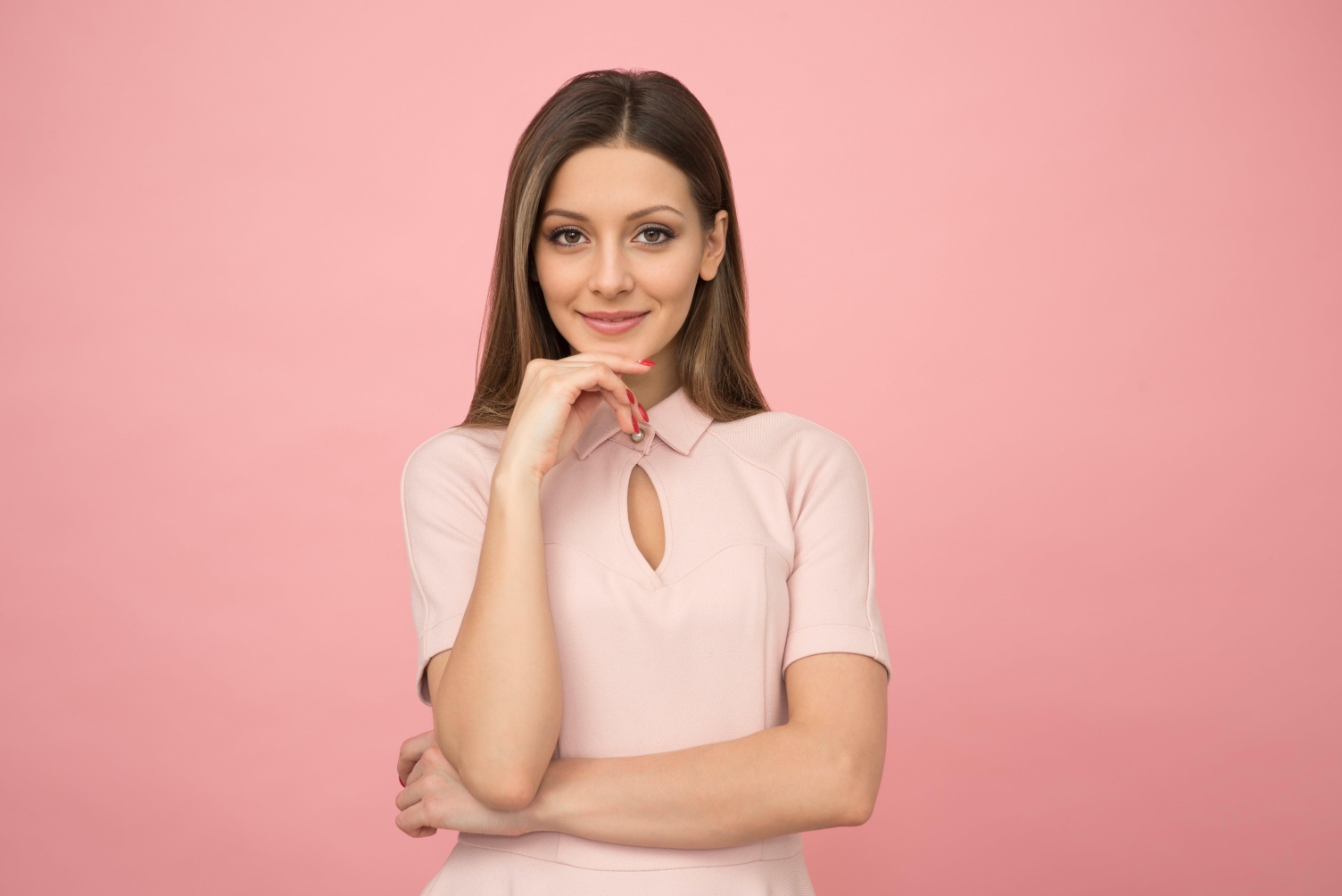} \end{tabular}
\\
\end{tabular}
\caption{Sample facial anonymization results in full-scene images, all performed using our proposed VerA. VerA works on aligned faces, therefore, we crop the aligned face from the full image as input and replace it by the anonymized face at the same location, following the standard in-place anonymization procedure.}
\label{fig:anon_inplace_full} 
\end{figure*}

\endgroup

\clearpage

\end{document}